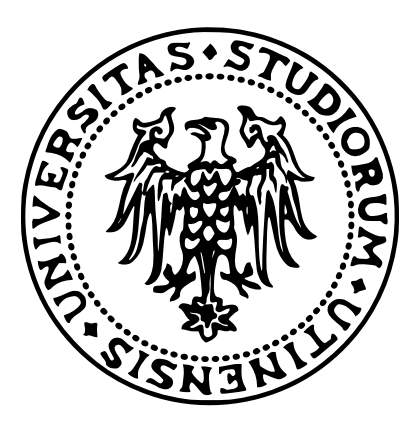

University of Udine

Department of Mathematics, Computer Science and Physics
Ph.D. Course in Computer Science, Mathematics and Physics

Dissertation

# TIMELINE-BASED PLANNING: EXPRESSIVENESS AND COMPLEXITY

Candidate
Nicola Gigante

Supervisor
Prof. Angelo Montanari, Ph.D.

Cosupervisors
Andrea Orlandini, Ph.D.
Prof. Mark Reynolds, Ph.D.

Cycle XXXI

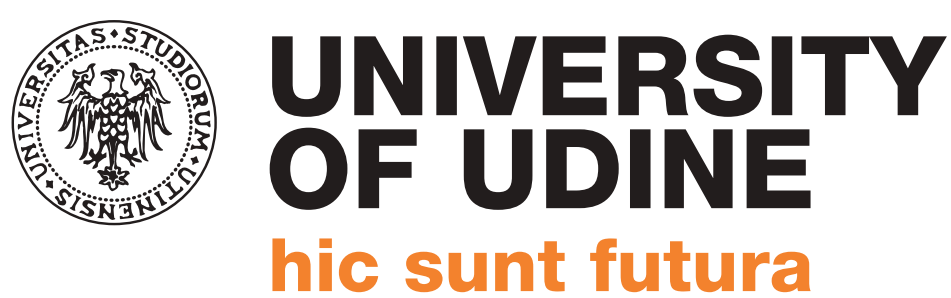

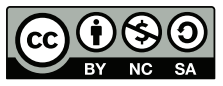

Department of Mathematics, Computer Science, and Physics
University of Udine
Via delle Scienze 208
Udine 33100 (UD), Italy

nicola.gigante@uniud.it

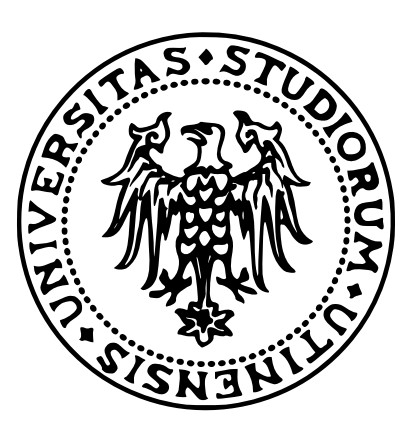

Università degli Studi di Udine

Dipartimento di Scienze Matematiche, Informatiche e Fisiche
Corso di Dottorato in Informatica e Scienze Matematiche e Fisiche

Dissertazione

# TIMELINE-BASED PLANNING: EXPRESSIVENESS AND COMPLEXITY

Candidato
Nicola Gigante

Supervisore
Prof. Angelo Montanari

Cosupervisori
Dott. Andrea Orlandini
Prof. Mark Reynolds

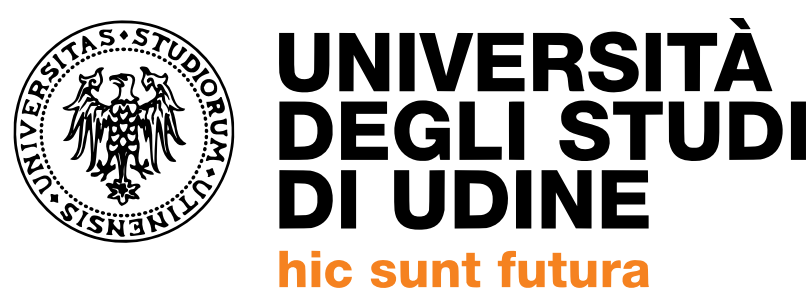

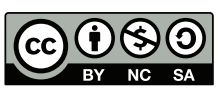

Dipartimento di Scienze Matematiche, Informatiche e Fisiche
Università degli Studi di Udine
Via delle Scienze 208
Udine 33100 (UD), Italia

nicola.gigante@uniud.it

# ABSTRACT

Automated planning is an area of artificial intelligence which aims at developing systems capable of autonomously reason about how to obtain a given goal, suitably interacting with their environment. *Timeline-based planning* is an approach originally developed in the context of space mission planning and scheduling, where problem domains are modelled as systems made of a number of independent but interacting components, whose behaviour over time, the *timelines*, is governed by a set of temporal constraints. This approach is different from the perspective adopted by more common planning languages, which reason in terms of which *actions* one or more agents can execute in order to obtain their goals.

Timeline-based systems have been successfully deployed in a number of complex real-world tasks, from mission planning and control to on-board autonomy for space exploration, over the past twenty years. However, despite this practical success, a thorough theoretical understanding of the paradigm was missing.

This thesis addresses this issue, by providing the first detailed account of formal and computational properties of the timeline-based approach to planning. In particular, it focuses on *expressiveness* and *computational complexity* issues. At first, we compare the expressiveness of timeline-based and action-based planning languages, showing that a particularly restricted variant of the formalism is already expressive enough to compactly capture action-based *temporal planning* problems. Then, we move to the characterisation of the problem in terms of computational complexity, showing that finding a solution plan for a timeline-based planning problem is EXPSPACE-complete. We also show that finding *infinite* solution plans is EXPSPACE-complete as well, and that finding solution plans of bounded length is NEXPTIME-complete.

Then, we approach the problem of timeline-based planning with *uncertainty*, that is, problems that include external components whose behaviour is not under the control of the planned system. We analyse the state-of-the-art approach to these problems, based on the concept of *flexible plans*, identifying some key issues, and then we propose an original solution based on *timeline-based games*, a game-theoretic interpretation of the problem as a two-players game where the controller, in order to win, has to execute a plan guaranteeing to satisfy the problem constraints independently from the behaviour of the environment. We show that this approach is strictly more general than the current one based on flexible plans, and we show that the problem of deciding whether a winning strategy for such games exists belongs to 2EXPTIME.

In the last part of the thesis, we provide a characterisation of the expressiveness of timeline-based languages in *logical* terms. We show that timeline-based planning problems can be expressed by *Bounded* TPTL *with Past* (TPTL$_b$+P), a variant of the classic *Timed Propositional Temporal Logic* (TPTL). We introduce TPTL$_b$+P, showing that, while TPTL *with Past* (TPTL+P) is known to be *non-elementary*, the satisfiability problem for TPTL$_b$+P is EXPSPACE-complete. Then, we describe a tableau method for TPTL and TPTL$_b$+P, extending a *one-pass tree-shaped* tableau recently introduced for *Linear Temporal Logic* (LTL), which is presented with a conceptually easier proof of its completeness based on a novel model-theoretic argument, and extended in order to support *past modalities*.

# SOMMARIO

La *pianificazione automatica* (automated planning) è un'area dell'intelligenza artificiale che mira allo sviluppo di sistemi in grado di ragionare autonomamente per perseguire determinati obbiettivi, interagendo di conseguenza con l'ambiente circostante. Il planning *basato su timeline* (timeline-based planning) è un approccio originariamente sviluppato per la gestione di missioni spaziali, secondo cui i problemi vengono modellati come sistemi composti da una moltitudine di componenti indipendenti, ma interagenti tra loro. Il comportamento di tali componenti nel tempo, descritto dalle *timeline*, è governato da un insieme di vincoli temporali. Questo approccio adotta una prospettiva differente da quella dei linguaggi di pianificazione più comunemente usati, che lavorano in termini di *azioni* (action-based) che uno o più agenti possono eseguire per ottenere i propri obbiettivi.

Negli ultimi vent'anni, i sistemi timeline-based sono stati impiegati con successo in molti scenari concreti, dal controllo di satelliti fino a missioni di esplorazione spaziale. Nonostante questo successo applicativo, manca una completa comprensione del paradigma dal punto di vista formale.

Questa tesi intende riempire tale vuoto, fornendo la prima dettagliata analisi delle proprietà formali e computazionali del timeline-based planning. In particolare, ci concentriamo su questioni di *espressività* e *complessità computazionale*. Inizialmente, compariamo l'espressività dei linguaggi di modellazione timeline-based e action-based, mostrando che una particolare restrizione del formalismo è già sufficientemente espressiva da catturare compattamente il planning *temporale* action-based. Dopodiché, caratterizziamo il problema in termini di complessità computazionale, mostrando che trovare una soluzione per un problema di timeline-based planning è EXPSPACE-completo. Mostriamo, inoltre, che la ricerca di soluzioni *infinite* è anch'essa EXPSPACE-completa, e che, dato un limite alla lunghezza delle soluzioni, il problema diventa invece NEXPTIME-completo.

Successivamente, approcciamo il problema del timeline-based planning con *incertezza*, in cui sono presenti delle componenti esterne il cui comportamento è fuori dal controllo del sistema. Analizziamo l'attuale approccio, basato sul concetto di *piano flessibile*, identificando alcune criticità, che approcciamo proponendo il concetto di *timeline-base game*. In questa interpretazione del problema basata sulla teoria dei giochi, il problema è visto come un gioco a due giocatori, in cui il controllore del sistema vince se riesce ad eseguire un piano garantendo il soddisfacimento dei vincoli, a prescindere dal comportamento dell'ambiente. Mostriamo la maggiore generalità di questo approccio rispetto all'attuale basato su piani flessibili, e dimostriamo che decidere se esiste una strategia vincente per tali giochi appartiene alla classe 2EXPTIME.

Nell'ultima parte della tesi, caratterizziamo l'espressività dei linguaggi di modellazione timeline-based in termini *logici*. Mostriamo che i problemi di timeline-based planning possono essere espressi dalla *Bounded* TPTL *with Past* (TPTL$_b$+P), una variante della classica logica *Timed Propositional Temporal Logic* (TPTL). Introduciamo TPTL$_b$+P, mostrando che, mentre la soddisfacibilità di TPTL *with Past* (TPTL+P) è nota essere *non elementare*, il problema per TPTL$_b$+P è EXPSPACE-completo. Dopodiché, descriviamo un metodo a *tableau* per TPTL e TPTL$_b$+P, estendendo un tableau *ad albero* recentemente introdotto per la *Linear Temporal Logic* (LTL), che viene presentato con una diversa e concettualmente più semplice dimostrazione di completezza, ed esteso per supportare *operatori temporali al passato*.

# ACKNOWLEDGEMENTS

This thesis is the result of three years of hard work and it would not have been possible at all without the precious help of many people. My gratitude goes first of all to my parents, and my mother in particular, for their endless understanding and support, and of course to my beloved Fita: her unconditional love and esteem are my most precious resources and motivation.

As everything written here is the result of team-working, I want to thank all those who made it possible. First of all, I am greatly thankful to my supervisor Angelo Montanari, who has been a wise and understanding guide during all my journey. I have learnt from him quite everything I know about how to do this job, and to him also goes the merit of identifying the satisfactory research direction that we pursued together. Sincere thanks to Mark Reynolds, for his support and his very welcoming hospitality during my visit in the wonderful Perth, one of the best parts of my life. Thanks to Andrea Orlandini and Marta Cialdea Mayer, whose work put the basis for mine. They have been very helpful coauthors and an invaluable guide through the planning community. Everybody else I worked with, including Dario Della Monica, Pietro Sala, and Guido Sciavicco, merit a big thanks for their precious advice. I also want to thank Simone Fratini and Nicolas Markey for carefully reviewing my thesis and giving me many precise and insightful suggestions, which helped me substantially improve the manuscript.

Many thanks to all the people from Rome that I met around the world, including Amedeo Cesta, Riccardo Rasconi, Angelo Oddi, Andrea Orlandini, and Alessandro Umbrico, for their friendly support and company, and in particular Simone Fratini, for his invaluable role as a mentor at the ICAPS '17 Doctoral Consortium. I have learnt a lot from that experience, and it gave me a fundamental motivation boost.

Finally, I want to thank all those people who shared this path with me. Thanks to Alberto Molinari, surprising travel mate since the very first conferences, and to Marta Fiori Carones for the many hours spent in stimulating conversations. Special thanks to Manlio Valenti for all the timely scheduled coffees, to Tobia Dondè and Davide Liessi for the relaxing and grammatically sound lunches, and to Andrea Viel, Andrea Brunello, Marco Peressotti, and all the other present and past colleagues. And, of course, thanks to Nicola Prezza, for his uncompressible friendship. Life is a circle: best wishes to Luca Geatti, who is starting his own journey right now.

Research is not just a job. You need to keep believing in what you are looking for. Thanks to all those who helped me believe in what I do.

# I

# INTRODUCTION

# 1 INTRODUCTION

This dissertation provides a detailed theoretical investigation of *timeline-based planning*, studying the computational complexity of the involved decision problems and the expressiveness of timeline-based modelling languages. In the timeline-based approach to automated planning, problem domains are modelled as systems made of independent, but interacting, components, whose behaviour over time, represented by the *timelines*, is governed by a set of temporal constraints. Introduced in the context of planning and scheduling for space missions, it has been successfully applied in a number of complex real-world scenarios, but its formal and theoretical properties are not yet fully understood. This thesis fills this gap, by providing a first detailed account of computational complexity and expressiveness issues. This introductory chapter gives a brief overview of the history and relevant literature about automated planning in general, and timeline-based planning in particular, introduces the topic of temporal logic, which will play an important role in later chapters, and provides a detailed account of the motivations behind the work, surveying its novel contributions.

# 1 AUTOMATED PLANNING

Many technological applications are nowadays driven by *artificial intelligence*. The resurgence in popularity and the appearance of many successful applications of AI in the last decade is commonly attributed to the development of *deep learning* techniques, on one hand, and to the rise of sufficient computational power that allows such techniques to be fed with huge amounts of data, on the other. However, not all the application scenarios, not even the most popular ones, can be approached in their full generality with the exclusive use of learning techniques: many complex technological applications require to combine learning with some kind of *reasoning*. This is the case, for example, of *autonomous vehicles*, one of the current flagship applications of AI, which, besides great computer vision and perception challenges, requires as well to address reasoning and decision making issues, from motion planning problems to the interpretation and enforcement of road signals, to the vehicle-vehicle and road-vehicle cooperation needed to ensure driving safety and security [125].

Located under the broad umbrella of *artificial intelligence*, the discipline of *automated planning* studies the design and development of such *autonomous systems*, capable of *reasoning* about how to achieve their goals, and how to *act* accordingly, given a description of themselves and of their environment.

The adoption of a *model* of the world, that is, an high level description of how the system is supposed to work and to interact with its environment, and the use of general reasoning techniques on top of such model, are fundamental features of the artificial intelligence approach to these scenarios. The starting point is the abstract modelling of the world, which requires the application of expert knowledge of the application domain, and then the planning process is performed in the most domain-independent way possible, applying *general* reasoning techniques, which are, at least in principle, directly applicable to different scenarios, represented by different input models.

This *model-based* approach distinguishes AI planning from domain-specific techniques, where ad-hoc algorithmic solutions and software systems are developed to solve the specific problem at hand, which, however, require non-trivial modifications when the problem changes.

Using an internal representation of the world, to reason and decide how to act, appears to be similar to how *intelligent* beings reason in their everyday lives. This is not a coincidence, and indeed the automated planning field can trace its roots back to the very early days of artificial intelligence research, with the work of Turing [133], at first, and especially McCarthy and Hayes [98], later, which were mostly interested in the philosophical question of whether a machine could imitate human behaviour.

In their seminal work, McCarthy and Hayes defined the *situation calculus*, a first-order logical formalism able to reason in terms of *actions* that, when performed, affect the truth value of some predicates, called *fluents*, which describe the state of the world. This philosophical investigation had substantial technological impact. Indeed, just a few years later, the first practical general planning system, the *STanford Research Institute Problem Solver* [57] (STRIPS) was introduced, which borrowed most of its core concepts from situation calculus. In STRIPS, the *fluents*, *i.e.*, time-varying variables, describing the world are predicates ranging over finite domains, whose truth values form the current *state* of the system. Then, a number of actions are available to affect the state of the world. Actions may have preconditions, *i.e.*, logical formulae which have to be true for the action to be applicable to the current state, and their effects are expressed in terms of how they change the fluents truth value. Given an initial state, a *solution plan* to a STRIPS problem consists in a sequence of actions that manage to reach a state satisfying the goal condition.

STRIPS was a turning point in the history of automated planning research, so much that forty years later most state-of-the-art planning systems are still based on the state/action dichotomy, much research effort is still being devoted to efficient reasoning techniques for STRIPS-like planning, and most contemporary systems even share with STRIPS some pieces of its input syntax, in the form of the *Planning Domain Description Language* [99] (PDDL), the standard language used in the *international planning competitions* (IPC).

From the point of view of the employed solving techniques, STRIPS inherited the work done a few years earlier on the *General Problem Solver* program [107], which in turn was built on top of earlier work on theorem proving in first-order logic [106]. From a scientific point of view, it is remarkable that a research line that nowadays finds so many important application domains can trace its root back to philosophical argumentation, and that its first developments were possible only thanks to the earlier development of much more abstract fields, including not only *computability theory* – the concept of *universal machine* was fundamental in Turing's conception of the possibility of artificial intelligence itself [133] – but also earlier work on the *foundations of mathematics*, including *formal logic* and *proof theory*.

The performance of early systems, of course, suffered from the limited capabilities of the available hardware, and from the lack of solid search heuristics. Such gaps were rapidly filled in the subsequent years, with the most important turning points being the introduction of the GRAPHPlan planning system, based on the *planning graph* data structure [20], and the FF heuristic search system [74]. With the introduction of the *planning as satisfiability* approach [79], the advancements of fast planning techniques intersected with the contemporary development of fast solving techniques for the *boolean satisfiability problem* (SAT) [19, 52, 96]. In this approach, planning problems are encoded into SAT formulae, exploiting fast SAT solvers combined with clever encodings [80, 120, 140]. At the state of the art, we can say that STRIPS-like planning, despite its high theoretical complexity, *can* be efficiently solved.

This success brought the planning community to raise the bar of expressiveness, investigating more flexible and expressive extensions of the base language, in order to make the problem more easily applicable to real-world scenarios. A natural extension was that of adding *time* as a first-class concept of the modelling language. Complex temporal specifications, known as *temporally extended goals* [8], were first introduced instead of the simple reachability objectives of STRIPS problems. Then the problem of *temporal planning* emerged, where time is treated explicitly by assigning a *duration* to each action, thus increasing the complexity of the reasoning process because of the possibility of overlapping of concurrent actions. Nevertheless, the problem has been tackled by many planning systems, and the PDDL 2.1 language introduced syntax to express temporal problems for the 2002 IPC [59], together with the ability to talk about the consumption and management of *resources*. Note that the field of *scheduling* algorithms studies a form of temporal reasoning as well, but temporal planning concerns about the question of *what* to do in addition to *when*, and is therefore an harder problem, in general.

Going in a different direction, the PDDL+ language introduced support for modelling *hybrid systems* [60], mixing discrete and continuous dynamics, using *hybrid automata* [73] as the underlying semantic framework. Inspired by SAT encodings for classical planning, PDDL+ problems have been encoded into *SAT Modulo Theory* (SMT) [26]. In complex real-world scenarios, the reasoning process often cannot be limited to the actions and decisions of a single agent, but has to confront with how the surrounding environment reacts to those actions, handling *exogenous events*, *i.e.*, changes in the state of the system that do not have a *causal* relation with the actions of the agents, and, in general, *nondeterministic behaviour*. This consideration has led to the study of planning problems over nondeterministic domains, both under full observability of the current state (FOND planning, see *e.g.*, [49]) and under partial observability [118], and *probabilistic* domains [93]. Recently, temporal planning with *uncontrollable* action durations has been considered [46].

## 2 TIMELINE-BASED PLANNING

While this thesis was being written, the *European Space Agency* announced the discovery of a pocket of «liquid water buried under layers of ice and dust in the south polar region of Mars» [3]. This is the first time the presence of liquid water on the red planet was given conclusive evidence, and opens new scenarios both for the search for present or past life forms – subglacial water pockets on Earth are rich of microbial life – and for the organisation of future human exploration missions [110]. The measurements that led to the discovery come from the MARSIS instrument (*Mars Advanced Radar for Subsurface and Ionosphere Sounding*), a low-frequency radar mounted on *Mars Express*, a spacecraft in orbit around Mars since 2003.

Controlling the operations of a spacecraft orbiting a planet two hundreds and sixty million kilometres away from Earth cannot be considered a trivial task. Many different teams of scientists compete on Earth for a time slot of usage of the many instruments mounted on the spacecraft [55]. Each one has to be operated while the orbiter is in the right position, pointed in the right direction, and in general when the operating conditions are compatible with the scientific task at hand. For example, MARSIS measurements made for the above experiment were performed only when the satellite was on the night side of the planet, in order to «minimize ionospheric dispersion of the signal.» Given the limited storage and processing resources of the spacecraft hardware, the collected data must be transmitted back to Earth as soon as possible, so to free space for new measurements. However, transmitting consumes energy, hence everything has to be done while taking care of not drying out the batteries, especially when the spacecraft is not under direct Sun exposure. Furthermore, transmission of data is only possible when one of Earth's deep space ground stations is visible, and when the usage of that ground station is allocated to the same spacecraft. The choice of the ground station, and other environmental factors such as *e.g.*, *space weather*, affect the transmission speed, and consequently how much data can be transmitted in the given time slot, how much energy is required for the transmission, and thus how much data can be collected at that particular stage. All of this has to be controlled while guaranteeing the basic functionality of the spacecraft, performing the scheduled maintenance tasks, and so on.

This kind of challenges is common to the operation of any space exploration mission, being it a probe, an orbiter, or a planetary rover, but also to the operation of scientific, meteorological or telecommunication satellites orbiting Earth. An evident feature of the constraints and requirements exemplified above is the important role assumed by *temporal reasoning*, that is, reasoning about time of execution of tasks and about how the execution time and duration of such tasks affect the others. It came natural, then, that when space agencies started to integrate *automated planning* techniques into their workflows, an important part of the effort was devoted to the integration between planning and *scheduling* technologies.

*Timeline-based planning* is an approach to planning born from the integration of planning and scheduling concepts in the context of space operations, designed to specifically address the issues outlined above. The approach was first proposed by Muscettola et al. [105], and deployed for the first time shortly after in the HSTS system [103], that was used to schedule and control the operations of the *Hubble Space Telescope*. The major features that make the timeline-based approach ideal for this kind of applications can be identified as its strong focus on *temporal reasoning* and its mostly *declarative* nature.

This is achieved by changing the modelling perspective, when compared to *action-based* planning paradigms *à la* STRIPS. In timeline-based planning, there is no explicit separation between states, actions and goals. Rather, the domain is modelled as a set of independent, but interacting, components, whose behaviour over time, described by the *timelines*, is governed by a set of temporal constraints called *synchronisation rules*. The solution to a problem is then a set of timelines describing a possible behaviour of the system components that satisfies all the rules. This point of view turns out to be more declarative than common action-based languages such as PDDL, since the modelling task is focused on *what* has or has not to happen, instead of what the agent has to *do* to obtain the same results. Furthermore, the modelling of the system can be subdivided between multiple knowledge engineers and domain experts, since the timelines of separate system components can be separately modelled, and the resulting models can better reflect the architecture of the combined system.

Another important feature of timeline-based systems, present since the first incarnations [103], is the ability to integrate the planning phase with the *execution* of the plan. Timeline-based planning domains often model *real-time* systems, whose constraints heavily depends on the precise timing of execution of the tasks. However, ensuring precise timing is often not possible, because of the inherent *temporal uncertainty* that arises in the interaction with the environment. This kind of uncertainty is taken into account, and the controller executing the plan can handle it, by the use of *flexible plans*, *i.e.*, sets of different plans that differ in the execution time of the tasks.

Since its inception, the timeline-based approach to planning has been adopted and deployed in a number of systems developed by space agencies on both sides of the Atlantic. The work by Muscettola, who later worked for NASA, led to the development of two major systems, EUROPA [129] and its successor EUROPA 2 [12, 15]. In addition, NASA's Jet Propulsion Laboratory developed the ASPEN system [37], which was successfully deployed for the *Earth Orbiter One* experiment [39]. ASPEN was also used to plan the scientific operations of the *Rosetta* mission [38], which operated over ten years a spacecraft that, launched in 2004, successfully travelled five million kilometres across the solar system to reach the comet *67P/Churyumov Gerasimenko* in 2014, landed a probe on its surface, and orbited around it until its dismissal two years later.

On the European side,[1] the timeline-based paradigm was notably implemented in the MEXAR2 system [28, 30, 29], which has been deployed by the mission planning team of the same *Mars Express* mission cited above. Later, the same concepts were built into APSI-TRF, the *Timeline Representation Framework* of ESA's *Automated Planning and Scheduling Initiative* [2, 32, 64], currently used in many of the space agency's operations. All of these systems were in use as a support for *mission planning* tasks, but the approach was also applied to *on-board* autonomy [39, 63, 104]. Recently, the timeline-based approach was implemented by the PLATINUm system [135, 134], a more general purpose framework which was employed in cooperative and assistive robotics [31, 35].

# 3 TEMPORAL LOGICS

Every complex software and hardware system needs to be designed carefully to avoid bugs, security issues, and in general, deviations from the original specification. However, not all bugs are created equal. In *business-critical* and *safety-critical* systems, a bug can cause severe losses or even damages to equipment or injuries to people. This kind of systems thus need to be designed more carefully than others, and the field of *formal methods* aims at developing mathematical techniques apt to guarantee as much as possible the correctness and the safety of complex systems. In *model checking*, one of the most successful approaches to formal verification, a model of the system is checked against a formal specification by automated techniques that can certify the adherence of the model to the specification, or otherwise provide a counterexample, *i.e.*, an example execution of the system where the desired properties do not hold.

Over the years, *temporal logics* has emerged as the most common specification language in the formal verification field. Temporal logics refers to a broad family of logics, usually *modal logics*, which can predicate about facts evolving in time. Introduced by Prior [114] to address philosophical questions «about the relationship between tense a modality» [70], the formalism of temporal logic was recognised as an ideal specification language for software systems by Pnueli [113], which introduced *Linear Temporal Logic* (LTL), the simplest and most common kind of temporal logic, which is still nowadays the de-facto standard specification language for most formal verification systems. Once again, twentieth-century philosophy comes to rescue to contemporary computer science. LTL is interpreted over *linear temporal structures*, meaning that the evolution of the system over time is represented as a discrete sequence of states. If, in contrast, one considers all the possible evolutions of the systems as a unique entity, we obtain the concept of *branching time*, and logics such as *Computation Tree Logic* (CTL), and its extension CTL*.

[1] Although Rosetta was an ESA mission, science planning was handled by NASA's JPL.

Besides formal verification, temporal logic is also often adopted in AI systems where *temporal reasoning* is involved, as we will do in the next chapters. As an example, the specification of planning with *temporally extended goals*, proposed by Bacchus and Kabanza [8], is made by evaluating a linear temporal logic formula over the trace of execution of the plan. Later, LTL has been proved to be able to capture STRIPS-like planning [42].

Details on LTL and other specific temporal logics will be introduced as needed in the following chapters. For a detailed introduction of the history and contemporary applications of temporal logics, we refer the reader to the survey by Goranko and Galton [70], and to Chittaro and Montanari [40] for a survey on the role of temporal reasoning in AI.

## 4 EXPRESSIVENESS AND COMPLEXITY OF TIMELINE-BASED PLANNING

Action-based planning *à la* STRIPS, and most of its successors, have been not only developed from an algorithmic and technological perspective, but have also been thoroughly studied from a theoretical perspective, starting from the earliest work on situation calculus [98] as a firm formal background.

Over the years, every major action-based planning paradigm has been classified from the point of view of the *computational complexity*, starting from classic STRIPS-like planning which has been proved to be PSPACE-complete by Bylander [24]. In light of this theoretical intractability result, Bäckström [9] looked for tractable fragments, showing a polynomial-time restriction of the problem yet with a useful degree of expressiveness. Then, in contrast to planning for temporally extended goals, which is still PSPACE-complete [53], temporal planning has been proved to be strictly more complex, being an EXPSPACE-complete problem [119]. The addition of other features or the relaxation of some modelling assumption correspond to an inevitable increase in computational complexity. Indeed, planning on a nondeterministic domain has been proved to be EXPTIME-complete [91] with full observability, 2EXPTIME-complete with partial observability [118], and EXPSPACE-complete for *conformant planning* (*i.e.*, with *no* observability at all). In the probabilistic case, the general problem of finding a plan that reaches the goal with a certain probability is even *undecidable* [93], although easier restrictions exist with complexities ranging from EXPSPACE- down to NP-complete [92, 102, 117].

From a formal perspective, the PDDL language, at least in its basic variants, has been given a standard well-defined semantics [59, 99], and the semantics of many of its extensions are based on, or have been related to, well-defined formal models such as hybrid automata for PDDL+ [60] and *Partially Observable Markov Decision Processes* (POMDPs) for probabilistic variants [93]. Some work has been done to relate planning languages to *logic*. Besides the various SAT and SMT encodings, that in addition to their practical relevance also provide a connection between certain PDDL fragments and corresponding logical formalisms, a more direct connection has also been given with *temporal logic*, proving that STRIPS-like planning and temporal planning can be captured by Linear Temporal Logic and suitable extensions [42, 48].

In contrast to the above impressive body of theoretical work, and despite the practical success of the approach deployed in many complex real-world problems, little work has been done on timeline-based planning from a formal and theoretical perspective. The concept of timelines and the main features of the paradigm have been characterised by different authors. The description of the early DDL.1 modelling description language by Cesta and Oddi [34] comes with a well-defined semantics. Then, Frank and Jónsson [62] formally describe the constraint-based interval planning paradigm that underlies the EUROPA system, and Chien et al. [36] describe the general characteristics of the timeline-based approach adopted by a large number of space mission planning systems currently in use, comparing their differences and common features. Taking a relatively different perspective, Frank [61] studied the concept of timeline from a knowledge engineering point of view. These works, however, did not aim at providing a unifying description of the timeline-based approach that could be taken as a starting point for further theoretical investigation. A first step in this direction was provided by a semantically well-founded framework for timeline-based planning by Cimatti et al. [47], which however did not consider temporal flexibility, considered instead by Bernardini [14]. The latter work, however, used the formalism of *simple temporal networks* (STN) to represent temporal constraints, and still did not address the issue of controllability of flexible plans. Controllability issues were considered in many ways [43, 108, 109], but still depending on the corresponding notions coming from the world of *STNs with uncertainty* (STNU) [138]. The first comprehensive formal framework defining timeline-based planning was introduced by Cialdea Mayer et al. [44], including a uniform treatment of the integration of planning and execution through temporal flexibility, with a formal account, independent from STNUs, of the notions of controllability of flexible plans. Such framework was then taken as the foundational platform for the PLATINUm system [134].

Still, a comprehensive theoretical understanding of the paradigm is missing. In particular, timeline-based planning lacks a solid formal understanding of its formal and computational characteristics, such as the computational complexity of the involved problems, and the expressiveness of its modelling languages, both in logical terms and in contrast to the more common action-based ones.

This dissertation addresses these issues, by providing the first thorough theoretical investigation of the paradigm.

# 5 CONTRIBUTIONS OF THE THESIS

The goal of our investigation is twofold:

1. providing a detailed theoretical understanding of the timeline-based approach to planning, including the *computational complexity* of the involved decision problems;

2. comparing the *expressiveness* of timeline-based and action-based modelling languages, providing a bridge between the two worlds.

In pursuing these objectives, we take as starting point the formal framework introduced by Cialdea Mayer et al. [44], which comes as a representative of the many actual modelling languages employed by timeline-based systems. This formal framework is recalled in Chapter 2, providing all the necessary background for what follows, concluding this first introductory part.

In the second part of the thesis, Chapter 3 addresses the *expressiveness* of timeline-based planning languages, by comparing it with action-based planning. In particular, in sight of the explicit focus of this paradigm on temporal reasoning, the natural candidate for this comparison is *temporal planning*. Timeline-based planning problems are thus compared with temporal planning problems, represented by the formal temporal planning language introduced by Rintanen [119] to prove his computational complexity results. We identify a greatly restricted fragment of the general timeline-based language, which is however already expressive enough to capture temporal planning problems. The result is shown by providing a polynomial-size encoding of temporal planning problems into timeline-based ones, that preserves the solution plans.

Later, Chapter 4 addresses the more involved issue of the *computational complexity* of finding solution plans for timeline-based planning problems. We study the general formalism, without any artificial restriction, but without considering *uncertainty*, which is left for later. In this context, the plan existence problem for timeline-based planning problems is proved to be EXPSPACE-complete. While the hardness for the EXPSPACE class comes directly from the encoding of action-based temporal planning problems provided earlier, the inclusion in the class is proved by exhibiting a decision procedure that runs using at most exponential space.

Such a procedure exploits the concept of *rule graphs*, a graph-theoretic representation of synchronisation rules that allows us to easily manipulate, decompose and reason about timeline-based planning problems, providing useful insights into the structure of the problem. Much space is devoted to the development of this concept, which allows us to prove an upper bound on the size of the solution plans of timeline-based problems, and to build a data structure that can represent such plans compactly enough, in order to be used in our exponential-space decision procedure.

The complexity of two interesting variants of the problem is also studied. First, the problem of finding a solution plan of a given maximum length is proved to be NEXPTIME-complete, leveraging much of the framework built for the general result, while proving the NEXPTIME-hardness by a reduction from a specific type of tiling problems. Then, we define the problem of the existence of timeline-based planning problems over *infinite plans*, and prove it to be EXPSPACE-complete as well, with different automata-theoretic technique.

Then, Chapter 5 extends the picture by reintroducing the concept of *uncertainty*, *i.e.*, studying the problem when the modelled system has to account for the behaviour of the surrounding environment. Timeline-based systems are especially good at integrating planning with execution, by handling the *temporal uncertainty* inherent in the interaction with the environment. However, the current approach, based on *flexible plans*, also faces some limitation when the temporal uncertainty is not the only level of nondeterminism needed to correctly model the problem. On one hand, the focus on temporal uncertainty enforces some systems, such as PLATINUm, to employ a feedback loop including a re-planning phase for handling non-temporal mismatches between the expected and actual environment behaviour. Such re-planning phase can be costly and reduces the reactivity and autonomy of the system. On the other hand, we observe that the *syntax* of timeline-based planning languages is able to express situations that require general nondeterminism to be handled, while apparently only focusing on temporal uncertainty, hence showing that flexible plans cannot be considered a complete semantics for such problems.

To tackle these issues, we propose a *game-theoretic* interpretation of timeline-based planning with uncertainty, introducing the concept of *timeline-based game*, a two-players game where the controller has to execute a plan that satisfies the problem constraints independently from the action of the other player, which represents the surrounding environment. We show that this approach is strictly more general than the current one based on flexible plans, and we prove that the problem of finding a winning strategy for such games belongs to the 2EXPTIME complexity class. Notably, both the definition of the game, and the algorithm provided to decide the existence of winning strategies, heavily exploits the conceptual framework of *rule graphs* introduced in Chapter 4.

The third part of the thesis studies other expressiveness issues, but focusing on the relationship between timeline-based planning and *temporal logics*.

First, in Chapter 6, we introduce the topic of *tableau methods*, a long-studied paradigm for solving the satisfiability problem of various logics. We study in particular the novel kind of tableau methods for *Linear Temporal Logic* introduced by Reynolds [116]. The method, in contrast to earlier ones, produces a pure tree-shaped structure, where each branch only needs a single pass to be either accepted or rejected. In this context, we provide a number of contributions:

1. we recall Reynolds's *one-pass tree-shaped* tableau method for LTL, but proving its soundness and, in particular, its completeness, with a novel proof technique employing a model-theoretic argument which is conceptually simpler than the combinatorial proof provided in the original presentation of the method;

2. we report the results of the experimental evaluation of an implementation of the method, which shows how the simpler tree-shaped rule-based structure allows the tableau to be efficiently implementable and easily parallelisable, becoming competitive with other LTL satisfiability tools.

3. we *extend* the method to support *past operators*, obtaining a one-pass tree-shaped tableau for the resulting LTL+P logic, with full proofs of the soundness and completeness of the extended system.

Lastly, Chapter 7 addresses the issue of characterising the expressiveness of timeline-based planning from a logical point of view. Logical characterisations of action-based planning exist [42], which shows how classical planning problems can be captured by LTL formulae. We pursue a similar result for timeline-based planning, but LTL is not enough for the task at hand. Hence, we consider *Timed Propositional Temporal Logic* (TPTL), a real-time extension of LTL, which sports most of the features needed to express timeline-based planning problems. However, as synchronisation rules can arbitrarily talk about the future or the past of the current time point, any encoding of timeline-based planning into TPTL would need the use of *past operators*. Unfortunately, adding past operators to TPTL is unfeasible, as the satisfiability problem for the resulting TPTL+P logic is known to be *non-elementary*. To circumvent the issue, we isolate a specific fragment of TPTL+P, called *Bounded* TPTL *with Past* ($\mathsf{TPTL}_b$+P), which we show to be expressive enough to capture (most of) timeline-based planning, while still having an EXPSPACE-complete satisfiability problem. The complexity of the satisfiability problem for $\mathsf{TPTL}_b$+P is proved by exhibiting a graph-shaped tableau method, but then a one-pass tree-shaped method *à la* Reynolds is shown, which extends the one for LTL+P described in Chapter 6.

## 5.1 PUBLICATIONS

The contents of this work have been published in the proceedings of a number of international conferences, see the List of Publications at page 173. In general, however, the whole material has been extensively revisited for this dissertation and many results have been extended and completed.

In particular, the contents of Chapter 3 have been published in [TIME 2016], and Chapter 4 draws from [ICAPS 2017] for the main complexity results. The concept of *rule graph* was exploited for the first time in [ICAPS 2017], and glimpsed already in [TIME 2016], but Chapter 4 provides a complete and formally rigorous treatment of the concept with full details of all the proofs. Furthermore, the automata-theoretic approach to the complexity of the problem over infinite timelines comes from the recent [KR 2018]. Then, the work published at [TIME 2018] is reported in Chapter 5.

In the second part of the thesis, Chapter 6 presents the results published in [IJCAI 2016] (for the experimental evaluation of the tableau implementation), and in [LPAR-21] (for the extension to past operators). The different model-theoretic proof of the completeness of the one-pass tree-shaped tableau for LTL is novel. Finally, in Chapter 7, the $\mathsf{TPTL}_b$+P logic and its use to encode timeline-based planning problem come from [IJCAI 2017], while the one-pass tree-shaped tableau for the logic has been reported in [GandALF 2018].

# TIMELINE-BASED PLANNING 2

This chapter introduces timeline-based planning problems in full details. In contrast to the world of action-based planning, where the STRIPS-inspired PDDL language has been unanimously adopted as a de-facto standard modelling language, timeline-based planning has not converged over a single formalism. The many systems developed during the decades-long history of the approach adopted different languages with different features and often different semantics. Our work is based on a recently introduced formal framework that captures many common features of timeline-based systems, enabling the development of the results reported in the next chapters.

## 1 INTRODUCTION

The action-based planning has long since converged over PDDL as the de-facto standard modelling language, mainly thanks to its adoption by the *international planning competition* [99]. This standardisation, together with the strong starting formal ground of the early days [98], eased cross-fertilisation between actors in the field.

In contrast, timeline-based planning systems evolved quite independently from each other, each one developing its own features and characteristics, while respecting the main philosophy behind the approach. This resulted in many concrete languages being adopted by the different systems. In the early days, a timeline-based *Domain Description Language* (DDL.1) [33] was described, embedding the main ideas behind the approach. Nevertheless, the IxTeT system [68], introduced no much later, immediately adopted a different input syntax. Systems developed at NASA, such as EUROPA 2, adopted the *New Domain Description Language* (NDDL) [12], but even inside NASA itself, the ASPEN system was developed around yet another different input language called *ASPEN Modeling Language* (AML) [37]. ESA systems based on APSI adopt the DDL.3 modelling language. Lately, the *Action Notation Modeling Language* (ANML) was proposed by Smith et al. [128] as an extension to both AML and NDDL, which also incorporate action-based elements to get the best of both worlds.

It is evident how this proliferation of languages with different syntaxes and semantics made difficult for the timeline-based planning community to progress towards the kind of formally-grounded understanding of the approach whose development is the aim of this thesis. Recently, some attempts were made to provide a formal background to the approach [14, 47, 61]. However, these early attempts missed to formalise some of the most important features of timeline-based systems such as *flexible plans*, together with the associated *controllability* issues. The present work is based on a clean and comprehensive formal framework describing timeline-based planning problems, including uncertainty, temporal flexibility, and controllability issues, recently provided by Cialdea Mayer et al. [44]. The formal language introduced in their work abstracts over most features supported by the concrete syntax of different languages, with a well-defined semantics that provides the ideal starting point for our investigation.

This chapter provides a detailed account of timeline-based planning as defined by Cialdea Mayer et al. [44], which all the subsequent chapters will build upon. However, while taking it as a starting point, our presentation of their framework is tailored to our needs, as the original presentation had, in part, different goals. For example, one of the stated goals was that of defining all the relevant concepts in such a way to isolate what the *executor* of the plans needed to store to be able to completely do its job without storing the whole problem domain. This leads to a certain degree of redundancy in some definitions, that we can omit. A second, and maybe the most important, difference of our presentation is the separation between *timeline-based planning problems*, which do not admit any uncertainty, and timeline-based planning problems *with uncertainty*, which reintroduce all the relevant bits to handle the interaction with the environment. While the original presentation of the framework had the explicit goal of providing a uniform definition that accounted for uncertainty and temporal flexibility, in our computational complexity and expressiveness study it is convenient to separate the basic *satisfiability* problem of finding a plan to a deterministic problem from the *synthesis* problem of finding a strategy to cope with the environment behaviour. For this reason, the former concept is defined first (Section 2), and extended later to define the latter (Section 3). The results regarding the two are also reported in different chapters (mostly Chapters 4 and 5). Other changes in our presentation are there for ease of exposition and uniformity with the whole material. In any case, care has been taken to ensure that the given definitions are equivalent to the originals in any detail that could affect expressiveness or computational complexity, such as succinctness of representation and syntactic restrictions or limitations.

Given the long history of timeline-based planning and the number different systems developed over the years, no formal framework can claim to be completely exhaustive, and here we are explicitly making a few assumptions. Most importantly, the general framework from Cialdea Mayer et al. is time-domain-agnostic, being equivalently definable over *discrete*, *dense*, or even *continuous* time domains. In this work, we are instead exclusively focused on timeline-based problems over *discrete* time domains, *i.e.*, time is a discrete linear order, and time stamps are integer numbers. As a matter of fact, most of the systems, despite often supporting the specification of fractional values as a syntactic convenience, reason over discrete domains, by discretising time over a convenient granularity. Nevertheless, the formal properties of the approach when applied to dense domains is of independent interest, and recently some work appeared studying this case, from a perspective similar to ours [21, 22]. Another important limitation of our investigation is that we do not consider the concept of *resource handling*. Reasoning about the consumption of resources, such as fuel or energy, is an important part of any real-world task, and most timeline-based systems supports modelling resources of different kinds. The formal framework we base our work on has been recently extended to represent resource handling [135], and the extension of our results to include this feature is an important future step in this line of research.

### 1.1 BASIC NOTATION

In what follows, we denote as $\mathbb{N}$ and $\mathbb{Z}$ the sets of, respectively, natural numbers and integers. We denote as $\mathbb{N}_+ = \mathbb{N} \setminus \{0\}$ the set of *positive* natural numbers, and as $\mathbb{N}_{+\infty} = \mathbb{N} \cup \{+\infty\}$ the set of natural numbers augmented with an infinitary value. Sequences (either finite or infinite) of elements $x_0, x_1, \ldots$ are denoted as $\overline{x} = \langle x_0, x_1, \ldots \rangle$. Given a sequence $\overline{x} = \langle x_0, x_1, \ldots \rangle$ and an index $k \in \mathbb{N}$, the sequence $\overline{x}_{\geq k} = \langle x_k, x_{k+1}, \ldots \rangle$ is the suffix of $\overline{x}$ starting from the $k$th element. Analogously we define $\overline{x}_{>k}$, $\overline{x}_{\leq k}$ and $\overline{x}_{<k}$. Given $i, j \in \mathbb{N}$, the subsequence between the $i$th and $j$th elements is denoted as $\overline{x}_{[i \ldots j]} = \langle x_i, \ldots, x_j \rangle$. Given two sequences $\overline{x} = \langle x_0, \ldots, x_n \rangle$ and $\overline{y} = \langle y_0, y_1, \ldots \rangle$, with $\overline{x}$ finite and $\overline{y}$ either finite or infinite, we denote as $\overline{xy} = \langle x_0, \ldots, x_k, y_0, y_1, \ldots \rangle$ the juxtaposition of $\overline{x}$ and $\overline{y}$. The length of a finite sequence $\overline{x} = \langle x_0, \ldots, x_{n-1} \rangle$ is denoted as $|\overline{x}| = n$. The empty sequence is denoted as $\varepsilon$, with $|\varepsilon| = 0$.

# 2 TIMELINE-BASED PLANNING

This section formally defines timeline-based planning problems. As already mentioned, here we mostly deal with problems that are not concerned with any uncertainty in the interaction with the external environment. Such ingredient will be added in Section 3.

## 2.1 TIMELINES

In our setting, interesting properties of the modelled system are represented by *state variables*, which range over finite domains.

**Definition 2.1** — State variable.
*A* state variable *is a tuple* $x = (V_x, T_x, D_x, \gamma_x)$, *where:*

- $V_x$ *is the* finite domain *of* $x$*;*
- $T_x : V_x \to 2^{V_x}$ *is the* value transition function *of* $x$*;*
- $D_x : V_x \to \mathbb{N} \times \mathbb{N}_{+\infty}$ *is the* duration function *of* $x$*;*
- $\gamma_x : V_x \to \{\mathsf{c}, \mathsf{u}\}$ *is the* controllability tag.

Intuitively, the transition function specifies which values $T_x(v)$ the variable can hold immediately after point in time where $x = v$. The duration function $D_x$ maps any value $v \in V_x$ into a pair of non-negative integers $(d^{x=v}_{min}, d^{x=v}_{max})$, which respectively specify the minimum and maximum duration of any time interval where $x = v$. The maximum duration can be infinite ($d^{x=v}_{max} = +\infty$), in which case there is no upper bound to how long the variable can hold the given value. The controllability tag comes into play when handling *uncertainty*. Intuitively, it states whether the *duration* of any time interval (of any *token*, more precisely, as defined below), where a variable $x$ holds a given value $v$ is controllable by the system ($\gamma_x(v) = \mathsf{c}$) or not ($\gamma_x(v) = u$).

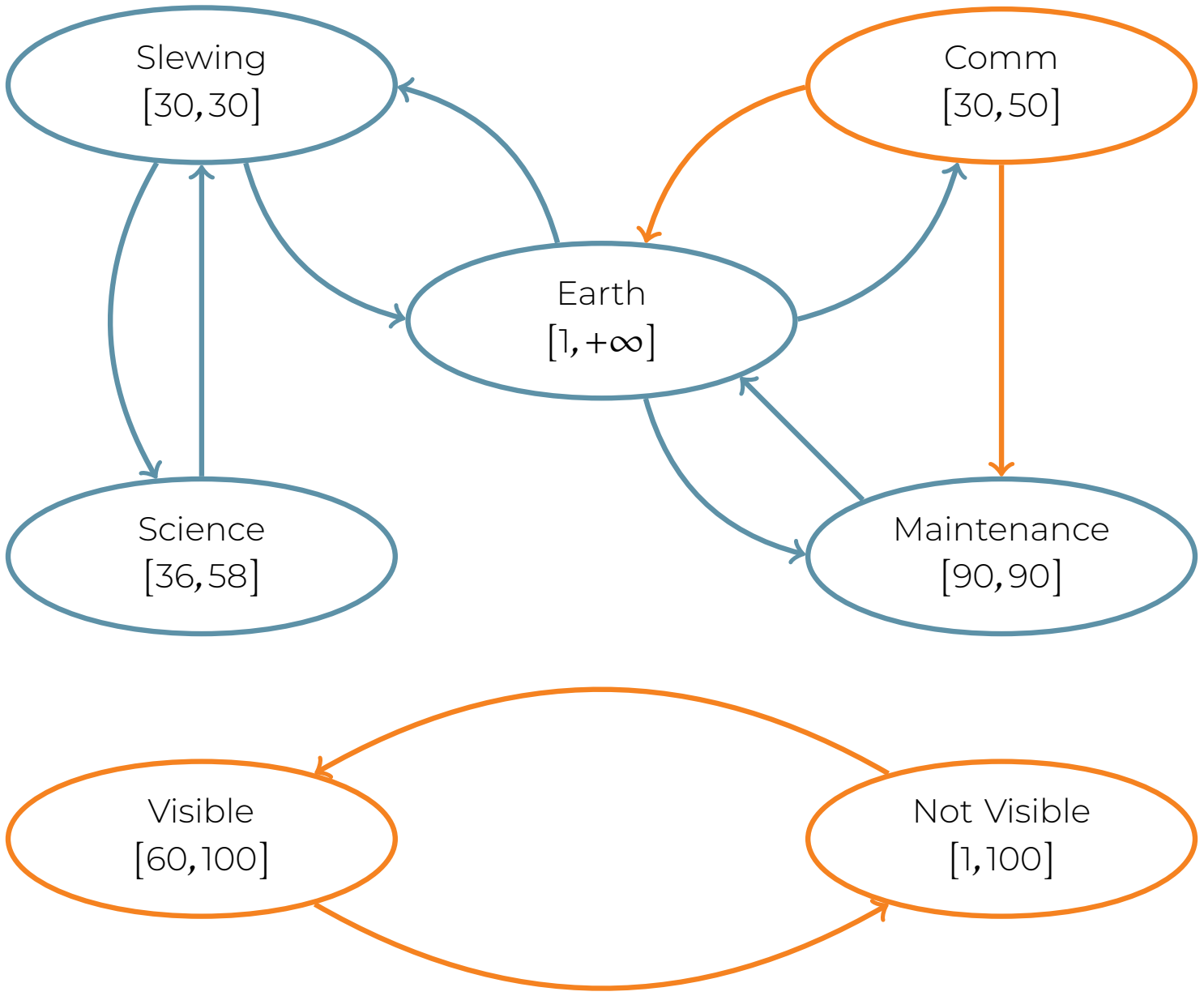

**Figure 2.1:** Values of the example state variables $x_p$ (above) and $x_v$ (below) visualised as state machines. Uncontrollable values are marked in orange.

Two example state variables are depicted in Figure 2.1, belonging to a domain concerning the operations of a satellite orbiting a planet (a scenario conceptually similar to the *Mars Express* mission). The first variable $x_p$, represents the pointing mode of the satellite, *i.e.*, whether it is pointing towards Earth, doing maintenance, doing scientific measurements, slewing between the direction facing Earth and the direction facing the underlying planet, or whether it is transmitting some communications. The domain of the variables thus consists of the five depicted values, and the transition function states which task can follow each other, being visualisable as a state machine. The minimum and maximum durations for each value are reported inside the bubbles. The second variable $x_v$ represents the visibility window of the Earth ground station, which determines when the station is visible for transmitting. In this example $\gamma_{x_p}(\mathsf{Comm}) = \mathsf{u}$, *i.e.*, the Comm value is *uncontrollable*, meaning that the system can decide when to start communicating but cannot decide nor predict how much time will be required by the transmission. All the values of the variable $x_v$ are uncontrollable since, of course, the satellite cannot decide when the ground stations are visible or not. In what follows, the controllability tag will be ignored, and considered again in Section 3.

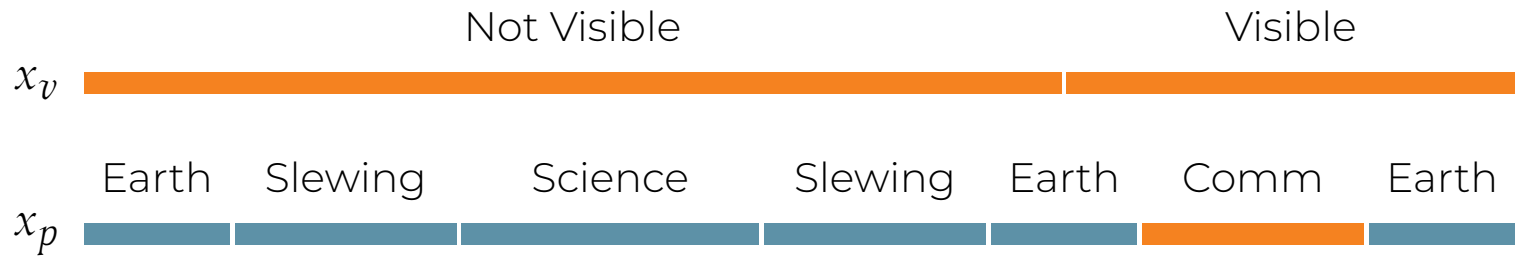

**Figure 2.2:** Example timelines for the variables $x_v$ (above) and $x_p$ (below)

The evolution over time of the values of each state variable is represented by the *timelines*, which are the core concept of the whole formalism.

**Definition 2.2** — Timeline.
*A* token *for a state variable $x$ is a triple $\tau = (x, v, d)$, where $v \in V_x$ is the value held by the variable, and $d \in \mathbb{N}_+$ is the* duration *of the token. A* timeline *for a state variable $x$ is a* finite *sequence $\overline{\tau} = \langle \tau_1, \ldots, \tau_k \rangle$ of tokens for $x$.*

A timeline thus represents how a state variable changes over time in terms of a sequence of time intervals where the value of the variable keeps the same value. Note that $d \in \mathbb{N}_+$, *i.e.*, the duration of tokens cannot be zero. Some notation will come useful to manipulate timelines and tokens. For any token $\tau_i = (x, v_i, d_i)$ in a timeline $\overline{\tau} = \langle \tau_1, \ldots, \tau_k \rangle$, we define $\mathsf{val}(\tau_i) = v_i$ and $\mathsf{duration}(\tau_i) = d_i$. Moreover, we can define the functions $\mathsf{start\text{-}time}(\overline{\tau}, i) = \sum_{j=1}^{i-1} d_j$ and $\mathsf{end\text{-}time}(\overline{\tau}, i) = \mathsf{start\text{-}time}(\overline{\tau}, i) + d_i$, for all $1 \le i \le k$, hence mapping each token $\tau_i$ to the corresponding $[\mathsf{start\text{-}time}(\overline{\tau}, i), \mathsf{end\text{-}time}(\overline{\tau}, i))$ time interval (right extremum excluded). When there is no ambiguity about which timeline we refer to, we write $\mathsf{start\text{-}time}(\tau_i)$ and $\mathsf{end\text{-}time}(\tau_i)$ to denote, respectively, $\mathsf{start\text{-}time}(\overline{\tau}, i)$ and $\mathsf{end\text{-}time}(\overline{\tau}, i)$. Note that two consecutive tokens can hold the same value, in which case they are still treated as two distinct entities. This is sometimes forbidden in timeline-based systems, but we do not need to impose this restriction.

Figure 2.2 shows two example timelines for the state variables $x_p$ and $x_v$. Note how the sequence of values in each timeline obeys the transition function of the two variables as depicted in Figure 2.1.

The time span of a timeline is called *horizon*.

**Definition 2.3** — Horizon of a timeline.
*The* horizon *of a timeline $\overline{\tau} = \langle \tau_1, \ldots, \tau_k \rangle$ is defined as $\mathsf{H}(\overline{\tau}) = \mathsf{end\text{-}time}(\tau_k)$.*
*The* horizon *of an empty timeline $\overline{\tau}$ is defined as $\mathsf{H}(\overline{\tau}) = 0$.*

Given a set of state variables $\mathsf{SV}$, the set of all the timelines that can be formed over the variables in $\mathsf{SV}$ is denoted as $\mathbb{T}_{\mathsf{SV}}$.

A *plan*, in timeline-based planning, is a set of timelines describing the evolution of the considered set of state variables. More precisely:

**Definition 2.4** — Plan.
*Let* $\mathsf{SV}$ *be a set of state variables. A* plan *over* $\mathsf{SV}$ *is a function* $\pi : \mathsf{SV} \to \mathbb{T}_{\mathsf{SV}}$ *that maps each variable to the timeline describing its behaviour, such that all the timelines have the same horizon,* i.e., $\mathsf{H}(\pi(x)) = \mathsf{H}(\pi(x'))$ *for all* $x, x' \in \mathsf{SV}$.

We denote as $\mathsf{H}(\pi)$ the horizon of the plan, *i.e.*, the horizon of all the timelines in the plan.

## 2.2 SYNCHRONISATION RULES

Given a system described by a set of state variables, its behaviour over time is governed by a set of temporal constraints called *synchronisation rules*. The particular syntactic structure of these rules is what shapes the computational properties of timeline-based planning, as shown in the next chapters.

In what follows, let us choose an arbitrary set $\mathcal{N} = \{a, b, \ldots\}$ of *token names*. The basic building blocks of the syntax of synchronisation rules are the atomic temporal relations, also called *atoms*.

**Definition 2.5** — Atomic temporal relation.
*A* term *over* $\mathcal{N}$ *is an expression matching the following grammar:*

$$\langle term \rangle := t \mid \mathsf{start}(a) \mid \mathsf{end}(a)$$

*where* $t \in \mathbb{N}$ *and* $a \in \mathcal{N}$. *Terms of the form* $t \in \mathbb{N}$ *are called* timestamps.

*An* atomic temporal relation, *or* atom, *over* $\mathcal{N}$ *is an expression of the form:*

$$\langle atom \rangle := \langle term \rangle \leq_{[l,u]} \langle term \rangle$$

*where* $l \in \mathbb{N}$ *and* $u \in \mathbb{N}_{+\infty}$.

For brevity, the subscript of atoms is omitted if $l = 0$ and $u = +\infty$. Atomic relations can be subdivided in different kinds depending on their form.

**Definition 2.6** — Qualitative, unbounded and pointwise atoms.
*Let* $\alpha \equiv T \leq_{[l,u]} T'$ *be an atomic temporal relation. Then,* $\alpha$ *is said to be* unbounded *if* $u = +\infty$ *and* $T'$ *is not a timestamp, and* bounded *otherwise. Moreover, an atom is* pointwise *if one of its terms is a timestamp.*

If $a$ and $b$ are two token names, then examples of atomic relations are $\mathsf{start}(b) \leq 5$, $\mathsf{start}(a) \leq_{[3,7]} \mathsf{end}(b)$, and $\mathsf{start}(a) \leq_{[0,+\infty]} \mathsf{start}(b)$. Intuitively, a token name $a$ refers to a specific token in a timeline, and $\mathsf{start}(a)$ and $\mathsf{end}(a)$ to its endpoints. Then, an atom such as $\mathsf{start}(a) \leq_{[l,u]} \mathsf{end}(b)$ constrains $a$ to start before the end of $b$, and the distance between the two endpoints to be comprised between the lower and upper bounds $l$ and $u$. Atomic relations are grouped into quantified clauses called *existential statements*.

**Definition 2.7** — Existential statement.
*Given a set* SV *of state variables, an* existential statement *over* SV *is a statement of the following form:*

$$\begin{aligned}\langle \textit{ex. statement}\rangle &:= \langle \textit{quantifier prefix}\rangle \,.\, \langle \textit{clause}\rangle \\ \langle \textit{quantifier prefix}\rangle &:= \exists a_1[x_1 = v_1]a_2[x_2 = v_2]\ldots a_n[x_n = v_n] \\ \langle \textit{clause}\rangle &:= \langle \textit{atom}\rangle \wedge \langle \textit{atom}\rangle \wedge \ldots \wedge \langle \textit{atom}\rangle\end{aligned}$$

*where* $n \in \mathbb{N}$, $a_1, \ldots, a_n \in \mathcal{N}$, $x_1, \ldots, x_n \in \mathsf{SV}$, *and* $v_i \in V_{x_i}$ *for all* $1 \leq i \leq n$.

In an existential statement $\mathcal{E} \equiv \exists a_1[x_1 = v_1]\ldots a_n[x_n = v_n] \,.\, \mathcal{C}$, all the token names appearing in the atoms inside $\mathcal{C}$ that do not appear in the quantifier prefix are said to be *free* in $\mathcal{E}$, and all those that do appear are said to be *bound*. An existential statement is *closed* if it does not contain free token names. Note that the quantifier prefix may as well be empty.

Finally, the syntax of *synchronisation rules* is defined as follows.

**Definition 2.8** — Synchronisation rules.
*Given a set of state variables* SV, *a* synchronisation rule *over* SV *is an expression matching the following grammar:*

$$\begin{aligned}\langle \textit{body}\rangle &:= \langle \textit{ex. statement}\rangle \vee \ldots \vee \langle \textit{ex. statement}\rangle \\ \langle \textit{rule}\rangle &:= a_0[x_0 = v_0] \rightarrow \langle \textit{body}\rangle \\ \langle \textit{rule}\rangle &:= \top \rightarrow \langle \textit{body}\rangle\end{aligned}$$

*where* $a_0 \in \mathcal{N}$, $x_0 \in \mathsf{SV}$, $v_0 \in V_{x_0}$, *and the only token name appearing free in the body is* $a_0$, *and only in rules of the first form.*

In rules of the first form, the quantifier in the head is called *trigger*, and rules of the second form are called *triggerless rules*. Intuitively, a synchronisation rule demands that whenever a token exists that satisfies the trigger, then at least one of the disjuncted existential statements must be satisfied, *i.e.*, there must exist other tokens as specified in the quantifier prefix such that the corresponding clause is satisfied. Triggerless rules have a trivial universal quantification, which means they only demands the existence of some tokens, as specified by the existential statements. As an example, consider the timelines in Figure 2.2, and the following synchronisation rule:

$$\begin{aligned}&a[x_p = \mathsf{Comm}] \rightarrow \exists b[x_v = \mathsf{Visible}] \,.\, \mathsf{start}(b) \leq \mathsf{start}(a) \wedge \mathsf{end}(a) \leq \mathsf{end}(b) \\ &a[x_p = \mathsf{Science}] \rightarrow \exists b[x_p = \mathsf{Slewing}]c[x_p = \mathsf{Comm}] \,.\, \\ &\qquad\qquad \mathsf{end}(a) = \mathsf{start}(b) \wedge \mathsf{end}(b) = \mathsf{start}(c)\end{aligned}$$

The first rule expresses an essential guarantee for the satellite system represented by the two example variables, namely that when the spacecraft is communicating with Earth, the ground station is visible. The timelines in the figure satisfy this constraint, since the time interval corresponding to the execution of the token where $x_p = \mathsf{Comm}$ is contained in the one of the token where $x_v = \mathsf{Visible}$. The second rule instructs the system to transmit data back to Earth after every measurement session, interleaved by the required slewing operation, and is as well satisfied in the example. A triggerless rule might instead be used to state the goal of the system, namely to perform some scientific measurement at all:

$$\top \longrightarrow a[x_p = \mathsf{Science}]$$

| Allen's relation | Syntax | Desugaring |
|---|---|---|
| | $a$ meets $b$ | $\mathsf{end}(a) = \mathsf{start}(b)$ |
| | $a$ before $b$ | $\mathsf{end}(a) \leq \mathsf{start}(b)$ |
| | $a$ after $b$ | $\mathsf{end}(b) \leq \mathsf{start}(a)$ |
| | $a \subseteq b$ | $\mathsf{start}(b) \leq \mathsf{start}(a) \wedge \mathsf{end}(a) \leq \mathsf{end}(b)$ |
| | $a$ overlaps $b$ | $\mathsf{start}(a) \leq \mathsf{start}(b) \wedge \mathsf{end}(a) \leq \mathsf{end}(b) \wedge \mathsf{start}(b) \leq \mathsf{end}(a)$ |

**Table 2.1:** Allen's interval relations expressed in terms of atomic temporal relations

Some simple syntactic sugar can be defined on top of the basic syntax. A strict version of unbounded atoms can be obtained by writing $T < T'$ to mean $T \leq_{[1,+\infty]} T'$. Then, one can require two endpoints to coincide in time by writing $\mathsf{start}(a) = \mathsf{start}(b)$ instead of $\mathsf{start}(a) \leq_{[0,0]} \mathsf{start}(b)$, and two tokens to coincide by writing $a = b$ instead of $\mathsf{start}(a) = \mathsf{start}(b) \wedge \mathsf{end}(a) = \mathsf{end}(b)$. More generally, all the Allen's interval relations [4] can be expressed in terms of these basic temporal relations, hence we can introduce abbreviations for each of them, as listed in Table 2.1.

Moreover, to constrain the duration of a token we can write $\mathsf{duration}(a) = t$, $\mathsf{duration}(a) \leq t$ and $\mathsf{duration}(a) \geq t$ instead of, respectively, $\mathsf{start}(a) \leq_{[t,t]} \mathsf{end}(a)$, $\mathsf{start}(a) \leq_{[0,t]} \mathsf{end}(a)$, and $\mathsf{start}(a) \leq_{[t,+\infty]} \mathsf{end}(a)$.

Note that, in contrast to the relations shown in Table 2.1, one cannot express, within a single existential statement, the *disjointness* of two tokens, *i.e.*, that one token appears either strictly after or strictly before another without overlap. That would need a disjunction while clauses are purely conjunctive, and disjunction is admitted only at top level, between whole existential statements. In relation to this issue, it is worth to note that the syntax of synchronisation rules as defined above does not include *negation*. One can easily negate an unbounded atom $T \leq T'$ by writing $T' < T$, but the negation bounded atoms cannot generally be expressed without some sort of disjunction. The absence of negation and the limited use of disjunctions are the most important syntactic restrictions of synchronisation rules.

## 2.3 TIMELINE-BASED PLANNING PROBLEMS

We will now formally define the semantics of synchronisation rules, to back up the intuition built in the previous sections, and then timeline-based planning problems will be defined. Let us start with atomic relations.

**Definition 2.9** — Semantics of atomic relations.
*An* atomic evaluation *is a function* $\lambda : \mathcal{N} \to \mathbb{N}^2$ *that maps each token name* $a$ *to a pair* $\lambda(a) = (s, e)$ *of natural numbers. Given a term* $T$ *and an atomic evaluation* $\lambda$*, the* evaluation *of* $T$ *induced by* $\lambda$*, denoted* $[\![T]\!]_\lambda$*, is defined as follows:*

- $[\![t]\!]_\lambda = t$ *for any* $t \in \mathbb{N}$*;*
- *for any* $a \in \mathcal{N}$*, if* $\lambda(a) = (s, e)$*, then* $[\![\mathsf{start}(a)]\!]_\lambda = s$ *and* $[\![\mathsf{end}(a)]\!]_\lambda = e$*.*

*Given an atomic temporal relation* $\alpha \equiv T \leq_{[l,u]} T'$ *and an atomic evaluation* $\lambda$*, we say that* $\lambda$ satisfies $\alpha$*, written* $\lambda \models \alpha$*, if and only if* $l \leq [\![T']\!]_\lambda - [\![T]\!]_\lambda \leq u$*.*

Given a *clause* $\mathcal{C} \equiv \alpha_1 \wedge \ldots \wedge \alpha_k$, by extension we write $\lambda \models \mathcal{C}$ if $\lambda \models \alpha_i$ for all $1 \leq i \leq k$. Atomic evaluations are extracted from tokens when trying to satisfy a whole existential statement.

**Definition 2.10** — Semantics of existential statements.
*Let* $\pi$ *be a plan over a set of state variables* $\mathsf{SV}$*, and consider an existential statement* $\mathcal{E} \equiv \exists a_1[x_1 = v_1] \ldots a_n[x_n = v_n] \,.\, \mathcal{C}$*. A function* $\eta : \mathcal{N} \to \mathsf{tokens}(\pi)$ *mapping any token name to a token belonging to the plan* $\pi$ *is called* token mapping.

*We say that* $\pi$ satisfies $\mathcal{E}$ *with the token mapping* $\eta$*, written* $\pi \models_\eta \mathcal{E}$*, if* $\eta(a_i) = \tau_i$ *such that* $\tau_i \in \pi(x_i)$ *and* $\mathsf{val}(\tau_i) = v_i$*, for all* $1 \leq i \leq n$*, and* $\lambda \models \mathcal{C}$ *for an atomic evaluation* $\lambda$ *such that* $\lambda(a_i) = (\mathsf{start\text{-}time}(\tau_i), \mathsf{end\text{-}time}(\tau_i))$ *for all* $1 \leq i \leq n$*.*

A whole synchronisation rule is satisfied by a plan if, whenever the trigger is satisfied, at least one of its existential statements is satisfied as well.

**Definition 2.11** — Semantics of synchronisation rules.
*Let* $\pi$ *be a plan and let* $\mathcal{R} \equiv a_0[x_0 = v_0] \to \mathcal{E}_1 \vee \ldots \vee \mathcal{E}_m$ *be a synchronisation rule.*

*We say that* $\pi$ satisfies $\mathcal{R}$*, written* $\pi \models \mathcal{R}$*, if for any token* $\tau_0 \in \pi(x_0)$*, if* $\mathsf{val}(\tau_0) = v_0$ *then there is at least one of its existential statements* $\mathcal{E}_i$ *and a token mapping* $\eta$ *such that* $\eta(a_0) = \tau_0$ *and* $\pi \models_\eta \mathcal{E}_i$*.*

*For a* triggerless *rule* $\mathcal{R} \equiv \top \to \mathcal{E}_1 \vee \ldots \vee \mathcal{E}_m$*,* $\pi \models \mathcal{R}$ *if there exist one* $\mathcal{E}_i$ *and a token mapping* $\eta$ *such that* $\pi \models_\eta \mathcal{E}_i$*.*

We can finally define the notion of *timeline-based planning problems*, and of which *solution plans* we are looking for.

**Definition 2.12** — Timeline-based planning problem.
*A* timeline-based planning problem *is a pair* $P = (\mathsf{SV}, S)$, *where* $\mathsf{SV}$ *is a set of state variables, and* $S$ *is a set of synchronisation rules over* $\mathsf{SV}$.

We consider the *size* $|P|$ of a problem $P$ to be the length of any reasonable representation of $P$, with a *binary* encoding of numeric parameters.

**Definition 2.13** — Solution plan.
*Let* $P = (\mathsf{SV}, S)$ *be a* timeline-based planning problem, *and let* $\pi$ *be a* plan *over* $\mathsf{SV}$. *Then,* $\pi$ *is a* solution plan *for* $P$ *iff* $\pi \models \mathcal{R}$ *for all synchronisation rules* $\mathcal{R} \in S$.

The set of all the solution plans of a timeline-based planning problem $P$ is denoted as $\mathsf{plans}(P)$. Hence, given a timeline-based planning problem $P$, *our* problem is that of finding whether there exists a plan $\pi \models P$, *i.e.*, if $\mathsf{plans}(P) \neq \emptyset$. This *plan existence problem* will be the main subject of most of the thesis.

It is worth to note that in the original definition by Cialdea Mayer et al. [44], timeline-based planning problems include a component $H$ which is a *bound* on the horizon of the aimed solutions. Here we omit this component, and consider a more general problem where no bound on the solutions horizon is set in advance. Nevertheless, the bounded-horizon variant is interesting since many application scenarios require or can take advantage of a known temporal horizon for the plan. Hence, we also define this variant, that will be studied in future chapters together with the more general one.

**Definition 2.14** —Timeline-based planning problem with bounded horizon.
*A timeline-based planning problem with* bounded horizon *is a tuple* $P = (\mathsf{SV}, S, H)$ *where* $\mathsf{SV}$ *is a set of state variables,* $S$ *is a set of synchronisation rules over* $\mathsf{SV}$, *and* $H \in \mathbb{N}_+$ *is a positive integer. A plan* $\pi$ *over* $\mathsf{SV}$ *is a* solution plan *for* $P$ *if it is a solution plan for the timeline-based planning problem* $P' = (\mathsf{SV}, S)$ *and* $\mathsf{H}(\pi) \leq H$.

Here we have defined the *deterministic* variant of the problem, where there is no support for modelling the uncertainty coming from the interaction with the external world. Section 3 defines *timeline-based planning problems with uncertainty*, which account for this important feature.

# 3 TIMELINE-BASED PLANNING WITH UNCERTAINTY

This section extends the definitions provided above with the notion of *temporal uncertainty*, defining *timeline-based planning problems with uncertainty*. The capability of handling such type of uncertainty, integrating the planning and execution phases, is one of the key features of timeline-based planning systems. The state-of-the-art approach to this issue among current timeline-based systems revolves around the notion of *flexible timelines* and, consequently, *flexible plans*.

$\overline{\tau}_{x_p}$: Earth, Slewing, Science, Slewing, Earth — 110 120 140 150 181 200 215 233

$\overline{\tau}'_{x_p}$: Earth, Slewing, Science, Slewing, Earth — 115 148 185 220

**Figure 2.3:** Example of flexible timeline $\overline{\tau}_{x_p}$ and one of its instances $\overline{\tau}'_{x_p}$, for the variable $x_p$ of Figure 2.1.

## 3.1 FLEXIBLE TIMELINES

A flexible timeline abstracts multiple different timelines that differ only for the precise timings of start and end of the tokens therein, embodying some *temporal uncertainty* about the events described by the timeline.

**Definition 2.15** — Flexible token.
*A* flexible token *for a state variable $x$ is a tuple $\tau = (x, v, [e, E], [d, D])$, where $v \in V_x$, $[e, E] \in \mathbb{N} \times \mathbb{N}$ is the interval of possible* end times *of the token, and $[d, D] \in \mathbb{N} \times \mathbb{N}_+$ is the interval of possible token* durations.

**Definition 2.16** — Flexible timeline.
*A* flexible timeline *for a state variable $x$ is a* finite *sequence $\overline{\tau} = \langle \tau_1, \ldots, \tau_k \rangle$ of flexible tokens $\tau_i = (x, v_i, [e_i, E_i], [d_i, D_i])$ for $x$, where $[e_1, E_1] = [d_1, D_1]$, $e_i \geq e_{i-1} + d_i$, and $E_i \leq E_{i-1} + D_i$.*

Hence, flexible timelines provide an uncertainty range for the end time and duration of each flexible token of the timeline. Note that each flexible token reports a range of its *end time*, rather than its start time, because in this way it can explicitly constrain its horizon. Tokens and timelines as defined in Definition 2.2 are also called *scheduled tokens* and *scheduled timelines*, when the context requires disambiguation. Similarly to the notation used for scheduled timelines, set of all the possible flexible timelines for the set of state variables SV is denoted as $\mathbb{F}_{SV}$.

Given a state variable $x = (V_x, T_x, D_x, \gamma_x)$, consider the *controllability tag* $\gamma_x$, which has been ignored in Section 2. The controllability tag tells, for each *value* of the domain of each variable, if the duration of tokens that hold the given value are under the control of the planner or not. Hence, a value $v \in V_x$ is said to be *controllable* if $\gamma_x(v) = \mathsf{c}$, and *uncontrollable* if, otherwise, $\gamma_x(v) = \mathsf{u}$.

Given a flexible timeline $\overline{\tau} = \langle \tau_1, \ldots, \tau_k \rangle$, with $\tau_i = (x, v_i, [e_i, E_i], [d_i, D_i])$, a scheduled timeline $\overline{\tau}' = \langle \tau'_1, \ldots, \tau'_k \rangle$, with $\tau_i = (x, v'_i, d'_i)$ is an *instance* of $\overline{\tau}$ if $d_i \leq d'_i \leq D_i$ and $e_i \leq \text{end-time}(\tau'_i) \leq E_i$. Figure 2.3 shows an example of flexible timeline for the example state variable $x_p$ of Figure 2.1, and one of its instances.

We can now define the concept of *flexible plan*, which is an object more involved than just a set of flexible timelines.

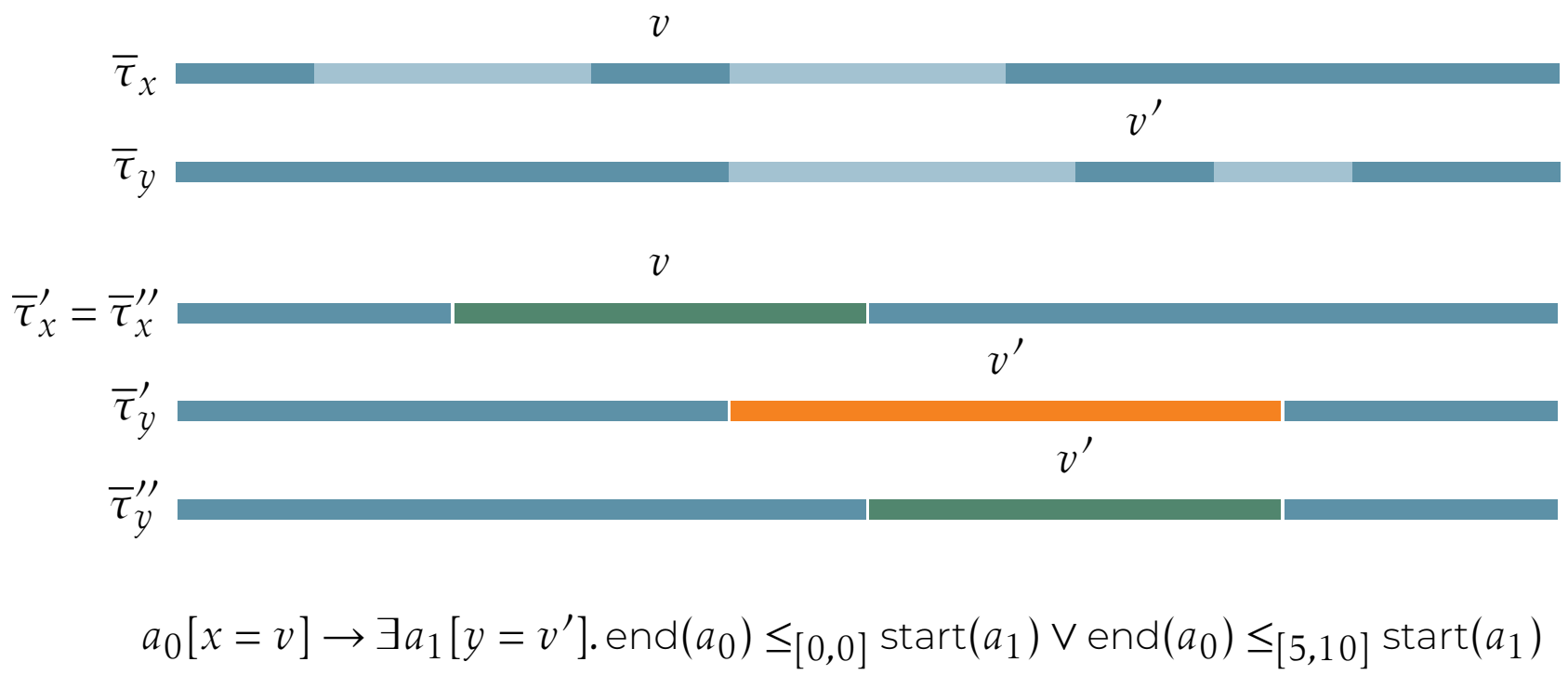

**Figure 2.4:** Example flexible timeline for two state variables $x$ and $y$, where not all the possible instances satisfy the above synchronisation rule.

**Definition 2.17** — Flexible plan.
*Given a set of state variables* SV, *a* flexible plan *over* SV *is a pair* $\Pi = (\pi, \mathsf{R})$, *where* $\pi : \mathsf{SV} \rightarrow \mathbb{F}_{\mathsf{SV}}$ *is a function providing a flexible timeline* $\pi(x)$ *for each state variable* $x$, *and* R *is a set of* atoms *(Definition 2.5) using as token names the set of tokens of the timelines in* $\pi$.

Intuitively, the flexible plan $\Pi = (\pi, \mathsf{R})$ represents a set of instances of the flexible timelines of $\pi$ which, additionally, satisfy the constraints imposed by the atoms included in R.

**Definition 2.18** — Instances of flexible plans.
*Let* $\Pi = (\pi, \mathsf{R})$ *be a flexible plan over* SV. *A plan* $\pi'$ *is an* instance *of* $\Pi$ *if* $\pi'(x)$ *is an instance of* $\pi(x)$ *for any* $x \in \mathsf{SV}$, *and all the atoms* $T \in \mathsf{R}$ *are satisfied by the atomic evaluation* $\lambda$ *such that* $\lambda(\tau) = (\mathsf{start\text{-}time}(\tau), \mathsf{end\text{-}time}(\tau))$ *for all token* $\tau$ *of* $\pi(x)$ *for any* $x \in \mathsf{SV}$.

To understand the need for the R component in Definition 2.17, consider Figure 2.4, which shows flexible timelines $\overline{\tau}_x = \langle \tau^x_0, \tau^x_1, \tau^x_3 \rangle$ and $\overline{\tau}_y = \langle \tau^y_0, \tau^y_1, \tau^y_3 \rangle$ for two variables $x$ and $y$, that have to be constrained by the shown synchronisation rule. The lower part of the picture shows some example instances of the flexible timelines. Given how the token $\tau^x_1$ is instantiated, not all the possible instances of the timeline for $y$ are valid with regards to the considered rule. The first example instantiation, namely $\overline{\tau}'_y$, violates the rule, while the second satisfies it. This happens because a simple set of flexible timelines misses the key information that $\tau^x_1$ cannot start before $\tau^y_1$. A flexible plan satisfying such a rule would then have to provide additional constraints ensuring this fact, such as $\mathsf{R} = \{\mathsf{end}(\tau^x_1) = \mathsf{start}(\tau^y_1)\}$ or $\mathsf{R}' = \{\mathsf{end}(\tau^x_1) \leq_{[5,10]} \mathsf{start}(\tau^y_1)\}$.

## 3.2 TIMELINE-BASED PLANNING PROBLEMS WITH UNCERTAINTY

We can now define timeline-based planning problems *with uncertainty*, as an extension of the timeline-based planning problems of Definition 2.12. We first provide the definition of problems and of *flexible solution plans*, and then discuss in detail their meaning and structure.

**Definition 2.19** — Timeline-based planning problem with uncertainty.
*A* timeline-based planning problem with uncertainty *is defined as a tuple* $P = (\mathsf{SV}_C, \mathsf{SV}_E, S, O)$ *where:*

1. $\mathsf{SV}_C$ *and* $\mathsf{SV}_E$ *are sets of, respectively, the* controlled *and* external *variables;*

2. $S$ *is a set of synchronisation rules over* $\mathsf{SV}_C \cup \mathsf{SV}_E$*;*

3. $O = (\pi_E, \mathsf{R}_E)$ *is a flexible plan, called the* observation, *specifying the behaviour of external variables.*

**Definition 2.20** — Flexible solution plan.
*Let* $P = (\mathsf{SV}_C, \mathsf{SV}_E, S, O)$*, with* $O = (\pi_E, \mathsf{R}_E)$*, be a timeline-based planning problem with uncertainty. A* flexible solution plan *for* $P$ *is a flexible plan* $\Pi = (\pi, \mathsf{R})$ *over* $\mathsf{SV}_C \cup \mathsf{SV}_E$ *such that:*

1. $\Pi$ *agrees with* $O$, i.e., $\pi(x) = \pi_E(x)$ *for each* $x \in \mathsf{SV}_E$*, and* $\mathsf{R}_E \subseteq \mathsf{R}$*;*

2. *the plan does not restrict the duration of uncontrollable tokens,* i.e., *for any state variable* $x$ *and any flexible token* $\tau = (x, v, [e, E], [d, D])$ *in* $\pi(x)$*, if* $\gamma_x(v) = \mathsf{u}$*, then* $d = d^{x=v}_{min}$ *and* $D = d^{x=v}_{max}$*;*

3. *any* instance *of* $\pi$ *is a solution plan for the timeline-based planning problem* $P' = (\mathsf{SV}_C \cup \mathsf{SV}_E, S)$*, and there is at least one such instance.*

The definitions above are worth a detailed explanation. Timeline-based planning problems with uncertainty consider two different sources of uncertainty: the behaviour of *external variables*, and the duration of *uncontrollable tokens*. In contrast to the simple problems without uncertainty defined in Definition 2.12, the set of state variables is split into the *controlled variables* $\mathsf{SV}_C$ and the *external variables* $\mathsf{SV}_E$. The behaviour of external variables cannot be constrained by the planner in any way, hence any solution plan is constrained to replicate the flexible timelines given by the *observation* $O$, which is a flexible plan describing their behaviour. It is worth to note that being $O$ a *flexible* plan, there is temporal uncertainty on the start and end time of the involved tokens, but the behaviour of the variables is otherwise known beforehand to the planner. Despite the name, borrowed from [44], the *observation* $O$ is more an *a priori* description of how the external variables will behave during the execution of the plan, up to the given temporal uncertainty on the precise timing of the events. The intended role of the external variables, then, is not much that of independent components interacting independently with the planned system, but rather, of external entities useful to represents given facts and invariants that the planner has to account for during the search for a solution.

As an example, consider a satellite seeking the right time to transmit data to Earth. When modelling this scenario as a timeline-based planning problem with uncertainty, the window of visibility of Earth's ground stations can be represented as an external variable with a suitable *observation*: the exact timing of when each station will effectively become visible is probably going to be uncertain up to some flexibility interval, but otherwise, the visibility and availability slots of each ground station to be used by that satellite have probably been already scheduled for the next months to come, and the planner does not have to account for any variability in that regard.

The second source of temporal uncertainty considered comes from tokens holding *uncontrollable values*. The duration of such tokens cannot be decided by the planner, hence their minimum and maximum duration in the flexibility range of the timeline has to coincide with that specified by the duration function of the variable. The planner can, however, decide which tokens to start and when, on *controlled* variables, even if $\gamma_x(v) = \mathsf{u}$. The uncontrollability is thus specifically limited to the duration of the token. It is worth to note how the formalism has intentionally been tailored to consider only *temporal uncertainty*, both with regards to external variables and to uncontrollable tokens.

## 3.3 CONTROLLABILITY OF FLEXIBLE PLANS

Timeline-based planning is an approach specifically targeted at the integration between the planning phase and the execution of the plan. In this regard, it is important to ensure that, once a flexible plan is found for a timeline-based planning problem with uncertainty, the plan can be effectively executed. This is not a trivial requirement given the presence of uncontrollable tokens, whose duration is decided during execution and is unknown beforehand. Indeed, Definition 2.17 ensures that any scheduled instance of the plan is a solution for the problem, but does not guarantee that (1) such an instance exists for any possible choice of the duration of uncontrollable tokens, and (2) at any time during the execution, the correct choice to keep following an instance of the plan depends only on events already happened and information already known.

For this reason, we have to speak about the *controllability* of flexible plans, *i.e.*, the property of being effectively executable by a controller. There are three major kinds of controllability properties that one may want to ensure on a flexible plan, depending on the application, informally defined as follows.

*Weak controllability* For any possible choice of the duration of uncontrollable tokens, there is an instance of the flexible plan respecting that choice.

*Strong controllability* There is a single way of instantiating controllable tokens that results into a valid instance of the flexible plan, no matter which is the duration of uncontrollable ones.

*Dynamic controllability* A strategy exists to choose how to instantiate each token, which, at any given point in time, can keep the execution in a valid instance of the plan, based only on what happened before that time.

The formal definitions that will follow are borrowed from Cialdea Mayer et al. [44] as all the rest of this chapter. However, concepts and terminology comes from further back to the works on *simple temporal networks with uncertainty* (STNU) [138], which face very similar problems.

Intuitively, *weak controllability* just ensures that the plan can be executed if the environment behaviour is known *beforehand*. This is clearly an unrealistic assumption, but can arise for example in the context of embedded devices with very little processing power but the possibility of storing in some form of ROM how to behave in correspondence of a pre-defined set of situations. *Strong controllability*, on the other hand, requires a single sequence of choices to always work regardless of the environment behaviour, which is still a quite rare luxury, but is nevertheless a useful guarantee to enforce in many safety-critical scenarios, as for example to ensure the existence of blind reset-to-home sequences in robotics. Finally, *dynamic controllability*, the concept which we are mostly interested in, designates a properly reactive scenario, where the controller can step-by-step decide what to do responding to how the environment behaved up to that point, ensuring in any the satisfaction of the problem constraints.

Let us now recap in formal terms the above concepts. The full and comprehensive formal framework can be found in [44]. Here, we just give a few compact definitions useful to frame the results of the next chapters.

Recall that the set of tokens of all the timelines of a plan is denoted as $\mathsf{tokens}(\pi)$. We extend this notation denoting as $\mathsf{tokens}(\Pi)$ the set of tokens of all the timelines of a flexible plan, and specifically, as $\mathsf{tokens}_U(\Pi)$ the set of *uncontrollable* tokens of $\Pi$, and as $\mathsf{tokens}_C(\Pi)$ the set of *controllable* tokens of $\Pi$.

**Definition 2.21** — Situation.
*Given a flexible plan* $\Pi = (\pi, \mathsf{R})$*, a* situation *for* $\Pi$ *is a function* $\omega : \mathsf{tokens}_U(\Pi) \to \mathbb{N}$ *assigning a duration to each uncontrollable token of* $\Pi$.

A situation represents the choices of the environment regarding the duration of uncontrollable tokens, both of controlled and external variables. Given a flexible plan $\Pi = (\pi, \mathsf{R})$, we denote as $\omega(\Pi)$ the set of instances of $\Pi$ where the duration of uncontrollable tokens corresponds to what dictated by $\omega$.

Regarding the external variables, any considered situation should respect the observation given by the planning problem. In other words, since the problem provides the behaviour of the external variable up to temporal uncertainty, and the executor is allowed to assume that things will evolve as stated, we only consider situations that fall into the uncertainty left by the problem.

**Definition 2.22** — Relevant situation.
*Given a timeline-based planning problem* $P = (\mathsf{SV}_C, \mathsf{SV}_E, S, O)$, *with* $O = (\pi_E, \mathsf{R}_E)$ *and a flexible plan* $\Pi = (\pi, \mathsf{R})$, *a situation* $\omega$ *is said to be* relevant *if any instance of* $\Pi$ *in* $\omega(\Pi)$ *satisfies the constraints of* $\mathsf{R}_E$.

Let us denote as $\Omega_\Pi$ the set of *relevant* situations for $\Pi$. If situations represent the decisions of the environment regarding the duration of uncontrollable tokens, then *scheduling functions* are the corresponding counterpart of the controller, deciding how to execute the whole plan.

**Definition 2.23** — Scheduling function.
*Given a timeline-based planning problem* $P = (\mathsf{SV}_C, \mathsf{SV}_E, S, O)$ *and a flexible plan* $\Pi = (\pi, \mathsf{R})$ *for* $P$, *a* scheduling function *for* $\Pi$ *is a map* $\theta : \mathsf{tokens}(\Pi) \to \mathbb{N}$ *providing an* end time *for each token in* $\Pi$, *such that the resulting scheduled plan* $\theta(\Pi)$ *is an instance of* $\Pi$.

Let us denote as $\mathcal{T}_\Pi$ the set of scheduling functions for a flexible plan $\Pi$. Hence, the task of the controller is that of deciding which scheduling function to apply in any given situation. This formalises the notion of *execution strategy*. Note that, while situations assign *durations* to tokens, a scheduling function assigns their *end times*, including those of uncontrollable tokens. Of course, it is not the job of the controller to decide the duration of uncontrollable tokens, hence any chosen scheduling function must agree with the current situation.

**Definition 2.24** — Execution strategy.
*Given a flexible plan* $\Pi = (\pi, \mathsf{R})$, *an* execution strategy *for* $P$ *is a map* $\sigma : \Omega_\Pi \to \mathcal{T}_\Pi$ *such that, given* $\theta = \sigma(\omega)$ *for any* $\omega \in \Omega_\Pi$, *if* $\tau$ *is an* uncontrollable *token of* $\theta(\Pi)$, *then* $\mathsf{duration}(\tau) = \omega(\tau)$.

Finally, we can now formally define the different concepts of controllability introduced above. We start from the first two simpler ones.

**Definition 2.25** — Weak and strong controllability.
*Let* $\Pi$ *be a flexible plan. Then,* $\Pi$ *is:*

1. weakly controllable *if there exists any execution strategy* $\sigma$ *for* $\Pi$*;*
2. strongly controllable *if there is any execution strategy* $\sigma$ *for* $\Pi$ *such that* $\sigma(\omega) = \sigma(\omega')$ *for each* $\omega, \omega' \in \Omega_\Pi$.

Now, to formally define *dynamic controllability*, we need a way to talk about what happened *up to a certain time*, during the execution, or, equivalently, to represent which information the execution strategy can rely on to inform its decisions. To this aim, given a scheduling function $\theta$, let $\theta_{<t}$, for some $t \in \mathbb{N}$, be a function mapping any token $\tau$ such that $\theta(\tau) < t$ to its duration. It is, in other words, a partial situation, ignoring any token that does not end before time $t$. We can now define *dynamic execution strategies*, and *dynamically controllable* flexible plans, *i.e.*, plans that admit such strategies.

**Definition 2.26** — Dynamic controllability.
*Let $\Pi$ be a flexible plan. Let $\sigma$ be an execution strategy for $\Pi$, let $\omega, \omega' \in \Omega_{\Pi}$ be two relevant situations, and $\tau \in \mathsf{tokens}_C(\Pi)$ be a controllable token in $\Pi$. Moreover, let $\sigma(\omega) = \theta$, $\sigma(\omega') = \theta'$, and $t = \theta(\tau)$. Then:*

1. *$\sigma$ is a* dynamic execution strategy *if $\theta_{<t} = \theta'_{<t}$ implies $\theta(\tau) = \theta'(\tau)$;*

2. *$\Pi$ is* dynamically controllable, *if it has a dynamic execution strategy.*

# 4 CONCLUSIONS

This chapter provided an overview of the formal framework representing timeline-based planning, as introduced by Cialdea Mayer et al. [44], which will act as the foundation of the work shown in subsequent chapters. Our presentation differs for a few aspects that will ease the exposition of what follows, but in such a way to be equivalent with respect to all its interesting properties.

As mentioned, the formalism presented here does not consider the handling of *resources*, which are nevertheless an important aspect of any real-world problem and are usually quite well supported by timeline-based systems [54]. Recently, the framework has been expanded to consider this feature [135], hence considering it in our theoretical investigation is one of the front-facing future developments of our work.

# II

# TIMELINE-BASED PLANNING

# EXPRESSIVENESS OF TIMELINE-BASED PLANNING 3

We begin our investigation of theoretical properties of timeline-based planning by answering some questions about the *expressiveness* of timeline-based planning languages. In particular, we compare timeline-based planning with a representative of action-based *temporal planning* languages, showing that the former is expressive enough to capture the latter.

# 1 INTRODUCTION

Timeline-based planning has evolved through the years in relative isolation with regards to more mainstream approaches to planning. As a result, there are at present no results directly comparing timeline-based and action-based planning languages in terms of expressiveness. The question of which domains can be compactly expressed in a formalism instead of the other is both of practical relevance and of theoretical interest. Given the exclusive focus of timeline-based planning on *temporal reasoning*, the ideal candidate for this comparison is clearly the paradigm of *temporal planning*, usually represented by the features introduced in PDDL 2.1, *i.e.*, actions with a given *duration*, that can execute concurrently, possibly overlapping in time [59]. This kind of planning problems have been proved to be EXPSPACE-complete by Rintanen [119]. Indeed, earlier work provided a translation of temporal PDDL into NDDL, the timeline-based modelling language adopted by the EUROPA 2 system [16]. However, the translation was specific to the two languages.

In this chapter, instead, we compare the two approaches using the formal framework described in Chapter 2 as a representative of a broad family of timeline-based languages. In his complexity analysis of temporal planning, Rintanen [119] defined a streamlined formal language for describing temporal planning domains, which subsumes the key temporal features of PDDL 2.1. Here, we take this language as a representative of action-based temporal planning, for our expressiveness comparison with timeline-based languages. In particular, we show that any problem expressed in Rintanen's language can be expressed by a suitable timeline-based planning problem, as per Definition 2.12.

In this and in most of the next chapters, we choose to consider preferably *compact* translations, *i.e.*, such that the size of the resulting problem is at most polynomial in the size of the translated one. From a pure expressiveness perspective, any kind of translation would suffice, but it would be of very little interest. Instead, here we provide a compact translation that, moreover, can be produced in polynomial time, hence having also a precise computational meaning: a *reduction* between the two plan existence problems, that allows us to obtain a result of EXPSPACE-hardness for timeline-based planning, providing a starting point for the developments of Chapter 4.

This translation is provided in Section 3. Before diving into it, the next section provides a number of useful results investigating the expressiveness of the synchronisation rules syntax. In particular, we show how a number of features of the formalism can be regarded as syntactic sugar. These results provide some hints about how expressive the language of synchronisation rules is, with their limitations and their potential. Moreover, they simplify the exposition of all the subsequent chapters.

# 2 EXPRESSIVENESS OF SYNCHRONISATION RULES

The language of synchronisation rules used in timeline-based planning problems is at the same time highly constrained from a syntactic point of view, and quite rich in expressive power. Because of this syntactic richness, it can be difficult at times to recognise which features are really important and which are just convenient. In Chapter 2, we already pointed out some syntactic sugar that can be put on top of the basic syntax, and Section 3 will study the expressiveness of the formalism when compared with common action-based planning languages. Here, we point out a few results, instrumental for the next chapters, which show how certain syntactic features included in the base formalism are not essential and can be regarded as syntactic sugar as well.

Since the following results require the translation of some problems into others, we need to define precisely which kinds of translation we are considering. Intuitively, we say that a problem *embeds* into another if all the solutions of the former can be extracted from the solutions of the latter.

**Definition 3.1** — Embedding of problems.
*Let* $P = (\mathsf{SV}, S)$ *and* $P' = (\mathsf{SV}', S')$ *be two timeline-based planning problems. We say that* $P$ embeds *into* $P'$ *(and* $P'$ *can be* projected *into* $P$*) if there exists a* polynomially computable projection function $f : \mathsf{plans}(P') \to \mathsf{plans}(P)$ *such that for any* $\pi \in \mathsf{plans}(P)$ *there exists a* $\pi' \in \mathsf{plans}(P')$ *such that* $f(\pi') = \pi$ (i.e., $f$ *is* surjective*)*.

Two problems are *equivalent* if one can be embedded into the other and *vice versa* with a *bijective* projection function.

Hence, some syntactic features of the formalism can be considered as not essential if a problem that uses such feature can be embedded into one that does not. An essential requirement of Definition 3.1 is that of considering only translations for which a *polynomially computable* projection function is available. This restriction is essential to ensure the definition is meaningful: otherwise, any problem would be embeddable into any other problem with a set of solutions of the same cardinality. That would technically fit the definition, despite being completely useless. However, there are no *a priori* restrictions on the size of one problem compared to the other. Although all the following results provide polynomial translations, a translation producing, *e.g.*, an exponentially bigger problem (while perhaps easier to solve), could very well be interesting to study, from the point of view of the expressive power.

Pointwise atoms are simple examples of such non-essential features.

**Theorem 1** — Removal of pointwise atoms.
Let $P = (\mathsf{SV}, S)$ be a timeline-based planning problem. Then, $P$ can be translated, in polynomial time, into an *equivalent* problem $P' = (\mathsf{SV}', S')$ that does not make use of pointwise atoms in any synchronisation rule of $S'$.

*Proof.* $P'$ is built from $P$ by adding an extra state variable $x_\bot = (V_{x_\bot}, D_{x_\bot}, T_{x_\bot})$ such that $V_{x_\bot} = \{\bot\}$, $T_{x_\bot}(\bot) = \varnothing$, and $D_{x_\bot} = (0, +\infty)$. The extra variable has thus only one value, whose tokens cannot have any successor but can be arbitrarily long. Hence, any solution of $P'$ can only have a timeline for $x_\bot$ consisting of a single token with $x_\bot = \bot$ starting at time 0 and ending at the end of the whole solution. Now, let $\mathcal{E} \equiv \exists a_1[x_1 = v_1] \ldots a_n[x_n = v_n] \,.\, \mathcal{C}$ be an existential statement where $\mathcal{C}$ contains a pointwise atom $\alpha \equiv T \leq_{[l,u]} t$ (resp., $\alpha \equiv t \leq_{[l,u]} T$). We can get rid of $\alpha$ by adding $a_\bot[x_\bot = \bot]$ to the quantifier prefix of $\mathcal{E}$, and replacing $\alpha$ with the atom $\mathsf{start}(x_\bot) \leq_{[t-u,t-l]} T$ (resp., $\mathsf{start}(x_\bot) \leq_{[t+l,t+u]} T$), where $t-u$ is understood to be zero (resp., $+\infty$) if $u = +\infty$. Any solution plan $\pi$ for $P$ can be extracted from the corresponding solution $\pi'$ of $P'$ by simply ignoring the timeline for the extra $x_\bot$ variable, hence $P$ embeds into $P'$, and $\pi'$ can be obtained back from $\pi$ by adding the only possible timeline for $x_\bot$, hence $P'$ embeds into $P$ as well, and the two problems are *equivalent*. ■

A similar argument allows us to avoid *triggerless rules.*

■ **Theorem 2** — Removal of triggerless rules.
Let $P = (\mathsf{SV}, S)$ be a timeline-based planning problem. Then, $P$ can be translated, in polynomial time, into an *equivalent* problem $P' = (\mathsf{SV}', S')$ such that $S'$ does not contain any triggerless rule.

*Proof.* Similarly to the proof of Theorem 1, we can add to $\mathsf{SV}'$ a variable $x_\bot$ such that $V_{x_\bot} = \{\bot\}$ $T_{x_\bot}(\bot) = \varnothing$, and $D_{x_\bot} = (0, +\infty)$, such that the timeline for $x_\bot$ is constrained to only have a single token as long as the entire plan. Then, any triggerless rule $\mathcal{R} \equiv \top \to \mathcal{E}_1 \vee \ldots \vee \mathcal{E}_n$ in $S$ can be translated into the triggered one $\mathcal{R}' \equiv a_\bot[x_\bot = \bot] \to \mathcal{E}_1 \vee \ldots \vee \mathcal{E}_n$, which adds an always-satisfied trigger. Then, the projection function is trivially built by simply ignoring the timeline for $x_\bot$, and is easily invertible, hence the two problems are equivalent. ■

Another simple translation can flatten the duration function of state variables, by expressing the same constraints using synchronisation rules.

■ **Definition 3.2** — Trivial duration function.
*Let $x = (V_x, T_x, D_x)$ be a state variable. The duration function $D_x$ is* trivial *if $D_x(v) = (0, +\infty)$ for all $v \in V_x$.*

■ **Theorem 3** — Removal of non-trivial duration functions.
Let $P = (\mathsf{SV}, S)$ be a timeline-based planning problem. Then, $P$ can be translated, in polynomial time, into an *equivalent* problem $P' = (\mathsf{SV}', S')$ such that any state variable $x \in \mathsf{SV}'$ has a trivial duration function.

*Proof.* The problem $P' = (\mathsf{SV}', S')$ can be obtained by adding to $P = (\mathsf{SV}, S)$ a number of synchronisation rules, one for each $v \in V_x$ in each variable $x \in \mathsf{SV}$ such that $D_x(v) = (d^{x=v}_{min}, d^{x=v}_{max})$, with $d^{x=v}_{min} \neq 0$ and $d^{x=v}_{max} \neq +\infty$ and then setting $D_x(v) = (0, +\infty)$. The added rule has the following form:

$$a[x = v] \to \mathsf{duration}(a) \geq d^{x=v}_{min} \wedge \mathsf{duration}(a) \leq d^{x=v}_{max}$$

where the second conjunct is omitted if $d_{max}^{x=v} = +\infty$. Then, a solution for $P'$ is directly a solution for $P$ as well, and *vice versa*, hence the projection function is the identity, up to renaming of variables, and the two problems are equivalent. A few words are worth spending on the time complexity of this translation. Since it iterates over all the values of the domain of each variables, one may at first suppose that an exponential number of iterations are needed. However, the input contains the representation of the transition function $T_x$ of each variable $x$, that has to specify some subset of $V_x$ for each $v \in V_x$. This implies that the input length is of the same order of magnitude of the domains cardinality, and thus iterating over each element of such domains takes linear time. ■

In contrast to the removal of duration functions shown in Theorem 3, removing transition functions is trickier.

■ **Definition 3.3** — Trivial transition function.
*Let $x = (V_x, T_x, D_x)$ be a state variable. The transition function $T_x$ is* trivial *if $T_x(v) = V_x$ for all $v \in V_x$.*

■ **Theorem 4** — Removal of non-trivial transition functions.
Let $P = (\mathsf{SV}, S)$ be a timeline-based planning problem. Then, $P$ can be translated, in polynomial time, into an *equivalent* problem $P' = (\mathsf{SV}', S')$ such that any state variable $x \in \mathsf{SV}'$ has a trivial transition function.

*Proof.* We need to express the fact that for each token where a certain variable $x$ holds a given value $v$, either the plan ends or the subsequent token has to hold one of the allowed values. However, we cannot reference the end of the plan in that way, without using a non-trivial transition function as in the proof of Theorem 1. Hence, we have to go the other way around: for any token for $x$, either this is the first token, or it is an allowed successor of its predecessor.

More formally, we build $P'$ by turning the transition functions into trivial ones and adding the following rule for each $x \in \mathsf{SV}$ and each $v \in V_x$:

$$a[x = v] \rightarrow \mathsf{start}(a) = 0 \lor \bigvee_{\substack{v' \in V_x \\ v \in T_x(v')}} \exists b[x = v'] \,.\, b \mathrel{\mathsf{meets}} a$$

Here, again, the solutions for $P'$ are trivially solutions for $P$ as well, and *vice versa*, and the two problems are equivalent. ■

It is worth to note that, in contrast to Theorem 3, the translation shown in Theorem 4 makes use of pointwise atoms (albeit a rather specific one). Without them, there is no way for a rule to reference the start of the whole plan, which is needed in order to encode transition functions where $T_x(v) = \varnothing$ for some $v$.

Hence, while both Theorems 1 and 4 hold, they cannot be combined to remove the use of *both* pointwise atoms and non-trivial transition functions. The same applies to Theorem 2 as well, which does not seem to be provable without the use of non-trivial transition functions.

All the translations above involve a very simple manipulation of variables, with the addition of some rules at most, and the solutions of the translated problems were mostly already solutions for the original one. The following is an example of a more involved translation where the shape of the problem is changed more evidently, motivating the generality of Definition 3.1.

**Theorem 5** — Restriction to binary variables.
Let $P = (\mathsf{SV}, S)$ be a timeline-based planning problem. Then, $P$ can be translated, in polynomial time, into an *equivalent* problem $P' = (\mathsf{SV}', S')$ that contains only binary variables, *i.e.*, $V_x = \{0,1\}$ for each $x \in \mathsf{SV}'$.

*Proof.* Let $P = (SV, S)$ be a timeline-based planning problem, with a state variable $x = (V_x, T_x, D_x) \in \mathsf{SV}$ with domain $V_x = \{v_1, \ldots, v_k\}$. We can construct an equivalent problem $P' = (\mathsf{SV}', S')$ where $x$ has been replaced by a suitable number of binary state variables $x_1 = (V_1, T_1, D_1), \ldots, x_k = (V_k, T_k, D_k)$. Note that we cannot simply replace each value of the variables domains with their binary encoding, since manipulating that would require to apply a universal quantification over multiple variables at once, which is not possible. Instead, we need to use a unary encoding, with one variable for each possible value that $x$ can hold. Synchronisation rules are then introduced to ensure that $x_i = 1$ holds in a solution of $P'$ if and only if $x = v_i$ holds in the corresponding solution for $P$. However, as before, iterating over the values of the variables domains can be done in linear time thanks to the presence of the transition functions as part of the input.

As a first step, a set of rules is introduced to ensure mutual exclusion between the binary variables, so for all $v_i \in V$:

$$a_i[x_i = 1] \rightarrow \exists \overbrace{a_0[x_0 = 0] \ldots a_k[x_k = 0]}^{\text{except } a_i} . \bigwedge_{\substack{j=0 \\ j \neq i}}^{k} a_i = a_j$$

Transitions and duration functions $T_x$ and $D_x$ have to be transferred to the new variables. By Theorem 3 we can suppose, without loss of generality, that the duration function is trivial. The transition function has to be encoded through the use of additional synchronisation rules. Hence, for each $i = 1, \ldots, k$ we set $T_i(v) = \{0,1\}$, and we add the following rule:

$$a_i[x_i = 1] \rightarrow \bigvee_{\substack{j=1,\ldots,k \\ v_j \in T(v_i)}} \exists a_j[x_j = 1] . \mathsf{end}(a_i) \leq_{[0,0]} \mathsf{start}(a_j)$$

All the synchronisation rules of the original problem have to be translated to speak about the newly introduced binary variables. So each appearance of a token quantifier of the form $a[x = v_i]$, both in a trigger or in an existential statement, is replaced with one of the form $a[x_i = 1]$. For example:

$$a[x = v_1] \rightarrow \exists b[x = v_2] \,.\, \mathcal{C}$$

is replaced by:

$$a[x_1 = 1] \rightarrow \exists b[x_2 = 1] \,.\, \mathcal{C}$$

The projection function to obtain a solution for $P$ given one for $P'$ is less trivial than the other cases, but is not complex. Each group of timelines for variables $x_1, \ldots, x_k$ is translated into a timeline for $x$ with the value given by the binary variable currently set. The encoding is two-way, hence the opposite projection is possible as well, and the two problems are equivalent. ■

These results, although simple, show both the features and the limitations of the synchronisation rules syntax. The absence of negation, the limited use of disjunctions, and the fixed $\forall\exists^*$ quantification structure, concur to obtain a peculiar tool, and will be crucial for the work shown in the following chapters.

# 3 TIMELINES VS ACTIONS

This section studies the relationship between timeline-based planning languages, represented by the formal language defined in Chapter 2, and action-based planning languages, which is commonly represented by the PDDL [99] language. The goal is to compare the expressiveness of the two approaches to the modelling of planning problems.

We focus on deterministic timeline-based planning problems, as defined in Section 2 of Chapter 2, as uncertainty will be added to the picture in Chapter 5. Given the explicit focus on temporal reasoning put by timeline-based planning, the natural action-based counterpart consists in *temporal planning languages* such as PDDL 2.1 [59], which adds to PDDL the explicit concept of time, and *durative actions*, *i.e.*, actions with a specified time duration.

Rather than directly studying PDDL itself, however, we take a more streamlined language as a representative of action-based planning languages, which is formally easier to define. The considered language was introduced by Rintanen [119] to study the computational complexity of the problem of plan existence for such kinds of planning domains. First, Rintanen's language will be recalled, and then we will show how any temporal planning problem can be expressed by a suitable timeline-based planning problem, with a translation preserving the set of solutions.

## 3.1 RINTANEN'S TEMPORAL PLANNING LANGUAGE

Here we recall the definition of the action-based temporal planning language that will be used in our comparison with timeline-based planning. In the following, we will refer to problems in this language simply as *temporal planning problems*. It can be thought of as an extension of a classical STRIPS-like planning language, where preconditions of actions can involve any state traversed during the execution of the plan rather than only the current one. The approach to the modelling of actions is, in some sense, opposite to that of PDDL. Instead of modelling the future effects of something that is being done now (*e.g.*, starting an action), the language focuses on what has to happen at the current time point if some preconditions involving the past hold (*i.e.*, the action started).

Preconditions are expressed as formulae whose syntax is defined as follows.

**Definition 3.4** — Precondition formula.
*Let $\Sigma$ be a set of* proposition letters. *A* precondition formula *over $\Sigma$ is any expression produced by the following grammar:*

$$\phi := p \mid \neg\phi \mid \phi \wedge \phi \mid \phi \vee \phi \mid [i..j]\phi$$

*where $p \in \Sigma$ and $i, j \in \mathbb{Z}$ with $i \leq j$.*

The set of all the precondition formulae over $\Sigma$ is denoted as $\mathcal{F}_\Sigma$. Precondition formulae are made of standard propositional logic connectives with the addition of a temporal operator $[i..j]\phi$ that states that $\phi$ must hold in all time points *from $i$ to $j$ steps from now*. The formula $[i..i]\phi$ is written in short as $[i]\phi$. In Rintanen's original definition, precondition formulae were restricted to talk about the past only. For example, the formula $[-5](p \wedge [3]q)$ contains the subformula $[3]q$, but it refers only to time points in the past as it is equivalent to $[-5]p \wedge [-2]q$. Our encoding does not need this restriction.

**Definition 3.5** — Temporal planning problem.
*Let $\Sigma$ be a set of proposition letters. A* literal *over $\Sigma$ is either $p$ or $\neg p$, with $p \in \Sigma$. A* temporal planning problem *over $\Sigma$ is a tuple $\mathcal{P} = (A, I, O, R, D, G)$, where:*

1. *$A \subseteq \Sigma$ is a finite set of proposition letters, called* fluents;[1]
2. *$I \subseteq A$ is a subset of $A$ representing the fluents true at the* initial state;
3. *$O \subseteq \Sigma$ is a finite set of proposition letters, called* actions *(or* operators*);*
4. *$R : O \to \mathcal{F}_\Sigma$ is a function that maps each action to its* precondition;
5. *$D$ is the* domain, i.e., *a finite set of rules of the form $r = (\phi, E)$, where $\phi$ is a precondition formula and $E$ is a set of literals over $A$;*
6. *$G \in \mathcal{F}$ is a precondition formula that specifies the* goal *condition.*

Intuitively, a rule $r = (\phi, E)$ states that whenever $\phi$ holds at time point $i$, the literals in $E$ become true at time point $i+1$, while all the other proposition letters preserve their truth. The truth of precondition formulae at each time point depends both on the truth value of fluents and on which actions are being executed (*i.e.*, the corresponding proposition letters hold) at that point, hence by choosing different actions we obtain different outcomes, depending on which rules are triggered. An action $o$ can be executed at any given time point $i$ only if it is *applicable*, *i.e.*, if its precondition $R(o)$ is true at $i$. It can be seen [119] that this language can model durative actions, conditional effects, and many other common features of PDDL 2.1.

To formally define this intuitive meaning, let us first give a semantics to precondition formulae. Given a temporal planning problem $\mathcal{P} = (A, I, O, R, D, G)$, let $\Sigma = A \cup O$. A *trace* for $\mathcal{P}$ is a finite sequence $\overline{\sigma} = \langle \sigma_1, \ldots, \sigma_n \rangle$ of *states* $\sigma_i \subseteq \Sigma$, meaning that the proposition letter $p \in \Sigma$ holds at time $i$ if and only if $p \in \sigma_i$.

**Definition 3.6** — Semantics of precondition formulae.
*Let $\mathcal{P} = (A, I, O, R, D, G)$ be a temporal planning problem, $\phi$ a precondition formula over $\Sigma$, $\overline{\sigma}$ a trace for $\mathcal{P}$, and $t \in \mathbb{N}$. We define when $\overline{\sigma}$* satisfies *$\phi$ at time $t$, written $\overline{\sigma} \models_i \phi$, as follows:*

1. $\overline{\sigma} \models_i p$ *iff* $i \geq 0$ *and* $p \in \sigma_i$;[2]
2. $\overline{\sigma} \models_i \neg\phi$ *iff* $\overline{\sigma} \not\models_i \phi$;
3. $\overline{\sigma} \models_i \phi_1 \wedge \phi_2$ *iff* $\overline{\sigma} \models_i \phi_1$ *and* $\overline{\sigma} \models_i \phi_2$;
4. $\overline{\sigma} \models_i \phi_1 \vee \phi_2$ *iff* $\overline{\sigma} \models_i \phi_1$ *or* $\overline{\sigma} \models_i \phi_2$;
5. $\overline{\sigma} \models_i [j..k]\phi$ *iff* $\overline{\sigma} \models_{i+h} \phi$ *for all* $j \leq h \leq k$.

Precondition formulae predicate over future and past states looking both at which fluents held and which actions were executing at any given time. Apart from the temporal operator, the semantics follows that of propositional logic. In particular, we can define some common shorthands, such as $\top \equiv p \vee \neg p$ for some $p \in \Sigma$, and $\bot \equiv \neg\top$.

A *plan* over some set of actions $O$ is a sequence $\overline{O} = \langle O_1, \ldots, O_n \rangle$ of *sets* of actions $O_i \subseteq O$. How a plan $\overline{O}$ for a problem $\mathcal{P}$ gets executed is defined as follows.

**Definition 3.7** — Execution trace of a plan.
*Let $\mathcal{P} = (A, I, O, R, D, G)$ be a temporal planning problem, and let $\overline{O} = \langle O_1, \ldots, O_n \rangle$ be a plan over $O$. The* execution trace *of $\overline{O}$ for $\mathcal{P}$ is a trace $\overline{\sigma} = \langle \sigma_1, \ldots, \sigma_{n+1} \rangle$ defined as follows, for each $0 \leq i \leq n$:*

---

[1] Elements of $A$ were called *state variables* in [119], but we change terminology here to not conflict with state variables of timeline-based planning problems.

[2] The original semantics given in [119] defines as well the truth value of fluents at negative time points , giving them the same truth value as the initial state. This arbitrary choice, however, is not essential in the proofs given there, nor for ours, so we adopt a more standard semantics.

1. $a \in \sigma_0$ *iff* $a \in I$;
2. $o \in \sigma_i$ *iff* $o \in O_i$;
3. $\overline{\sigma} \models_i R(o)$ *for all* $o \in \sigma_i$, i.e., *the preconditions of actions are satisfied;*
4. $a \in \sigma_{i+1}$ *if there exists a rule* $r = (\phi, E) \in D$ *such that* $\overline{\sigma} \models_i \phi$ *and* $a \in E$, *or if* $a \in \sigma_i$ *and there is no such rule such that* $\neg a \in E$;
5. $a \notin \sigma_{i+1}$ *if there exists a rule* $r = (\phi, E) \in D$ *such that* $\overline{\sigma} \models_i \phi$ *and* $\neg a \in E$, *or if* $a \notin \sigma_i$ *and there is no such rule such that* $a \in E$;

Note that Items 4 and 5 of Definition 3.7 are defined in such a way that if there is no enabled rule affecting a given fluent, its truth value remains unchanged from the previous state. Note that the kind of planning problems that we are defining are *deterministic*, hence the two conditions conflict if two rules affect the same fluent in opposite ways. In this case, it means some actions have conflicting effects, and the execution trace for such a plan does not exist.

**Definition 3.8** — Solution plan.
*A* solution plan *for a temporal planning problem* $\mathcal{P} = (A, I, O, R, D, G)$ *is a plan* $\overline{O} = \langle O_1 \ldots, O_n \rangle$ *that, when executed, reaches a state that satisfies the goal* $G$, i.e., *such that the execution trace* $\overline{\sigma}$ *of* $\overline{O}$ *over* $\mathcal{P}$ *exists and* $\overline{\sigma} \models_{n+1} G$.

The original motivation for the introduction of this language was the study of the computational complexity of temporal planning problems. In this regard, Rintanen [119] proves that, given a temporal planning problem $\mathcal{P} = (A, I, O, R, D, G)$, deciding whether there exists a solution plan for $\mathcal{P}$ is EXPSPACE-complete.

## 3.2 CAPTURING TEMPORAL PLANNING WITH TIMELINES

We will now show how any temporal planning problem, as per Definition 3.5, can be polynomially encoded into a timeline-based planning problem, preserving the solutions.

The first step is to introduce a simplification of the syntax of Rintanen's temporal planning language, which can be applied without changing its complexity nor its expressive power. In particular, we show how temporal operators of the form $[i]\phi$ are sufficient to *compactly* express any general temporal formula $[i..j]\phi$, *i.e.*, that any problem can be rewritten to only use temporal operators of the former kind with at most a polynomial increase in size. As remarked previously, we are concerned to find translations that preserve solutions, *i.e.*, such that there is a bijection between the solutions of the translation and of the original problem, when restricted to the original alphabet. Hence given two temporal planning problems $\mathcal{P} = (A, I, O, R, D, G)$ and $\mathcal{P}' = (A', I', O', R', D', G')$, $\mathcal{P}'$ is a translation of $\mathcal{P}$ if, for any plan $\overline{O}'$ over $O'$, $\overline{O}'$ is a solution plan for $\mathcal{P}'$ if and only if its restriction to $O$ is a solution plan for $\mathcal{P}$.

**Lemma 3.9** — Simplification of temporal operators.
*Any temporal planning problem can be compactly translated into another one that only makes use of temporal formulae of the form* $[i]\phi$.

*Proof.* Let $\mathcal{P} = (A, I, O, R, D, G)$ be a temporal planning problem. We will translate it into a problem $\mathcal{P}' = (A', I', O', R', D', G')$ whose precondition formulae only makes use of temporal operators of the form $[i]\phi$, equal to $P$ excepting for what follows. First, observe that $[i..j]\phi \equiv [j][i-j..0]\phi$ and that $[0]\phi \equiv \phi$, thus we can suppose *w.l.o.g.* that all the occurrences of temporal operators are either already of the simple form $[i]\phi$ or of the form $[i..0]\phi$, for $i < 0$. For any formula $\phi$ that appears inside an occurrence of a temporal operator, let $[k_1..0]\phi, \ldots, [k_n..0]\phi$ be all such occurrences, and let $k = \mathsf{max}\{-k_1, \ldots, -k_n\} + 1$. The core idea is to encode a counter that increments at each step through all the execution of the plan, from zero up to a maximum of $k$ (and stays at $k$ afterwards), but resets to zero every time $\neg\phi$ holds. Then, to know if $[k_i..0]\phi$ holds it is sufficient to check if the counter is greater than $-k_i$.

The value of the counter $c^\phi$ for the formula $\phi$, in short only $c$ from now, is represented in binary notation by additional *actions* $c_0, \ldots, c_{w-1} \in O'$ ($c_0$ the least significant), where $w = \lceil \mathsf{log}_2(k+1) \rceil + 1$. What follows will use a few shorthands for basic formulae that assert useful facts about the counter:

1. The formula $c = n$, for $n < k$, asserts the current value of the counter, and is simply a conjunction of literals asserting the truth value of the single bits of $n$. The formula $c \neq n$ is a shorthand for $\neg(c = n)$.

2. The formula $c < n$ compares the current value of $c$ with a constant value $n$. This shorthand can be defined recursively on the number $w$ of bits of the counter. We can suppose *w.l.o.g.* that $n$ can be represented in at most the same $w$ bits as $c$, otherwise any formula $c < n$ with $n > 2^w - 1$ can be replaced by $\top$. Let $\langle b_0 \ldots b_{w-1} \rangle$ be the bits of $n$ (represented in the formulae as $\top$ for 1, $\bot$ for 0). For $w = 1$, $c_0 < b_0$ is just $\neg c_0 \wedge b_0$. For $w > 1$, $\langle c_0 \ldots c_{w-1} \rangle < \langle b_0 \ldots b_{w-1} \rangle$ is defined as:

$$(c_{w-1} < b_{w-1}) \vee (c_{w-1} \longleftrightarrow b_{w-1} \wedge \langle c_0 \ldots c_{w-2} \rangle < \langle b_0 \ldots b_{w-2} \rangle)$$

   Then, $c < n$ is just $\langle c_0 \ldots c_{w-1} \rangle < \langle b_0 \ldots b_{w-1} \rangle$. Moreover, shorthands $c > n$, $c \geq n$ and $c \leq n$ are defined as one may expect.

3. The formula $\mathsf{inc}(c)$ asserts that the counter has incremented its value since the previous step, *i.e.*, if $c$ currently holds the value $n$, then at the previous step it held the value $n - 1$, and vice versa. Again, it can be defined recursively on the number $w$ of bits. For $w = 1$, $\mathsf{inc}(c_0)$ is simply $[-1]c_0 \longleftrightarrow \neg c_0$. For $w > 1$, $\mathsf{inc}(\langle c_0 \ldots c_{w-1}\rangle)$ is defined as:

$$\bigwedge \begin{cases} [-1]c_0 \longleftrightarrow \neg c_0 & \text{the least-significant bit (lsb) increments} \\ [-1]c_0 \quad \rightarrow \mathsf{inc}(\langle c_1 \ldots c_{w-1}\rangle) & \text{if the lsb is set, it carries to the other bits} \\ \neg[-1]c_0 \quad \rightarrow \bigwedge_{i=1}^{w-1} (c_i \longleftrightarrow [-1]c_i) & \text{otherwise, they remain unchanged} \end{cases}$$

With these formulae in place we can write a rule that enforces the counter to increase at each step if less than $k$, stay still when it reaches $k$, and reset to zero whenever $\neg\phi$ holds. For this purpose we introduce an additional *fluent* $f_c \in A'$ that we will set to true at the initial state and that we require to be true in the goal condition. In other words, we define $I'(f_c) = 1$ and $G' \equiv G \wedge f_c$. This flag will be set to false by the following rule, to invalidate the plan whenever the counter does *not* behave as intended. The rule is thus $(\neg\psi, \{\neg f_c\})$ where $\psi$ is the following formula:

$$\begin{aligned} \psi \equiv &([-1](\phi \wedge c < k) \rightarrow \mathsf{inc}(c)) \wedge && (1) \\ &([-1](\phi \wedge c = k) \rightarrow c = k) \wedge && (2) \\ &([-1]\neg\phi \rightarrow c = 0) && (3) \end{aligned}$$

The first clause says that if at the previous step $\phi$ was true and the counter had not reached its maximum value, then an increment took place. The second clause says that if $\phi$ was true but the counter reached the maximum value, it stayed the same. Finally, the third clause states that if $\phi$ *did not* hold at the previous step, then the counter had to be reset to zero. Since this rule set $f_c$ to false whenever $\psi$ is *false*, any plan containing a sequence of states where the counter does not behave as wanted is rejected. However, any plan that was valid before is still valid now, when the truth values for the new actions and the new fluents are added accordingly. With the counter in place, we can rewrite any formula of the form $[k_i..0]\phi$ with the formula $c_\phi > -k_i$, stating that the steps passed since the last time $\phi$ was false are more than $-k_i$.

Note that this encoding only adds a constant number of rules and a single new fluent for each formula $\phi$ that appears inside a temporal operator. The size of the precondition formula for the new rule is polynomial in the number of *bits* used to represent $k_0, \ldots, k_n$, thus polynomial in the size of the input. ■

Note that Lemma 3.9, as an immediate consequence, implies that it is possible to obtain a *negated normal form* for precondition formulae, that is, for each formula $\phi$ there is an equivalent one $\mathsf{nnf}(\phi)$ where negations are only applied to proposition letters (either fluents or actions). $\mathsf{nnf}(\phi)$ can be obtained as for propositional logic as far as boolean connectives are concerned, and by observing that $\neg[i]\phi \equiv [i]\neg\phi$. Similarly, note that $[i](\phi_1 \wedge \phi_2) \equiv [i]\phi_1 \wedge [i]\phi_2$ and $[i](\phi_1 \vee \phi_2) \equiv [i]\phi_1 \vee [i]\phi_2$, hence any formula can also be translated to an equivalent one where all the temporal operators are pushed down to be applied only to literals.

Now we can show how to encode any temporal planning problem into a corresponding timeline-based planning problem. As always, we are interested in translations that produce problems of size polynomial in the size of the original problem, and that preserve all its solutions.

**Theorem 6** — Timelines can capture action-based temporal planning.
Let $\mathcal{P} = (A, I, O, R, D, G)$ be a temporal planning problem. A timeline-based planning problem $P = (\mathsf{SV}, S)$ can be built in polynomial time such that there is a one-to-one relation between solution plans for $\mathcal{P}$ and solution plans for $P$.

*Proof.* Let $\mathcal{P} = (A, I, O, R, D, G)$ be an temporal planning problem. Thanks to Lemma 3.9, we can assume *w.l.o.g.* that all the temporal operators that appear in precondition formulae for rules in $D$ and all action preconditions in $R$ are of the form $[i]\phi$. Moreover, we can assume that all the aforementioned formulae are in *negated normal form*, and that all the temporal operators are pushed down to literals. We will now translate $\mathcal{P}$ into a timeline-based problem $P = (\mathsf{SV}, S)$ in a solution-preserving way. Let $F$ be a set of formulae built as follows:

1. $\phi \in F$ for each *subformula* $\phi$ of each precondition formula from the rules in $D$ and of each precondition $R(o)$, with $o \in O$;
2. for each $p \in A \cup O$, $p \in F$ and $\neg p \in F$;
3. for each $p \in A$, $[\pm 1]p \in F$ and $[\pm 1]\neg p \in F$;
4. for each rule $(\phi, E) \in D$, $[-1]\phi \in F$;
5. for each formula $\phi \in F$, $\mathsf{nnf}(\neg\phi) \in F$.

The set of state variables $\mathsf{SV}$ for $P$ contains a state variable $x_\phi$ for each $\phi \in F$. Each of these state variables is boolean (*i.e.*, their domain is the set $\{0, 1\}$), and its duration is fixed to a unitary length, that is, $D(v) = (1, 1)$ for each $v \in \{0, 1\}$. The transition function does not impose any constraint, so $T(v) = \{0, 1\}$ for $v \in \{0, 1\}$.

For each $p \in A \cup O$, the value of state variables $x_p$ will describe the execution trace of $\mathcal{P}$ at a given time point, with each unitary token corresponding to a single state. A set of suitable synchronisation rules will ensure that each $x_\phi$ state variable will be true (resp., false) only when the corresponding formula $\phi$ would be true (resp., false) given the truth values of literals. To implement this behaviour for conjunctions, for each formula $\phi \wedge \psi$ appearing in $F$, there will be two rules as follows:

$$
\begin{aligned}
a[x_{\phi\wedge\psi} = 1] &\rightarrow \exists b[x_\phi = 1] c[x_\psi = 1] \;.\; a = b \wedge a = c \\
a[x_{\phi\wedge\psi} = 0] &\rightarrow \exists b[x_{\mathsf{nnf}(\neg(\phi\wedge\psi))} = 1] \;.\; a = b
\end{aligned}
$$

The first rule above ensures that whenever we have an interval where $\phi \wedge \psi$ holds, then both $\phi$ and $\psi$ hold over that interval. The second rule handles the case where the formula is false, and it delegates the work to the rules governing the variable for its negation. The negated formula does not appear directly because all the formulae in $F$ are in negated normal form, hence a rule to handle negation is unnecessary. Negations are instead handled at the bottom level on literals, with rules connecting the tokens of $x_p$, for each letter $p$, with the tokens of its negation $x_{\neg p}$. So for each literal $\ell$ over letters $p \in A \cup O$ we have:

$$a[x_\ell = 1] \rightarrow \exists b[x_{\overline{\ell}} = 0] \,.\, a = b$$
$$a[x_\ell = 0] \rightarrow \exists b[x_{\overline{\ell}} = 1] \,.\, a = b$$

The rules for disjunctions are symmetrical to conjunctions:

$$\begin{aligned} a[x_{\phi\vee\psi} = 1] \rightarrow & \exists b[x_\phi = 1] \,.\, a = b \\ & \vee \exists b[x_\psi = 1] \,.\, a = b \\ a[x_{\phi\vee\psi} = 0] \rightarrow & \exists b[x_{\mathsf{nnf}(\neg(\phi\vee\psi))} = 1] \,.\, a = b \end{aligned}$$

The last kind of formula to handle is the temporal operator. For a formula $[i]\ell$, for some literal $\ell$, the rules have to ensure that whenever the corresponding variable is true in an interval, then $\ell$ holds at $i$ time steps *after* that point (or *before* if $i$ is negative). This is easily expressed as:

$$a[x_{[i]\ell} = 1] \rightarrow \exists b[x_\ell = 1] \,.\, \mathsf{start}(a) \leq_{[i,i]} \mathsf{start}(b)$$

if $i \geq 0$, and similarly, if $i < 0$:

$$a[x_{[i]\ell} = 1] \rightarrow \exists b[x_\ell = 1] \,.\, \mathsf{start}(b) \leq_{[i,i]} \mathsf{start}(a)$$

With this infrastructure in place, the timelines of the problem now encode the truth of all the formulae involved in the description of the execution trace of any solution plan for $\mathcal{P}$, so it is possible to encode the rules of the problem itself. Recall that each rule $(\phi, E)$ specifies that every time the precondition $\phi$ is satisfied, literals in $E$ must be true at the next step. This is equivalent to say that every time $\phi$ is satisfied, the formula $\bigwedge_{\ell \in E}[1]\ell$ holds, and it can thus be expressed as follows, where $\ell_1, \ldots, \ell_n \in E$:

$$a[x_\phi = 1] \rightarrow \exists a_1[x_{[1]\ell_1} = 1] \ldots a_n[x_{[1]\ell_n} = 1] \,.\, a = a_1 \wedge \cdots \wedge a = a_n$$

Since we are encoding a *deterministic* planning problem, it is also implicit that every literal not explicitly changed by a rule has to preserve its truth value. Additional synchronisation rules are required to ensure this *inertia*. These rules say that if a literal holds a given value at the current time point, it either had the same value at the previous step, or a precondition of some rule involving it was true, causing its change. A special case is needed for the first time point, which has not a predecessor. In detail, for every literal $\ell$ over $A$, let $\phi_1,\ldots,\phi_n$ be the preconditions of all the rules $r_i = (\phi_i, E_i)$ such that $\ell \in E_i$. Then, the inertia for literal $\ell$ can be expressed as follows:

$$
\begin{aligned}
a[x_\ell = 1] \;\rightarrow\; & \mathsf{start}(a) = 0 \\
\bigvee\; & \exists b[x_{[-1]\ell} = 1]\,.\, a = b \\
\textstyle\bigvee_{i=1}^{n} & \exists b[x_{[-1]\phi_i} = 1]\,.\, a = b
\end{aligned}
$$

In a similar way, it is possible to encode preconditions of actions, so that when actions are performed their preconditions are ensured to hold. For each action $o \in O$, we have:

$$a[x_o = 1] \rightarrow \exists b[x_{R(o)} = 1]\,.\, a = b$$

At this point, the domain of the problem is completely encoded by the synchronisation rules of the timeline-based planning problem, and it is now sufficient to express the initial state and the goal condition. Let $\ell_1,\ldots,\ell_n \in I$ be the literals asserted by the initial state, and $G$ be the formula that describes the goal condition. They are encoded by the following triggerless synchronisation rules:

$$
\begin{aligned}
&\top \rightarrow \exists a_1[x_{\ell_1} = 1]\ldots a_n[x_{\ell_n} = 1]\,.\, \mathsf{start}(a_1) = 0 \wedge \ldots \wedge \mathsf{start}(a_n) = 0 \\
&\top \rightarrow \exists a[x_G = 1]\,.\, \top
\end{aligned}
$$

This step completes the encoding. The timelines for the variables $x_o$, with $o \in O$ from any solution plan $\pi$ for $P$ encode a solution plan $\overline{O}$ for the temporal planning problem $\mathcal{P}$. Moreover, it can be seen that building $P$ can be done in polynomial time as stated, since it only involves iterating over the elements of $F$ whose size is linear in the size of $\mathcal{P}$. In particular the size of the encoded problem is polynomial as well. ■

Theorem 6 proves that any temporal planning problem can be encoded by a timeline-based planning problem preserving all its solutions. In its statement we stressed on the fact that the encoding can be produced in polynomial time. Although complexity-theoretic considerations will be the topic of Chapter 4, the EXPSPACE-completeness of temporal planning shown by Rintanen [119] allows us to already state the following immediate consequence of Theorem 6.

■ **Corollary 3.10**

*Let $P$ be a timeline-based planning problem. Deciding whether there exists a solution plan for $P$ is* EXPSPACE*-hard.* ■

# 4 CONCLUSIONS

In this chapter, we started our formal investigation of timeline-based planning by showing that the formal language described in Chapter 2 is expressive enough to capture action-based temporal planning. This result compares the two paradigms, setting a starting point for subsequent developments.

It is important to note that the translation provided in the proof of Theorem 6 requires a quite restricted set of features among those supported by the general formalism. In particular, note the absence of *unbounded* atoms. On the other hand, since temporal planning problems can have solutions of length even doubly exponential in the size of the problem [119], a translation targeting timeline-based planning problems with bounded horizon (Definition 2.14) would not be possible. This fact is worth noting, since most timeline-based planning often considers the bounded horizon variant of the problem.

# 4 COMPLEXITY OF TIMELINE-BASED PLANNING

With basic expressiveness issues settled in the previous chapter, we move to study the computational complexity of the problem of plan existence for a given timeline-based planning problem. We prove that finding whether a solution plan exists is EXPSPACE-complete, and becomes NEXPTIME-complete if a bound on the solution horizon is given. In doing that, we make use of a conceptual framework based on the notion of *rule graph*, a structure that allows us to decompose and reason about synchronisation rules. The concept is developed here and extensively employed in the rest of the thesis. Then, we define and approach the problem of finding *infinite* solution plans, a variant that was not considered in literature before. The problem is proved to be EXPSPACE-complete as well, employing a different automata-theoretic argument, that sheds light on the problem from an alternative perspective.

# 1 INTRODUCTION

How hard a problem is to solve is maybe the most common question that immediately comes to mind to anyone that is confronted with a new computational task. Analysing the computational complexity of a problem answers this question from a precise formal perspective. For action-based planning formalisms, this issue has been addressed in many forms. The basic classical planning problem has been proved to be PSPACE-complete [24], but has also been studied in greater detail: fixed-parameter tractable fragments have been studied [11], and strict time and space bounds have been identified [10]. The complexity of most of the many extensions to the classical planning problem have been classified as well, from the EXPSPACE-completeness of temporal planning discussed in Chapter 3 [119], to the many variants of nondeterministic and probabilistic planning problems [91, 92, 93, 102, 118].

In contrast, timeline-based planning lacks any result regarding the computational complexity of its related decision problems. This chapter provides the first results in this direction.

In particular, we prove that finding a solution plan for timeline-based planning problems, as per Definition 2.12, is EXPSPACE-complete. Since we know already that the problem is EXPSPACE-hard from the encoding of action-based temporal planning shown in Chapter 3 (Corollary 3.10), it is sufficient here to show a decision procedure that can solve the problem in an exponential amount of space. To exhibit such procedure, we introduce and make use of the notion of *rule graphs*, a graph representation of synchronisation rules that allows us to better decompose and reason about rules. In particular, the decomposition of rule graphs into a particular kind of connected components allows us to define a particular data structure, called *matching record*, that can completely represent the current state of a plan in a compact way. Matching records allow us to derive a *small-model result*, that provides a doubly-exponential upper bound on the size of solution plans for a given timeline-based planning problem, and to exhibit a decision procedure that, thanks to such upper bound, can be proved to require at most an exponential amount of space. Rule graphs and matching records are essential in our proof, but they also provide a solid foundation to reason about synchronisation rules and their semantics, that will be exploited in subsequent chapters as well (Chapters 5 and 7, in particular).

Then, we consider two different variants of the problem. First, deciding whether a solution plan exists for timeline-based planning problems with *bounded horizon* (Definition 2.14) is proved to be NEXPTIME-complete. The inclusion in the class can be easily shown by adapting the procedure for the general case, while the hardness is proved by a reduction from a certain kind of *tiling problems*. Then, a different automata-theoretic perspective is explored, showing how a nondeterministic finite automaton can be built to accept only the language of words that suitably represent the solution plans of a given problem. Such construction provides an alternative proof of the above results, but by adapting it to build a Büchi automaton, we prove that finding an *infinite* solution plan to a given problem is EXPSPACE-complete as well.

The chapter is structured as follows. At first, we introduce *rule graphs* and all the surrounding conceptual framework in Section 2. Then, Section 3 defines matching records and uses them to prove the doubly exponential upper bound on the size of solution plans, that then is used to prove the EXPSPACE-completeness of the problem. The NEXPTIME-completeness for the case with bounded horizon is proved thereafter. Then, Section 4 provides the complexity of the problem over *infinite* solution plans exploiting a different automata-theoretic argument, and Section 5 wraps up the chapter with final remarks and ideas for future lines of work.

# 2 STRUCTURE OF SYNCHRONISATION RULES

This section defines and explores the concept of *rule graph*, which will be essential in the development of the following complexity analysis, and extensively employed in the following chapters as well.

Without loss of generality, in this chapter we will make a few assumptions that simplify the exposition considerably. In particular, we can assume that any considered timeline-based planning problem does not make use of *pointwise atoms* (Theorem 1 in Chapter 3), has no *triggerless rules* (Theorem 2), and only makes use of *trivial duration functions* (Theorem 3). Moreover, given a timeline-based planning problem $P = (\mathsf{SV}, S)$, we assume *w.l.o.g.* that any rule $\mathcal{R} \in S$ is *verbose* in the following sense.

**Definition 4.1** — Verbose rule.
*Let $\mathcal{R} \equiv a_0[x_0 = v_0] \rightarrow \mathcal{E}_1 \vee \ldots \vee \mathcal{E}_m$ be a synchronisation rule. $\mathcal{R}$ is said to be* verbose *if, in all existential statements $\mathcal{E}_i$ of $\mathcal{R}$, for any token name $a$ quantified in $\mathcal{E}_i$ including $a_0$, both endpoints* $\mathsf{start}(a)$ *and* $\mathsf{end}(a)$ *are mentioned in the body of $\mathcal{E}_i$.*

It can be seen that any synchronisation rule can be turned into a verbose one by at worst adding constraints of the form $\mathsf{start}(a) \leq \mathsf{end}(a)$ to its existential statements, without changing the meaning of the rule as a whole.

## 2.1 PLANS AS EVENT SEQUENCES

In what follows, it will come useful to represent plans in a form more suitable to manipulation. In particular, instead of focusing on the single timelines as building blocks, plans can be flattened over a single sequence of events that mark the start/end of tokens.

**Definition 4.2** — Event sequence.
*Let* SV *be a set of state variables. Let* $\mathcal{A}_{\mathsf{SV}}$ *be the set of all the terms, called* actions, *of the form* $\mathsf{start}(x,v)$ *or* $\mathsf{end}(x,v)$, *where* $x \in \mathsf{SV}$ *and* $v \in V_x$.

*An* event sequence *over* SV *is a sequence* $\overline{\mu} = \langle \mu_1, \ldots, \mu_n \rangle$ *of pairs* $\mu_i = (A_i, \delta_i)$, *called* events, *where* $A_i \subseteq \mathcal{A}_{\mathsf{SV}}$ *is a finite set of actions, and* $\delta_i \in \mathbb{N}_+$, *such that, for any* $x \in \mathsf{SV}$:

1. *for all* $1 \le i \le n$, *if* $\mathsf{start}(x,v) \in A_i$ *for some* $v \in V_x$, *then there are no* $\mathsf{start}(x,v')$ *in any* $\mu_j$ *before the closest* $\mu_k$, *with* $k > i$, *such that* $\mathsf{end}(x,v) \in A_k$ *(if any);*
2. *for all* $1 \le i \le n$, *if* $\mathsf{end}(x,v) \in A_i$ *for some* $v \in V_x$, *then there are no* $\mathsf{end}(x,v')$ *in any* $\mu_j$ *after the closest* $\mu_k$, *with* $k < i$, *such that* $\mathsf{start}(x,v) \in A_k$ *(if any);*
3. *for all* $1 \le i < n$, *if* $\mathsf{end}(x,v) \in A_i$ *for some* $v \in V_x$, *then* $\mathsf{start}(x,v') \in A_i$ *for some* $v' \in V_x$.
4. *for all* $1 < i \le n$, *if* $\mathsf{start}(x,v) \in A_i$ *for some* $v \in V_x$, *then* $\mathsf{end}(x,v') \in A_i$ *for some* $v' \in V_x$.

Intuitively, an event $\mu_i = (A_i, \delta_i)$ consists of a set $A_i$ of actions describing the start or the end of some tokens, happening $\delta_i$ time steps after the previous one. In an *event sequence*, events are collected to describe a whole plan.

Note that events where nothing happens are possible, although useless. It is unnecessary to forbid them, though.[1] The first two conditions of Definition 4.2 ensure a proper parenthesis structure between the start and the end of tokens that appear in the sequence. Item 1 ensures that there are no tokens starting before the end of the previous one, and Item 2 ensures the specular condition for the $\mathsf{end}(x,v)$ actions. The end action of a token is placed at the same time of the starting one because start/end pairs identify intervals with the left extremum included but the right one excluded. Then, Items 3 and 4 of Definition 4.2 ensure that the end (start) of a token is followed (preceded) by the start (end) of another excepting, most importantly, for the last (first) event in the sequence. In this way we avoid any gap in the description of each timeline of the represented plan.

Figure 4.1 shows an example event sequence associated with a plan made of two timelines. The variables $x$ and $y$ in the example can take a few integer values, say $V_x = V_y = \{0,1,2,3\}$. Observe how each transition from a token to the next is represented by a pair of start/end actions belonging to the same event.

It is important to note that, by Definition 4.2, a started token is not required to end before the end of the sequence, and a token can end without the corresponding starting action to ever have appeared before. In this case, the event sequence is said *open* for the variable $x$ that is missing the start/end event. In event sequences where this does not happen, called *closed*, both the endpoints of all tokens are specified.

[1] As a matter of fact, allowing empty events improves the exposition of Chapter 5.

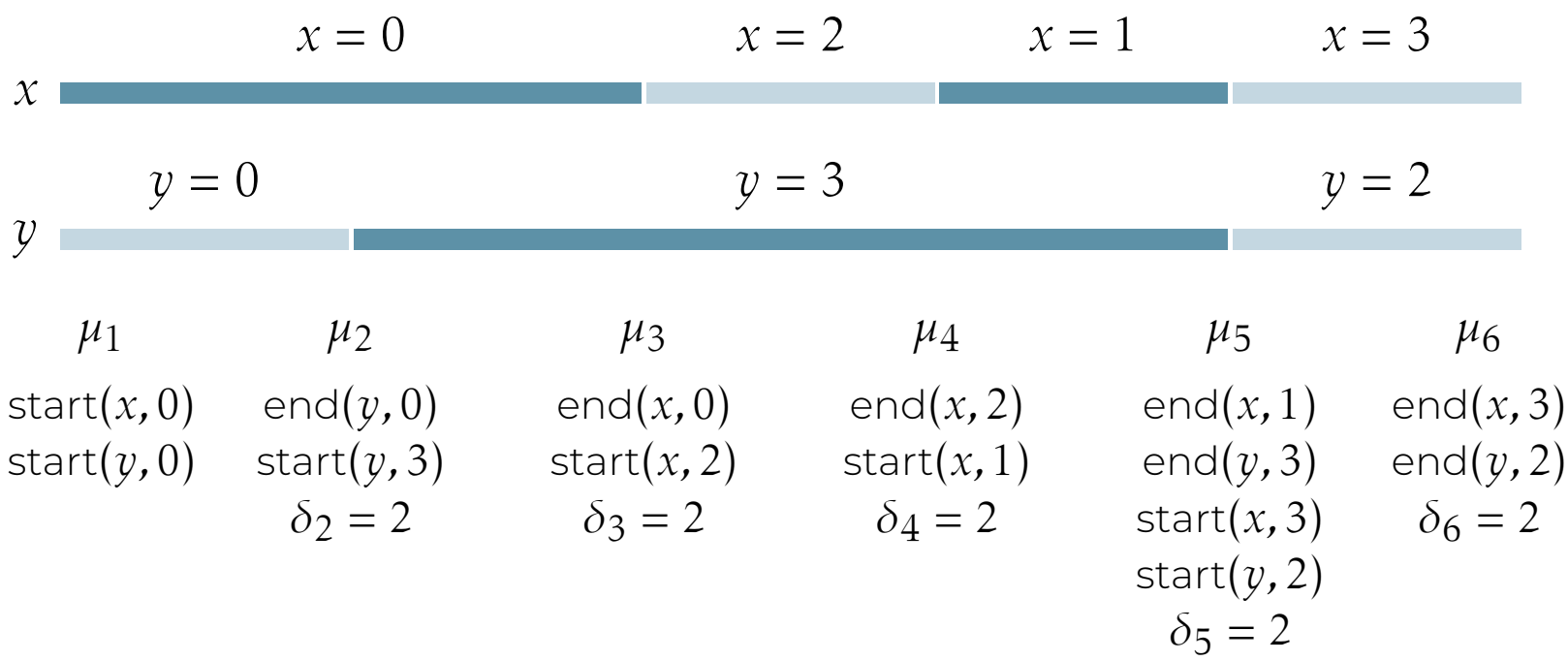

**Figure 4.1:** Example event sequence associated with two timelines

**Definition 4.3** — Open and closed event sequences.
*An event sequence* $\overline{\mu} = \langle \mu_1, \ldots, \mu_n \rangle$ *is* closed *on the right (left) for a variable* $x$ *if for each* $1 \leq i \leq n$, *if* $\mathsf{start}(x, v) \in A_i$ $(\mathsf{end}(x, v) \in A_i)$, *then there is a* $j > i$ $(j < i)$ *such that* $\mathsf{end}(x, v) \in A_j$ $(\mathsf{start}(x, v) \in A_j)$. *Otherwise,* $\overline{\mu}$ *is* open *on the right (left) for* $x$.

An event sequence is simply *open* or *closed* (to the right or to the left) if it is respectively open or closed (to the right or to the left) for any variable $x$. Note that the empty event sequence is closed on both sides for any variable. Moreover, on closed event sequences, the first event only contains $\mathsf{start}(x, v)$ actions and the last event only contains $\mathsf{end}(x, v)$ actions, and one for each variable $x$. By admitting open event sequences, we can represent plans that are under construction, for example during the execution of the decision procedure that is going to be presented in the next sections. Most importantly, a subsegment $\overline{\mu}_{[i\ldots j]}$ of an event sequence $\overline{\mu}$ is still, conveniently, a proper event sequence. Note that an event $\mu$ can be appended to an event sequence $\overline{\mu}$, *i.e.*, $\overline{\mu}\mu$ is a proper event sequence, only if $\overline{\mu}$ is open to the right. Similarly, two event sequences $\overline{\mu}_1$ and $\overline{\mu}_2$ can be concatenated into $\overline{\mu} = \overline{\mu}_1\overline{\mu}_1$ only if $\overline{\mu}_1$ is open to the right and $\overline{\mu}_2$ is open to the left, and the empty event sequence $\varepsilon$ can be concatenated to the right or to the left with any other.

Let us set some notation to talk about event sequences. Given an event sequence $\overline{\mu} = \langle \mu_1, \ldots, \mu_n \rangle$ over a set of state variables $\mathsf{SV}$, where $\mu_i = (A_i, \delta_i)$, we define $\delta(\overline{\mu}) = \sum_{1<i\leq n} \delta_i$, that is, $\delta(\overline{\mu})$ is the time passed between the start and the end of the sequence, its duration. The amount of time spanning a subsequence, written as $\delta_{i,j}$ when $\overline{\mu}$ is clear from context, is then $\delta(\overline{\mu}_{[i\ldots j]}) = \sum_{i<k\leq j} \delta_k$. If $\overline{\mu}$ is *closed to the left*, the first event properly defines a value for all the variables at the first time point, allowing us to define it for any time point throughout the sequence. Hence, for any $x \in \mathsf{SV}$, $v \in V_x$, and $1 \le i \le n$, we write $\mathsf{val}(\overline{\mu}, x, i) = v$ if and only if $\mathsf{start}(x, v) \in A_j$ where $j \le i$ and there is no $\mathsf{start}(x, v') \in A_k$ for any $v' \in V_x$ and any $j < k < i$.

Now, the direct correspondence between closed event sequences and plans can be easily defined.

**Definition 4.4** — Correspondence between event sequences and plans.
*Let $\overline{\mu} = \langle \mu_1, \ldots, \mu_n \rangle$ be a closed event sequence. Then, $\pi_{\overline{\mu}}$ is the* plan *where, for each $x \in \mathsf{SV}$, $\pi_{\overline{\mu}}(x) = \langle \tau_1, \ldots, \tau_k \rangle$ is a timeline such that $\mathsf{start}(x, v) \in A_i$ ($\mathsf{end}(x, v) \in A_i$) iff there is a $\tau_j$ such that $\mathsf{val}(\tau_j) = v$ and $\mathsf{start\text{-}time}(\tau_j) = \delta_{1,i}$ ($\mathsf{end\text{-}time}(\tau_i) = \delta_{1,i}$).*

Note how the essential assumption of not using pointwise atoms allows us to forget about any absolute time reference and reason only in terms of distance between events. In mapping an event sequence to its represented plan, the value of $\delta_1$ of the first event $\mu_1 = (A_1, \delta_1)$ is ignored, since it would represent the time passed after a non-existent previous event. By fixing an arbitrary value for $\delta_1$, the converse mapping from plans to event sequences can be also defined, hence we denote as $\overline{\mu}_\pi$ the event sequence such that $\pi_{\overline{\mu}_\pi} = \pi$.

Given this correspondence, the following sections will manipulate plans as event sequences. A token for a state variable $x$ is identified, inside an event sequence, by the pair of indices of the start and end events of the token. More formally, given $\overline{\mu} = \langle \mu_1, \ldots, \mu_n \rangle$, the pair $(i, j)$ of indices $1 \le i, j \le n$ (or the pair of the corresponding events) identifies a token $\tau \in \pi_{\overline{\mu}}(x)$ in the timeline for $x$ if $\mu_i = (A_i, \delta_i)$ and $\mu_j = (A_j, \delta_j)$ such that $\mathsf{start}(x, \mathsf{val}(\tau)) \in A_i$ and $\mathsf{end}(x, \mathsf{val}(\tau)) \in A_j$, and there are no other actions referring to $x$ in the events between $\mu_i$ and $\mu_j$. Considering again the example in Figure 4.1, we can say that the indices $(2, 5)$, and the corresponding events $\mu_2$ and $\mu_5$, identify the second token of the timeline for $y$, where $y = 3$. Moreover, if a rule $\mathcal{R} \equiv a_0[x_0 = v_0] \rightarrow \mathcal{E}_1 \vee \ldots \vee \mathcal{E}_m$ is triggered by some token $\tau$ of $\pi_{\overline{\mu}}(x_0)$, and $\tau$ is identified by two events $\mu_i$ and $\mu_j$, then we say that $\mathcal{R}$ is triggered by the event $\mu_i$, *i.e.*, we identify the triggering point of the rule with the start of the triggering token.

As a final remark, note that, by Definition 4.2, the same event cannot contain both $\mathsf{start}(x, v)$ and $\mathsf{end}(x, v)$ for the same $x$, *i.e.*, tokens cannot start and end at the same time. This is compatible with Definition 2.2, which excludes tokens of null duration.

## 2.2 THE RULE GRAPHS

We can now define the main theoretical tool behind our complexity analysis. The *rule graphs* are edge-labelled graphs that suitably represent the synchronisation rules of a problem in a way that makes them easier to manipulate.

**Definition 4.5** — Rule graphs.
*Let $\mathcal{E} \equiv \exists a_1[x_1 = v_1] \ldots a_k[x_k = v_k] \,.\, \mathcal{C}$ be one of the existential statements of a*

*synchronisation rule* $\mathcal{R} \equiv a_0[x_0 = v_0] \rightarrow \mathcal{E}_1 \vee \ldots \vee \mathcal{E}_m$.

*Then, the* rule graph *of* $\mathcal{E}$ *is an edge-labelled graph* $G_{\mathcal{E}} = (V, E, \beta)$ *where:*

1. *the set of nodes* $V$ *is made of* terms *(as per Definition 2.5) such that:*
   (a) $\mathsf{start}(a) \in V$ *or* $\mathsf{end}(a) \in V$ *if and only if* $a \in \{a_0, \ldots, a_k\}$*, for any* $a \in \mathcal{N}$*;*
   (b) *if the term* $T$ *is used in* $\mathcal{C}$*, then* $T \in V$*;*
2. $E \subseteq V \times V$ *is the edge relation such that, for each pair of nodes* $T, T' \in V$*, there is an edge* $(T, T') \in E$ *if and only if* $\mathcal{C}$ *contains an atom of the form* $T \leq_{[l,u]} T'$*, or* $T = \mathsf{start}(a_i)$ *and* $T' = \mathsf{end}(a_i)$ *for some* $0 \leq i \leq k$*;*
3. $\beta : E \rightarrow \mathbb{N} \times \mathbb{N}_{+\infty}$ *is the edge-labelling function, such that for each* $e \in E$*, if* $e$ *is associated with the atom* $T \leq_{[l,u]} T'$ *in* $\mathcal{C}$*, then* $\beta(e) = (l, u)$*, or* $\beta(e) = (0, +\infty)$ *otherwise, and* vice versa*, if* $\beta(e) \neq (0, +\infty)$*, then there is such an atom in* $\mathcal{C}$*.*

Figure 4.2 shows the rule graph for (the existential statement of) an example synchronisation rule. Unbounded edges, *i.e.*, those with $\beta(e) = (l, +\infty)$, are drawn as dashed arrows. Intuitively, a rule graph is an alternative representation of an existential statement where the endpoints of each quantified token are represented by nodes, and temporal constraints between such endpoints by labelled edges. Note that, in Definition 4.5 and in any manipulation of rule graphs, the token name $a_0$ used in the trigger of the synchronisation rule of a given existential statement is considered to be quantified in the statement as all the other token names involved.

Any existential statement has its rule graph, and the opposite connection is possible as well. Indeed, given any edge-labelled graph $G = (V, E, \beta)$ where $V$ is a set of terms, $E \subseteq V \times V$, and $\beta : E \rightarrow \mathbb{N} \times \mathbb{N}_{+\infty}$, we can associate an existential statement $\mathcal{E}$ such that $G_{\mathcal{E}} = G$. This close relationship allows us to identify the rule graphs with their existential statements, manipulating or analysing the graphs to affect the existential statements, and *vice versa*.

A rule graph *matches* over an event sequence if each node can be mapped to an event such that such temporal constraints are satisfied.

**Definition 4.6** — Rule graphs matching over event sequences.
*Let* $\overline{\mu} = \langle \mu_1, \ldots, \mu_n \rangle$ *be a (possibly open) event sequence, and let* $G_{\mathcal{E}} = (V, E, \beta)$ *be the rule graph of some existential statement* $\mathcal{E}$*. A* matching function *is a function* $\gamma : V \rightarrow [1, \ldots, n]$*, mapping each node* $T \in V$ *with an event* $\mu_{\gamma(T)}$ *in* $\overline{\mu}$*.*

*We say that* $G_{\mathcal{E}}$ matches $\overline{\mu}$ *with the matching function* $\gamma$*, written* $\overline{\mu}, \gamma \models G_{\mathcal{E}}$*, if:*

1. *for each node* $T \in V$*, with* $T = \mathsf{start}(a)$ *(resp.,* $T = \mathsf{end}(a)$*), if* $a$ *is quantified as* $a[x = v]$ *in* $\mathcal{E}$*, then the event* $\mu_{\gamma(T)} = (A_T, \delta_T)$ *is such that* $\mathsf{start}(x, v) \in A_T$ *(resp.,* $\mathsf{end}(x, v) \in A_T$*);*
2. *if both* $T = \mathsf{start}(a)$ *and* $T' = \mathsf{end}(a)$ *belong to* $V$ *for some token name* $a \in \mathcal{N}$*, then* $\gamma(T)$ *and* $\gamma(T')$ *identify the endpoints of a token for* $x$ *in* $\pi_{\overline{\mu}}(x)$*;*

$$a[x=0] \to \exists b[y=3]c[x=1]\,.\,\mathsf{start}(b) \leq_{[0,3]} \mathsf{end}(a) \land \mathsf{end}(b) = \mathsf{end}(c) \land \mathsf{start}(a) \leq \mathsf{end}(a) \land \mathsf{start}(c) \leq \mathsf{end}(c)$$

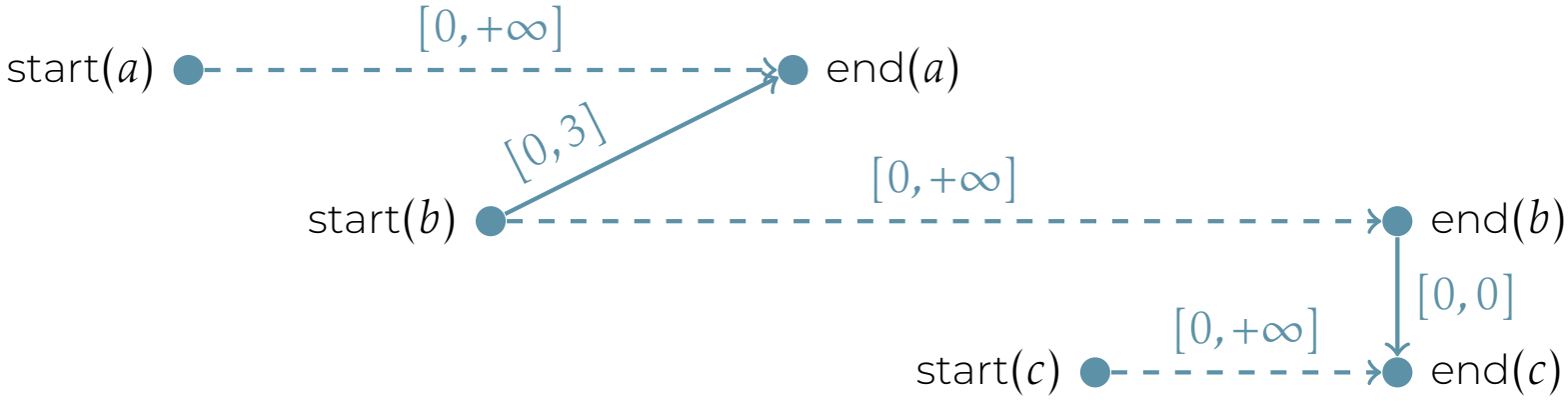

**Figure 4.2:** Rule graph of the existential statement of an example rule. Note that the rule is *verbose*, hence the rule graph contains both start and end nodes for each token name involved.

3. *if* $\mathsf{start}(a) \in V$ *but* $\mathsf{end}(a) \notin V$, *then there is no event in* $\overline{\mu}$ *that ends the token started at* $\gamma(\mathsf{start}(a))$*;*

4. *if* $\mathsf{end}(a) \in V$ *but* $\mathsf{start}(a) \notin V$, *then there is no event in* $\overline{\mu}$ *that starts the token ended at* $\gamma(\mathsf{end}(a))$*;*

5. *for any edge* $e \in E$, *if* $e = (T, T')$ *and* $\beta(e) = (l, u)$, *then* $l \leq \delta_{\gamma(T),\gamma(T')} \leq u$.

*We write* $\overline{\mu} \models G_{\mathcal{E}}$ *if there is some* $\gamma$ *for which* $\overline{\mu}, \gamma \models G_{\mathcal{E}}$.

As an example, if $\overline{\mu} = \langle \mu_1, \ldots, \mu_6 \rangle$ is the event sequence shown in Figure 4.1, then the rule graph of Figure 4.2 matches over $\overline{\mu}$ with a matching function $\gamma$ such that $\gamma(\mathsf{start}(a)) = 1$, $\gamma(\mathsf{start}(b)) = 2$, $\gamma(\mathsf{end}(a)) = 3$, $\gamma(\mathsf{start}(c)) = 4$, and $\gamma(\mathsf{end}(b)) = \gamma(\mathsf{end}(c)) = 5$.

Note that the use of Definition 4.6 requires an explicit association between the rule graph and its original existential statement, because the rule graph by itself does not include any information about the meaning of the token names mentioned in its nodes. For this reason we often keep this explicit association by referring to a rule graph as $G_{\mathcal{E}}$, where $\mathcal{E}$ is its original existential statement.

Now, recall how the satisfaction of existential statements is defined by Definition 2.10. As event sequences represent plans, if a rule graph matches over an event sequence, it follows that the corresponding existential statement is satisfied by the plan. Hence, we can rephrase whether a synchronisation rule is satisfied by a plan in terms of how its rule graphs match over the corresponding event sequence.

■ **Lemma 4.7** — Satisfaction of a synchronisation rule by an event sequence. *Let* $\mathcal{R} \equiv a_0[x_0 = v_0] \to \mathcal{E}_1 \lor \ldots \lor \mathcal{E}_m$ *be a synchronisation rule over a set of state variables* SV, *let* $\pi$ *be a plan over* SV, *and let* $\overline{\mu}_\pi = \langle \mu_1, \ldots, \mu_n \rangle$ *be the correspond-* *ing event sequence. Then, $\pi \models \mathcal{R}$ if and only if, for any $\mu_i = (A_i, \delta_i)$ such that* $\mathsf{start}(x_0, v_0) \in A_i$*, there is an $\mathcal{E}_k$ and a $\gamma$ such that $\overline{\mu}, \gamma \models G_{\mathcal{E}_k}$ and* $\gamma(\mathsf{start}(a_0)) = i$.

*Proof.* Let $\mathcal{E} \equiv a_1[x_1 = v_1] \ldots a_n[x_n = v_n] \,.\, \mathcal{C}$ be any existential statement of $\mathcal{R}$. At first, we need to show that there is a token mapping $\eta$ such that $\pi \models_\eta \mathcal{E}$ if and only if there is a matching function $\gamma$ such that $\overline{\mu}_\pi, \gamma \models G_{\mathcal{E}}$. Then, it directly follows that if the rule is triggered by an event $\mu_i$ that identifies the starting point of the trigger token $\tau$, and $\gamma(\mathsf{start}(a_0)) = i$, then $\eta(a_0) = \tau$.

($\longrightarrow$). Let $\eta$ be a token mapping such that $\pi \models_\eta \mathcal{E}$. By Definition 2.10, this means that $\eta(a_i) = \tau_i$ where $\tau_i \in \pi(x_i)$ and $\mathsf{val}(\tau_i) = v_i$ for all $1 \le i \le n$, and $\lambda \models \mathcal{C}$ for the atomic evaluation $\lambda$ where $\lambda(a_i) = (\mathsf{start\text{-}time}(\tau_i), \mathsf{end\text{-}time}(\tau_i))$ for all $1 \le i \le n$. From such $\eta$ and $\lambda$ we can build a matching function $\gamma$ for $G_{\mathcal{E}}$ as follows. For each token name $a_i$ quantified as $a_i[x_i = v_i]$ in $\mathcal{E}$, let $\eta(a_i) = \tau_i$ be a token identified in $\overline{\mu}_\pi = \langle \mu_1, \ldots, \mu_n \rangle$ by the two events $\mu_j = (A_j, \delta_j)$ and $\mu_k = (A_k, \delta_k)$. In particular, $\mathsf{start}(x_i, v_i) \in A_j$ and $\mathsf{end}(x_i, v_i) \in A_k$. Then, for each node $T = \mathsf{start}(a_i) \in V$ (resp., $T = \mathsf{end}(a_i) \in V$), we define $\gamma(T) = j$ (resp., $\gamma(T) = k$). In this way, Items 1 and 2 of Definition 4.6 are satisfied by construction. For Item 5, consider any edge $e \in E$, corresponding to an atom $T \le_{[l,u]} T'$ in $\mathcal{C}$. Now, observe that by construction we have $[\![T]\!]_\lambda = \delta_{1,\gamma(T)}$ and $[\![T']\!]_\lambda = \delta_{1,\gamma(T')}$ and $\beta(e) = (l, u)$. Since $\lambda \models \mathcal{C}$, we have $l \le [\![T']\!]_\lambda - [\![T]\!]_\lambda \le u$, and it follows that $l \le \delta_{1,\gamma(T')} - \delta_{1,\gamma(T)} \le u$, which means $1 \le \delta_{\gamma(T),\gamma(T')} \le u$, hence Item 5 of Definition 4.6 is satisfied.

($\longleftarrow$). On the converse, let $\gamma$ be a matching function such that $\overline{\mu}, \gamma \models G_{\mathcal{E}}$. We can build the corresponding token mapping $\eta$ in the following way. For each node $T = \mathsf{start}(a_i) \in V$ (or $\mathsf{end}(a_i) \in V$, interchangeably), consider the corresponding event $\mu_{\gamma(T)} = (A_{\gamma(T)}, \delta_{\gamma(T)})$ in $\overline{\mu}_\pi$, which, since $\mathsf{start}(x_i, v_i) \in A_{\gamma(T)}$ (or $\mathsf{end}(x_i, v_i) \in A_{\gamma(T)}$), marks the start (or end) of a certain token $\tau \in \pi(x_i)$. Then, we define $\eta(a_i) = \tau$. It is easy to confirm that, in this way, $\pi \models_\eta \mathcal{E}$ as per Definition 2.10. First, $\tau \in \pi(x_i)$ and $\mathsf{val}(\tau) = v_i$ by construction. Then, consider the atomic evaluation $\lambda$ such that $\lambda(a_i) = (\mathsf{start\text{-}time}(\tau), \mathsf{end\text{-}time}(\tau))$. For each atom $T \le_{[l,u]} T'$ in $\mathcal{C}$, we know that $G_{\mathcal{E}}$ contains an edge $e = (T, T') \in E$ with $\beta(e) = (l, u)$. Since $\overline{\mu}_\pi, \gamma \models G_{\mathcal{E}}$, by definition we know that $l \le \delta_{\gamma(T),\gamma(T')} \le u$. Observe again that by construction we have $[\![T]\!]_\lambda = \delta_{1,\gamma(T)}$ and $[\![T']\!]_\lambda = \delta_{1,\gamma(T')}$, hence $l \le [\![T']\!]_\lambda - [\![T]\!]_\lambda \le u$, implying that $\lambda \models \mathcal{C}$. ∎

It is useful to set up some notation to talk about rules triggered by an event sequence. Given an event sequence $\overline{\mu} = \langle \mu_1, \ldots, \mu_n \rangle$ and a rule graph $G_{\mathcal{E}} = (V, E, \beta)$, we write $\overline{\mu}, \gamma \models_i G_{\mathcal{E}}$ to denote the fact that $\overline{\mu}, \gamma \models G_{\mathcal{E}}$ and $\mathsf{start}(a_0) \in V$ implies $\gamma(\mathsf{start}(a_0)) = i$, that is, that $G_{\mathcal{E}}$ matches over $\overline{\mu}$ in agreement with the token $\mu_i$ that triggered the corresponding rule. Note that only the index of the triggering event is specified, but there is no ambiguity over which token name we are constraining since the trigger name of the rule of each existential statement is uniquely identified. Consequently, we also write $\overline{\mu} \models_i G_{\mathcal{E}}$ if $\overline{\mu}, \gamma \models_i G_{\mathcal{E}}$ for some $\gamma$, and given a synchronisation rule $\mathcal{R} \equiv a_0[x_0 = v_0] \rightarrow \mathcal{E}_1 \vee \ldots \vee \mathcal{E}_m$, we write $\overline{\mu} \models_i \mathcal{R}$ if there is an $\mathcal{E}_k$ such that $\overline{\mu} \models_i G_{\mathcal{E}_k}$, and $\overline{\mu} \models \mathcal{R}$ if for any event $\mu_i$ in $\overline{\mu}$ that triggers $\mathcal{R}$, it holds that $\overline{\mu} \models_i \mathcal{R}$. Finally, given a timeline-based planning problem $P = (\mathsf{SV}, S)$, we write $\overline{\mu} \models P$ if $\overline{\mu} \models \mathcal{R}$ for any $\mathcal{R} \in S$. Thanks to Lemma 4.7 we can conclude that this notation is well-defined, since $\overline{\mu} \models P$ if and only if $\pi_{\overline{\mu}} \models P$. Considering our examples, note how the event sequence $\overline{\mu}$ from Figure 4.1 and the graph $G$ from Figure 4.2 are such that $\overline{\mu} \models_1 G$, since the rule is triggered (only) by $\mu_1$ and, as noted before, matches correctly with $\gamma(\mathsf{start}(a)) = 1$.

Intuitively, Lemma 4.7 tells us that rule graphs are a faithful representation of existential statements. Rule graphs, however, are easier to manipulate, decompose, and reason about, mainly for two reasons: on one hand, because their graph structure allows us to reuse all the terminology and intuitive understanding of graphs, and on the other hand, the distinction of the endpoints of tokens, which are a single entity, into two different nodes of the graph, which can be manipulated separately, will be essential in our argument.

A first result that we can prove by reasoning on rule graphs is an upper bound on the distance between two consecutive events in an event sequence.

**Lemma 4.8** — Upper bound on the distance between events.
*Let $P = (\mathsf{SV}, S)$ be a timeline-based planning problem. Let $L$ and $U$ be the* maximum *lower and (finite) upper bounds of any edge in any rule graph of $P$, and let $d = \max(L, U) + 1$. If there is any event sequence $\overline{\mu}$ such that $\overline{\mu} \models P$, then there exists another event sequence $\overline{\mu}'$ such that $\overline{\mu}' \models P$ and the distance between two consecutive events of $\overline{\mu}'$ is at most $d$.*

*Proof.* Consider $\overline{\mu} = \langle \mu_1, \ldots, \mu_n \rangle$ and any pair of subsequent events $\mu_i$ and $\mu_{i+1}$ such that $\delta_{i+1} > d$ (we do not consider $\delta_1$, since, as said before, we can ignore its value). We define $\overline{\mu}' = \langle \mu'_1, \ldots, \mu'_n \rangle$ to be an event sequence equal to $\overline{\mu}$ excepting for the fact that $\delta'_{i+1} = d$. Observe that such edit does not change which rules are triggered in $\overline{\mu}'$, since the triggers are not affected by the duration of tokens. Now, for any time any rule $\mathcal{R} \in S$ is triggered in $\overline{\mu}$, there is an existential statement $\mathcal{E}$ of $\mathcal{R}$ and a matching function $\gamma$ such that $\overline{\mu}, \gamma \models G_{\mathcal{E}}$, where $G_{\mathcal{E}} = (V, E, \beta)$ is the rule graph of $\mathcal{E}$ (by Definition 2.11 and Lemma 4.7). We show that it also holds that $\overline{\mu}', \gamma \models G_{\mathcal{E}}$. To see this, suppose that there are nodes $T, T' \in V$ such that $\gamma(T) = i$ and $\gamma(T') = i{+}1$, and observe that there cannot be any edge $e = (T, T') \in E$ such that $\beta(e) = (l, u)$ with $u \neq +\infty$, otherwise we would have $\delta_{\gamma(T),\gamma(T')} > u$, violating the matching. Then, for any such edge it must be $u = +\infty$, then it still matches correctly with $\delta'_{i+1} = d$. Let $\beta(e) = (l, +\infty)$. Given that $l \leq d$ by definition, it still holds that $l \leq \delta_{i,i+1}$ as required to match the edge. Hence $\gamma$ is still a suitable matching function such that $\overline{\mu}', \gamma \models G_{\mathcal{E}}$, hence $\overline{\mu}' \models P$ as stated. ∎

What follows introduces some concepts and tools to reason about rule graphs. The next section makes use of them to carry on our complexity analysis.

## 2.3 BOUNDS ON RULE GRAPHS PATHS

It is useful to set some terminology regarding edges and paths of rule graphs. Let $G = (V, E, \beta)$ be a rule graph. An edge $e \in E$ is *unbounded* if $\beta(e) = (l, u)$ with $u = +\infty$, and is *bounded* otherwise. A rule graph is *bounded* if it is made of bounded edges only. A path $\overline{T} = \langle T_1, \dots, T_k \rangle$, where $(T_i, T_{i+1}) \in E$ for all $1 \leq i < k$, is said to be *bounded* if it is made only of bounded edges.

It is useful to consider also *undirected* paths, that is, sequences of nodes of $G$ which form a path if we ignore the direction of edges. Formally, an undirected path is a sequence of nodes $\overline{T} = \langle T_1, \dots, T_k \rangle$ such that either $(T_i, T_{i+1}) \in E$ or $(T_{i+1}, T_i) \in E$ for all $1 \leq i < k$. Note that, of course, a path of the graph is in particular also an undirected path. An *undirected edge* is an undirected path $\overline{T} = \langle T_1, T_2 \rangle$ of only two nodes. To distinguish the direction of an undirected edge $\overline{T} = \langle T_1, T_2 \rangle$ with respect to the real edge found in the graph, we define $\mathsf{sign}(\overline{T}) = +1$ if $e = (T_1, T_2) \in E$, and $\mathsf{sign}(\overline{T}) = -1$ if $e = (T_2, T_1) \in E$. We also extend the notation for the edge bounds by defining $\beta(\overline{T}) = \beta(e)$.

If two nodes are connected by an undirected path, we can compute the minimum and maximum distance between their corresponding events in any event sequence, generalising and combining the bounds of the single edges that form the path.

**Definition 4.9** — Lower and upper bounds on undirected paths.
*Let $G = (V, E, \beta)$ be a rule graph, and let $\overline{T}_k = \langle T_1, \dots, T_k \rangle$ be an* undirected *path of $G$. Then, let $\mathsf{lbound}(\overline{T}) \in \mathbb{Z} \cup \{-\infty\}$ and $\mathsf{ubound}(\overline{T}) \in \mathbb{Z} \cup \{+\infty\}$ be the two quantities recursively defined as follows:*

1. *if $k = 1$, $\mathsf{lbound}(\langle T_1 \rangle) = 0$ and $\mathsf{ubound}(\langle T_1 \rangle) = 0$;*
2. *if $k = 2$, given $\overline{T}_2 = \langle T_1, T_2 \rangle$, with $\beta(\overline{T}_2) = (l, u)$:*
$$\mathsf{lbound}(\overline{T}_2) = \min(\mathsf{sign}(\overline{T}_2) \cdot l, \mathsf{sign}(\overline{T}_2) \cdot u)$$
$$\mathsf{ubound}(\overline{T}_2) = \max(\mathsf{sign}(\overline{T}_2) \cdot l, \mathsf{sign}(\overline{T}_2) \cdot u)$$
3. *if $k > 2$, given $\overline{T}_{k-1} = \langle T_1, \dots, T_{k-1} \rangle$:*
$$\mathsf{lbound}(\overline{T}_k) = \mathsf{lbound}(\overline{T}_{k-1}) + \mathsf{lbound}(\langle T_k, T_{k+1} \rangle)$$
$$\mathsf{ubound}(\overline{T}_k) = \mathsf{ubound}(\overline{T}_{k-1}) + \mathsf{ubound}(\langle T_k, T_{k+1} \rangle)$$

*where, in the above context, it holds that $x \pm \infty = \pm\infty$, $x < +\infty$, and $x > -\infty$ for all $x \in \mathbb{Z}$, $-\infty < +\infty$, $\pm 1 \cdot +\infty = \pm\infty$, $+\infty + \infty = +\infty$ and $-\infty - \infty = -\infty$.*

**Lemma 4.10** — Correctness of the bounds.
*Let $G = (V, E, \beta)$ be a rule graph, and let $\overline{T} = \langle T_1, \dots, T_k \rangle$ be an* undirected *path of $G$. Then, for any event sequence $\overline{\mu}$ and any matching function $\gamma$ such that $\overline{\mu}, \gamma \models G$, it holds that $\mathsf{lbound}(\overline{T}) \leq \gamma(T_k) - \gamma(T_1) \leq \mathsf{ubound}(\overline{T})$.*

*Proof.* The proof goes by induction on the length of $\overline{T} = \langle T_1, \ldots, T_k \rangle$, following the structure of Definition 4.9. If $k = 1$, with $\mathsf{lbound}(\overline{T}) = 0$ and $\mathsf{ubound}(\overline{T}) = 0$, the thesis trivially holds. If $k = 2$, we have an undirected edge $\overline{T} = \langle T_1, T_2 \rangle$. Then, either $\overline{T}$ agrees with the direction of the edge in the graph, *i.e.*, $e = (T_1, T_2) \in E$, or not, *i.e.*, $e = (T_2, T_1) \in E$. In the first case, by definition, $\mathsf{lbound}(\overline{T}) = l$ and $\mathsf{ubound}(\overline{T}) = u$, where $\beta(\overline{T}) = (l, u)$. In the second case, the definition sets $\mathsf{lbound}(\overline{T}) = -u$ and $\mathsf{ubound}(\overline{T}) = -l$. In either case, the thesis follows immediately from Definition 4.6. Note that if the edge is unbounded, then either $\mathsf{ubound}(\overline{T}) = +\infty$ or $\mathsf{lbound}(\overline{T}) = -\infty$ depending on the direction, but in either case, the inequalities still hold, following the definition of matching of unbounded edges. Otherwise, if $k > 2$, consider any event sequence $\overline{\mu}$ and any $\gamma$ such that $\overline{\mu}, \gamma \models G$. Since $\mathsf{lbound}(\overline{T}_{k-1}) \leq \gamma(T_{k-1}) - \gamma(T_1)$ and $\mathsf{lbound}(\langle T_{k-1}, T_k \rangle) \leq \gamma(T_k) - \gamma(T_{k-1})$ by the induction hypothesis, it follows that $\mathsf{lbound}(\overline{T}_k) \leq \gamma(T_k) - \gamma(T_1)$. It holds also in the case that $\mathsf{lbound}(\langle T_{k-1}, T_k \rangle) = -\infty$, since in this case it follows that $\mathsf{lbound}(\overline{T}_k) = -\infty$ and this is coherent with the addition of an unbounded edge between $T_{k-1}$ and $T_k$, in the opposite direction relative to the path. In a similar way, the stated inequality holds for $\mathsf{ubound}(\overline{T}_k)$. ■

Intuitively, given an undirected path $\overline{T} = \langle T_1, \ldots, T_k \rangle$, the bounds $\mathsf{lbound}(\overline{T})$ and $\mathsf{ubound}(\overline{T})$ are the minimum and maximum *signed distance* between the two endpoints of the path, generalising the bounds given on the single edges to whole undirected paths. Their value is positive if the edges involved in the path constrain $T_1$ to be mapped before $T_k$, with the given bounds on their distance, and is negative if the terms are constrained to be mapped in the reverse order. If they have opposite signs, it means that both orders are possible, although still respecting the given distance constraints. The two bounds $\mathsf{lbound}(\overline{T})$ and $\mathsf{ubound}(\overline{T})$ also provide the maximum absolute distance between the endpoints of the path, denoted as $d_{max}(\overline{T}) = \max(|\mathsf{lbound}(\overline{T})|, |\mathsf{ubound}(\overline{T})|)$, which is well-defined only if both bounds are finite. We also define $d_{max}(T, T')$, for two nodes $T$ and $T'$, as the minimum $d_{max}(\overline{T})$ among all undirected paths $\overline{T}$ connecting $T$ and $T'$. That is, $d_{max}(T, T')$ is the maximum distance the two nodes can have when mapped on any event sequence.

With these concepts at hand, rule graphs can often be simplified to obtain smaller ones, with a smaller number of edges or with smaller bounds on their edges, while preserving the same matchings over any event sequence.

■ **Lemma 4.11** — Simplification of undirected paths.
*Any timeline-based planning problem $P = (\mathsf{SV}, S)$ can be translated, in polynomial time, into an equivalent one $P' = (\mathsf{SV}, S')$ such that, for each existential statement $\mathcal{E}$ of any rule $\mathcal{R} \in S$, if its rule graph $G_{\mathcal{E}} = (V, E, \beta)$ contains an* undirected *path $\overline{T} = \langle T_1, \ldots, T_k \rangle$ and an edge $e = (T_1, T_k) \in E$, then:*

1. $\mathsf{ubound}(\overline{T}) \geq l$ *and* $\mathsf{lbound}(\overline{T}) \leq u$, *where* $\beta(e) = (l, u)$*;*

2. *$l$ and $u$ are not both zero;*

3. $\mathsf{lbound}(\overline{T}) \le l \le u \le \mathsf{ubound}(\overline{T})$, *but either* $\mathsf{lbound}(\overline{T}) \neq l$ *or* $\mathsf{ubound}(\overline{T}) \neq u$.

*Proof.* Consider any event sequence $\overline{\mu}$ and any $\gamma$ such that $\overline{\mu}, \gamma \models G_{\mathcal{E}}$, and suppose there is an undirected path $\overline{T} = \langle T_1, \ldots, T_k \rangle$ and an edge $e = (T_1, T_k) \in E$. To prove Item 1, suppose by contradiction that $\mathsf{ubound}(\overline{T}) < l$, or $\mathsf{lbound}(\overline{T}) > u$, where $\beta(e) = (l, u)$. This assumption directly conflicts with the fact that $l \le \gamma(T_k) - \gamma(T_1) \le \mathsf{ubound}(\overline{T})$ and $\mathsf{lbound}(\overline{T}) \le \gamma(T_k) - \gamma(T_1) \le u$, which follows from Lemma 4.10 and Definition 4.6. Hence, either both $\mathsf{ubound}(\overline{T}) \le l$ and $\mathsf{lbound}(\overline{T}) \le u$, or there cannot be any such $\gamma$, *i.e.*, $\mathcal{E}$ is *unsatisfiable*. In this case, it can be removed from $\mathcal{R}$ obtaining an equivalent timeline-based planning problem. To prove Item 2, note that if $l = u = 0$, it means the only possible satisfying choice is to have $\gamma(T_k) = \gamma(T_1)$, *i.e.*, the two terms have to be mapped to the same event. In this case, we can express the same constraint by changing to zero the lower and upper bounds of each edge of $\overline{T}$, and removing $e$. Note that this creates new edges $e'$ with $l' = u' = 0$, but the total number of edges strictly decreases, hence the operation can be repeated only a finite number of times. As for Item 3, suppose that $\mathsf{lbound}(\overline{T}) > l$. Then, we have that the constraint $l \le \gamma(T_k) - \gamma(T_1)$ given by the edge is implied by the fact that $\mathsf{lbound}(\overline{T}) \le \gamma(T_k) - \gamma(T_1)$, which follows from $\overline{T}$. Then, the lower bound $l$ can be increased up to $\mathsf{lbound}(\overline{T})$, without changing how the graph can match over $\overline{\mu}$, *i.e.*, we can obtain an equivalent problem by setting $\beta(e) = (\mathsf{lbound}(\overline{T}), u)$. A similar argument ensures that we can, if needed, decrease the value of $u$ to ensure that $\mathsf{ubound}(\overline{T}) \ge u$. Hence, we have $\mathsf{lbound}(\overline{T}) \le l \le u \le \mathsf{ubound}(\overline{T})$. Now, suppose $\mathsf{lbound}(\overline{T}) = l$ and $\mathsf{ubound}(\overline{T}) = u$. This means that the edge imposes exactly the same constraint $\mathsf{lbound}(\overline{T}) \le \gamma(T_k) - \gamma(T_1) \le \mathsf{ubound}(\overline{T})$ imposed by the path. In this case, therefore, the edge is redundant and can be removed from the graph. It is easy to verify how any rule graph can be traversed to apply these simplifications in polynomial time. ■

Lemma 4.11 has some useful consequences, which will be exploited as needed. The first one is that we can assume *w.l.o.g.* the acyclicity of all the considered rule graphs.

■ **Lemma 4.12** — Acyclicity of rule graphs.
*Any timeline-based planning problem* $P = (\mathsf{SV}, S)$ *can be translated, in polynomial time, into an equivalent one* $P' = (\mathsf{SV}, S')$ *such that, for each existential statement* $\mathcal{E}$ *of any rule* $\mathcal{R} \in S$, *its rule graph* $G_{\mathcal{E}}$ *is acyclic.*

*Proof.* At first, take any timeline-based planning problem and translate it as stated in Lemma 4.11. Then, suppose by contradiction that a cycle is present in any rule graph $G_{\mathcal{E}}$ of the translated problem. The cycle can be seen as made of a path $\overline{T} = \langle T_k, \ldots, T_1 \rangle$ and an edge $e = (T_1, T_k)$ with $\beta(e) = (l, u)$. Since the edge goes from $T_1$ to $T_k$, if we look at the path as an undirected path $\overline{T} = \langle T_1, \ldots, T_k \rangle$, the upper bound of the path results in $\mathsf{ubound}(\overline{T}) = -\sum_{1\le i\le k-1} l_i$, where $l_i$ is the lower bound of the edge $e_i = (T_{i+1}, T_i)$. All such lower bounds are non-negative and $l \ge 0$ by definition, therefore $\mathsf{ubound}(\overline{T}) \le 0 \le l$. Moreover, $\mathsf{ubound}(\overline{T}) \ge l$ by Item 1 of Lemma 4.11, hence the only possible consequence is that $l = u = \mathsf{ubound}(\overline{T}) = 0$, but this cannot be because of Item 2 of Lemma 4.11. ■

Hence, from now on we will suppose *w.l.o.g.* that all the considered timeline-based planning problems lead to acyclic rule graphs where, in general, the conditions described by Lemma 4.11 hold.

## 2.4 DECOMPOSING RULE GRAPHS

Most of the next section will deal with rule graphs and their *subgraphs*. Given the rule graph $G_{\mathcal{E}} = (V, E, \beta)$ of an existential statement $\mathcal{E}$, a *subgraph* of $G_{\mathcal{E}}$ is a graph $G' = (V', E', \beta')$ such that $V' \subseteq V$, $E' \subseteq E$, and $\beta' = \beta|_{E'}$. We write $G' \sqsubseteq G_{\mathcal{E}}$. A trivial subgraph of any rule graph is the empty graph $G_{\varnothing} = (\varnothing, \varnothing, \varnothing)$. It is worth noting that a subgraph $G'$ of a rule graph $G_{\mathcal{E}}$ is still a proper rule graph, and can be associated with some existential statement $\mathcal{E}'$ such that $G' = G_{\mathcal{E}'}$, albeit probably not part of the considered planning problem. Hence, anything that can be said on rule graphs holds directly on their subgraphs, which are proper rule graphs themselves.

Note that the existential statement associated with a subgraph of a rule graph may not be verbose, *i.e.*, a subgraph may miss the $\mathsf{start}(a)$ or the $\mathsf{end}(a)$ nodes for some token name $a$. For this reason, it is worth to note that subgraphs of rule graphs may not match on the same event sequences, since it can be that $\overline{\mu} \models G_{\mathcal{E}}$ but $\overline{\mu} \not\models G_{\mathcal{E}'}$ for some $G_{\mathcal{E}'} \sqsubseteq G_{\mathcal{E}}$. That is because of Items 3 and 4 of Definition 4.6, which forbid a graph that misses start/end nodes to match if the corresponding tokens are not open in the event sequence. This is why we assumed to start with verbose rules in the first place, otherwise, the rule graphs of non-verbose existential statements would not ever be able to match on closed event sequences.

Moreover, note that a (directed) path $\overline{T} = \langle T_1, \ldots, T_k \rangle$ can be regarded as a subgraph of $G$, hence we can write $\overline{T} \sqsubseteq G$. Similarly, single nodes can be seen as simple subgraphs as well, hence we can write $T \sqsubseteq G$ when $T \in V$. Finally, if a rule graph $G = (V, E, \beta)$ matches on $\overline{\mu}$ with some matching function $\gamma$, we denote as $\overline{\mu}|_G^{\gamma}$ the subsequence of $\overline{\mu}$ covered by $G$, *i.e.*, $\overline{\mu}|_G^{\gamma} = \overline{\mu}_{[i\ldots j]}$ where $i = \mathsf{min}_{T\in V}(\gamma(T))$ and $j = \mathsf{max}_{T\in V}(\gamma(T))$.

A particular kind of subgraphs will play an important role in what follows.

■ **Definition 4.13** — Bounded components of rule graphs.
*Let $G_{\mathcal{E}}$ be the rule graph of some existential statement $\mathcal{E}$. A* bounded component *$B = (V_B, E_B, \beta_B)$ is a* maximal *subgraph of $G_{\mathcal{E}}$ where each node $T \in V_B$ can reach any other $T' \in V_B$ through a* bounded undirected path.

In other words, bounded components of a rule graph are maximal subgraphs connected by bounded edges. A bounded component $B \sqsubseteq G_{\mathcal{E}}$ is the *trigger component* of $\mathcal{E}$ if it contains the term $\mathsf{start}(a_0)$ where $a_0$ is the trigger token name of $\mathcal{E}$.

Suppose $\overline{\mu}, \gamma \models B$ for some event sequence $\overline{\mu}$ and some bounded component $B \sqsubseteq G_{\mathcal{E}}$. Since any two nodes $T$ and $T'$ of $B$ are connected by a undirected path $\overline{T} = \langle T, \ldots, T' \rangle$, we know $\mathsf{lbound}(\overline{T}) \leq \gamma(T') - \gamma(T) \leq \mathsf{ubound}(\overline{T})$, and, since $\overline{T}$ is *bounded*, we know $\mathsf{lbound}(\overline{T})$ and $\mathsf{ubound}(\overline{T})$ are *finite*. Moreover, Lemma 4.11 has the interesting consequence that a bounded component cannot contain any *unbounded* edge: since any two nodes $T$ and $T'$ are connected by a *bounded* undirected path $\overline{T} = \langle T, \ldots, T' \rangle$, the presence of an unbounded edge, either $e = (T, T') \in E$ or $e = (T', T) \in E$, would contradict either Item 1 or Item 3 of Lemma 4.11, depending on the sign of $\mathsf{lbound}(\overline{T})$ and $\mathsf{ubound}(\overline{T})$.

These observations allow us to define in the following way the maximum distance between any event involved in the matching of a bounded component.

**Definition 4.14** — Window of a bounded rule graph.
*Let $B \sqsubseteq G_{\mathcal{E}}$ be a bounded component of a rule graph $G_{\mathcal{E}}$. The* window *of $B$ is the quantity defined as:*

$$\mathsf{window}(B) = \max_{T, T' \sqsubseteq B} d_{max}(T, T')$$

In other words, the *window* of a bounded component $B$ is the maximum amount of time that can elapse between two events that are involved in the matching of $B$ over any event sequence. More precisely, this means that $\delta(\overline{\mu}|_B^{\gamma}) \leq \mathsf{window}(B)$ for any $\overline{\mu}$ and $\gamma$ such that $\overline{\mu}, \gamma \models B$. It is worth to note that this definition is well-defined thanks to the fact that no unbounded edge can be included in $B$, as observed above, and thus all the undirected paths connecting the nodes of $B$ have finite bounds.

If $\overline{B} = \{B_0, \ldots, B_n\}$ are the the bounded components of a rule graph $G_{\mathcal{E}}$, we can define the *window* of $G_{\mathcal{E}}$ as $\mathsf{window}(G_{\mathcal{E}}) = \sum_{0 \leq i \leq n} \mathsf{window}(B_i)$. The window of a rule graph does *not* give a bound on the distance of nodes of the graph, which is not possible because of unbounded edges, but gives us a bound on the size of any set of bounded components that *overlap* with each other.

**Lemma 4.15** — Matching gap.
*Let $G_{\mathcal{E}}$ be a rule graph, and let $\overline{\mu}$ be an event sequence such that $\overline{\mu}, \gamma \models G_{\mathcal{E}}$ for some $\gamma$. If $\delta(\overline{\mu}|_{G_{\mathcal{E}}}^{\gamma}) > \mathsf{window}(G_{\mathcal{E}})$, then there is a position $k$ in $\overline{\mu}|_{G_{\mathcal{E}}}^{\gamma}$ that is not covered by any* bounded *edge,* i.e., *there are no $T, T' \sqsubseteq G_{\mathcal{E}}$ such that the edge $(T, T') \in E$ is bounded and $\gamma(T) \leq k$ and $\gamma(T) > k$.*

*Proof.* For each bounded component $B_i \sqsubseteq G_{\mathcal{E}}$, let $s_i$ and $e_i$ be two positions such that $\overline{\mu}|_{B_i}^{\gamma} = \overline{\mu}_{[s_i \ldots e_i]}$, and order the components in a sequence $\overline{B} = \langle B_0, \ldots, B_n \rangle$ such that $s_i < s_{i+1}$ for all $0 \leq i < n$. If there is no position $k$ not covered by any unbounded edge, then $\delta(\overline{\mu}_{[s_i \ldots s_{i+1}]}) \leq \mathsf{window}(B_i)$ for all $0 \leq i < n$, hence $\delta(\overline{\mu}|^{\gamma}_{G_{\mathcal{E}}}) \leq \sum_{0 \leq i \leq n} \mathsf{window}(B_i) = \mathsf{window}(G_{\mathcal{E}})$. ■

Intuitively, $\mathsf{window}(G_{\mathcal{E}})$ gives us a sufficiently large size where we can find groups of components matching together, while being able to correctly verify the satisfaction of all the edges that connect them. Extending this consideration to any rule graph of a problem $P$, we denote $\mathsf{window}(P)$ as the maximum $\mathsf{window}(G_{\mathcal{E}})$ for all the existential statements $\mathcal{E}$ of all the rules of a problem $P$.

A consequence of their definition is that any two bounded components $B, B' \sqsubseteq G_{\mathcal{E}}$ of a rule graph are *disjoint*, *i.e.*, they do not have nodes in common. In particular, we can observe that the sets of nodes of the bounded components of a rule graph $G_{\mathcal{E}}$ forms a partition of all the nodes of $G_{\mathcal{E}}$, and that any unbounded edge of $G_{\mathcal{E}}$ must connect two different bounded components. Hence, by merging as follows all the components of a rule graph $G_{\mathcal{E}}$, and adding back all the unbounded edges between them, one gets back the whole $G_{\mathcal{E}}$.

■ **Definition 4.16** — Concatenation of subgraphs.
*Let $G_{\mathcal{E}'} = (V', E', \beta')$ and $G_{\mathcal{E}''} = (V'', E'', \beta'')$ be two* disjoint *subgraphs of a rule graph $G_{\mathcal{E}} = (V, E, \beta)$. The* concatenation *of $G_{\mathcal{E}'}$ and $G_{\mathcal{E}''}$, written $G_{\mathcal{E}'} \rightrightarrows G_{\mathcal{E}''}$, is a subgraph $(V_{\rightrightarrows}, E_{\rightrightarrows}, \beta_{\rightrightarrows})$ of $G_{\mathcal{E}}$, where:*

1. $V_{\rightrightarrows} = V' \cup V''$;
2. $E_{\rightrightarrows} = E' \cup E'' \cup (E \cap (V' \times V''))$, i.e., *the edges going from $G_{\mathcal{E}'}$ to $G_{\mathcal{E}''}$ are added;*
3. $\beta_{\rightrightarrows} = \beta|_{E_{\rightrightarrows}}$.

In other words, the concatenation of two subgraphs is the set-theoretic union of the two, that we can denote as $G_{\mathcal{E}'} \cup G_{\mathcal{E}''}$, plus all the edges that connect one to the other. It holds that $G_{\mathcal{E}'} \cup G_{\mathcal{E}''} \sqsubseteq G_{\mathcal{E}'} \rightrightarrows G_{\mathcal{E}''}$ but in general not *vice versa*. Note that the definition considers only the edges that go from one subgraph to the other. As it turns out, in the case of a rule graph decomposed into its bounded components, connecting only edges that go in one direction is sufficient to reconstruct the whole rule graph.

■ **Lemma 4.17** — Decomposition into concatenation of bounded components.
*Any rule graph $G_{\mathcal{E}}$ can be decomposed into a sequence of bounded components $\overline{B} = \langle B_1, \ldots, B_n \rangle$ such that $G_{\mathcal{E}} = B_1 \rightrightarrows B_2 \rightrightarrows \ldots \rightrightarrows B_n$.*

*Proof.* Consider the directed graph $\mathcal{B} = (V, E)$ such that $V$ is the set of bounded components of $G_{\mathcal{E}}$ and, for any two bounded components $B, B' \sqsubseteq G_{\mathcal{E}}$, $(B, B') \in E$ if and only if $B \neq B'$ and there is an edge (unbounded, as we know) from any node in $B$ to any node in $B'$. In other words $\mathcal{B}$ is the graph obtained by collapsing on a single node each bounded component, or, equivalently, collapsing any bounded edge of $G_{\mathcal{E}}$. Now, it is sufficient to show that $\mathcal{B}$ is acyclic. In this way, any topological ordering of $\mathcal{B}$ provides a sequence $\overline{B} = \langle B_1, \ldots, B_n \rangle$ of bounded components such that any edge between any two of them goes from $B_i$ to $B_j$ with $j > i$, and thus their concatenation $B_1 \rightrightarrows \cdots \rightrightarrows B_n$ corresponds to the whole $G_{\mathcal{E}}$. To prove the acyclicity of $\mathcal{B}$, at first note that it cannot contain self-loops by definition. Then, suppose by contradiction that it contains a cycle $\overline{B} = \langle B^k, \ldots, B^1 \rangle$ where $(B^{i+1}, B^i) \in E$ and $(B^1, B^k) \in E$. Consider the unbounded edge $e = (T_1, T_k)$ of $G_{\mathcal{E}}$, with $\beta(e) = (l, u)$, that connects the node $T_1$ of $B^1$ to the corresponding node $T_k$ of $B^k$. Then, $\overline{B}$ identifies one (but possibly many) undirected path $\overline{T} = \langle T_1, \ldots, T_k \rangle$ in $G_{\mathcal{E}}$, that connect $T_k$ to $T_1$ by traversing all the bounded components of $\overline{B}$ passing through the unbounded edges that form the cycle. We know that all the unbounded edges of $\overline{T}$ go in the same direction, from $T_{i+1}$ to $T_i$ for some $i$, because $\overline{B}$ is a cycle in $\mathcal{B}$. Hence, by Definition 4.9 it follows that $\mathsf{lbound}(\overline{T}) = -\infty$, but $\mathsf{ubound}(\overline{T})$ is finite. However, since $u = +\infty$, we have $\mathsf{ubound}(\overline{T}) < u$, which contradicts Item 3 of Lemma 4.11. ■

Thanks to Lemma 4.17 we can extend the partial order between the nodes of the graph to a pre-order between its bounded components. The graph concatenation operation has a strict connection with the matching relation of rule graphs over event sequences.

■ **Lemma 4.18** — Matching of graph concatenations.
*Let $G_{\mathcal{E}'}$ and $G_{\mathcal{E}''}$ be two disjoint subgraphs of $G_{\mathcal{E}}$, and suppose that the lower bound of any* unbounded *edge in $G_{\mathcal{E}}$ is zero. Let $\overline{\mu}$ be an event sequence such that $\overline{\mu}_{[i\ldots j]} \models G_{\mathcal{E}'}$ and $\overline{\mu}_{[j\ldots k]} \models G_{\mathcal{E}''}$, with $i \leq j \leq k$. Then, $\overline{\mu}_{[i\ldots k]} \models G_{\mathcal{E}'} \rightrightarrows G_{\mathcal{E}''}$.*

*Proof.* Let $\gamma'$ and $\gamma''$ be two matching functions such that $\overline{\mu}_{[i\ldots j]}, \gamma' \models G_{\mathcal{E}'}$ and $\overline{\mu}_{[i\ldots j]}, \gamma'' \models G_{\mathcal{E}''}$. We can define a matching function $\gamma$ such that $\gamma(T) = \gamma'(T)$ if $T \sqsubseteq G_{\mathcal{E}'}$ and $\gamma(T) = \gamma''(T)$ if $T \sqsubseteq G_{\mathcal{E}''}$ (this is well-defined since the two subgraphs are disjoint). Let us now check that $\overline{\mu}_{[i\ldots k]}, \gamma \models G_{\mathcal{E}'} \rightrightarrows G_{\mathcal{E}''}$. We already observed that, qualitatively, the fact that $G_{\mathcal{E}''}$ matches completely later than $G_{\mathcal{E}'}$ ensures that any unbounded edge added between the two is satisfied. Moreover, by Items 3 and 4 of Definition 4.6, we know that any two nodes $\mathsf{start}(a) \sqsubseteq G_{\mathcal{E}'}$ and $\mathsf{end}(a) \sqsubseteq G_{\mathcal{E}''}$ will be mapped in such a way that $\gamma(\mathsf{start}(a))$ and $\gamma(\mathsf{end}(a))$ correctly identify a proper token. ■

The assumption that the lower bounds in unbounded edges are zero made in the statement of Lemma 4.18 is essential to make it work as-is. For ease of exposition, in the following section we suppose that this restriction holds for any considered rule graph. However, the whole argument can be adapted easily to remove this restriction by taking care of such lower bounds when reasoning about the concatenation of subgraphs.

# 3 COMPLEXITY OF TIMELINE-BASED PLANNING

In this section we can leverage the conceptual framework of rule graphs defined previously to analyse the computational complexity of the plan existence problem for timeline-based planning. In particular, we prove that the problem is EXPSPACE-complete. Note, that EXPSPACE-hardness comes directly from the encoding of action-based temporal planning problems shown by Corollary 3.10 in Chapter 3. Hence, it is sufficient here to provide a decision procedure that can solve the problem using at most exponential space.

The section starts by providing a *small-model theorem*, that is, a result showing that any satisfiable timeline-based planning problem has a solution shorter than a given upper bound. Then, the decision procedure will be provided, whose complexity depends on said bound.

## 3.1 THE SMALL-MODEL THEOREM

The upper bound will be proved with a fairly standard contraction argument. We will show that, when a solution longer than the bound is found, then another shorter one can be built by suitably contracting it between specific points. Similar techniques are often used to prove upper bounds on the length of models for various temporal logics, *e.g.* [127].

The key step behind this kind of contraction arguments is the identification of some *local* state, *i.e.*, an object that can be defined at a specific point of the solution, that can finitely represent the current state of the system at that point. A long solution can then be cut between two repetitions of the same state. In the case of timeline-based planning problems, the current state cannot be represented only by the current value of state variables. Instead, apparently, the whole plan contributes to the current state of the system because any rule triggered anywhere can be satisfied by looking at something happened arbitrarily far in the past or that will happen arbitrarily far in the future. However, thanks to the decomposition of rule graphs into *bounded components*, such information can be finitely and compactly represented in a data structure, that we call *matching record*, which has size exponential in the size of the problem and uniquely identifies the state of the system.

**Definition 4.19** — Matching record.
*Let $P = (\mathsf{SV}, S)$ be a timeline-based planning problem and let $\overline{\mu} = \langle \mu_1, \ldots, \mu_n \rangle$ be an event sequence over* SV *closed to the left such that $\delta(\overline{\mu}) \geq 2\,\mathsf{window}(P)$.*

*The* matching record *of $\overline{\mu}$ is a tuple $[\overline{\mu}] = (\overline{\omega}, \Gamma, \Delta)$, where:*

1. *$\overline{\omega}$ is the shortest suffix $\overline{\mu}_{\geq h}$ of $\overline{\mu}$ that can be split into two subsequences spanning at least* window$(P)$ *time steps,* i.e., *$\overline{\omega} = \overline{\omega}_{-}\overline{\omega}_{+}$, where $\overline{\omega}_{-} = \overline{\mu}_{[h \ldots h_{+}-1]}$, $\overline{\omega}_{+} = \overline{\mu}_{\geq h_{+}}$, and both $\delta(\overline{\omega}_{-}) \geq \mathsf{window}(P)$ and $\delta(\overline{\omega}_{+}) \geq \mathsf{window}(P)$;*

2. $\Gamma$ *is a function that maps any* existential statement $\mathcal{E}$ *of any* $\mathcal{R} \in S$ *and a* $1 \leq k \leq |\overline{\omega}_-|$, *to the* maximal *subgraph* $\Gamma(\mathcal{E}, k)$ *of* $G_\mathcal{E}$ *such that:*

   (a) $\overline{\mu}_{\leq h+k}, \gamma \models \Gamma(\mathcal{E}, k)$ *for some matching function* $\gamma$*;*

   (b) $\Gamma(\mathcal{E}, k)$ *does not contain the trigger node* $\mathsf{start}(a_0)$*;*

   (c) *any edge going out of* $\Gamma(\mathcal{E}, k)$ *is* unbounded*;*

3. $\Delta$ *is a function that maps any rule* $\mathcal{R} \in S$ *to a set of* instances, *where an instance is a function* $I$ *that maps an* existential statement $\mathcal{E}$ *and a* $1 \leq k \leq |\overline{\omega}_-|$ *to a subgraph* $I(\mathcal{E}, k)$ *of* $G_\mathcal{E}$.

   $\Delta$ *is such that for each event* $\mu_i$ *in* $\overline{\mu}_{<h_+}$ *that triggers a rule* $\mathcal{R} \in S$, *there is an instance* $I \in \Delta(\mathcal{R})$ *such that, for each existential statement* $\mathcal{E}$ *and each* $1 \leq k \leq |\overline{\omega}_-|$, $I(\mathcal{E}, k)$ *is the* maximal *subgraph* $I(\mathcal{E}, k) \sqsubseteq G_\mathcal{E}$ *such that:*

   (a) $\overline{\mu}_{\leq h+k}, \gamma \models_i I(\mathcal{E}, k)$ *for some matching function* $\gamma$*;*

   (b) *any edge going out of* $I(\mathcal{E}, k)$ *is* unbounded.

*If instead* $\overline{\mu}$ *is empty, made of only one event, or* $\delta(\overline{\mu}) < 2\,\mathsf{window}(P)$, *then* $[\overline{\mu}] = \overline{\mu}$.

Intuitively, the matching record $[\overline{\mu}] = (\overline{\omega}, \Gamma, \Delta)$ of a long enough event sequence $\overline{\mu}$ stores three pieces of information: the recent history $\overline{\omega}$ of the plan, *i.e.*, a suffix of the event sequence that is worth remembering in detail, a record $\Gamma$ of all the pieces of rule graphs that matched in the past, and the record $\Delta$ of pieces of rule graphs that still have to be matched in the future to satisfy some previously triggered rule. The following result confirms the role of matching records: when an event sequence is represented by its matching record, we have sufficient information to decide whether the event sequence satisfies or not the given timeline-based planning problem.

**Lemma 4.20** — Matching record of a solution plan.
*Let* $P = (\mathsf{SV}, S)$ *be a timeline-based planning problem, and let* $[\overline{\mu}] = (\overline{\omega}, \Gamma, \Delta)$ *be the matching record of an event sequence* $\overline{\mu}$ *over* $\mathsf{SV}$.

*Then,* $\overline{\mu} \models P$ *if and only if, for each rule* $\mathcal{R} \equiv a_0[x_0 = v_0] \rightarrow \mathcal{E}_1 \vee \ldots \vee \mathcal{E}_m \in S$*:*

1. *if* $\mathcal{R}$ *is triggered by an event* $\mu_i$ *inside* $\overline{\omega}_+$, *there is an existential statement* $\mathcal{E}$ *of* $\mathcal{R}$, *a* $1 \leq k \leq |\overline{\omega}_-|$, *and two subgraphs* $G_{>k} \sqsubseteq G_\mathcal{E}$ *and* $G_{\leq k} \sqsubseteq \Gamma(\mathcal{E}, k)$, *such that:*

   (a) $\overline{\mu}_{>k} \models_i G_{>k}$*;*

   (b) $G_{\leq k}$ *and* $G_{>k}$ *are* disjoint *and* $G_{\leq k} \rightrightarrows G_{>k} = G_\mathcal{E}$*;*

2. *for every instance* $I \in \Delta(\mathcal{R})$ *there is at least one existential statement* $\mathcal{E}$, *a* $1 \leq k \leq |\overline{\omega}_-|$, *and two subgraphs* $G_{\leq k} \sqsubseteq I(\mathcal{E}, k)$ *and* $G_{>k} \sqsubseteq G_\mathcal{E}$ *such that:*

   (a) $\overline{\mu}_{>k} \models G_{>k}$*;*

   (b) $G_{\leq k}$ *and* $G_{>k}$ *are disjoint and* $G_{\leq k} \rightrightarrows G_{>k} = G_\mathcal{E}$.

*Proof* ($\longleftarrow$). Consider a timeline-based planning problem $P = (\mathsf{SV}, S)$, and let $[\overline{\mu}] = (\overline{\omega}, \Gamma, \Delta)$ be the matching record of an event sequence $\overline{\mu} = \langle \mu_1, \ldots, \mu_n \rangle$ over $\mathsf{SV}$, with $\mu_i = (A_i, \delta_i)$. We show that if the stated conditions hold, then $\overline{\mu} \models P$, *i.e.*, that for each rule $\mathcal{R} \equiv a_0[x_0 = v_0] \rightarrow \mathcal{E}_1 \vee \ldots \vee \mathcal{E}_m \in S$ triggered by an event $\mu_i$, there is an $\mathcal{E}_k$ such that $\overline{\mu} \models_i G_{\mathcal{E}_k}$. Recall that $\overline{\omega}_+ = \overline{\mu}_{\geq h_+}$. We distinguish two cases, depending on whether $\mu_i$ lies before $\overline{\omega}_+$ ($i < h_+$) or not ($i \geq h_+$).

If $i < h_+$, by Item 3 of Definition 4.19, there is a corresponding instance $I = \Delta(\mathcal{R})$. By Item 2 of this Lemma, there is at least an existential statement $\mathcal{E}$, an integer $k$ with $1 \leq k \leq |\overline{\omega}_-|$ and a subgraph $G_{>k} \sqsubseteq G_{\mathcal{E}}$ such that $\overline{\mu}_{\leq h+k}, \gamma_{\leq k} \models_i I(\mathcal{E}, k)$ for some $\gamma_{\leq k}$, and $\overline{\mu}> h + k, \gamma_{>k} \models G_{>k}$ for some $\gamma_{>k}$. Then, since $I(\mathcal{E}, k) \rightrightarrows G_{>k} = G_{\mathcal{E}}$, we can combine $\gamma_{\leq k}$ and $\gamma_{>k}$ to obtain a matching function $\gamma$ such that $\overline{\mu} \models_i G_{\mathcal{E}}$.

If $i \geq h_+$, since Item 1 holds, we know that there are an existential statement $\mathcal{E}$ of $\mathcal{R}$, a $k \in \mathbb{N}$ with $1 \leq k \leq n$, and two *disjoint* subgraphs $G_{\leq k} \sqsubseteq \Gamma(\mathcal{E}, k)$ and $G_{>k} \sqsubseteq G_{\mathcal{E}}$, such that $\overline{\mu}_{>k}, \gamma_{>k} \models_i G_{>k}$ for some $\gamma_{>k}$ and $G_{\leq k} \rightrightarrows G_{>k} = G_{\mathcal{E}}$.

By Item 2 of Definition 4.19, we know that $\overline{\mu}_{\leq k}, \gamma_{\leq k} \models \Gamma(\mathcal{E}, k)$ for some $\gamma_{\leq k}$, and so in particular $\overline{\mu}_{\leq k}, \gamma_{\leq k} \models G_{\leq k}$. However, we can argue that $\overline{\mu}_{\leq k}, \gamma_{\leq k} \models G_{\leq k}$ as well, because $G_{\leq k}$ and $G_{>k}$ are disjoint and if $G_{\leq k}$ contained some unpaired $\mathsf{start}(a)$ nodes that were not already unpaired in $\Gamma(\mathcal{E}, k)$, it could not be that $G_{\leq k} \rightrightarrows G_{>k} = G_{\mathcal{E}}$. Then, by combining $\gamma_{>k}$ and $\gamma_{\leq k}$ we can obtain the matching function $\gamma$ such that $\overline{\mu}, \gamma \models G_{\mathcal{E}}$ thanks to Lemma 4.15. Moreover, since the trigger node $\mathsf{start}(a_0)$ is not part of $G_{\leq k}$ (Item 2b of Definition 4.19), then $\mathsf{start}(a_0) \sqsubseteq G_{>k}$, and since $\gamma(\mathsf{start}(a_0)) = \gamma_{>k}(\mathsf{start}(a_0))$, we have $\overline{\mu}, \gamma_{\leq k} \models_i G_{\mathcal{E}}$.

($\longrightarrow$). We now suppose that the event sequence $\overline{\mu}$ is a solution plan for the timeline-based planning problem $P = (\mathsf{SV}, S)$ and prove that the stated conditions hold. As for Item 1, suppose $\mathcal{R}$ is triggered inside $\overline{\omega}_+$ by an event $\mu_i$. Since $\overline{\mu} \models P$, we know there is a $\gamma$ such that $\overline{\mu}, \gamma \models G_{\mathcal{E}}$ for some existential statement $\mathcal{E}$ of $\mathcal{R}$, and $\gamma(\mathsf{start}(a_0)) = i$. Now, let $G_{\mathcal{E}} = (V, E, \beta)$, and let $h \leq k < h_+$ be a position in $\overline{\omega}_-$ such that there is no *bounded* edge $e = (T, T') \in E$ overlapping $k$, *i.e.*, with $\gamma(T) \leq k$ and $\gamma(T') > k$. This position is guaranteed to exist by Lemma 4.15 since $\delta(\overline{\omega}_-) \geq \mathsf{window}(P)$. Hence, we can identify the two subgraphs $G_{\leq k} \sqsubseteq G_{\mathcal{E}}$ and $G_{>k} \sqsubseteq G_{\mathcal{E}}$ such that $\overline{\mu}_{\leq k} \models G_{\leq k}$ and $\overline{\mu}_{>k} \models G_{>k}$. Note that by construction the two subgraphs are disjoint, since we are splitting them on the gap at position $k$, we know $G_{\leq k} \rightrightarrows G_{>k} = G_{\mathcal{E}}$, and $G_{\leq k} \sqsubseteq \Gamma(\mathcal{E}, k)$ because $\Gamma(\mathcal{E}, k)$ is the *maximal* subgraph of $G_{\mathcal{E}}$ that matches over $\overline{\mu}_{\leq k}$.

As far as Item 2 is concerned, suppose a rule $\mathcal{R} \in S$ is triggered by an event $\mu_i$ with $i < h+$. Since $\overline{\mu} \models P$, we know there is a $\gamma$ such that $\overline{\mu}, \gamma \models G_{\mathcal{E}}$ for some existential statement $\mathcal{E}$ of $\mathcal{R}$. Similarly to the previous case, we can identify a position $k$ and two disjoint subgraphs $G_{\leq k} \sqsubseteq G_{\mathcal{E}}$ and $G_{>k} \sqsubseteq G_{\mathcal{E}}$ such that $\overline{\mu}_{\leq k} \models G_{\leq k}$ and $\overline{\mu}_{>k} \models G_{>k}$. Then, we know $G_{\leq k} \sqsubseteq I(\mathcal{E}, k)$ and $G_{\leq k} \rightrightarrows G_{>k} = G_{\mathcal{E}}$. ∎

It is easy to verify that the size of $[\overline{\mu}]$ is exponential in the size of the considered planning problem.

**Lemma 4.21** — Size of matching records.
*Let $P = (\mathsf{SV}, S)$ be a timeline-based planning problem and let $\overline{\mu}$ be an event sequence over $\mathsf{SV}$. Then, the size of $[\overline{\mu}]$ is at most* exponential *in the size of $P$.*

*Proof.* Let $[\overline{\mu}] = (\overline{\omega}, \Gamma, \Delta)$. The first component, $\overline{\omega}$, is the shortest suffix of $\overline{\mu}$ that can be split into two subsequences of $\mathsf{window}(P)$ span each. Thus, $\delta(\overline{\omega}) \le 2\,\mathsf{window}(P)$. However, since it is the shortest such suffix, and the distance between events is at most exponential in the size of $P$ by Lemma 4.8, we have that $\delta(\overline{\omega}) \in \mathcal{O}(\mathsf{window}(P))$. It is easy to see that $\mathsf{window}(P) \le 2^{|P|}$, hence there are at most an exponential number of events, and each such event $\mu = (A, \delta)$ has polynomial size, given that $\delta \le \mathsf{window}(P)$ and $A$ is a set of size at most polynomial, hence $|\overline{\omega}| \in \mathcal{O}(2^{|P|})$. Then, consider the size of $\Gamma$. A key observation is that $\Gamma(\mathcal{E}, k) \sqsubseteq \Gamma(\mathcal{E}, k+1)$ for each $1 \le k < |\overline{\omega}_{-}|$. This means that, even though the different values of $k$ are exponentially many, there are only a polynomial number of them where $\Gamma(\mathcal{E}, k)$ differs from $\Gamma(\mathcal{E}, k+1)$, and this allows us to represent $\Gamma$ using a polynomial amount of space. A similar argument holds for $I(\mathcal{E}, k)$ for each instance $I \in \Delta(\mathcal{R})$ for any $\mathcal{R}$. However, there can be an exponential number of instances in $\Delta(\mathcal{R})$, hence the size of $\Delta$ can be exponential in the size of $P$. Considering the three components together, the size of $[\overline{\mu}]$ is exponential in the size of the problem. ■

Finally, all the building blocks are in place to show the main result of this section, which paves the way for an exponential-space decision procedure for the plan existence problem for timeline-based planning.

**Theorem 7** — Small-model theorem for timeline-based planning problems.
$P = (\mathsf{SV}, S)$ be a timeline-based planning problem. If there is any solution plan $\overline{\mu} \models P$, then there is a solution plan $\overline{\mu}' \models P$ such that $\delta(\overline{\mu}') \in \mathcal{O}\left(2^{2^{|P|}}\right)$.

*Proof.* The proof adopts a standard contraction argument. We know from Lemma 4.21 that $[\overline{\mu}] \in \mathcal{O}(2^{|P|})$. Let $K \in \mathcal{O}(2^{|P|})$ be the actual maximum size, in bits, of the matching record of an event sequence over $\mathsf{SV}$. If $|\overline{\mu}| > 2^K$, there has to be at least two positions $i$ and $j$ such that $[\overline{\mu}_{\le i}] = [\overline{\mu}_{\le j}]$. Then, it can be checked that for any two event sequences $\overline{\mu}'$ and $\overline{\mu}''$ such that $[\overline{\mu}'] = [\overline{\mu}'']$, it holds that $[\overline{\mu}'\mu] = [\overline{\mu}''\mu]$ for any event $\mu$. Hence, it holds that $\overline{\mu} \models P$ if and only if $\overline{\mu}' = \overline{\mu}_{\le i}\overline{\mu}_{\ge j+1} \models P$. In other words, we can cut and remove the subsequence $\overline{\mu}_{[i+1\ldots j]}$ without changing the satisfaction of the rules of $P$, obtaining an event sequence $\overline{\mu}'$ such that $|\overline{\mu}'| < 2^K$ (or we can repeat the process until we obtain one). Recall that $K \in \mathcal{O}(2^{|P|})$. Since, thanks to Lemma 4.8, we supposed *w.l.o.g.* that the time distance $\delta_{k+1}$ between any two consecutive events $\mu_k$ and $\mu_{k+1}$ is bounded by some $d \in \mathcal{O}(2^{|P|})$, we obtain that $\delta(\overline{\mu}') < d \cdot 2^K$, hence at most doubly-exponential in the size of $P$. ■

## 3.2 THE DECISION PROCEDURE

Thanks to Theorem 7, and exploiting the machinery of matching records, we can now devise a decision procedure for deciding whether a given timeline-based planning problem admits a solution plan, using at most an exponential amount of space, proving the EXPSPACE-completeness of the problem.

Since we aim at proving an upper bound on the *space* complexity of the problem, we can employ a standard shortcut: the algorithm is designed to run, using at most an exponential amount of space, on a *nondeterministic* Turing machine. Then, the classic complexity theory result by Savitch [111, 123], which, in particular, implies that NEXPSPACE = EXPSPACE, ensures that a *deterministic* exponential-space procedure exists as well.

The first building block is a subprocedure that, given the matching record $[\overline{\mu}]$ of an event sequence and an event $\mu$, obtains the matching record $[\overline{\mu}\mu]$ of the event sequence resulting from appending the event at the end.

**Lemma 4.22** — Appending an event to a matching record.
*Let $P = (\mathsf{SV}, S)$ be a timeline-based planning problem, let $[\overline{\mu}]$ be the matching record of an event sequence $\overline{\mu}$, and let $\mu$ be any event applicable to $[\overline{\mu}]$. Then, the matching record $[\overline{\mu}\mu]$ can be built in* exponential time *in the size of $P$.*

*Proof.* Let $[\overline{\mu}] = (\overline{\omega}, \Gamma, \Delta)$ be the matching record of some event sequence $\overline{\mu} = \langle \mu_1, \ldots, \mu_n \rangle$. Let be clear that the construction does not receive $\overline{\mu}$ as input, but only $[\overline{\mu}]$. Then, we will build the matching record $[\overline{\mu}\mu]$ of the event sequence $\overline{\mu}\mu$. Let it be $[\overline{\mu}\mu] = (\overline{\omega}', \Gamma', \Delta')$. Let us show how the three components have to be updated to account for the incoming event.

At first, updating $\overline{\omega}$ is the easiest part of the procedure. In the case that $\delta(\overline{\omega}\mu) \leq 2\,\mathsf{window}(P)$, then $\overline{\omega}' = \overline{\omega}\mu$, otherwise, if $\overline{\omega} = \overline{\mu}_{\geq h}$ for some $h$, then $\overline{\omega}' = \overline{\mu}_{\geq h'}\mu$, where $h' \geq h$ is the least position in $\overline{\mu}$ such that $\delta(\overline{\omega}') \geq 2\,\mathsf{window}(P)$. In other words, the new event is appended, becoming part of the recent history stored by the matching record, and the oldest events are discarded, as many as possible while preserving the fact that $\delta(\overline{\omega}') \geq 2\,\mathsf{window}(P)$. With the updated $\overline{\omega}'$, recall that $\overline{\omega}'_-$ is the shortest prefix of $\overline{\omega}$ such that $\delta(\overline{\omega}'_-) \geq \mathsf{window}(P)$.

Then, $\Gamma$ is updated accordingly. Let $s = h' - h$ be the number of events discarded in the update of $\overline{\omega}$ to $\overline{\omega}'$. As a first step, all the components of $\Gamma$ are shifted back by $s$ positions, and updated to reflect the incoming event, *i.e.*, $\Gamma'(\mathcal{E}, k) = \Gamma(\mathcal{E}, k+s)$ for all $1 \leq k \leq |\overline{\omega}'_-| - s$. Then, $\Gamma'(\mathcal{E}, k)$ for the positions $k > |\overline{\omega}'_-| - s$ can be obtained by composing two parts. The first part $G_{[h' \ldots k'']}$ is the maximal concatenation of bounded components of $G_{\mathcal{E}}$ such that $\overline{\mu}_{[h' \ldots k]} \models G_{[h' \ldots k]}$ for some $h' > h$, and the second is $\Gamma'(\mathcal{E}, h')$, *i.e.*, we set $\Gamma'(\mathcal{E}, k) = \Gamma(\mathcal{E}, k' + s) \rightrightarrows G_{[h \ldots k]}$. Note that this composition correctly captures the definition of $\Gamma'(\mathcal{E}, k)$, since $\delta(\overline{\mu}_{[h \ldots k]}) \geq \mathsf{window}(P)$ for all $k > |\overline{\omega}'_-| - s$.

To update $\Delta$, we proceed as follows. First, for all $\mathcal{R} \in S$, the contents of each $I \in \Delta(\mathcal{R})$ are shifted back of $s$ positions, *i.e.*, $I'(\mathcal{E},k) = I(\mathcal{E},k+s)$ for all $1 \leq k \leq |\overline{\omega}_+|$. Then, new values for the newcomer positions $I(\mathcal{E},k)$ with $k > |\overline{\omega}'_-| - s$, are computed. This can be done similarly to $\Gamma$. Then, new instances have to be created for the new positions. So suppose the rule $\mathcal{R}$ is triggered by a position $i$, and let $\mathcal{E}$ be an existential statement of $\mathcal{R}$. Then we have a (possibly empty) subgraph $G \sqsubseteq G_{\mathcal{E}}$ such that $\overline{\mu}_{[h+k \ldots h+k']} \models_i G$, for some $k'$, and all the edges going out of $G$ are unbounded. Then, we add to $\Delta(\mathcal{R})$ a new instance $I'$ such that $I(\mathcal{E},k') = \Gamma'(\mathcal{E},k) \cup G$.

Now, it is time to check that these updates can be computed in *nondeterministic exponential time*, as stated. To see this, observe that a single rule graph can be matched over an event sequence in polynomial space, hence exponential time in particular, by nondeterministically guessing a matching function $\gamma$ and then checking the satisfaction of the match. Then, during the updates of $\Gamma$ and $\Delta$, an exponential number of such checks are needed. ■

Combining Corollary 3.10, Theorem 7, and Lemmata 4.20, 4.21 and 4.22, we can finally devise a decision procedure for the problem.

**Theorem 8** — Complexity of timeline-based planning.
Deciding whether a timeline-based planning problem admits a solution plan is EXPSPACE-complete.

*Proof.* We devise a *nondeterministic* decision procedure for the problem, that runs in nondeterministic exponential space. The procedure builds the matching record $[\overline{\mu}]$ of a satisfying event sequence $\overline{\mu}$ incrementally, starting from $[\overline{\mu}_0] = [\varepsilon]$ and, at each step $i \geq 0$, nondeterministically guessing the next event $\mu_i$ to obtain $[\overline{\mu}_{i+1}] = [\overline{\mu}_i \mu]$, thanks to Lemma 4.22. At each step, by Lemma 4.20 the procedure can check whether a solution for the problem has been found. If not, the procedure continues until a number of steps greater than the maximum bound computed in Theorem 7 is reached, in which case the nondeterministic computation branch is rejected. By Lemma 4.21, the size of matching records is exponential in the size of the problem, hence to maintain the current matching record and to count up the upper bound of Theorem 7, only exponential space is needed. Given the well-known result by Savitch [123] that NEXPSPACE = EXPSPACE, the nondeterministic procedure outlined here also gives us a *deterministic* procedure running in exponential space. ■

## 3.3 COMPLEXITY WITH BOUNDED HORIZON

As noted in Section 2 of Chapter 2, a variant of timeline-based planning problems that is relevant in practical applications is that of problems with *bounded horizon* (Definition 2.14), where the input is expected to provide an *a priori* bound on the duration of the interesting solutions. This section studies the complexity of this special case, proving it to be NEXPTIME-complete.

As will be shown later, the decision procedure for the general case shown above can be adapted in a straightforward way. To shown that the problem is NEXPTIME-hard, we employ a reduction from a variant of *tiling problem*.

**Definition 4.23** — Exponentially bounded square tiling problem.
*A* tiling structure *is a tuple* $\mathcal{T} = (T, t_0, H, V, n)$, *where* $T$ *is a set of elements called* tiles, $t_0 \in T$ *is the* initial *tile,* $H, V \subseteq T \times T$, *are the* horizontal *and* vertical adjacency relations, *and* $n \in \mathbb{N}_+$ *is a positive number, encoded in binary.*

*A* tiling *of the tiling structure* $\mathcal{T}$ *is a function* $f : [n] \times [n] \to T$, *mapping any position* $(i, j)$ *of the square of size* $n \times n$ *to a tile* $f(i, j) \in T$ *such that:*

1. $f(0,0) = t_0$
2. *for all* $x \in [n-1]$ *and* $y \in [n]$, $f(x,y)\ H\ f(x+1,y)$
3. *for all* $x \in [n]$ *and* $y \in [n-1]$, $f(x,y)\ V\ f(x,y+1)$

*The* exponentially bounded square tiling problem *is the problem of finding whether a given tiling structure admits a tiling.*

Tiling problems have been used for a long time as a source of reductions to study the computational complexity of many problems in logic and combinatorics [77, 84, 85, 122, 136, 139]. Quoting van Emde Boas [136], «it is the simplicity of the tiling problem combined with the very local structure of the tiling constraints which makes these problems attractive for use in reductions.» Furthermore, by variating the shape of the surface to be tiled (a square, a rectangle with an unbounded side, a quadrant, *etc.*), one can obtain complete problems for a broad range of complexity classes ranging from NP to EXPSPACE, to non-elementary, undecidable, and highly undecidable problems. Two-players variants of the problem such as *tiling games* have also been studied [41].

In particular, the *exponentially bounded square tiling problem* defined above is known to be NEXPTIME-hard [77], and allows us to give a simple and straightforward proof of NEXPTIME-completeness for our problem.

**Theorem 9** — Complexity of bounded horizon timeline-based planning.
Finding whether a timeline-based planning problem with bounded horizon admits a solution plan is NEXPTIME-complete.

*Proof.* Let us first shown how the problem can be solved in *nondeterministic exponential time*. A timeline-based planning problem with bounded horizon $P = (\mathsf{SV}, S, H)$ asks to find a solution plan $\pi$ for the problem $P' = (\mathsf{SV}, S)$ such that $\mathsf{H}(\pi) \le H$. Hence, we can employ the same procedure shown in the proof of Theorem 8, but taking care of stopping the search not when the doubly exponential upper bound of Theorem 7 is found, but after only $H$ steps. Since $H \in \mathcal{O}(2^{|P|})$, the algorithm then runs in nondeterministic exponential time.

Let us now show that the problem is NEXPTIME-hard, by reduction from the exponentially bounded square tiling problem defined in Definition 4.23. Let $\mathcal{T} = (T, t_0, H, V, n)$ be a tiling structure. We build a suitable timeline-based planning problem with bounded horizon $P = (\mathsf{SV}, S, H)$ such that $\mathcal{T}$ admits a tiling if and only if $P$ admits a solution plan.

The set $\mathsf{SV}$ consists of a single state variable $x$ with domain $V_x = T$, *i.e.*, one possible value for each tile in $T$. The transition function is $T_x(v) = V_x$, for each $v \in V_x$, and the duration function constrains every token to have unitary duration, that is, $D_x(v) = (1, 1)$ for each $v \in V_x$. By setting the horizon $H = n^2$, the values of $x$ over time represent the tilings of the $n \times n$ square in a row-major layout. The synchronisation rules can encode the tiling constraints as follows. First of all, the initial tile is put into place with a triggerless rule:

$$\top \rightarrow \exists a[x = t_0] \,.\, \mathsf{start}(a) = 0$$

The horizontal tiling relation $H$ is represented by the following rules. For each tile $t \in T$, there is a rule for the consistency of the $H$ relation on the right:

$$a[x = t] \rightarrow \mathsf{start}(a) = n^2 \vee \bigvee_{tHt'}^{t' \in T} \exists b[x = t'] \,.\, \mathsf{start}(a) \leq_{[1,1]} \mathsf{start}(b)$$

and one on the left:

$$a[x = t] \rightarrow \mathsf{start}(a) = 0 \vee \bigvee_{t'Ht}^{t' \in T} \exists b[x = t'] \,.\, \mathsf{start}(b) \leq_{[1,1]} \mathsf{start}(a)$$

These rules handle the satisfaction of the horizontal constraints in both directions. Both handle the special case for respectively the first/last tile, which cannot have anything on the right/left side. The encoding of vertical tiling constraints is similar. For each $t \in T$, the following rules are added:

$$a[x = t] \rightarrow \mathsf{start}(a) \leq n \vee \bigvee_{tVt'}^{t' \in T} \exists b[x = t'] \,.\, \mathsf{start}(a) \leq_{[n,n]} \mathsf{start}(b)$$

$$a[x = t] \rightarrow n^2 - n \leq \mathsf{start}(a) \vee \bigvee_{t'Vt}^{t' \in T} \exists b[x = t'] \,.\, \mathsf{start}(b) \leq_{[n,n]} \mathsf{start}(a)$$

It can be verified that this timeline-based problem with horizon correctly encodes the original tiling problem. Moreover, the encoding can be produced in polynomial time, since it only involves loops making one step for each elements of the tiling set $T$ and the relations $H$ and $V$. ■

# AN AUTOMATA-THEORETIC PERSPECTIVE

This section revisits the results given in previous sections from a different, *automata-theoretic* point of view. We will re-prove Theorem 8 by constructing a suitable *nondeterministic finite automaton* that can recognise solution plans for a given timeline-based planning problem.

This different perspective gives us two major contributions:

1. the resulting decision procedure, based on a reachability analysis of the automaton, is more likely to lead to practicable techniques since it can exploit all existing tools based on automata theory;

2. by extending the construction to *Büchi* automata, we can prove the computational complexity of the problem when interpreted over *infinite plans*, a generalisation never considered before in the literature.

The rest of the section will show the automata-theoretic construction in detail, providing an alternative proof of Theorem 8. Then, the problem of timeline-based planning on *infinite timelines* will be defined, and the complexity of the problem will be shown to still be EXPSPACE-complete by extending the automaton construction to obtain a Büchi automaton.

## CAPTURING TIMELINES WITH FINITE AUTOMATA

We now show how to build, for any given timeline-based planning problem $P = (\mathsf{SV}, S)$, a *nondeterministic finite automaton* (NFA) that accepts exactly those words that represent solution plans for $P$. We first have to choose a suitable word representation for plans. In the rest of the chapter we made extensive use of *event sequences* for this purpose, which, however, do not fit well as the input word of a finite automaton. The representation used by our automaton construction is, however, very similar.

### PLANS AS WORDS

Let $\Sigma = \{\sigma : \mathsf{SV} \to \mathcal{V} \cup \{\text{␣}\} \mid \sigma(x) \in V_x \cup \{\text{␣}\}\}$, *i.e.*, the alphabet is a set of functions assigning a legitimate value (or the special symbol ␣) to each variable in SV. Notice that we can equivalently define $\Sigma$ as the cartesian product $\times_{x \in \mathsf{SV}}(V_x \cup \{\text{␣}\})$, *i.e.*, a set of tuples with a value (or ␣) for each variable. Treating symbols as functions will be useful as a notation to define the automaton, while the tuple point of view helps the intuitive understanding of the construction. Elements of $\Sigma$ are called *token symbols*. Among them, we identify the set $\Sigma_S = \{\sigma \in \Sigma \mid \sigma(x) \neq \text{␣}\}$ of *starting token symbols*. Observe that $|\Sigma| \leq (|\mathcal{V}| + 1)^{|\mathsf{SV}|}$, that is, the number of token symbols is at most exponential in the size of $P$.

A one-to-one correspondence can be easily established between finite words $\overline{\sigma} = \sigma_0\sigma_1 \cdots \sigma_n \in \Sigma_S\Sigma^*$ and event sequences $\overline{\mu} = \langle \mu_1, \dots, \mu_m \rangle$. Each event $\mu_i = (A_i, \delta_i)$ corresponds to $\delta_i$ symbols $\sigma_j, \dots, \sigma_{j+\delta_i}$ in the word, with $\sigma_k(x) = \text{␣}$ for each $x \in \mathsf{SV}$ and each $j \leq k < j + \delta_i$, and for the last of them, $\sigma_{j+\delta_i}(x) = v$ if $\mathsf{start}(x, v) \in A_i$ and $\sigma_{j+\delta_i}(x) = v$ otherwise. In other terms, time is flattened, with a symbol for each time step instead of the jumps of $\delta_i$ time steps made by each event, and $\sigma_j(x) = v$ if a token with $x = v$ starts at time $j$, while $\sigma_j(x) = \text{␣}$ if no token for the variable $x$ is starting at time $j$ (and so the current token for $x$ has the value of the position $j' < j$ where $\sigma_{j'} \neq \text{␣}$).

## BLUEPRINTS

Intuitively speaking, the states of the automaton contain patterns of tokens, called *blueprints*, that track the satisfaction of the rules on the current word. Blueprints specify the qualitative information about the relative position of such tokens abstracting away most of the quantitative information about how far apart tokens are. Each blueprint is associated with a synchronisation rule of the problem, and a specific way to schedule tokens apt to satisfy such rule.

While reading the input word, the automaton nondeterministically matches the abstract tokens of a set of blueprints with the concrete tokens found in the plan. A word is accepted if every token that triggers a synchronisation rule is involved in some instantiation of a blueprint that satisfies the rule.

This intuitive description can be turned into a formal definition as follows. Let us define the following two quantities:

1. $N$ is the largest finite constant appearing in $P$ as bounds in any atom;
2. $M$ is the length of the largest existential prefix of an existential statement occurring inside a synchronisation rule of $P$.

Notice that $N$ is exponential in the size of $P$, since constants are expressed in binary, while $M \in \mathcal{O}(|P|)$. Then, let $K = 2 \cdot (N + 1) + 2 \cdot (N + 1) \cdot (M + 1)$.

**Definition 4.24** — Blueprints.
*A* blueprint *is a tuple* $B = (m, f_s, f_e, f_{\mathsf{SV}}, f_{\mathcal{V}}, \mathcal{P})$, *with* $m \in [M]$, $f_s, f_e : [m] \to [K]$, $f_{\mathsf{SV}} : [m] \to \mathsf{SV}$, $f_{\mathcal{V}} : [m] \to \mathcal{V}$, *and* $\mathcal{P} \subseteq \mathbb{N}$, *such that for all* $i \in [m]$:

1. $x - 1 \in \mathcal{P}$ *for each* $x + 1 \in [2(N + 1), K] \cap (\mathsf{Img}(f_s) \cup \mathsf{Img}(f_e))$;
2. $f_s(i) < f_e(i)$;
3. *if* $f_{\mathsf{SV}}(i) = (V_x, T_x, D_x)$, *then* $f_{\mathcal{V}}(i) \in V_x$;
4. *tokens in the blueprint are* disjoint, i.e., *for each* $j \in [m]$, *if* $f_{\mathsf{SV}}(i) = f_{\mathsf{SV}}(j)$, *then exactly one of the following holds:*

(a) $f_e(i) \le f_s(j)$,

(b) $f_s(i) \ge f_e(j)$,

(c) $[f_s(i), f_e(i)] = [f_s(j), f_e(j)]$ and $f_{\mathcal{V}}(i) = f_{\mathcal{V}}(j)$.

*We refer to $\mathcal{P}$ as the set of* pumping points *of $B$.*

Each blueprint is functional to the satisfaction of a synchronisation rule. Figure 4.3 shows two blueprints for the following synchronisation rule:

$$a_0[x_0 = v_0] \to \exists a_1[x_1 = v_3] a_2[x_0 = v_1] . \\ (\mathsf{start}(a_1) \le_{[1,+\infty]} \mathsf{start}(a_0) \wedge \mathsf{end}(a_0) \le_{[1,+\infty]} \mathsf{start}(a_2)) \qquad (\mathcal{R}_{ex})$$

The meaning of the rule is the following: for every token $a_0$ for $x_0$ with value $v_0$, there must exist a token $a_1$ for $x_1$ with value $v_3$ and a token $a_2$ for $x_0$ with value $v_1$ such that the starting point of token $a_1$ occurs strictly before the starting point of $a_0$ and the ending point of $a_0$ occurs strictly before the starting point of $a_2$. Such a rule is triggered by tokens $(1,6)$, $(6,8)$, and $(8,12)$ in the plan in Figure 4.3. Blueprint $B_1$ can be used to certify the fulfilment of such a rule when triggered by $(6,8)$ (resp., $(8,12)$), because $B_1$ satisfies $\mathcal{R}$ and there is an instantiation of $B_1$ which associates $a_0$ with $(6,8)$ (resp., $(8,12)$).

The blueprint $B_1$ is an abstraction representing all scenarios where the three tokens $a_0$ (for $x_0 = v_0$), $a_1$ (for $x_1 = v_3$), and $a_2$ (for $x_0 = v_1$) occur in the following relative positions: $a_1$ occurs entirely before $a_0$, which, in turn, occurs entirely before $a_2$. As already mentioned, information about how long and how far apart the tokens are is abstracted away, thanks to the presence of the *pumping points* (filled circles in the picture), which allow for stretching the distances through the addition of points. Roughly speaking, a pumping point can be replaced by one or more other points if needed in order to match the blueprint over the input word.

Such a blueprint can be instantiated in two different ways in the plan depicted in Figure 4.3:

1. by instantiating tokens $a_0, a_1, a_2$ in $B_1$ with tokens $(6,8)$, $(0,4)$, and $(14,18)$ in the plan, respectively;

2. by instantiating tokens $a_0, a_1, a_2$ in $B_1$ with tokens $(8,12)$, $(0,4)$, and $(14,18)$ in the plan, respectively.

Contrarily, blueprint $B_1$ cannot be instantiated by associating $a_0$ with $(1,6)$ in the plan because, even though the state variable and the value of $a_0$ match the ones of $(1,6)$, there is no token for $x_1$ ending before point 1 in the plan, meaning that the relative positioning of tokens imposed by $B_1$ cannot be fulfilled.

Now, let $\mathbb{B}$ be the set of blueprints, and note that, by Definition 4.24, it follows that $|\mathbb{B}| \in \mathcal{O}(M \cdot K^M \cdot K^M \cdot |\mathsf{SV}|^M \cdot |\mathcal{V}|^M \cdot 2^{2M})$, *i.e.*, the number of blueprints

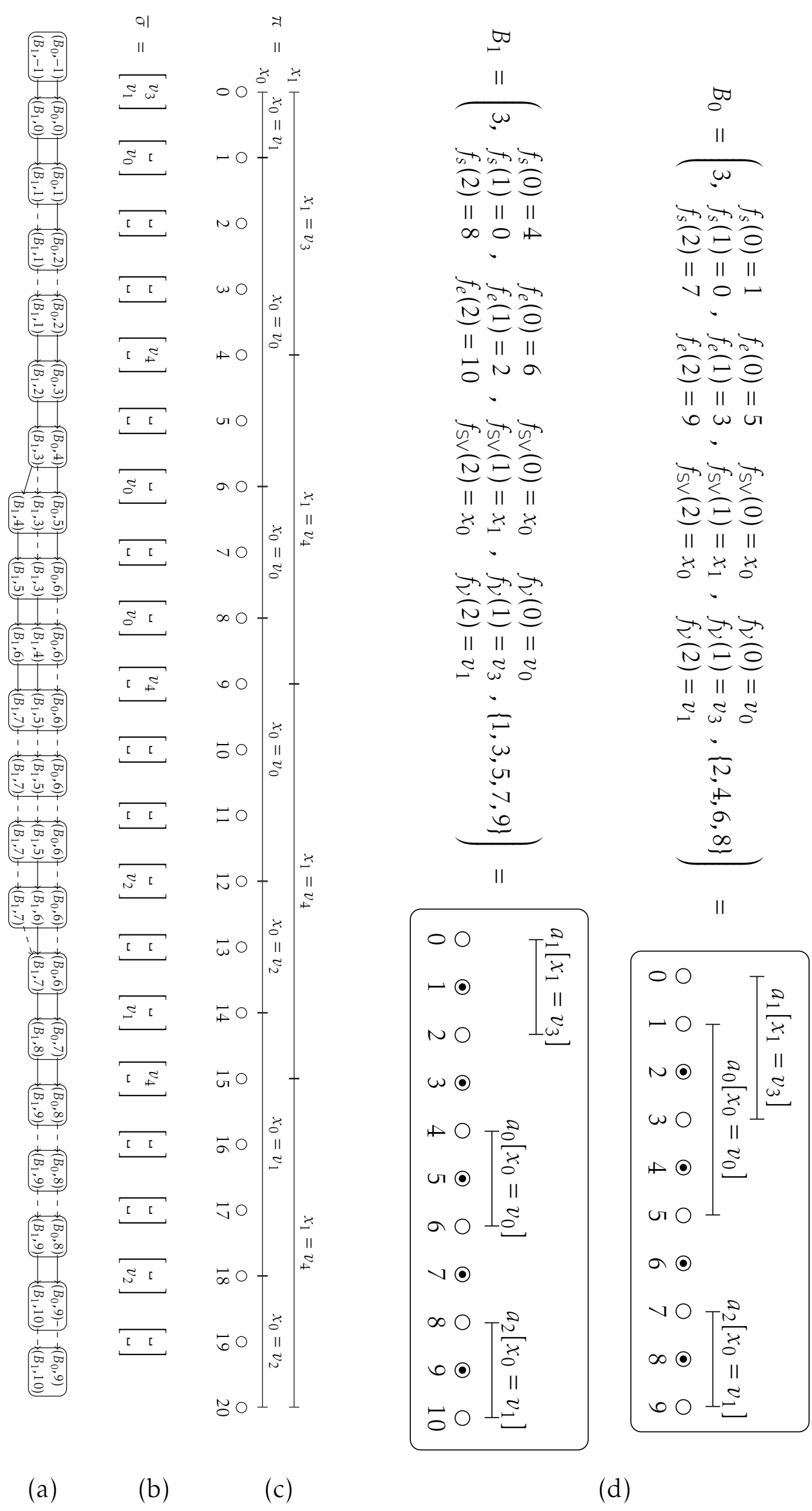

**Figure 4.3:** A run of the automaton (a) over a word $w$ (b) associated with the plan $\pi$ (c); states contain blueprints $B_0$ and $B_1$ (d).

is at most exponential in the size of $P$. We can now formally define how a word matches a given blueprint.

**Definition 4.25** — Fulfilment of rules by blueprints.
*Let* $\mathcal{R} = a_0[x_0 = v_0] \to \mathcal{E}_1 \vee \ldots \vee \mathcal{E}_k$ *be a synchronisation rule, and consider a blueprint* $B = (m, f_s, f_e, f_{\mathsf{SV}}, f_{\mathcal{V}}, \mathcal{P})$. *We say that* $B$ fulfils $\mathcal{R}$, *written* $B \models \mathcal{R}$, *if there is an existential statement* $\mathcal{E}_i = \exists a_1[x_1 = v_1] \ldots a_m[x_m = v_m] \,.\, \alpha_1 \wedge \cdots \wedge \alpha_h$ *such that:*

1. $f_{\mathsf{SV}}(j) = x_j$ *and* $f_{\mathcal{V}}(j) = v_j$ *for each* $0 \le j \le m$;
2. *each* $\alpha_j \equiv T \le_{[l,u]} T'$ *is satisfied by an atomic evaluation* $\lambda_j$ *such that:*
   (a) $\lambda_j$ *interprets* $\mathsf{start}(a_k)$ *and* $\mathsf{end}(a_k)$ *as, respectively,* $f_s(k)$ *and* $f_e(k)$;
   (b) *if* $u \neq +\infty$, *then* $[[\![T]\!]_{\lambda_j}, [\![T']\!]_{\lambda_j}] \cap \mathcal{P} = \varnothing$;

Blueprints are, in some sense, *static*, representing *a priori* a possible arrangement of tokens through the entire plan. A *viewpoint*, instead, tracks the current state of matching of a blueprint over the current word. More precisely, a viewpoint is a pair $(B, k)$, where $B$ is a blueprint and $-1 \le k \le K$ is the current point up to which the blueprint has correctly matched over the word that is being scanned by the automaton. States of the automaton are sets of viewpoints, as can be seen in the example run of the automaton shown in Figure 4.3a.

In other terms, if, after reading the token symbol $\sigma_i$ in the input word $\overline{\sigma}$, the automaton reaches a state that contains the viewpoint $(B, k)$; then, point $k$ of $B$ is matched with point $i$ of the input plan/word.

The set of viewpoints is denoted by $\mathbb{V}$. Its size is at most exponential in the size of $P$, that is, $|\mathbb{V}| \in \mathcal{O}(K \cdot |\mathbb{B}|)$. We say that a viewpoint $(B, k)$, with $B = (m, f_s, f_e, f_{\mathsf{SV}}, f_v, \mathcal{P})$, is *closed* if $k \ge \max(\mathrm{Img}(f_e))$, and is *initial* if $k = -1$.

Since blueprints are a compact representation of (parts of) a plan, a point in a blueprint may correspond to several points in the plan. In particular, this is the case with pumping points, that is, if $k$ is a pumping point of $B$, then the automaton computation can (nondeterministically) move to a state where viewpoint $(B, k)$ is left unchanged, while the input head move to from $\sigma_i$ to $\sigma_{i+1}$, meaning that the $k$th point of the blueprint is associated with both points $i$ and $i+1$ of the plan. In this way, several points of the plan are abstracted together.

Being in a pumping point $k$ is a necessary condition for a viewpoint to be allowed to *pump* the blueprint at point $k$, that is, to keep its pointer still while the input head moves to the next token symbol and the computation moves to the next step; however, it is not a sufficient one. To define when a viewpoint can pump the blueprint, we introduce the notions of *active* variables (and their values) in a viewpoint, as well as the one of *pumping symbol* for a viewpoint.

Let $\mathsf{value} : \mathbb{V} \times \mathsf{SV} \to \mathcal{V} \cup \{\bot\}$ be a partial function that returns the *value* of a state variable $x$ in a viewpoint $V$, defined as:

$$\mathsf{value}(V,x) = \begin{cases} v & \exists i \begin{cases} f_s(i) \le k < f_e(i) \\ f_{\mathsf{SV}}(i) = x \\ f_v(i) = v \end{cases} \\ \bot & \text{otherwise} \end{cases}$$

Thanks to the disjointness of the tokens for $x$ in the blueprint (condition 4 in Definition 4.24), value is well defined. This disjointness condition directly follows from the definition of timeline (Definition 2.2), and plays a major role in the construction. A variable $x \in \mathsf{SV}$ is *active* in $V$ if $\mathsf{value}(V,x)$ is defined.

**Definition 4.26** — Pumping symbol.
*Let $B$ be a blueprint with set of pumping points $\mathcal{P}$. A token symbol $\sigma \in \Sigma$ is a* pumping symbol *for a viewpoint $V = (B,k)$, if $k \in \mathcal{P} \cup \{-1\}$ and $\sigma(x) = \llcorner\!\lrcorner$ for every $x$ that is active in $V$.*

Intuitively, a viewpoint $V$ is allowed to pump when the next symbol read by the automaton is a pumping symbol for $V$. The second part of the condition ($\sigma(x) = \llcorner\!\lrcorner$ for all active variables $x$) reflects the fact that a point in a blueprint cannot hide a point that correspond in the input plan to the end of the active token: this would cause the active token to span two or more tokens in the plan, thus violating the (qualitative) structure of the plan itself.

Besides pumping, a viewpoint can evolve during the computation by *synchronising* with the plan, that is, the pointer of the viewpoint moves along the blueprint as the input head moves along the plan. In this case, we say that the viewpoint *steps* upon reading a given token symbol. This is captured by means of the following notions.

A viewpoint $V = (B,k)$, with $B = (m, f_s, f_e, f_{\mathsf{SV}}, f_v, \mathcal{P})$, is said to *start* (*end*) a state variable $x$ if there is an $i$ such $f_{\mathsf{SV}}(i) = x$ and $f_s(i) = k$ ($f_e(i) = k$).

**Definition 4.27** — Step symbol.
*Let $V = (B,k)$ be a viewpoint. A token symbol $\sigma \in \Sigma$ is a* step symbol *for $V$ if:*

1. *for all $x \in \mathsf{SV}$, if $(B,k+1)$ starts $x$, then it holds that $\mathsf{value}((B,k+1),x) = \sigma(x)$;*
2. *for all $x \in \mathsf{SV}$ with $x$ being active in $V$, $(B,k+1)$ ends $x$ if and only if $\sigma(x) \neq \llcorner\!\lrcorner$.*

The evolution of viewpoints along computations can be described by means of a ternary relation $\rightarrow \in \mathbb{V} \times \Sigma \times \mathbb{V}$, defined as follows: $(V,\sigma,V') \in \rightarrow$, with $V = (B,k)$, if and only if one of the following holds:

1. $\sigma$ is a pumping symbol for $V$ and $V' = V$,
2. $\sigma$ is a step symbol for $V$ and $V' = (B,k+1)$,
3. $V$ is closed and $V' = V$.

In the following, we use the more intuitive notation $V \to V'$ for $(V, \sigma, V') \in \to$. It is worth noticing that the notions of pumping and step symbols are not mutually exclusive, meaning that a symbol can be both a pumping and a step symbol for a given viewpoint. Consequently, a viewpoint $V = (B, k)$ can pump and step at the same time, thus leading to a state containing the two viewpoints $V_1 = V$ (obtained through pumping) and $V_2 = (B, k+1)$ (obtained through stepping). Roughly speaking, in such a case the viewpoint splits into two *copies*, that is, two viewpoints with the same blueprint but different pointers. Such a splitting corresponds to different instantiations of the same blueprint in order to satisfies a synchronisation rule that is triggered by different tokens in the plan.

Figure 4.3a depicts how viewpoints based on blueprints $B_0$ and $B_1$ (shown in Figure 4.3d) evolve with respect to the word $\overline{\sigma}$ (Figure 4.3b): stepping is represented through solid arrows, pumping is represented through dashed ones. Let us focus on blueprint $B_1$. As already pointed out, it can be used to certify the fulfilment of synchronisation rule $\mathcal{R}_{ex}$ shown above, when triggered by $(6, 8)$ and $(8, 12)$, through different instantiations: $a_0$ is associated with $(6, 8)$ in one of them, while it is associated with $(8, 12)$ in the other. This is done by exploiting the above described splitting mechanism. At the beginning, after reading the first token symbol ($\sigma_0 = \{x_0 \mapsto v_1, x_1 \mapsto v_3\}$), the viewpoint $V_0 = (B_1, 0)$ is created, thus establishing the correspondence between point 0 in $B_1$ and point 0 in the plan. It is easy to check that $\sigma_1 = \{x_0 \mapsto v_0, x_1 \mapsto \llcorner\!\lrcorner\}$ is a step symbol (but it is not a pumping one) for $V_0$, thus the computation leads to viewpoint $V_1 = (B_1, 1)$, and so on. Before reading the token symbol $\sigma_6 = \{x_0 \mapsto v_0, x_1 \mapsto \llcorner\!\lrcorner\}$, the state contains viewpoint $V_3 = (B_1, 3)$; since $\sigma_6$ is both a pumping and a step symbol for $V_3$, viewpoint $V_3$ splits into $V_3' = V_3$ and $V_4 = (B_1, 4)$. Viewpoint $V_4$ (created through stepping) establishes a correspondence between point 4 in $B_1$ and point 6 in the plan, thus commencing the instantiation of token $(4, 6)$ in $B_1$ with $(6, 8)$ in the plan (which is a trigger for $\mathcal{R}_{ex}$). On the other hand, viewpoint $V_3'$ (created through pumping) will be used to fulfil $\mathcal{R}_{ex}$ when it will be triggered by token $(8, 12)$ in the plan. To this end, $V_3'$ will pump point 3 until token symbol $\sigma_8 = \{x_0 \mapsto v_0, x_1 \mapsto \llcorner\!\lrcorner\}$ is reached; at that point, it steps to viewpoint $V_4' = (B_1, 4)$, thus paving the way to relating token $(4, 6)$ in $B_1$ with $(8, 12)$ in the plan (which is another trigger for $\mathcal{R}_{ex}$).

## THE AUTOMATA CONSTRUCTION

We are now ready to define the NFA we are looking for. As a matter of fact, it is obtained by intersecting two simpler automata, $\mathcal{E}_P$ and $\mathcal{A}_P$. The first checks that the word respects the basic syntactic requirements needed to correspond to an actual event sequence over SV, while the second actually exploits the machinery defined above to accept only words that represent solutions for the problem.

As in the rest of the chapter, we can suppose to consider only timeline-based planning problems that do not use *triggerless rules* nor *pointwise atoms*, and that only contain *trivial duration functions*. This means that the $\mathcal{E}_P$ automaton only has to check the transition function, so it is fairly straightforward to build it in such a way that a word $\overline{\sigma}$ is accepted by $\mathcal{E}_P$ if and only if $\overline{\sigma}$ correctly represents an event sequence over SV.

Let us therefore focus on $\mathcal{A}_P$, which does the greatest part of the work. It is defined as $\mathcal{A}_P = (2^{\mathbb{V}}, \Sigma, Q_0, \mathcal{F}, \Delta)$, where:

1. $2^{\mathbb{V}}$ is the finite set of states (states are sets of viewpoints), whose size is $2^{|\mathbb{V}|}$, that is, at most doubly exponential in the size of $P$;

2. $\Sigma$ is the input alphabet;

3. $Q_0 \subseteq 2^{\mathbb{V}}$ is the set of initial states, such that $\Upsilon \in Q_0$ if and only if $k = -1$ for all $(B, k) \in \Upsilon$;

4. $\mathcal{F} \subseteq 2^{\mathbb{V}}$ is the set of final states, defined as:

$$\mathcal{F} = \{\Upsilon \subseteq 2^{\mathbb{V}} \mid V \text{ is closed for all } V \in \Upsilon\}$$

5. $\Delta \subseteq 2^{\mathbb{V}} \times \Sigma \times 2^{\mathbb{V}}$ is the transition relation, which reflects the aforementioned transition $\rightarrow$. Roughly speaking, a set of viewpoints $\Upsilon$ evolves into another one, say it $\Upsilon'$, upon reading token symbol $\sigma$ if, and only if, each viewpoint $V \in \Upsilon$ evolves into a viewpoint $V' \in \Upsilon'$ according to the $\rightarrow$ relation and, vice versa, each viewpoint $V' \in \Upsilon'$ is the image of a viewpoint $V \in \Upsilon$ with respect to relation $\rightarrow$. Formally, $(\Upsilon, \sigma, \Upsilon') \in \Delta$ if and only if:

   (a) if $\Upsilon \in Q_0$, then $\sigma$ is a starting symbol;
   (b) for each $V \in \Upsilon$, there exists $V' \in \Upsilon'$ such that $V \rightarrow V'$;
   (c) for each $V' \in \Upsilon'$, there exists $V \in \Upsilon$ such that $V \rightarrow V'$;
   (d) if there is a synchronisation rule $R = a_0[x_0 = v_0] \rightarrow \mathcal{E}_1 \vee \ldots \vee \mathcal{E}_h$ in $S$ such that $\sigma(x_0) = v_0$, then there exists $(B, k) \in \Upsilon'$ such that $B \models R$, $(B, k)$ starts $x$, and $\mathsf{value}((B, k), x) = v_0$.

For every token in the plan that triggers a synchronisation rule, condition 5d forces the existence of a viewpoint $(B, k)$ to be used to satisfy that rule when triggered by that token. In particular, this condition forces the split of a viewpoint in situations analogous to the one described above, that is, when the blueprint of the viewpoint must be used to satisfy a rule triggered by two different tokens of the plan: indeed, if the viewpoint were not split, then one of the two triggers would never be instantiated and the computation would halt when the token symbol corresponding to the beginning of such token is read, without reaching a final state. Instead, satisfaction of triggerless rules is guaranteed by the presence of suitable viewpoints in every initial state belonging to $Q_0$.

It is worth pointing out that $\mathcal{A}_P$ has two sources of nondeterminism: one is represented by the nondeterministic choice of the set of viewpoints contained in the initial states, while the other is given by viewpoints that can both pump and step in correspondence of some token symbol.

It can be seen that $|\mathcal{A}_P| \in \mathcal{O}(|\Sigma| \cdot 2^{|\mathbb{V}|})$, that is, the size of $\mathcal{A}_P$ is at most doubly exponential in the one of $P$. After checking that the construction above actually captures the semantics of synchronisation rules, we proved the following result.

**Theorem 10** — Recognising solution plans through NFAs.
Let $P$ be a planning problem. A *nondeterministic* finite automaton $\mathcal{A}$ of size at most doubly exponential in the size of $P$ can be built that accepts a non-empty language if and only if $P$ admits a solution plan.

The automaton has doubly exponential size, but combining a classic reachability procedure (which runs in logarithmic space in the size of the automaton), with an on-the-fly generation of the nodes of the automaton, we can devise from this construction a decision procedure for the problem that actually runs in exponential space. This can be then seen as another proof of Theorem 8.

## 4.2 TIMELINE-BASED PLANNING ON INFINITE TIMELINES

In this section, we show that the construction described in the previous section can be suitably adapted to deal with *infinite timelines*. By extending the problem to infinite timelines, one can express *recurrent* goals, such as requiring certain facts to hold infinitely often during the execution of the system. Despite the existence of a number of natural application scenarios, to the best of our knowledge, the case of infinite plans has not been investigated in the context of timeline-based planning. Here, we show that the automata-theoretic approach described above can be naturally extended, by building a Büchi automaton instead of a normal NFA.

Formally, the definition of infinite timelines and infinite solution plans to a large extent coincides with the standard one: besides changing Definition 2.2 to consider *infinite* sequences of tokens, all the remaining definitions remain unchanged. The representation of plans through event sequences remains mostly unchanged, by considering infinite sequences of events. As a consequence, the encoding of plans as words is unchanged as well.

The automata construction shares most of its structure with the finite case. The $\mathcal{E}_P$ automaton can be defined in the same way, as it only has to take care of admitted transitions. Hence, we can directly turn to the definition of the infinite-word version of the $\mathcal{A}_P$ automaton, that we will call $\mathcal{B}_P$.

Given $\Sigma, Q_0$ and $\mathcal{F}$ are defined exactly as they have been defined in the previous section for $\mathcal{A}_P$, the automaton is defined as $\mathcal{B}_P = (2^{\mathbb{V}} \times 2^{\mathbb{V}}, \Sigma, Q_0 \times Q_0, 2^{\mathbb{V}} \times \mathcal{F}, \Delta_\omega)$, where the set of states $2^{\mathbb{V}} \times 2^{\mathbb{V}}$ is a set of *pairs* of what were states of $\mathcal{A}_P$, and the transition relation $\Delta_\omega \subseteq (2^{\mathbb{V}} \times 2^{\mathbb{V}}) \times \Sigma \times (2^{\mathbb{V}} \times 2^{\mathbb{V}})$ is defined in such a way that $((\Upsilon, \overline{\Upsilon}), \sigma, (\Upsilon', \overline{\Upsilon}')) \in \Delta_\omega$ if and only if:

1. $(\Upsilon, \sigma, \Upsilon') \in \Delta$;
2. if $\overline{\Upsilon} \notin \mathcal{F}$, then $(\overline{\Upsilon}, \sigma, \overline{\Upsilon}') \in \Delta$;
3. if $\overline{\Upsilon} \in \mathcal{F}$, then $\overline{\Upsilon}' = \Upsilon'$.

Intuitively, the automaton runs two copies of $\mathcal{A}_P$ in parallel. The first computation, whose state is represented by $\Upsilon$, proceeds exactly like a run of $\mathcal{A}_P$ and its goal is to ensure that every token that triggers a synchronisation rule is involved in some instantiation of a blueprint that satisfies the rule. However, due to the possible presence of *recurrent goals*, that is, rules imposing constraints to be verified infinitely often along the plan, such a component might never happen to reach states in $\mathcal{F}$.

For this reason, we introduce a second copy of $\mathcal{A}_P$, that is, the second component $\overline{\Upsilon}$, that works on separate, adjacent *chunks* of the input word, focusing each time on (previously triggered) synchronisation rules that have not yet been fulfilled at the beginning of the current chunk. When all these synchronisation rules (ignoring further synchronisation rules that might be triggered after the beginning of the chuck) have been satisfied (*i.e.*, $\overline{\Upsilon} \in \mathcal{F}$), the state $(\Upsilon, \overline{\Upsilon})$ is final for the automaton $\mathcal{B}_P$. With the next transition, leading to a new state $(\Upsilon', \overline{\Upsilon}')$, a new chunk is started by taking a *snapshot* of the first state component $\Upsilon'$, that is, $\Upsilon'$ is copied into $\overline{\Upsilon}'$. It is worth noticing that final states in $\mathcal{B}_P$ are determined only by the second component $\overline{\Upsilon}$, that is, $(\Upsilon, \overline{\Upsilon})$ is a final state of $\mathcal{B}_P$ if and only if $\overline{\Upsilon}$ is a final state of $\mathcal{A}_P$, and that the state that follows a final one along a run of $\mathcal{B}_P$, say it $(\Upsilon, \overline{\Upsilon})$, satisfies $\Upsilon = \overline{\Upsilon}$.

Moreover, note that finding a recurrent solution for a timeline-based planning problem (if any) is more general than finding a solution for it. As a matter of fact, it is possible to show that a timeline-based planning problem $P$ can be translated into a timeline-based planning problem $P'$ that only admits recurrent solutions, for which there exists a correspondence between solutions for $P$ and solutions for $P'$ such that the former ones correspond to finite prefixes of the latter.

Finally, observe that the considerations made in the previous section on the size and the final state reachability check for the automaton $\mathcal{A}_P \sqcap \mathcal{E}_P$ (obtained by intersecting $\mathcal{A}_P$ and $\mathcal{E}_P$) still hold for the size and the final state recurrent reachability check for automaton $\mathcal{B}_P$, thus yielding the following complexity characterisation for the problem of finding (infinite) recurrent solutions for timeline-based planning problems.

**Theorem 11** — Complexity with infinite solution plans.
Deciding whether a given timeline-based planning problem admits an infinite solution plan is EXPSPACE-complete.

## 5 CONCLUSIONS AND FURTHER WORK

In this chapter, we proved the first results regarding the computational complexity of timeline-based planning problems.

Besides the results *per se*, the conceptual framework of rule graphs introduced and used to prove them is also of independent interest, as proved by how the concept will be used in the following chapters. Nevertheless, the consequences of their definition still needs to be further explored. For example, Lemma 4.11 provides some criteria apt to simplify the structure of the synchronisation rules of a problem, removing some cases of redundant edges. However, rules are only considered in isolation, and the ability to identify edges that are redundant because of the interactions with other rules would have an important practical impact. Preprocessing techniques apt to reducing the magnitude of edge bounds are also an interesting development.

Our complexity analysis may be further refined in the future. In particular, finding tractable fragments is an important task for problems of such an high complexity. Since an exponential jump in the complexity comes from the succinct representation of edge bounds, the complexity of the problem with limited bounds and/or with only unbounded edges is conjectured to decrease to PSPACE. A complete parameterised complexity analysis, in the style of the work done by Bäckström et al. [11] on classical planning, would perfectly complete the picture.

The infrastructure based on matching records, built to show Theorem 8, may also be adapted in the future to solve related problems. In particular, matching records may form the basis of a monitoring procedure, enabling the construction of runtime verification tools [83] for timeline-based systems.

Section 4 used an automata-theoretic argument to prove that the problem over *infinite* solution plans is EXPSPACE-complete as well. Although interesting, this automata-theoretic approach is still in its infancy. Indeed, the shown decision procedure exploits the fact that the doubly-exponentially sized automaton resulting from the construction can be generated on-the-fly during a nondeterministic procedure. However, while theoretically this ensures that a deterministic procedure with the same space bound exists, the resulting algorithm will hardly be applicable in practice. In order to leverage the existing huge corpus of automata-theoretic research and software tools, a more expressive and more succinct class of automata (with a consequently harder emptiness problem) needs to be found to encode the problem into a more succinct automaton that can then be effectively manipulated.

As a last remark, note that the problem of timeline-based planning over infinite plans, handled in many systems in terms of repeated problems with bounded horizon, has never been explicitly studied in the literature. However, the problem has quite natural uses cases, deserving to be further investigated.

# TIMELINE-BASED PLANNING WITH UNCERTAINTY 5

The ability of properly integrating planning and execution is one of the flagship features of timeline-based planning systems. Current systems employ the notion of *flexible plan* to handle the *temporal uncertainty* that inherently arises when interacting with the environment in order to execute a plan. In this chapter, after discussing some limitations of the current approach, we propose and study the concept of *timeline-based games*, our take at timeline-based planning problems with uncertainty.

# 1 INTRODUCTION

The most important feature of existing timeline-based planning systems is certainly that of integrating the planning and execution phases under a unified framework. Issues related to uncertainty have been ignored in Chapters 3 and 4, in order to simplify the starting point for the discussion, providing their foundational results about expressiveness and complexity of timeline-based planning languages and problems. It is now time to add uncertainty back to the picture.

Rather than simply extending our results to the flexible timelines setting, a path which would certainly be worth exploring, this chapter takes a more proactive approach. First, in the rest of this section, we highlight a few limits of the current approach to uncertainty based on flexible plans. Then, Section 2 introduces *timeline-based games*, an extension to timeline-based planning problems with uncertainty that addresses the considered issues. Then, Section 4 addresses the problem of finding a winning strategy for a given timeline-based game, showing the complexity of the problem.

## 1.1 MOTIVATIONS

The whole discussion revolves around the notion of *nondeterminism*. The design of most timeline-based planning systems, and of the formal framework by Cialdea Mayer et al. [44] in particular, has been intentionally tailored to the handling of *temporal uncertainty*, *i.e.*, uncertainty about *when* things will happen, disregarding general forms of nondeterminism, *i.e.*, uncertainty about *what* will happen. Indeed, as defined in Section 3 of Chapter 2, flexible plans are intrinsically sequential objects, that cannot represent any choice about how the execution of the plan can proceed if not regarding the timing of events. This is, as stated multiple times, an intentional design choice of these systems.

In the meantime, the action-based planning community studied how to handle general nondeterminism quite extensively in the past years, following different approaches such as, for instance, reactive planning systems [13], deductive planning [130], POMDPs [78], model checking [49], and, especially, fully observable nondeterministic planning (FOND planning) [100, 101], which was also recently solved considering temporally extended goals [25, 112].

On the other hand, these approaches to nondeterministic action-based planning do not support flexible plans and temporal uncertainty, and do not account for controllability issues. Recently, SMT-based techniques have been exploited to deal with uncontrollable durations in strong temporal planning [46], but dynamic controllability issues are not addressed.

It seems therefore that the two worlds have evolved in different and incomparable ways. On one side, timeline-based planning supports specifically temporal uncertainty but does not consider general nondeterminism. On the other side, action-based planning considers general nondeterminism but does not explicitly support temporal uncertainty.

However, the explicit focus of timeline-based planning on temporal uncertainty does not imply that handling general nondeterminism would be useless in the common application scenarios of these systems. As explained in Section 3 of Chapter 2, the external variables in timeline-based planning problems with uncertainty are used more to express known facts about what will happen, rather than components of a full-fledged external entity running alongside the planned system. To this end, planning problems include a flexible plan, the *observation*, describing the behaviour of external variables up to the given temporal flexibility. The definition of the various forms of controllability then assumes that the behaviour of the environment follows what is stated by the observation. This is perfectly fine in some scenarios but limiting in others. For example, in collaborative robotics domains where the PLATINUm planning system was designed to be deployed [134], the controlled system has to cooperate with human agents, hence a true reactive behaviour is required and strong assumptions about the environment choices are not available. To cope with this need, many timeline-based systems have employed a feedback loop between the planning and execution phases, which includes a *failure manager* that senses when the execution is deviating from the assumed observation, and triggers a *re-planning* phase if necessary, devising a new flexible plan and a dynamic execution strategy that can be used to resume execution. However, the re-planning phase can be expensive to perform on-the-fly, limiting the real-time reactivity of the system.

Moreover, even when ignoring the issue noted above, the relationship between temporal uncertainty, nondeterminism, and timeline-based planning languages turns out to be more complex than anticipated. As a matter of fact, even explicitly focusing on temporal uncertainty, timeline-based planning languages are still able to express scenarios where handling nondeterminism in a more general way is required. To see what we mean, consider a timeline-based planning problem with uncertainty $P = (\mathsf{SV}_C, \mathsf{SV}_E, S, O)$, with a single controlled state variable $x \in \mathsf{SV}_C$ with $V_x = \{v_1, v_2, v_3\}$, $\mathsf{SV}_E = \varnothing$, and $S$ consisting of the following synchronisation rules:

$$\begin{aligned}
a[x = v_1] \to {} & \exists b[x = v_2] \,.\, \mathsf{end}(a) \le_{[0,0]} \mathsf{start}(b) \land \mathsf{start}(a) \le_{[0,5]} \mathsf{end}(a) \\
& \lor \exists c[x = v_3] \,.\, \mathsf{end}(a) \le_{[0,0]} \mathsf{start}(c) \land \mathsf{start}(a) \le_{[6,10]} \mathsf{end}(a) \\
\top \to {} & \exists a[x = v_1] \,.\, \mathsf{start}(a) = 0
\end{aligned}$$

Suppose that $D_x(v) = [1,10]$ for all $v \in V_x$, and that tokens where $x = v_1$ are uncontrollable, *i.e.*, $\gamma_x(v_1) = \mathsf{u}$ and $\gamma_x(v_2) = \gamma_x(v_3) = \mathsf{c}$. The rules require the controller to start the execution with a token where $x = v_1$, followed by a token where either $x = v_2$ or $x = v_3$ depending on the duration of the first token. This scenario is, intuitively, trivial to control. The system must execute $x = v_1$ as a first token due to the second rule. Then, the environment controls its duration, and the system simply has to wait for the token to end, and then execute either $x = v_2$ or $x = v_3$ depending on how long the first token lasted. However, there are no flexible plans that represent this simple strategy, since each given plan must fix the value of every token in advance. To guarantee the satisfaction of the rules, the value to assign to $x$ on the second token must be chosen during the execution, but this is not possible because of the exclusively sequential nature of flexible plans. In this case, therefore, the problem would be considered as unsolvable, even if the goals stated by the rules seem simple to achieve.

The problem above stems from the inherently sequential nature of flexible plans, which cannot represent the need for a choice to be made during execution other than regarding the timings of events. However, the example shows how the syntax of the language supports the modelling of scenarios where making qualitative choices depending on the environment nondeterministic behaviour is needed. Note that this is a different situation to that of *deterministic* action-based languages such as PDDL. In these languages, nondeterminism is not supported and simply cannot enter the picture. To support modelling nondeterministic behaviour, PDDL has to be extended with syntactic elements useful for the purpose (*e.g.*, the `anyof` keyword for nondeterministic effects). In this case, instead, the basic syntax of the language is sufficient to express such scenarios, but their possible solutions cannot be represented. In logical terms, one may say that dynamically controllable flexible plans do not provide a *complete semantics* for timeline-based planning with uncertainty. One may suppose that this expressive power comes from *disjunctions* in synchronisation rules, which allows us to compose the above example, but results such as Theorem 6 of Chapter 3 show how their presence is essential even to express simple *deterministic* scenarios, hence the gap cannot be filled by removing them.

It can be seen that scenarios like the one above would immediately arise when trying to encode any kind of nondeterministic action-based problem such as *fully observable nondeterministic* (FOND) planning problems. Hence, extending to nondeterministic planning the results of Chapter 3 is impossible. The key point, however, is that a syntactic representation of a FOND planning problem would be perfectly feasible, similarly to the encoding for classical planning gave in Theorem 6, which, however, would lack a proper semantics, corresponding to FOND *policies*, to express its solutions.

## 1.2 CONTRIBUTIONS

In this chapter, we propose and study an extension to timeline-based planning problems with uncertainty, called *timeline-based games*, which addresses both the issues outlined above by treating temporal uncertainty and general nondeterminism in a uniform way. Timeline-based games are two-player turn-based perfect-information games where the players play by executing the start and end endpoints of tokens, building a set of timelines. The first player, representing the controller, wins the game if it can manage to build a solution plan for a given timeline-based planning problem, independently from the behaviour of the second player, which represents the environment.

In Section 2, after defining the structure of these games, we show that they can capture the semantics of timeline-based planning problems with uncertainty, in the sense that for any such problem there is a game where the controller has a winning strategy if and only if the problem admits a dynamically controllable flexible plan. Moreover, we show that they strictly subsume the approach based on flexible plans, by showing how the above example can be modelled into a timeline-based game that admits a winning strategy for the controller.

Then, we address the problem of finding a winning strategy for such games, showing, in Section 4, that the problem of deciding whether the controller has a winning strategy for a given timeline-based planning game is in 2EXPTIME. The decision procedure heavily exploits the framework of *rule graphs* defined in Chapter 4. Whether this upper bound is strict is still an open question.

# 2 TIMELINE-BASED GAMES

This section introduces the *timeline-based games*, our game-theoretic approach to the handling of uncertainty in timeline-based planning. We first describe their general mechanism, including the winning condition, and then go in detail on how they relate to dynamically controllable flexible plans and the issues brought up in Section 1.

Intuitively, a timeline-based game is a turn-based, two-player game played by the controller, *Charlie*, and the environment, *Eve*. By playing the game, the players progressively build the timelines of a *scheduled plan* (see Definition 2.4). At each round, each player makes a move deciding which tokens to start and/or to stop and at which time. Both players are constrained by the set $\mathcal{D}$ of *domain* rules, which describe the basic rules governing the world. Domain rules replace the *observation* carried over by timeline-based planning problems with uncertainty (Definition 2.19), but generalise them allowing one to freely model the interaction between the system and the environment. Note that domain rules are not intended to be *Eve*'s (nor *Charlie*'s) *goals*, but, rather, a set of background facts about how the world works that can be assumed to hold at any time. Since neither player can violate $\mathcal{D}$, the strategy of each player may safely assume the validity of such rules. In addition, *Charlie* is responsible for satisfying the set $\mathcal{S}$ of *system* rules, which describe the rules governing the controlled system, including its goals. *Charlie* wins if, assuming *Eve* behaves according to the domain rules, he manages to construct a plan satisfying the system rules. In contrast, *Eve* wins if, while satisfying the domain rules, she prevents *Charlie* from winning, either by forcing him to violate some system rule, or by indefinitely postponing the fulfilment of his goals.

## 2.1 THE GAME ARENA

Let us immediately start by defining *timeline-based games* themselves.

**Definition 5.1** — Timeline-based game.
*A* timeline-based game *is a tuple* $G = (\mathsf{SV}_C, \mathsf{SV}_E, \mathcal{S}, \mathcal{D})$, *where* $\mathsf{SV}_C$ *and* $\mathsf{SV}_E$ *are the sets of, respectively, the* controlled *and the* external *variables, and* $\mathcal{S}$ *and* $\mathcal{D}$ *are two sets of synchronisation rules, respectively called* system *and* domain *rules, involving variables from both* $\mathsf{SV}_C$ *and* $\mathsf{SV}_E$.

During the game, the current state of the play can be seen as a partially built plan, where at any given time some tokens will be waiting to be completed. A *partial plan* is such a plan when each timeline might be incomplete. In Chapter 4 we already met objects apt to represent partially built plans: the event sequences (Definition 4.2). Event sequences closed on the left (Definition 4.3) exactly represent a plan that is being built going forward in time.

**Definition 5.2** — Partial plan.
*Let* $G = (\mathsf{SV}_C, \mathsf{SV}_E, \mathcal{S}, \mathcal{D})$ *be a timeline-based game. A* partial plan *for G is an event sequence* $\overline{\mu}$ *over* $\mathsf{SV}_C \cup \mathsf{SV}_E$, closed on the left.

Let $\mathbf{\Pi}_G$ be the set of all possible partial plans for $G$, or simply $\mathbf{\Pi}$ when there is no ambiguity. It is worth stressing again that the plan being built by the players, represented by the partial plan, is a *scheduled* plan, not a flexible one. The uncertainty is moved to the ignorance about what the next moves of *Eve* will be at each step. Partial plans can be either open or closed on the right depending on the particular moment of the game, but they are always closed on the left. Since there is no ambiguity, we will simply say *open* or *closed* to mean open or closed on the *right*. Recall that $\delta(\overline{\mu})$ denotes the duration of $\overline{\mu}$, that is, the distance in time between the last and the first events of the sequence, hence in our settings it can be interpreted as the time elapsed from the start of the game.

Since $\varepsilon$ counts as a closed event sequence and $\delta(\varepsilon) = 0$, the *empty* partial plan $\varepsilon$ is good starting point for the game. Players thus incrementally build a partial plan, starting from $\varepsilon$, by playing actions that specify which tokens to start and/or end, producing an event that extends the event sequence, or complementing the already existing last event of the sequence. Recall from Definition 4.2 that actions are terms of the form $\mathsf{start}(x,v)$ or $\mathsf{end}(x,v)$, where $x \in \mathsf{SV}$ and $v \in V_x$, and that the set of possible actions over $\mathsf{SV}$ is denoted as $\mathcal{A}_{\mathsf{SV}}$, here just $\mathcal{A}$ for simplicity. Actions of the former kind are called *starting* actions, and those of the latter kind are called *ending* actions. Then, we partition all the available actions into those that are playable by either of the two players.

**Definition 5.3** — Partition of player actions.
*The set $\mathcal{A}$ of available actions over the set of state variables $\mathsf{SV} = \mathsf{SV}_C \cup \mathsf{SV}_E$ is partitioned into the set $\mathcal{A}_C$ of* Charlie*'s actions, and the set $\mathcal{A}_E$ of* Eve*'s actions, defined as follows:*

$$\mathcal{A}_C = \underbrace{\{\mathsf{start}(x,v) \mid x \in \mathsf{SV}_C,\ v \in V_x\}}_{\textit{start tokens on } \text{Charlie's } \textit{timelines}} \cup \underbrace{\{\mathsf{end}(x,v) \mid x \in \mathsf{SV},\ v \in V_x,\ \gamma_x(v) = \mathsf{c}\}}_{\textit{end controllable tokens}}$$

$$\mathcal{A}_E = \underbrace{\{\mathsf{start}(x,v) \mid x \in \mathsf{SV}_E,\ v \in V_x\}}_{\textit{start tokens on } \text{Eve's } \textit{timelines}} \cup \underbrace{\{\mathsf{end}(x,v) \mid x \in \mathsf{SV},\ v \in V_x,\ \gamma_x(v) = \mathsf{u}\}}_{\textit{end uncontrollable tokens}}$$

Hence, players can start tokens for the variables that they own, and end the tokens that hold values that they control. It is worth to note that, in contrast to the original definition of timeline-based planning problems with uncertainty (Definition 2.19), Definition 5.3 admits cases where $x \in \mathsf{SV}_E$ and $\gamma_x(v) = \mathsf{c}$ for some $v \in V_x$, that is, cases where *Charlie* may control the duration of a variable that belongs to *Eve*. This situation is symmetrical to the more common one where *Eve* controls the duration of a variable that belongs to *Charlie* (*i.e.*, uncontrollable tokens), and we have no needs to impose any asymmetry.

Actions are combined into *moves* that can start/end multiple tokens at once.

**Definition 5.4** — Moves for *Charlie*.
*A* move $\mathsf{m}_C$ *for* Charlie *is a term of the form* $\mathsf{wait}(\delta_C)$ *or* $\mathsf{play}(A_C)$*, where* $\delta_C \in \mathbb{N}$ *and* $\varnothing \neq A_C \subseteq \mathcal{A}_C$ *is a set of either all* starting *or all* ending *actions.*

**Definition 5.5** — Moves for *Eve*.
*A* move $\mathsf{m}_E$ *for* Eve *is a term of the form* $\mathsf{play}(A_E)$ *or* $\mathsf{play}(\delta_E, A_E)$*, where* $\delta_E \in \mathbb{N}$ *and* $A_E \subseteq \mathcal{A}_E$ *is a set of either all* starting *or all* ending *actions.*

Two different aspects of the mechanics of the game influence the above definitions. First, moves such as $\mathsf{play}(A_C)$ and $\mathsf{play}(\delta_E, A_E)$ can play either only $\mathsf{start}(x,v)$ or only $\mathsf{end}(x,v)$ actions. A move of the former kind is called a *starting* move, while a move of the latter kind is called an *ending* move. Note that empty moves $\mathsf{play}(\delta_E, \varnothing)$ can be considered both starting or ending moves. Moreover, we consider $\mathsf{wait}$ moves as *ending* moves. In some sense, starting and ending moves have to be alternated during the game.

Second, the two players can play the two different sets of moves defined above, hence we denote as $\mathcal{M}_C$ the set of moves playable by *Charlie*, and as $\mathcal{M}_E$ the set of moves playable by *Eve*. *Charlie* can choose to play some actions to start/end a set of tokens, by playing a $\mathsf{play}(A_C)$ move, or to do nothing and wait a certain amount of time by playing a $\mathsf{wait}(\delta_C)$ move. *Charlie* plays first at each round, as will be formally stated later, hence *Eve* can reply to *Charlie*'s move by playing a $\mathsf{play}(A_E)$ move in response to a $\mathsf{play}(A_C)$ move by *Charlie*, and a $\mathsf{play}(\delta_E, A_E)$ move in response to a $\mathsf{wait}(\delta_C)$ move by *Charlie*. If *Charlie* plays a $\mathsf{play}(A_C)$ move, the given actions are applied *immediately*, for some specific sense defined later, and *Eve* replies by specifying what happens to her variables at the same time point. Instead, if *Charlie* plays a $\mathsf{wait}(\delta_C)$ move to wait some amount of time $\delta_C$, there is no reason why *Eve* should be forced to wait the same amount of time without doing nothing, so she can play a $\mathsf{play}(\delta_E, A_E)$ move, specifying an amount $\delta_E \leq \delta_C$, so that actions in $A_E$ will be applied accordingly, *interrupting* the wait of *Charlie* who can then timely reply to *Eve*'s actions. This is formalised as the following notion of *round*.

**Definition 5.6** — Round.
*A* round $\rho$ *is a pair* $(\mathsf{m}_C, \mathsf{m}_E) \in \mathcal{M}_C \times \mathcal{M}_E$ *of moves such that:*

1. $\mathsf{m}_C$ *and* $\mathsf{m}_E$ *are either both* starting *or both* ending *moves;*
2. *either* $\rho = (\mathsf{play}(A_C), \mathsf{play}(A_E))$, *or* $\rho = (\mathsf{wait}(\delta_C), \mathsf{play}(\delta_E, A_E))$ *with* $\delta_E \leq \delta_C$*;*

A *starting* (*ending*) round is one made of both starting (ending) moves. Note that since *Charlie* cannot play empty play moves and wait moves are considered ending moves, each round is unambiguously either a starting or an ending round. We can now define how a round is applied to the current partial plan to obtain the new one.

**Definition 5.7** — Outcome of rounds.
*Let* $\overline{\mu} = \langle \mu_1, \ldots, \mu_n \rangle$ *be a partial plan, with* $\mu_n = (A_n, \delta_n)$, *let* $\rho = (\mathsf{m}_C, \mathsf{m}_E)$ *be a round,* $\delta_E$ *and* $\delta_C$ *the time increments of the moves* ($\delta_C = \delta_E = 1$ *for* $\mathsf{play}(A)$ *moves), and let* $A_E$ *and* $A_C$ *be the set of actions of the two moves* ($A_C$ *is empty if* $\mathsf{m}_C$ *is a* wait *move).*

*The* outcome *of* $\rho$ *on* $\overline{\mu}$ *is the event sequence* $\rho(\overline{\mu})$ *defined as follows:*

1. *if* $\rho$ *is a starting round, then* $\rho(\overline{\mu}) = \overline{\mu}_{<n}\mu'$, *where* $\mu' = (A_n \cup A_C \cup A_E, \delta_n)$*;*
2. *if* $\rho$ *is an ending round, then* $\rho(\overline{\mu}) = \overline{\mu}\mu'$, *where* $\mu' = (A_C \cup A_E, \delta_E)$*;*

*We say that* $\rho$ *is* applicable *to* $\overline{\mu}$ *if:*

a) *the above construction is well-defined,* i.e., $\rho(\overline{\mu})$ *is a valid event sequence;*

b) $\rho$ *is an ending round if and only if* $\overline{\mu}$ *is open for all variables.*

We say that a single move by either player is applicable to $\overline{\mu}$ if there is a move for the other player such that the resulting round is applicable to $\overline{\mu}$.

Together, Definitions 5.6 and 5.7 finally define the mechanics of the game, that can now be fully clarified. The game starts from the empty partial plan $\varepsilon$, and players play in turn, composing a round from the move of each one, which is applied to the current partial plan to obtain the new one. Let $\overline{\mu}$ be the current partial plan. At each step of the game, both players can either only stop the execution of a set of tokens, by playing an ending round, or start the execution of a set of others, by playing a starting round (Item 1 of Definition 5.6). This does *not* mean that at each time point in the constructed plan only one of the two things can happen, but that the ending and starting actions of each events are contributed separately in two phases. When a starting round is played, its actions are added to the last event of the round (and indeed, since no time amount need to be specified, note that starting rounds can only be made of $\mathsf{play}(A)$ moves). In contrast, when an ending round is played, the corresponding actions form an event that is appended to $\overline{\mu}$, obtaining that $\delta(\rho(\overline{\mu})) > \delta(\overline{\mu})$. Then, the next round, which must be a starting round by Item b) of Definition 5.7, can start the new tokens following the ones that were just closed. Note that Items a) and b) of Definition 5.7 together ensure that a) the played actions make sense with regards to the current partial plan being built (such as the fact that token can be closed only if it was open *etc.*, see Definition 4.2), and b) that time cannot stall, by forcing starting rounds to be immediately followed by ending ones.

## 2.2 THE WINNING CONDITION

Since we defined the mechanics of the game, we need now to define the winning condition for the players, for which we define the notion of *strategy*.

**Definition 5.8** — Strategy for *Charlie*.
*A* strategy for Charlie *is a function* $\sigma_C : \mathbf{\Pi} \to \mathcal{M}_C$ *that maps any given partial plan* $\overline{\mu}$ *to a move* $\mathsf{m}_C$ *applicable to* $\overline{\mu}$.

**Definition 5.9** — Strategy for *Eve*.
*A* strategy for Eve *is a function* $\sigma_E : \mathbf{\Pi} \times \mathcal{M}_C \to \mathcal{M}_E$ *that maps a partial plan* $\overline{\mu}$ *and a move* $\mathsf{m}_C \in \mathcal{M}_C$ *applicable to* $\overline{\mu}$, *to a* $\mathsf{m}_E$ *such that* $\rho = (\mathsf{m}_C, \mathsf{m}_E)$ *is applicable to* $\overline{\mu}$.

A sequence $\overline{\rho} = \langle \rho_0, \ldots, \rho_n \rangle$ of rounds is called a *play* of the game. A play is said to be *played according to* some strategy $\sigma_C$ for *Charlie*, if, starting from the initial partial plan $\overline{\mu}_0 = \varepsilon$, it holds that $\rho_i = (\sigma_C(\Pi_{i-1}), \mathsf{m}^i_E)$, for some $\mathsf{m}^i_E$, for all $0 < i \leq n$, and to be played according to some strategy $\sigma_E$ for *Eve* if $\rho_i = (\mathsf{m}^i_C, \sigma_E(\Pi_{i-1}, \mathsf{m}^i_C))$, for all $0 < i \leq n$. It can be seen that for any pair of strategies $(\sigma_C, \sigma_E)$ and any $n \geq 0$, there is a unique run $\overline{\rho}_n(\sigma_C, \sigma_E)$ of length $n$ played according both to $\sigma_C$ and $\sigma_E$.

It is worth to note that, according to our definition of strategy, *Charlie* can base his decisions only on the previous rounds of the game, not including *Eve*'s move at the current round. However, *Charlie* can still react *immediately*, in some sense, to decide which token to start after an uncontrollable one closed by *Eve*, because of the alternation between starting and ending rounds. Hence *Charlie* can choose the starting actions of an event depending on the ending actions of that same event, but the contrary is not true: after *Eve* closes a token, *Charlie* has to wait at least one time step to react to that move with an *ending* action. This design choice is crucial to replicate and capture the semantics of dynamically controllable flexible plans, as will be detailed in Section 3.

As for the winning condition, we have to formalise the intuition given at the beginning of the section, regarding the role of domain rules and system rules. *Charlie* wins if, *assuming* domain rules are never broken, he manages to satisfy the system rules no matter how *Eve* plays.

Thus, let $G = (\mathsf{SV}_C, \mathsf{SV}_E, \mathcal{S}, \mathcal{D})$ be a planning game. To evaluate the satisfaction of the two sets of rules over the current partial plan, we proceed as follows. First, we define from $G$ two timeline-based planning problems (as for Definition 2.12), $P_\mathcal{D} = (\mathsf{SV}, \mathcal{D})$ and $P_\mathcal{S} = (\mathsf{SV}, \mathcal{S})$. Then, given a partial plan $\overline{\mu}$, we consider the scheduled plan $\pi_{\overline{\mu}'}$ corresponding to an event sequence $\overline{\mu}'$ obtained by closing $\overline{\mu}$ at time $\delta(\overline{\mu})$, *i.e.*, completing the last event of $\overline{\mu}$ in such a way to close any open token. Then, we say that a partial plan $\overline{\mu}$, and the play $\overline{\rho}$ such that $\overline{\mu} = \overline{\rho}(\varepsilon)$, are *admissible*, if $\pi_{\overline{\mu}'} \models P_\mathcal{D}$, *i.e.*, if the partial plan satisfies the domain rules, and are *successful* if both $\pi_{\overline{\mu}'} \models P_\mathcal{D}$ and $\pi_{\overline{\mu}'} \models P_\mathcal{S}$, *i.e.*, if the partial plan satisfies both the domain and system rules.

**Definition 5.10** — Admissible strategy for *Eve*.
*A strategy $\sigma_E$ for* Eve *is* admissible *if for each strategy $\sigma_C$ for* Charlie, *there is a $k \geq 0$ such that the play $\overline{\rho}_k(\sigma_C, \sigma_E)$ is admissible.*

**Definition 5.11** — Winning strategy for *Charlie*.
*Let $\sigma_C$ be a strategy for* Charlie. *We say that $\sigma_C$ is a* winning strategy *for* Charlie *if for any* admissible *strategy $\sigma_E$ for* Eve, *there exists an $n \geq 0$ such that the play $\overline{\rho}_n(\sigma_C, \sigma_E)$ is successful.*

We say that *Charlie wins* the game $G$ if he has a winning strategy, while *Eve wins* the game if a winning strategy does not exist.

To see an example, consider a timeline-based game $G = (\mathsf{SV}_C, \mathsf{SV}_E, \mathcal{S}, \mathcal{D})$ with two variables $x \in \mathsf{SV}_C$ and $y \in \mathsf{SV}_E$, $V_x = V_y = \{go, stop\}$, unit duration, and the sets of rules defined as follows:

$$\mathcal{S} = \left\{ \begin{array}{c} a[x = stop] \to \exists b[y = stop] \,.\, \mathsf{end}(b) = \mathsf{start}(a) \\ \top \to \exists a[x = stop] \,.\, \top \end{array} \right\}$$

$$\mathcal{D} = \left\{ \quad \top \to \exists a[y = stop] \,.\, \top \quad \right\}$$

Here, *Charlie*'s ultimate goal is to realise $x = stop$, but this can only happen after *Eve* realised $y = stop$. This is guaranteed to happen, since we consider only admissible strategies. Hence, the winning strategy for *Charlie* only chooses $x = go$ until *Eve* chooses $y = stop$, and then wins by executing $x = stop$. If $\mathcal{D}$ was instead empty, a winning strategy would not exist since a strategy that never chooses $y = stop$ would be admissible. This would therefore be a case where *Charlie* loses because *Eve* can indefinitely postpone his victory.

# 3 TIMELINE-BASED GAMES AND FLEXIBLE PLANS

Let us compare now the concept of *dynamic controllability* of flexible plans, as defined in [44], with the existence of winning strategies for timeline-based planning games, and show the greater generality of the latter concept.

The first step is to back the claim of this greater generality. Hence, we prove that given a flexible solution plan for a timeline-based planning problem with uncertainty, we can reduce the problem of the dynamic controllability of the plan to the existence of a winning strategy for a particular game.[1] To this aim, we need a way to represent as a game any given planning problem with uncertainty together with its flexible plan. Intuitively, this can be done by encoding the observations $\mathcal{O}$ into suitable domain rules. The game associated with a problem therefore mimics the exact setting described by the problem. What follows shows how such a game is built and which relationship exists between its winning strategies and dynamically controllable flexible plans for the original problem.

**Theorem 12** — Winning strategies vs dynamic controllability.
Let $P$ be a timeline-based planning problem with uncertainty, and suppose that $P$ admits a flexible solution plan $\Pi$. Then, a timeline-based game $G_{P,\Pi}$ can be built, in polynomial time, such that $\Pi$ is dynamically controllable if and only if *Charlie* has a winning strategy for $G_{P,\Pi}$.

*Proof.* Let $P = (\mathsf{SV}_C, \mathsf{SV}_E, S, O)$ be a timeline-based planning problem with uncertainty and a flexible solution plan $\Pi = (\pi, \mathsf{R})$ for $P$. We can build an equivalent timeline-based game $G_{P,\Pi} = (\mathsf{SV}_C, \mathsf{SV}_E, \mathcal{S}, \mathcal{D})$, by keeping $\mathsf{SV}_C$ and $\mathsf{SV}_E$ unchanged, and then suitably encoding the *observation* $O$ and the flexible plan $\Pi$ into, respectively, the set of domain rules $\mathcal{D}$ and of system rules $\mathcal{S}$. In this way, *Eve*'s behaviour will be constrained to follow what is dictated by the observation, replicating the semantics of timeline-based planning problems with uncertainty, and the behaviour of *Charlie* will follow by construction what is stated by the flexible plan.

To proceed, let $\mathsf{SV}_E = \{x_1, \ldots, x_n\}$, $O = (\pi_E, \mathsf{R}_E)$, and $\overline{\tau}_i = \pi_E(x_i) = \langle \tau^i_1, \ldots, \tau^i_{k_i} \rangle$, for some $k_i$ and all $x_i \in \mathsf{SV}_E$, with $\tau^i_j = (x_i, v^i_j, [e^i_j, E^i_j], [d^i_j, D^i_j])$. Finally, let $\mathsf{R}_E = \{\alpha_1, \ldots, \alpha_m\}$. The set $\mathcal{D}$ can encode the whole observation by a single triggerless rule stating that: (1) the tokens $\tau^i_j$ are required to exist, (2) their (a) position in the sequence, and (b) the end time and duration flexibility ranges correspond to the plan, and (3) the atoms in $\mathsf{R} = \{\alpha_1, \ldots, \alpha_{|\mathsf{R}|}\}$ are satisfied.

[1] Because of the great computational complexity mismatch, this is of course not a way to solve the dynamic controllability problem in practice, but shows the greater generality of the formalism.

Such a rule can be written as follows:

$$\top \to \exists \tau^1_1[x_1 = v^1_1], \ldots, \exists \tau^n_{k_n}[x_n = v^n_{k_n}]\,. \tag{1}$$

$$\wedge \bigwedge_{\substack{1\le i\le n\\1\le j<k_i}} \mathsf{end}(\tau^i_j) = \mathsf{start}(\tau^i_{j+1}) \tag{2a}$$

$$\wedge \bigwedge_{\substack{1\le i\le n\\1\le j\le k_i}} e^i_j \le \mathsf{end}(\tau^i_j) \le E^i_j \wedge d^i_j \le \mathsf{duration}(\tau^i_j) \le D^i_j \tag{2b}$$

$$\wedge \bigwedge_{1\le i\le m} \alpha_i \tag{3}$$

Adding the rule above to the set $\mathcal{D}$ of domain rules will ensure that any admissible play of the game follows the observation $O$. In a completely similar way we can encode the flexible plan $\Pi$ into a rule to add to the system rules $\mathcal{S}$. Note that by definition of flexible plan, following the plan satisfying $\mathsf{R}$ is sufficient to satisfy the set $S$ of problem rules, which thus can be discarded and be replaced with the single rule that encodes the plan. Now we can argue that *Charlie* has a winning strategy for $G_{P,\Pi}$ if and only if $\Pi$ is dynamically controllable.

($\longrightarrow$). Suppose that there exists a dynamic execution strategy $\sigma$ for $\Pi$. Then we can obtain a winning strategy for $G_{P,\Pi}$, by combining the flexible plan with the control strategy. Note that the flexible plan, encoded in the system rules, already constrains any winning play of the game to follow the plan, giving *Charlie* no freedom about which tokens to start after the end of any other. However, the plan does not provide the exact timing of the start of each token, of which only the flexibility interval is known. Hence, *Charlie* needs to be able to react to the end of each uncontrollable token by starting the next token immediately after. For this reason, it is important that, as noted in Section 2.2, *Charlie* has the ability to decide which tokens to start in a starting round of the game based on which were closed in the previous ending round. On the contrary, to make *Charlie* decide when to *end* each token, his winning strategy can just mimic the dynamic execution strategy of the plan.

($\longleftarrow$). On the other hand, if a winning strategy for $G$ exists, we need to ensure that it can be translated into a dynamic execution strategy for $\Pi$. Besides the translation itself which can be constructed easily, the important point is to guarantee that winning strategies for the game do not have more expressive power than dynamic control strategies for the plan (in general they do, but not in this specific game which is already constrained to replicate a single flexible plan). The critical issue here is that Definition 2.26 requires that the end time of any token $\tau$ chosen by the control strategy only depends on the end times of uncontrollable tokens that ended *before* the end time of $\tau$. In other words, the control strategy cannot decide to end a token at time $t$ based on the fact that other tokens ended at time $t$, but only based on those ended at time $t-1$ or before. This corresponds exactly to the semantics of our game, where *Charlie* cannot choose which tokens to end in an ending round base on which uncontrollable tokens will be ended by *Eve* at the same round, but has to wait to the next ending round, at least one time step later. ■

Theorem 12 shows that given a flexible solution plan $\Pi$, we can decide its dynamic controllability by looking for a winning strategy for the game $G_{P,\Pi}$. In a more general setting, given the timeline-based planning problem with uncertainty $P = (\mathsf{SV}_C, \mathsf{SV}_E, S, O)$, in a completely similar way we can build a game $G_P = (\mathsf{SV}_C, \mathsf{SV}_E, \mathcal{S}, \mathcal{D})$ such that the existence of a dynamically controllable flexible plan for $P$ implies the existence of a winning strategy for $G_P$. This is done by encoding the observation $O$ into the set of domain rules $\mathcal{D}$ exactly as done in Theorem 12, but setting $\mathcal{S} = S$, without constraining the game to any specific plan. Then if a plan exists, and it is dynamically controllable, with arguments similar to Theorem 12 it can be checked that a winning strategy for $G_P$ must exist as well.

■ **Corollary 5.12** — Generality of timeline-based games.
*Let $P$ be a timeline-based planning problem with uncertainty. Then, a timeline-based game $G_P$ can be built, in polynomial time, such that if $P$ admits a dynamically controllable flexible solution plan, then* Charlie *has a winning strategy for $G_P$.* ■

The converse is not true, however, because winning strategies for timeline-based games are strictly more expressive than timeline-based planning problems with uncertainty, hence there can be some problems $P$ that do *not* have any dynamically controllable flexible plan, but there is a winning strategy for $G_P$. This is the case with the example problem discussed in Section 1.1, which has an easy winning strategy when seen as a game, while it has no dynamically controllable flexible plan. We can encode the example problem $P$ with the game $G_P$, in which the shown synchronisation rules are included as system rules, the set of domain rules is empty (since there are no external variables and thus the observation is empty as well). The winning strategy is as simple as promised: after playing $\mathsf{start}(x, v_1)$ at the beginning, *Charlie* only has to wait for *Eve* to play $\mathsf{end}(x, v_1)$, and then play $\mathsf{start}(x, v_2)$ or $\mathsf{start}(x, v_3)$ according to the current timestamp. Therefore, one can prove the following theorem.

■ **Theorem 13**
There exists a timeline-based planning problem with uncertainty $P$ such that there are no dynamically controllable flexible plans for $P$, but *Charlie* has a winning strategy for the associated planning game $G_P$. ■

# FINDING WINNING STRATEGIES

In previous sections we motivated and shown the definition of *timeline-based games*, and shown how the existence of a winning strategy for such a game subsumes the existence of dynamically controllable flexible plans for the equivalent timeline-based planning problem with uncertainty. In this section, we show how such a winning strategy can be found, providing the complexity of the problem of establishing whether it exists.

The proof employs an encoding of timeline-based games into *concurrent game structures* (CGS) [7], and in particular, into the specific subclass of *turn-based synchronous game structures*. The encoding of the game into one such structure allows us to decide the existence of a winning strategy by expressing the winning condition in a *extended alternating-time temporal logic* (ATL*) [7] formula. In the rest of the section, a brief introduction to CGSs and the ATL* logic will be provided, and then the encoding of timeline-based games into CGSs will be shown and used to prove the final result.

## 4.1 ALTERNATING-TIME TEMPORAL LOGIC

In the field of *model checking* [50], two of the logics most commonly used as a specification language are *linear temporal logic* (LTL) and the *computation tree logic* (CTL), a fragment of the more general CTL*. In contrast to LTL, which is a temporal logic interpreted over *linear-time* domains (see more on this front in Chapter 6), CTL and CTL* adopt a *branching-time* semantics: CTL/CTL* formulae are interpreted over trees (usually the result of the unrolling of some state transition system) that represent the different possible future evolutions of a system, and formulae can quantify over paths of such trees. For example, the CTL* formula $\mathsf{A}\phi$ demands $\phi$ to be true when interpreted over *all* the paths of the tree rooted at the current state, while $\mathsf{E}\phi$ dually states that there is at least one such path where $\phi$ holds.

*Alternating-time temporal logic* (ATL) [7], and its extension ATL*, can be seen as generalisations of CTL and CTL* where the system used to interpret the formulae is the transition system resulting from the composition of multiple agents instead of a single one. In this context, paths of the system corresponds to the trace of the parallel execution of the components of the system. When we assign goals to each player or to subsets of the players, we can see the structure as the state space of a *game*, and ATL/ATL* formulae have the ability to quantify over paths played according to specific *strategies* adopted by each player. In particular, a formula of the form $\langle\!\langle A \rangle\!\rangle \phi$ is true whenever there is a set of strategies for the set of players $A$ such that $\phi$ is true on all the paths played according to said strategies. If $\mathcal{P}$ is the set of *all* the players of the system, then $\langle\langle\mathcal{P}\rangle\rangle\phi$ is equivalent to $\mathsf{E}\phi$ in CTL*, hence ATL/ATL* are proper extensions of CTL/CTL*. The ability of quantifying over paths played according to specific strategies makes these logics particularly suitable for expressing the winning condition of timeline-based games. Let us now recall the syntax and semantics of these logics.

Let $\mathcal{P} = \{1, \ldots, k\}$ be a finite set of *players*, and $\Sigma$ a finite set of propositions. Then, an ATL* formula over $\Sigma$ and $\mathcal{P}$ is a *state formula* $\phi$ defined as follows:

$$\phi := p \mid \neg\phi_1 \mid \phi_1 \vee \phi_2 \mid \langle\langle \mathsf{A} \rangle\rangle \psi \qquad \text{state formulae}$$

$$\psi := \phi \mid \neg\psi \mid \psi_1 \vee \psi_2 \mid \mathsf{X}\psi_1 \mid \psi_1 \,\mathcal{U}\, \psi_2 \qquad \text{path formulae}$$

where $\phi$, $\phi_1$ and $\phi_2$ are state formulae, $\psi$, $\psi_1$ and $\psi_2$ path formulae, and $A \subseteq \mathcal{P}$ is a set of players. Besides standard boolean connectives, with shortcuts such as $\psi_1 \wedge \psi_2 \equiv \neg(\neg\psi_1 \vee \neg\psi_2)$ and $\psi_1 \to \psi_2 \equiv \neg\psi_1 \vee \psi_2$, we have the *tomorrow* ($\mathsf{X}\psi$) and *until* ($\psi_1 \mathcal{U} \psi_2$) temporal operators as in LTL and CTL*, and, more importantly, the *existential* strategy quantifying operator $\langle\langle \mathsf{A} \rangle\rangle \psi$, from which we can define the *universal* dual $[\![\mathsf{A}]\!]\psi = \neg\langle\langle \mathsf{A} \rangle\rangle\neg\psi$. Note the subdivision into path and state formulae: path formulae need to be interpreted over a path, while state formulae are interpreted over a single state. ATL is the fragment of ATL* where temporal operators can only appear directly nested inside a strategy quantifying operator.

ATL and ATL* formulae are interpreted over *concurrent game structures*, which can represent a vast variety of concurrent games. In our setting, it is sufficient to consider the specific subclass of *turn-based synchronous game structures*.

**Definition 5.13** — Turn-based synchronous game structure.
*A* turn-based synchronous game structure *is a tuple* $S = \langle \mathcal{P}, Q, \Sigma, \nu, \lambda, R \rangle$, *where:*

1. $\mathcal{P} = \{1, \ldots, k\}$ *is the set of players;*
2. $Q$ *is the finite set of* states*;*
3. $\Sigma$ *is the finite set of* propositions*;*
4. $\nu : Q \to 2^{\Sigma}$ *provides the set* $\nu(q)$ *of propositions true at any state* $q \in Q$*;*
5. $\lambda : Q \to \mathcal{P}$ *is a function telling which player owns any given state;*
6. $R \subseteq Q \times Q$ *is the transition relation.*

Turn-based synchronous game structures, simply called *game structures* from now, represent games where players play in turn, not concurrently, and indeed each state $q \in Q$ is owned by the player $\lambda(q)$, who plays when the game reaches one of their states. To define how ATL/ATL* formulae are interpreted over game structures we need to define the notion of strategy. A path of the game is an infinite sequence of states $\overline{q} = \langle q_0, q_1, \ldots \rangle$ such that $(q_i, q_{i+1}) \in R$ for all $i \geq 0$. Given a player $a \in \mathcal{P}$, a *strategy* for $a$ is a function $f_a : Q^+ \to Q$ that maps any non-empty finite prefix $\overline{q} = \langle q_0, \ldots, q_n \rangle$ of a path (the history of the game play), where $\lambda(q_n) = a$, to the next state $f_a(q_n)$ chosen among the successors of $q_n$. A play such that $q_{i+1} = f_a(q_i)$ for any $q_i$ such that $\lambda(q_i) = a$ is said to be *played according to* the strategy $f_a$. Given a set of players $A \subseteq \mathcal{P}$, and a set of strategies $F_A$, one for each $a \in A$, the sequence $\overline{q}$ is played according to $F_A$ if it is played according to all the strategies in $F_A$.

Then, given a game structure $S = \langle \mathcal{P}, Q, \Sigma, \nu, \lambda, R \rangle$, and an ATL* (state) formula $\phi$ over propositions $\Sigma$ and players $\mathcal{P}$, we define the *satisfaction* of $\phi$ over a state $q \in Q$ of $S$, written $S, q \models \phi$, as follows:

1. $S, q \models p$ iff $p \in \nu(q)$;
2. $S, q \models \neg\phi$ iff $S, q \not\models \phi$;
3. $S, q \models \phi_1 \vee \phi_2$ iff $S, q \models \phi_1$ or $S, q \models \phi_2$;
4. $S, q \models \langle\!\langle A \rangle\!\rangle \psi$ iff there exists a set of strategies $F_A$, one for each $a \in A$, such that $S, \overline{q} \models \psi$ for all paths $\overline{q} = \langle q, \ldots \rangle$ played from $q$ according to $F_A$;

where the satisfaction of a path formula $\psi$ over a state sequence $\overline{q} = \langle q_0, q_1, \ldots \rangle$, written $S, \overline{q} \models \psi$, is defined as follows:

5. $S, \overline{q} \models \phi$ iff $S, q_0 \models \phi$, for a state formula $\phi$;
6. $S, \overline{q} \models \neg\psi$ iff $S, \overline{q} \not\models \psi$;
7. $S, \overline{q} \models \psi_1 \vee \psi_2$ iff $S, \overline{q} \models \psi_1$ or $S, \overline{q} \models \psi_2$;
8. $S, \overline{q} \models \mathsf{X}\psi_1$ iff $S, \overline{q}_{\geq 1} \models \psi_1$;
9. $S, \overline{q} \models \psi_1 \, \mathcal{U} \, \psi_2$ iff there exists a position $i \geq 0$ such that $S, \overline{q}_{\geq i} \models \psi_2$, and $S, \overline{q}_{\geq j} \models \psi_1$ for all $0 \leq j < i$.

Note that the existential strategy quantifier $\langle\!\langle A \rangle\!\rangle$ contains an element of universal quantification on the *paths* played according to the existentially quantified set of strategies. In the restricted context of *turn-based* structures that we are considering, the games are *determined*, which means the absence of a winning strategy for a player implies the existence of a winning strategy for the antagonist. In other terms, as also noted in [7], over turn-based structures it holds that $\langle\!\langle A \rangle\!\rangle \phi = [\![ \mathcal{P} \setminus A ]\!] \phi$.

## 4.2 FINDING STRATEGIES VIA ATL* MODEL CHECKING

We can now finally show how the existence of a winning strategy can be decided for timeline-based games, by encoding them into suitable game structures.

From the definitions given in Section 2 and, in particular, the definitions of strategies for the two players (Definitions 5.8 and 5.9), it can be seen that a timeline-based game provides an implicit representation for a potentially infinite state space composed by all possible partial plans $\mathbf{\Pi}$. Therefore, in order to encode a timeline-based game into a game structure, we need to reduce it to a finite state space. The key observation here is that, although each synchronisation rule can potentially look arbitrarily far in the past and in the future, a finite representation of the history of the game is possible. We met already the same problem in Chapter 4, when solving the *satisfiability* problem for timeline-based planning problems. Indeed, here we can reuse the framework built for the satisfiability problem.

Recall, from Section 3.1, that given a timeline-based planning problem $P = (\mathsf{SV}, S)$ and any event sequence $\overline{\mu}$, it is possible to build a structure $[\overline{\mu}]$, called *matching record*, such that an effective algorithm exists to tell whether or not $\overline{\mu}$ is a solution for $P$ (Lemma 4.20). Furthermore, given an event $\mu$, it is possible to effectively build the matching record $[\overline{\mu}\mu]$. This can therefore be the mechanism apt to reduce to a finite size the state space of our games. Given a timeline-based game $G = (\mathsf{SV}_C, \mathsf{SV}_E, \mathcal{S}, \mathcal{D})$, a game structure can be built over the set of all possible matching records for the timeline-based planning problem $P = (\mathsf{SV}_C \cup \mathsf{SV}_E, \mathcal{S} \cup \mathcal{D})$. Suitable edges connect the states according to the applicable moves available to the players. Then, a simple $\mathsf{ATL}^*$ formula expressing the winning condition of Definition 5.11 can be model-checked against the game structure to decide the existence of the winning strategy.

**Theorem 14** — Complexity of finding winning strategies.
Let $G = (\mathsf{SV}_C, \mathsf{SV}_E, \mathcal{S}, \mathcal{D})$ be a timeline-based game. Whether *Charlie* has a winning strategy for $G$ can be decided in *doubly-exponential time*.

*Proof.* As anticipated, we will reduce the problem to the model checking of a suitable $\mathsf{ATL}^*$ formula over a game structure that encodes the state space of the game. Given a timeline-based game $G = (\mathsf{SV}_C, \mathsf{SV}_E, \mathcal{S}, \mathcal{D})$, the starting point for building the corresponding game structure is the set $\mathcal{M}$ of all the *matching records* for the timeline-based planning problem $P = (\mathsf{SV}_C \cup \mathsf{SV}_E, \mathcal{S} \cup \mathcal{D})$. First, observe that by construction, if an event sequence $\overline{\mu}$ is such that $\overline{\mu} \models P$, then it also holds that $\overline{\mu} \models P_{\mathcal{S}} = (\mathsf{SV}_C \cup \mathsf{SV}_E, \mathcal{S})$ and $\overline{\mu} \models P_{\mathcal{D}} = (\mathsf{SV}_C \cup \mathsf{SV}_E, \mathcal{D})$. Observe, moreover, that from a matching record $[\overline{\mu}]$ built for $P$ we can decide, in the same way as shown in Lemma 4.20, whether $\overline{\mu} \models P_{\mathcal{S}}$ or $\overline{\mu} \models P_{\mathcal{D}}$, by focusing on a subset of the rules of $P$ and ignoring the others. Since the matching record $[\overline{\mu}]$ by definition includes as-is a suffix of $\overline{\mu}$, and the last event in particular, moves applicable to $\overline{\mu}$ when seen as a partial plan are identifiable by looking at $[\overline{\mu}]$ alone. Let $[\mathbf{\Pi}]$ be the set of matching records $[\overline{\mu}]$ for $\overline{\mu} \in \mathbf{\Pi}$. Then, we can encode the state space of the game as the turn-based synchronous game structure $S_G = \langle \mathcal{P}, Q, \Sigma, \nu, \lambda, R \rangle$ defined as follows:

1. the set of players is $\mathcal{P} = \{1, 2\}$, where player 1 represents *Charlie* and player 2 represents *Eve*;

2. $Q \subseteq [\Pi] \cup ([\Pi] \times \mathcal{M}_C)$ is the set of states, which is partitioned into the set $Q_1 = [\Pi]$ and the set $Q_2 \subseteq [\Pi] \times \mathcal{M}_C$ of pairs $([\overline{\mu}], \mathsf{m}_C)$ where $[\overline{\mu}]$ is a matching record and $\mathsf{m}_C$ is a *Charlie* move applicable to $[\overline{\mu}]$;

3. $\Sigma = \{d, w\}$ is a set of two propositions;

4. the valuation $\nu$ is such that $\nu(q) = \varnothing$ for all $q \in Q_2$, for all $[\overline{\mu}] \in Q_1$, $d \in \nu([\overline{\mu}])$ iff $q \models P_{\mathcal{D}}$ and $w \in \nu([\overline{\mu}])$ iff $[\overline{\mu}] \models P_{\mathcal{D}}$ and $[\overline{\mu}] \models P_{\mathcal{S}}$;

5. $\lambda(q) = 1$ if $q \in Q_1$ and $\lambda(q) = 2$ if $q \in Q_2$;

6. the transition relation is bipartite, relating only states from $Q_1$ to states from $Q_2$ or *vice versa*, and is defined as follows:

   (a) $([\overline{\mu}], ([\overline{\mu}'], \mathsf{m}_C)) \in R$ if and only if $[\overline{\mu}] = [\overline{\mu}']$;

   (b) $(([\overline{\mu}], \mathsf{m}_C), [\overline{\mu}']) \in R$ if and only if there is an *Eve*'s move $\mathsf{m}_E$ such that the round $\rho = (\mathsf{m}_C, \mathsf{m}_E)$ is applicable to $\overline{\mu}$, and $[\rho(\overline{\mu})] = [\overline{\mu}']$.

Hence, the game structure follows the game by connecting two states if there is a corresponding move of the game. The partition into two kind of states represents the division in turns where *Charlie* plays first without knowing *Eve*'s move, and then *Eve* moves knowing both the current partial plan and *Charlie*'s move. Note that general strategies for game structures as defined above are stateful, since they are functions $f : Q^+ \to Q$ which account for the entire history up to the last state. However, by construction, a single state of $S_G$ already sums up all the necessary information about the history of the game that brought the players to reach that state, hence *w.l.o.g.* we can consider *state-less* strategies $f : Q \to Q$, *i.e.*, this game is *positional*. Then, because of how the states are partitioned and how the transition relation is defined, we can see that a strategy for player 1 consists in a function $f_1 : [\Pi] \to \mathcal{M}_C$ which, given any state $[\overline{\mu}] \in Q_1$ belonging to player 1, chooses *Charlie*'s move $\mathsf{m}_C$, selecting the next state $([\overline{\mu}], \mathsf{m}_C)$. Similarly, a strategy for player 2 (*i.e.*, *Eve*) is a function $f_2 : [\Pi] \times \mathcal{M}_C \to \mathcal{M}_E$, which from a state $([\overline{\mu}], \mathsf{m}_C) \in Q_2$, which belongs to player 2, selects a move $\mathsf{m}_E$ for *Eve*, and consequently the next state $[\overline{\mu}'] \in Q_1$ such that $[\overline{\mu}'] = [\rho(\overline{\mu})]$ for $\rho = (\mathsf{m}_C, \mathsf{m}_E)$. Hence we can notice that strategies for player 1 and 2 in $S_G$ corresponds to strategies for *Charlie* and *Eve*, respectively, in $G$. Now, since we labelled each state $[\overline{\mu}] \in Q_1$ either with $d$ or $w$ depending on whether $[\overline{\mu}]$ was *admissible* or *winning*, we can find whether a strategy for *Charlie* exists by model-checking over $S_G$ the following $\mathsf{ATL}^*$ formula $\phi$:

$$\phi \equiv \langle\!\langle 1 \rangle\!\rangle(\mathsf{F}d \to \mathsf{F}w)$$

Note that $\phi$ corresponds to $\neg\langle\!\langle 2 \rangle\!\rangle(\mathsf{F}d \wedge \mathsf{G}\neg w)$, *i.e.*, we are asking whether player 2 *cannot* maintain $w$ false (*i.e.*, system rules unsatisfied), while behaving well (*i.e.*, domain rules satisfied). Hence, it can be seen that a winning strategy for *Charlie* exists if and only if the above formula holds on $S_G$. Since the size of $[\overline{\mu}]$ is exponential in the size of $G$ (by Lemma 4.21), the size of $Q$ is doubly exponential in the size of $G$. Now, we know that ATL* model checking over a fixed-size formula can be solved in polynomial time in the size of the structure [7], and each node can by itself be built in exponential time as noted in the proof of Lemma 4.22. Hence we obtained a procedure to solve the problem running in *doubly-exponential time*. ■

# 5 CONCLUSIONS AND OPEN QUESTIONS

This chapter provided our take at the problem of timeline-based planning with uncertainty. Flexible plans, as currently adopted by most state-of-the-art timeline-based systems, are inherently sequential, and are designed to handle temporal uncertainty without considering further sources of nondeterminism. However, here we have shown that this exclusive focus is in contrast with the syntax of the modelling language, which can express problems that require to handle general nondeterminism. The sequentiality of flexible plans also forces some systems, such as PLATINUm [134], to employ a feedback loop including a re-planning phase to handle any non-temporal mismatch between the expected and actual behaviours of the environment.

To overcome these issues, we introduced the notion of *timeline-based games*, that generalise timeline-based planning problems with uncertainty as defined by Definition 2.19. We have shown that the approach is strictly more general than the current one based on dynamic controllability of flexible plans, and then we provided a procedure, running in doubly-exponential time, to decide whether a winning strategy exists for a given game.

We believe that the approach introduced here is worth exploring in a number of different directions. First of all, of course, a lower bound on the complexity of the problem of the existence of winning strategies is needed, to settle the complexity issue. We conjecture that, similarly to what done in Section 3.3 for timeline-based planning problems with bounded horizon, a reduction from *domino-tiling games* [41] can be used to obtain a matching lower bound, showing that the problem is 2EXPTIME-complete. This would solve the problem of finding whether a winning strategy exists for a given game, but then, the problem of the *synthesis* of controllers to implement such strategies will have to be addressed as well.

Then, to test the applicability of the approach, the decision procedure shown here must be turned into a form that can be implemented in a reasonably efficient way. Although the framework of ATL* model-checking has come useful to prove Theorem 14, an ad-hoc algorithm to check the existence of the strategy, with a symbolic representation of the game structure, may be the needed ingredients. In this regard, it is worth to note that the automata-theoretic construction shown in Section 4 of Chapter 4 might be an alternative underlying platform for the construction of the concurrent game structure.

Once the basic questions about the approach will be answered, it can be foreseeably extended in a number of directions. First of all, note that the definitions of strategies and of winning condition given here are the ones apt to precisely capture the semantics of dynamically controllable flexible plans as defined by Cialdea Mayer et al. [44], but are not the only ones that can be sensibly defined over the same game arena. In particular, the game might be defined over infinite executions, letting *Charlie* win if it can satisfy the system rules infinitely often. Then, a number of extensions of the game arena can be imagined as well, such as multiple-players extensions, with players belonging to teams competing with each other, but that cooperates inside the same team to obtain a common goal while also trying to achieve the personal goals of each. Then, variants with partial observability of the timelines belonging to other players can be defined. Furthermore, a distributed version of the game might be defined, where there is not a single unique clock and communication happens through message passing.

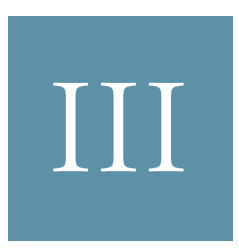

# TIMELINE-BASED PLANNING AND TEMPORAL LOGIC

# TABLEAU METHODS FOR LINEAR TEMPORAL LOGICS 6

The last part of this dissertation draws a connection between timeline-based planning and temporal logics. This chapter introduces *tableau methods*, a class of satisfiability checking techniques that plays a role in this connection. A recently introduced *one-pass tree-shaped* tableau method for LTL satisfiability checking is studied, providing a novel and more insightful completeness proof. Then, after reviewing an experimental evaluation of the method, we extend it to support *past operators*. The presented work opens the way for the adaptation of the method to the $\mathsf{TPTL_b{+}P}$ logic used in the next chapter to capture timeline-based planning problems.

# INTRODUCTION

*Linear Temporal Logic* (LTL) is a propositional modal logic originally introduced as a specification language for the behaviour of reactive systems [113]. Since its inception, LTL has become the de-facto standard language to express properties of systems in the field of *formal verification* of hardware and software systems [50]. Its success is due mainly to its simplicity, which leads to specifications fairly similar to their natural language counterparts. In this context, the *model checking* problem for LTL formulae (deciding if a given structure satisfies a formula) has been thoroughly studied over the last years. However, with the specifications and the modelled systems constantly growing in size and complexity, it becomes more and more difficult for knowledge engineers to be confident in the correctness of the specifications, which thus need to be checked against common modelling errors. For example, checking a system against a valid specification (*i.e.*, a formula which is trivially true in every structure), is useless at the least, and could be severely harmful at worst: the positive answer from the model checking procedure may convince the designers of the system of its safety while potentially life-threatening bugs may instead be present. For this reason, a growing importance is being given to the *sanity check* of specifications, which is an application of the *satisfiability checking* problem, that is, deciding whether there exists a model satisfying a given formula.

The satisfiability problem for LTL and other temporal logics also has interesting applications in artificial intelligence, and in planning in particular. As it turns out, it can be proved [42] that classical STRIPS-like planning problems can be expressed as LTL formulae, whose satisfiability corresponds to the existence of a solution plan for the problem. Furthermore, with the release of PDDL 3 [67] supporting *temporally extended goals* [8], LTL (or rather, the LTL-inspired syntactic extensions to PDDL) becomes the modelling language used to express goals, background theory, and control knowledge.

In the case of LTL, satisfiability checking can be reduced to model checking, and both problems are PSPACE-complete [127]. Nevertheless, a great research effort has been dedicated over the years to techniques specifically targeted at satisfiability checking of LTL formulae. Besides model checkers that directly support satisfiability checking, like, for instance, NuSMV [45], many tools have been developed based on a wide range of different techniques. An interesting approach is to reduce the problem to the emptiness problem of *Büchi automata*, an approach adopted by tools like Aalta [86]. An alternative approach is *temporal resolution*, which was pioneered by Cavalli and Cerro [27] and Venkatesh [137], and later employed by Fisher et al. [58]. Such an approach is also at the core of the *labelled superposition* technique by Suda and Weidenbach [132]. Recently, SAT-based techniques have also gained attention [88]. A number of surveys exist comparing the performance of tools based on these and other techniques [87, 121, 124].

This chapter focuses on *tableau methods*, another class of decision procedures for LTL satisfiability, which were among the first to be investigated [141]. Originally devised for propositional logic [18] and later adapted to many other logics during the last half of a century [51], tableau-based techniques provide a useful theoretical tool to reason about the proof theory of the considered logic, as they are tightly related to cut-free Gentzen-style sequent calculi for the given logic. Often – this is the case for LTL – tableau methods provide the easiest to understand decision procedures for the logic.

In contrast to the tableau method for propositional and first-order logic, most of early tableaux for LTL produce *graph-shaped* structures [89, 94, 141]. After building the tableau itself, whose nodes are labelled with sets of formulae, the decision procedure traverses it to look for a specific kind of infinite path witnessing the existence of a model for the formula. This procedure turns out to be very easy to show and to prove correct, but also makes it impractical because of the huge size of the resulting graphs. Aiming at solving this issue, *incremental* tableaux were proposed [71, 81] which traverse only a limited part of the graph by ignoring nodes that are not effectively reachable from the initial ones, and creates the traversed nodes on the fly during the traversal. However, the size of the traversed part of the graph can still grow significantly. A different approach was proposed by Schwendimann [126], with a *one-pass* tableau system which works by building a graph structure which is almost a tree if not for a number of back-edges.

More recently, Reynolds introduced a one-pass tableau method for LTL which builds a purely *tree-shaped* structure [116]. Similarly to Schwendimann's, Reynolds' tableau only requires a single pass to decide about the acceptance or the rejection of a given branch of the search tree. Unlike Schwendimann's, however, Reynolds' system is a purely tree-shaped search procedure, where any branch can be explored independently from any other. Its rule-based tree search architecture combines the simplicity of classic declarative tableaux with the efficiency of a one-pass system. Despite its simplicity, or maybe thanks to it, the system also turns out to be efficient in practice. Indeed, a tool that implements the system (briefly described in Section 4) has shown superior performance with regards to other tableau methods, and competitive with other tools adopting different techniques. Moreover, it has shown to be very easily and effectively parallelised [97].

In this chapter, we study Reynolds' one-pass tree-shaped tableau method for LTL, introducing a few contributions. First, we provide a clean exposition of the tableau method itself, with full proof of its soundness and completeness. In contrast to the original exposition of the system [116], we employ a novel proof technique for the completeness result. The new proof is based on an argument which is conceptually simpler and more insightful, as it provides a greater understanding of the role of the most critical rule of the system. Then, we discuss some experimental results obtained with the aforementioned implementation of the system, that show how the simplicity of its rule based tree search pays off in terms of efficiency and ease of parallelisation.

Finally, we exploit this simplicity to extend the tableau system to support *past temporal operators*, hence providing a one-pass tree-shaped tableau method for *Linear Temporal Logic with Past* (LTL+P). It is well known that if we interpret the logic on structures with a definite starting point in time, then past modalities do not add any expressive power, *i.e.*, any formula with past modalities has an initially equivalent future-only counterpart [65, 66, 115]. Nevertheless, LTL+P is interesting for a number of reasons [89, 90]. Most importantly, many relevant properties are easier to express using past modalities, allowing specifications to match more closely the way they are expressed in natural language. Furthermore, if we consider the size of specifications, past-time modalities do add expressive power, as LTL+P is exponentially more succinct than LTL [95]. The extension to past operators of Reynolds' tableau system is thus a natural step, and, moreover, it opens the path to provide a one-pass tree-shaped tableau method for the timed temporal logics used in Chapter 7 to capture timeline-based planning problems.

The simplicity of the system allows us to extend it to LTL+P (and to other logics in the next chapter) in a very modular way: only three rules have to be added, while any existing rule remains unchanged, and applying the system to a future-only formula leads to the same computation that would result from the future-only tableau. The novel proof technique used to prove the completeness of the system for LTL can be directly extended to the LTL+P case as well. The contribution is thus twofold: while providing evidence of how easily extendable the system is, we also further improve its extensibility by providing an easier proof technique to build such extensions upon.

The chapter is structured as follows. The rest of this section recalls the syntax and semantics of LTL and LTL+P, and provides a basic exposition of the classic graph-shaped tableau, for comparison with the one-pass tree-shaped one. Then, Section 2 provides a clean exposition of the tableau system itself, and Section 3 provides full proofs of its soundness and completeness, adopting our novel, simpler proof technique. Then, Section 4 reports the results of the experiments on the aforementioned implementation. Finally, in Section 5 shows the extension of the system LTL+P also proving its soundness and completeness. Section 6 concludes the chapter discussing the obtained results and potential for future work.

## 1.1 LINEAR TEMPORAL LOGIC

Let us now recall the precise syntax and semantics of this well-known logic. *Linear Temporal Logic* (LTL) is a propositional temporal logic interpreted over infinite, discrete linear orders. *Linear Temporal Logic with Past* (LTL+P) extends LTL with the addition of temporal operators able to talk about what happened in the *past* respect to the current time. We will now briefly recall the syntax and semantics of LTL+P, which encompasses that of LTL as well.

Syntactically, LTL can be seen as an extension of propositional logic with the addition of the *tomorrow* ($\mathsf{X}\phi$, *i.e.*, at the *next* state $\phi$ holds) and *until* ($\phi_1 \mathcal{U} \phi_2$, *i.e.*, $\phi_2$ will eventually hold and $\phi_1$ will hold *until* then) temporal operators. LTL+P is obtained from LTL by adding the *yesterday* ($\mathsf{Y}\phi$, *i.e.*, at the *previous* state $\phi$ holds) and *since* ($\phi_1 \mathcal{S} \phi_2$, *i.e.*, there was a past state where $\phi_2$ held, and $\phi_1$ has held *since* then) *past* temporal operators. Formally, given a set $\Sigma$ of proposition letters, LTL+P formulae over $\Sigma$ are generated by the following grammar:

$$
\begin{array}{lr}
\phi := p \mid \neg\phi \mid \phi_1 \vee \phi_2 \mid \phi_1 \wedge \phi_2 & \text{propositional connectives} \\
\qquad \mid \mathsf{X}\phi_1 \mid \phi_1 \mathcal{U} \phi_2 \mid \phi_1 \mathcal{R} \phi_2 \mid \mathsf{F}\phi_1 \mid \mathsf{G}\phi_1 & \textit{future}\text{ temporal operators} \\
\qquad \mid \mathsf{Y}\phi_1 \mid \phi_1 \mathcal{S} \phi_2 \mid \phi_1 \mathcal{T} \phi_2 \mid \mathsf{P}\phi_1 \mid \mathsf{H}\phi_1 & \textit{past}\text{ temporal operators}
\end{array}
$$

where $p \in \Sigma$ and $\phi_1$ and $\phi_2$ are LTL+P formulae. Most of the temporal operators of the language can be defined in terms of a small number of basic ones. In particular, the *release* ($\phi_1 \mathcal{R} \phi_2 \equiv \neg(\neg\phi_1 \mathcal{U} \neg\phi_2)$), *eventually* ($\mathsf{F}\phi \equiv \top \mathcal{U} \phi$), and *always* ($\mathsf{G}\phi \equiv \neg\mathsf{F}\neg\phi$) future operators can all be defined in terms of the *until* operator, while the *triggered* ($\phi_1 \mathcal{T} \phi_2 \equiv \neg(\neg\phi_1 \mathcal{S} \neg\phi_2)$), *once* ($\mathsf{P}\phi \equiv \top \mathcal{S} \phi$), and *historically* ($\mathsf{H}\phi \equiv \neg\mathsf{P}\neg\phi$) past operators can all be defined in terms of the *since* operator. However, in our setting, it is useful to consider them as primitive.

A *model* for an LTL+P formula $\phi$ is an infinite sequence of sets of proposition letters, *i.e.*, $\overline{\sigma} = \langle \sigma_0, \sigma_1, \ldots \rangle$, where each *state* $\sigma_i \subseteq \Sigma$ represents the proposition letters that hold in the state. Given a model $\overline{\sigma}$, a position $i \geq 0$, and an LTL+P formula $\phi$, we inductively define the *satisfaction* of $\phi$ by $\overline{\sigma}$ at position $i$, written as $\overline{\sigma}, i \models \phi$, as follows:

1. $\overline{\sigma}, i \models p$ iff $p \in \sigma_i$;
2. $\overline{\sigma}, i \models \neg\phi$ iff $\overline{\sigma}, i \not\models \phi$;
3. $\overline{\sigma}, i \models \phi_1 \vee \phi_2$ iff $\overline{\sigma}, i \models \phi_1$ or $\overline{\sigma}, i \models \phi_2$;
4. $\overline{\sigma}, i \models \phi_1 \wedge \phi_2$ iff $\overline{\sigma}, i \models \phi_1$ and $\overline{\sigma}, i \models \phi_2$;
5. $\overline{\sigma}, i \models \mathsf{X}\phi$ iff $\overline{\sigma}, i+1 \models \phi$;
6. $\overline{\sigma}, i \models \mathsf{Y}\phi$ iff $i > 0$ and $\overline{\sigma}, i-1 \models \phi$;
7. $\overline{\sigma}, i \models \phi_1 \mathcal{U} \phi_2$ iff there exists $j \geq i$ such that $\overline{\sigma}, j \models \phi_2$, and $\overline{\sigma}, k \models \phi_1$ for all $k$, with $i \leq k < j$;

8. $\overline{\sigma}, i \models \phi_1 \mathcal{S} \phi_2$ iff there exists $j \leq i$ such that $\overline{\sigma}, j \models \phi_2$, and $\overline{\sigma}, k \models \phi_1$ for all $k$, with $j < k \leq i$;
9. $\overline{\sigma}, i \models \phi_1 \mathcal{R} \phi_2$ iff either $\overline{\sigma}, j \models \phi_2$ for all $j \geq i$, or there exists $k \geq i$ such that $\overline{\sigma}, k \models \phi_1$ and $\overline{\sigma}, j \models \phi_2$ for all $i \leq j \leq k$;
10. $\overline{\sigma}, i \models \phi_1 \mathcal{T} \phi_2$ iff either $\overline{\sigma}, j \models \phi_2$ for all $0 \leq j \leq i$, or there exists $k \leq i$ such that $\overline{\sigma}, k \models \phi_1$ and $\overline{\sigma}, j \models \phi_2$ for all $i \geq j \geq k$

We say that $\overline{\sigma}$ *satisfies* $\phi$, $\overline{\sigma} \models \phi$, if it satisfies the formula at the first state, *i.e.*, if $\overline{\sigma}, 0 \models \phi$. This is often called *initial* satisfiability, as opposed to the notion of *global* satisfiability, where a formula is said to be satisfied by a model if it holds at some position of the model. Without loss of generality, we restrict our attention to initial satisfiability. It can be indeed easily shown that $\phi$ is globally satisfiable if $\mathsf{F}\phi$ is initially satisfiable. We will also make use of the notion of *initial* equivalence between formulae, *i.e.*, we say that two formulae $\phi$ and $\psi$ are *equivalent* ($\phi \equiv \psi$) if and only if they are satisfied by the same set of models at the initial state.

A *literal* $\ell$ is a formula of the form $p$ or $\neg p$, where $p \in \Sigma$. Since we defined the *release* and *triggered* operators as primitive instead of defining them in terms of the *until* and *since* operators, any LTL+P formula can be turned into its *negated normal form*: a formula $\phi$ can be translated into an equivalent one $\mathsf{nnf}(\phi)$ where *negations* appear only in literals, by exploiting the duality between the *until/release* and *since/triggered* pairs of operators, and classic *De Morgan* laws (the *tomorrow* and *yesterday* operators are their own duals). In the whole chapter, we will assume *w.l.o.g.* that all the considered formulae are in negated normal form.

## 1.2 GRAPH-SHAPED TABLEAUX

This section recalls how classic graph-shaped tableau systems for LTL and LTL+P work. The presentation follows the one given in [89]. Presenting a graph-shaped tableau system will allow us to better highlight the differences with the one-pass tree-shaped system that is the topic of this chapter.

The *closure* of $\phi$ is the set of formulae, some of which are subformulae of $\phi$, that are sufficient to be considered to decide the satisfiability of $\phi$.

**Definition 6.1** — Closure.
*The* closure *of an* LTL+P *formula* $\phi$ *is the set* $\mathcal{C}(\phi)$ *of formulae defined as follows:*

1. $\phi \in \mathcal{C}(\phi)$;
2. *if* $\psi \in \mathcal{C}(\phi)$, *then* $\mathsf{nnf}(\neg\phi) \in \mathcal{C}(\phi)$;
3. *if* $\circ\phi \in \mathcal{C}(\phi)$, *with* $\circ \in \{\mathsf{X}, \mathsf{Y}\}$, *then* $\phi \in \mathcal{C}(\phi)$;

4. *if* $\phi_1 \circ \phi_2 \in \mathcal{C}(\phi)$*, with* $\circ \in \{\wedge, \vee\}$*, then* $\{\phi_1, \phi_2\} \subseteq \mathcal{C}(\phi)$*;*

5. *if* $\phi_1 \circ \phi_2 \in \mathcal{C}(\phi)$*, with* $\circ \in \{\mathcal{U}, \mathcal{R}\}$*, then* $\{\phi_1, \phi_2, \mathsf{X}(\phi_1 \circ \phi_2)\} \subseteq \mathcal{C}(\phi)$*;*

6. *if* $\phi_1 \circ \phi_2 \in \mathcal{C}(\phi)$*, with* $\circ \in \{\mathcal{S}, \mathcal{T}\}$*, then* $\{\phi_1, \phi_2, \mathsf{Y}(\phi_1 \circ \phi_2)\} \subseteq \mathcal{C}(\phi)$*;*

The *tableau* for $\phi$ is a graph where the nodes, called *atoms* are maximally consistent sets of formulae from the closure of $\phi$.

**Definition 6.2** — Atom of the graph-shaped tableau.
*An* atom *for an* LTL+P *formula* $\phi$ *is a subset* $\Delta \subseteq \mathcal{C}(\phi)$ *of the closure of* $\phi$ *such that:*

1. *for each* $p \in \Sigma$*,* $p \in \Delta$ *if and only if* $\neg p \notin \Delta$*;*

2. *if* $\phi_1 \vee \phi_2 \in \Delta$*, then* $\phi_1 \in \Delta$ *or* $\phi_2 \in \Delta$ *(or both);*

3. *if* $\phi_1 \wedge \phi_2 \in \Delta$*, then both* $\phi_1 \in \Delta$ *and* $\phi_1 \in \Delta$*;*

4. *if* $\phi_1 \,\mathcal{U}\, \phi_2 \in \Delta$*, then* $\phi_2 \in \Delta$ *or* $\{\phi_1, \mathsf{X}(\phi_1 \,\mathcal{U}\, \phi_2)\} \subseteq \Delta$*;*

5. *if* $\phi_1 \,\mathcal{S}\, \phi_2 \in \Delta$*, then* $\phi_2 \in \Delta$ *or* $\{\phi_1, \mathsf{Y}(\phi_1 \,\mathcal{S}\, \phi_2)\} \subseteq \Delta$*;*

6. *if* $\phi_1 \,\mathcal{R}\, \phi_2 \in \Delta$*, then* $\{\phi_1, \phi_2\} \subseteq \Delta$ *or* $\{\phi_2, \mathsf{X}(\phi_1 \,\mathcal{R}\, \phi_2)\} \subseteq \Delta$*;*

7. *if* $\phi_1 \,\mathcal{T}\, \phi_2 \in \Delta$*, then* $\{\phi_1, \phi_2\} \subseteq \Delta$ *or* $\{\phi_2, \mathsf{Y}(\phi_1 \,\mathcal{T}\, \phi_2)\} \subseteq \Delta$*.*

We can then formally define the tableau for $\phi$.

**Definition 6.3** — Graph-shaped tableau for a LTL formula.
*The* tableau *for an* LTL+P *formula* $\phi$ *is a graph* $G_\phi = (V, E)$*, where* $V \subseteq 2^{\mathcal{C}(\phi)}$ *is the set of all the* atoms *for* $\phi$*, and* $E \subseteq V \times V$ *is such that, for any edge* $(\Delta, \Delta') \in E$*:*

1. $\mathsf{X}\phi_1 \in \Delta$ *if and only if* $\phi_1 \in \Delta'$*;*

2. $\mathsf{Y}\phi_1 \in \Delta'$ *if and only if* $\phi_1 \in \Delta$*.*

Intuitively, atoms represent sets of formulae that can possibly be true at the same time in any given state of a model for $\phi$. An edge between two atoms $\Delta$ and $\Delta'$ is present in the tableau if $\Delta'$ represents a state that can be a successor of $\Delta$ in a model for $\phi$, respecting the requests imposed by *tomorrow* and *yesterday* operators. The semantics of the temporal operators such as, for example, the *until* operator $\phi_1 \,\mathcal{U}\, \phi_2$, is implemented by *expanding* them into a present and a future part, that is then postponed to the next state. In the case of the *until* operator, if $\phi_1 \,\mathcal{U}\, \phi_2$ holds in any given state, then by the semantics of the operator it follows that either $\phi_2$ holds now, or $\phi_1$ holds and $\phi_1 \,\mathcal{U}\, \phi_2$ holds again at the next state. Hence, if an atom $\Delta$ contains $\phi_1 \,\mathcal{U}\, \phi_2$, then it has to contain either $\phi_2$ or $\phi_1$ and $\mathsf{X}(\phi_1 \,\mathcal{U}\, \phi_2)$. The constraint on the edge relation given in Definition 6.3 then forces $\phi_2 U \phi_2$ to appear in any successor of $\Delta$.

An atom $\Delta$ is called *initial* if $\phi \in \Delta$ and there is no $\mathsf{Y}\phi \in \Delta$. From an infinite sequence of atoms $\overline{\Delta} = \langle \Delta_0, \Delta_1, \ldots \rangle$ one can extract a state sequence $\overline{\sigma} = \langle \sigma_0, \ldots \rangle$ by stating that $\sigma_i = \Delta_i \cap \Sigma$ for all $i \geq 0$. Any path in the tableau starting from an initial atom hence represents a state sequence which is a candidate model for $\phi$. The path is effectively a model for the formula if the satisfaction of the formulae requested by the *until* operators are not postponed forever.

As proved in [127], any satisfiable LTL (and LTL+P, as they are expressively equivalent) formula has a *periodic* model, *i.e.*, a model $\overline{\sigma} = \overline{\sigma}_1 \overline{\sigma}_2^{\omega}$, where $\overline{\sigma}_1$ is the finite *prefix* and $\overline{\sigma}_2$ is the *period*, which repeats infinitely. As a consequence, the satisfiability of $\phi$ can be characterised by the presence in $G_\phi$ of a cycle, reachable from an initial atom, where for any $\phi_1 \mathcal{U} \phi_2$ in any atom $\Delta$ of the cycle, $\phi_2 \in \Delta'$ in some other $\Delta'$ in the cycle. Rather than explicitly looking for such a cycle in the graph, however, it is sufficient to look for a *self-fulfilling strongly connected component* (self-fulfilling SCC) of $G_\phi$, *i.e.*, an SCC of $G_\phi$, reachable from an initial atom, such that for any *until* operator present in any atom of the component, another atom of the component contains the requested formula. In contrast to the search for cycles, the SCC decomposition can be done in linear time in the size of the graph.

**Theorem 15** — Lichtenstein and Pnueli [89], Theorem 3.16.
Let $\phi$ be an LTL+P formula. Then, $\phi$ is satisfiable if and only if $G_\phi$ contains a strongly connected component $C$, reachable from an initial atom, such that for any $\Delta \in C$ and any $\phi_1 \mathcal{U} \phi_2 \in \Delta$, there is a $\Delta'$ such that $\phi_2 \in \Delta'$.

The satisfiability checking procedure then works by building the tableau structure $G_\phi$ from $\phi$, and then applying the SCC decomposition to look for self-fulfilling SCCs. A few observation can be made on this method. First, regarding the support for past operator, it is worth to note that removing Items 5 and 7 from Definition 6.2 and Item 2 from Definition 6.3 yields a tableau system for pure-future LTL. Note that the definition of self-fulfilling SCC does not involve formulae requested by the *since* operator, as those cannot be postponed forever in the past, since each path has a definite starting point.

From a practical standpoint, this method is difficult to implement in an efficient way. As the number of possible different atoms is exponential in the size of the formula, the graph is in the worst case exponentially large. Several optimisations are possible on top of this basic procedure. For example, by generating the graph structure incrementally while looking for the fulfilling SCC, it is possible to obtain notable speedups, but that would be still not optimal, since the LTL satisfiability problem is PSPACE-complete [127]. The non-optimal computational complexity is a limitation shared by all the tableau methods that will be discussed in this and the next chapter.

# 2 THE ONE-PASS TREE-SHAPED TABLEAU FOR LINEAR TEMPORAL LOGIC

This section recalls Reynolds' one-pass tree-shaped tableau for LTL [116]. The next section will then show the soundness and completeness of the system, exploiting our major contribution on this front which is the simpler overhauled completeness proof.

A tableau for $\phi$ is a tree $T$ where each node $u$ is labelled by a subset $\Gamma(u)$ of the closure $C(\phi)$ and the label of the root node $u_0$ contains only $\phi$, *i.e.*, $\Gamma(u_0) = \{\phi\}$. The tableau is built recursively from the root, at each step applying one from a set of *rules* to a leaf of the tree. Each rule can add one or two children to the current node, advancing the construction of the tree, or close the current branch by *accepting* (✓) or *rejecting* (✗) the current node. At the end of the section, the construction is proved to terminate with a *complete* tableau, *i.e.*, one where all the leaves are either ticked or crossed. Once a complete tableau is obtained, the formula is recognised as satisfiable if there is at least one accepted branch. Given two nodes $u$ and $v$, we write $u \leq v$ to mean that $u$ is an ancestor of $v$, and $u < v$ to mean that $u$ is a *proper* ancestor of $v$, *i.e.*, $u \leq v$ and $u \neq v$.

The construction of a branch of the tree can be seen as the search for a model of the formula in a state-by-state way. At each step, *expansion rules* are applied first, building a possible assignment of proposition letters for the current state, which is then checked for the absence of contradictions. Next, the *termination rules* are checked to possibly detect if the construction can be stopped. At the end, information about the current state is used to determine the next one, and the construction proceeds by executing a temporal *step*.

The *expansion rules* look for a specific formula into the label, creating one or two children whose labels are obtained by replacing the target formula with some others. Table 6.1 shows the expansion rules with the following notation: each rule looks for a formula $\phi \in \Gamma(u)$, where $u$ is the current node, and two children $u'$ and $u''$ of $u$ are created with labels $\Gamma(u') = \Gamma(u) \setminus \{\phi\} \cup \Gamma_1(\phi)$ and $\Gamma(u'') = \Gamma(u) \setminus \{\phi\} \cup \Gamma_2(\phi)$, where $\Gamma_1(\phi)$ and $\Gamma_2(\phi)$ are defined in Table 6.1. If the corresponding expansion rule has an empty $\Gamma_2(\phi)$, a single child is added. The rules for the primitive operators such as the *until* operator would be sufficient, but Table 6.1 shows as well some derived rules for operators such as the *eventually* and *always* operators, for ease of exposition. Note that a common feature of all the expansion rules is that the expanded formulae are entailed by their replacement, *i.e.*, both $\Gamma_1(\phi) \models \phi$ and $\Gamma_2(\phi) \models \phi$.

After a finite number of applications of expansion rules, the construction will eventually reach a node whose label contains only *elementary* formulae, *i.e.*, only formulae of the forms $p$ or $\neg p$, for some $p \in \Sigma$, or $\mathsf{X}\alpha$ for some $\alpha \in C(\phi)$ (while $p$ and $\neg p$ cannot both belong to the same label, it can be the case that

| Rule | $\phi \in \Gamma$ | $\Gamma_1(\phi)$ | $\Gamma_2(\phi)$ |
|---|---|---|---|
| DISJUNCTION | $\alpha \vee \beta$ | $\{\alpha\}$ | $\{\beta\}$ |
| UNTIL | $\alpha\, \mathcal{U}\, \beta$ | $\{\beta\}$ | $\{\alpha, \mathsf{X}(\alpha\, \mathcal{U}\, \beta)\}$ |
| RELEASE | $\alpha\, \mathcal{R}\, \beta$ | $\{\alpha, \beta\}$ | $\{\beta, \mathsf{X}(\alpha\, \mathcal{R}\, \beta)\}$ |
| EVENTUALLY | $\mathsf{F}\alpha$ | $\{\alpha\}$ | $\{\mathsf{XF}\alpha\}$ |
| CONJUNCTION | $\alpha \wedge \beta$ | $\{\alpha, \beta\}$ | |
| ALWAYS | $\mathsf{G}\alpha$ | $\{\alpha, \mathsf{XG}\alpha\}$ | |

**Table 6.1:** Tableau expansion rules. When a formula $\phi$ of one of the types shown in the table is found in the label $\Gamma$ of a node $u$, one or two children $u'$ and $u''$ are created with the same label as $u$ excepting for $\phi$ which is replaced, respectively, by the formulae from $\Gamma_1(\phi)$ and $\Gamma_2(\phi)$.

both $\mathsf{X}\alpha$ and $\mathsf{X}\neg\alpha$ belong to it, in which case the contradiction will be detected at a later stage). Such a label is called a *poised label* and the node is called a *poised node*. A *poised branch* $\overline{u} = \langle u_0, \ldots, u_n \rangle$ is a branch of the tableau where $u_n$ is a poised node. Intuitively, a poised node represents a guess for the truth of elementary formulae at the current state. Once a poised node has been obtained, the search can proceed to the next state, by applying the STEP rule:

STEP Given a poised branch $\overline{u} = \langle u_0, \ldots, u_n \rangle$, a child $u_{n+1}$ is added to $u_n$, with $\Gamma(u_{n+1}) = \{\alpha \mid \mathsf{X}\alpha \in \Gamma(u_n)\}$.

It is worth to make a couple of observations at this point. First of all, note that the expansion rules can be seen as a more schematic way of expressing the definition of *atom* of a graph-shaped tableau as defined in Definition 6.2, and the STEP rule corresponds to Item 1 of Definition 6.3. At the same time, note that some combinations are missing. In a graph-shaped tableau, an atom may exist containing a disjunction $\alpha \vee \beta$ and both its disjuncts, while in this system, one child of a node containing $\alpha \vee \beta$ contains $\alpha$, and the other contains $\beta$, but there is no child necessarily containing both (unless both have been expanded by other means). During the construction, adding either formula to the opposite child would still result in locally consistent labels, but would be redundant, since once one of the two disjuncts has been chosen to hold, the truth of the other is irrelevant. A single node of the one-pass tree-shaped system can thus be viewed as a representative of a set of atoms from the graph-shaped tableau that ignores some formulae whose truth is irrelevant in that particular point of the model. Similarly, a branch of the tree can be seen as a set of paths of the graph-shaped tableau which differ only regarding the truth value of these irrelevant formulae.

Since the STEP rule represents an advancement in time, (some of the) poised nodes will be used later to extract a model of $\phi$ from a successful tableau branch.

Before this can happen, the label of each poised node has to be checked for the absence of *contradictions*. If the check succeeds, then the STEP rule can be applied to advance to the next temporal state; otherwise, the branch is crossed.

CONTRADICTION If $\{p, \neg p\} \subseteq \Gamma(u_n)$, for some $p \in \Sigma$, then $u_n$ is *crossed*.

The rules introduced so far allow us to build a tentative model for the formula step-by-step, but we introduced only rules that can reject wrong branches, thus we still need to specify how to recognise good branches corresponding to actual models. The first obvious case in which we have to stop the construction is when there is nothing left to do:

EMPTY Given a branch $\overline{u} = \langle u_0, \ldots, u_n \rangle$, if $\Gamma(u_n) = \emptyset$, then $u_n$ is *ticked*.

Since LTL models are in general infinite, only tableaux for simple formulae will have branches satisfying the EMPTY rule. Others, *e.g.*, $\mathsf{GF}p$, at any state require something to happen in the future. Thus, some criteria are needed to ensure that the construction can find models exhibiting infinitary behaviours, while guaranteeing that the expansion of every branch eventually terminates.

For this reason, the system includes a pair of *termination rules*, the LOOP rule and the PRUNE rule, which are checked at each poised node. Note that the potentially infinite expansion of the branches is caused by the recursive nature of most of the expansion rules. The UNTIL rule (and the derived EVENTUALLY rule) is the most critical one. In the expansion of a formula of the form $\alpha\,\mathcal{U}\,\beta$, the rule tries to either fulfil the request for $\beta$ immediately or to postpone its fulfilment to a later step by adding $\mathsf{X}(\alpha\,\mathcal{U}\,\beta)$ to the label. A formula of the form $\mathsf{X}(\alpha\,\mathcal{U}\,\beta)$ is called X-*eventuality*. If an X-eventuality appears in a label, it means that some pending request still needs to be fulfilled in the future, and some criterion has to be used to avoid postponing its fulfilment indefinitely. The same arguments hold for formulae of the form $\mathsf{XF}\beta$, but since the EVENTUALLY rule can be derived from the UNTIL rule, we focus on the latter. Given a branch $\overline{u} = \langle u_0, \ldots, u_n \rangle$, an X-eventuality of the form $\mathsf{X}(\psi_1\,\mathcal{U}\,\psi_2)$ appearing in $\Gamma(u_i)$ for some $0 \leq i \leq n$ is said to be *requested* in $u_i$; moreover, we say that it is *fulfilled* in $u_k$, with $i < k$, if $\beta \in \Gamma(u_k)$ and $\alpha \in \Gamma(u_j)$ for all $i \leq j < k$. Moreover, we say that that it is fulfilled in a subsequence $\overline{u}_{[i\ldots j]}$ if it is fulfilled in $u_k$ for some $i < k \leq j$.

The LOOP and PRUNE rules, checked in this order, allow us to respectively handle the case of a branch that is repeating by successfully fulfilling recurrent requests and the case of a branch that is indefinitely postponing the fulfilment of an eventuality that is impossible to fulfil. They are defined as follows.

LOOP If there is a poised node $u_i < u_n$ such that $\Gamma(u_i) = \Gamma(u_n)$, and all the X-eventualities requested in $u_i$ are fulfilled in $\overline{u}_{[i+1\ldots n]}$, then $u_n$ is *ticked*.

PRUNE If there are two positions $i < j \leq n$, such that $\Gamma(u_i) = \Gamma(u_j) = \Gamma(u_n)$, and among the X-eventualities requested in these nodes, all those fulfilled in $\overline{u}_{[j+1\ldots n]}$ are fulfilled in $\overline{u}_{[i+1\ldots j]}$ as well, then $u_n$ is *crossed*.

Intuitively, the LOOP rule is responsible for recognising when a model for the formula has been found, while the PRUNE rule rejects branches that were doing redundant work without managing to fulfil any new eventuality. Then, if none of these two rules have closed the branch, the current state is ready and the construction can advance in time by applying the STEP rule. The PRUNE rule was the main novelty of this tableau system when introduced by Reynolds [116], and is the main focus of the novel completeness proof of Section 3.

The rules described above are repeatedly applied to the leaves of any open branch until all branches have been either ticked or crossed. This process is guaranteed to terminate.

**Theorem 16** — Termination.
Given a formula $\phi$, the construction of a (complete) tableau $T$ for $\phi$ will always terminate in a finite number of steps.

*Proof.* To start with, we observe that the tree has a finite branching degree, as any rule of the system creates at most two children. Thus, by König's lemma, for the construction to proceed forever the tree should contain at least one infinite branch. This, however, cannot be the case, because of the finite number of possible different labels and of different X-eventualities: after enough steps, any branch has to contain either two occurrences of the same label triggering the LOOP rule, or three occurrences of the same label triggering the PRUNE rule. Note that the LOOP rule is never triggered if the formula is unsatisfiable. However, if this is the case, the PRUNE rule will always eventually be because the set of X-eventualities is finite and the number of X-eventualities encountered along a branch from a given node is non-decreasing. ■

As for the computational complexity of the procedure, it can be seen that the whole decision procedure runs using at most an exponential amount of space: only a single branch at the time is needed to be kept in memory, and any branch cannot be longer than an exponential upper bound, given by the number of possible different labels that can appear on a branch before triggering either the LOOP or the PRUNE rule. The running time of the procedure is therefore at most doubly exponential in the size of the formula. The procedure is thus not optimal with regards to the complexity of LTL satisfiability, which is PSPACE-complete. However, on one hand, this is in line with the classic graph-shaped tableau-based decision procedure highlighted in Section 1.2 and with the one-pass system by Schwendimann [126]. On the other hand, the procedure can be turned into a PSPACE one by suitably exploiting nondeterminism, and then employing the well-known fact that PSPACE =NPSPACE [123]. However, such theoretically optimal version would not be of any practical use.

To get an intuitive understanding of how the tableau works for typical formulae, take a look at Figure 6.1, where two example tableaux are shown.

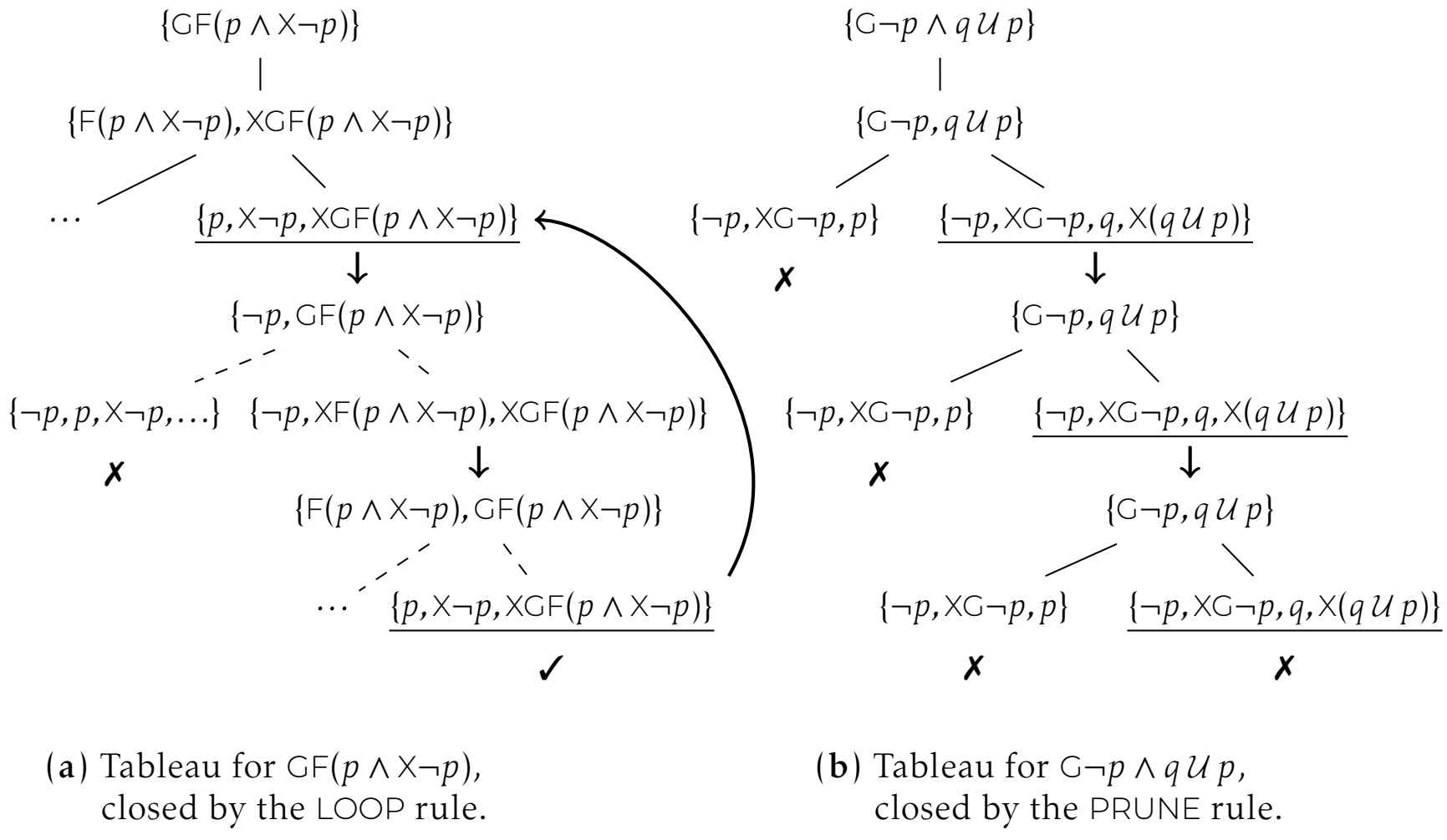

(**a**) Tableau for $\mathsf{GF}(p \land \mathsf{X}\neg p)$, closed by the LOOP rule.

(**b**) Tableau for $\mathsf{G}\neg p \land q\,\mathcal{U}\,p$, closed by the PRUNE rule.

**Figure 6.1:** Example tableaux for two formulae, involving the LOOP and PRUNE rules. Dashed edges represent subtrees collapsed to save space, bold arrows represent the application of a STEP rule to a poised label.

Figure 6.1a shows part of the tableau for the liveness formula $\mathsf{GF}(p \land \mathsf{X}\neg p)$. This formula requires something, *i.e.*, $p \land \mathsf{X}\neg p$, to happen infinitely often. As we go down any branch, we can see that the request $\mathsf{XGF}(p \land \mathsf{X}\neg p)$ is present at any poised label, propagated by the corresponding expansion rule. Then, any time $\mathsf{F}(p \land \mathsf{X}\neg p)$ is added to a label, the branch forks to choose between adding $p \land \mathsf{X}\neg p$ immediately or postponing it. When the request is fulfilled, $p$ cannot hold twice in a row, and this is handled by the CONTRADICTION rule, that fires when the wrong choice is made. Then, the LOOP rule is triggered by the rightmost branch, which is repeating the same label for the second time. The looping arrow does not represent a real edge, since otherwise it would not be a tree, but it is just a way to highlight to which label the loop is jumping to.

Figure 6.1b shows an example of application of the PRUNE rule in the tableau for the formula $\mathsf{G}\neg p \land q\,\mathcal{U}\,p$. The formula is unsatisfiable, not directly because of a propositional contradiction, but rather because the eventuality requested by $q\,\mathcal{U}\,p$ cannot be realised for the presence of the $\mathsf{G}\neg p$ component. The expansion of the *until* operator will then try to realise $p$ at each step, each time resulting into a propositional contradiction. The rightmost branch would then continue postponing the X-eventuality forever, if not for the PRUNE rule which crosses the branch after the third repetition of the same label (with no X-eventuality fulfilled in between).

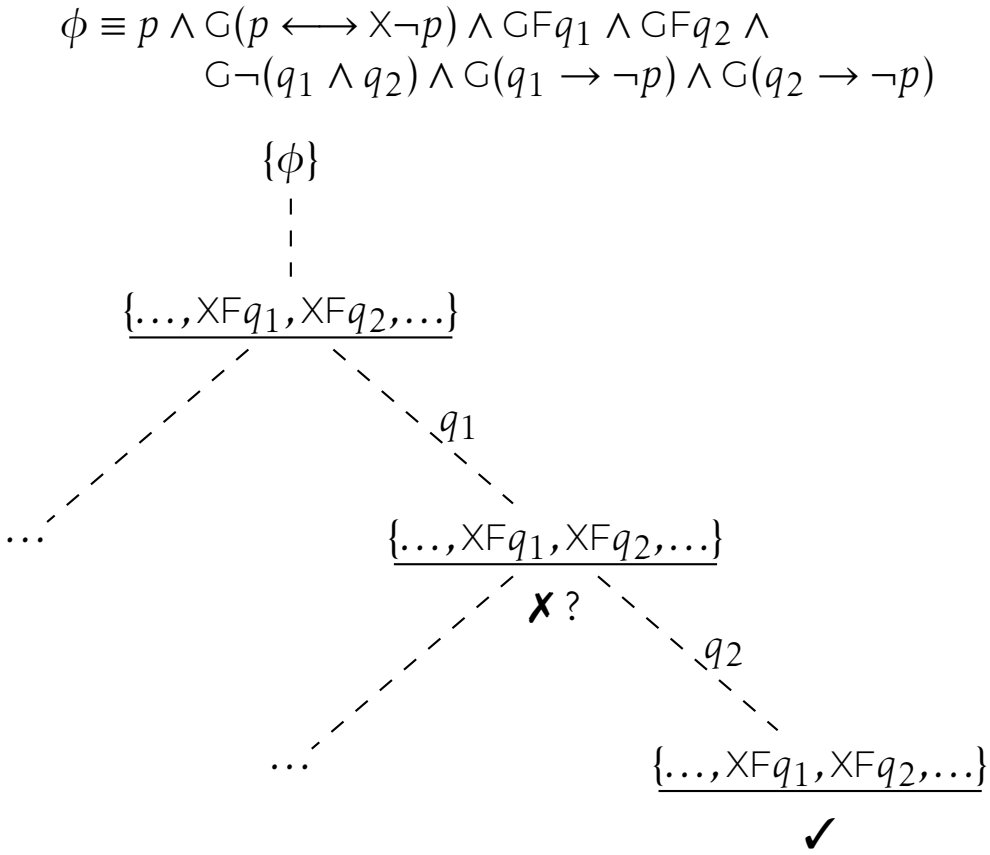

**Figure 6.2:** Example of why the PRUNE rule waits for *three* repetitions of the same label.

One may wonder why the PRUNE rule needs to look for the third occurrence of a label before triggering. An enlightening example is given in Figure 6.2. The formula $\phi$ shown in the figure requires $q_1$ and $q_2$ to appear infinitely often, but never at the same time, thus forcing a kind of (not necessarily strict) alternation between the two. Developing the tableau for $\phi$, one can see that the requests $\mathsf{XF}q_1$ and $\mathsf{XF}q_2$ will be permanently present in the labels, including, in particular, after realising one of the two. Thus, crossing the branch after the second occurrence of the label would be wrong, since the repetition of the label alone does not imply that the branch is doing wasteful. Indeed, after the first repetition, the branch can continue making different choices, realising the other request, and can be closed by the LOOP rule after having realised both.

## 3 SOUNDNESS AND COMPLETENESS

This section proves that the tableau system for LTL described earlier is *sound* and *complete*. In comparison with known proofs for Reynolds' one-pass tree-shaped tableau for LTL, our proof of the completeness of the system adopts a new more insightful argument, which characterises the behaviour of the PRUNE rule by identifying a specific class of models, called *greedy models*, that are found by the system. Orthogonally, proof also of course handles the YESTERDAY rule, added in this chapter to support past operators.

The two proofs are independent, but related by the common use of the concept of *pre-model*, which is an abstract representation of one or more models of a formula. The concept of pre-model is not new in the expositions of tableau methods, but here we define it in such a way to cope with the fact, noted above, that the labels of our tableau can avoid to contain formulae that are not relevant for the satisfiability of the formula in a given state. The rest of the section is organised as follows. In Section 3.1, we introduce the concept of pre-model, and of *greedy pre-model*, setting up the necessary machinery used in the subsequent proofs. Then, Sections 3.2 and 3.3 exploit this machinery to prove the system to be sound and complete, respectively.

## 3.1 PRE-MODELS

Pre-models are an abstract representation of one or more models of a formula that differ for some details that are not relevant to the satisfaction of the formula. Pre-models are made of basic objects called *atoms*.

**Definition 6.4** — Atom.
*An* atom *for an* LTL *formula* $\phi$ *is a set* $\Delta \subseteq \mathcal{C}(\phi)$, *such that:*

1. $\{p, \neg p\} \not\subseteq \Delta$, *for any* $p \in \Sigma$;
2. *if* $\psi \in \Delta$, *then either* $\Gamma_1(\psi) \subseteq \Delta$ *or* $\Gamma_2(\psi) \neq \varnothing$ *and* $\Gamma_2(\psi) \subseteq \Delta$, *where* $\Gamma_1(\psi)$ *and* $\Gamma_2(\psi)$ *are defined in Table 6.1;*
3. *for each* $\psi, \psi' \in \mathcal{C}(\phi)$, *if* $\psi \in \Delta$ *and* $\psi \models \psi'$, *then* $\psi' \in \Delta$, i.e., $\Delta$ *is closed by logical deduction.*

Our definition of atom follows the expansion rules as defined in Table 6.1, similarly to Definition 6.2. In addition to that, however, the set is closed by logical entailment. Another difference is that our atoms are *incomplete, i.e.*, they do not necessarily specify the truth of all the formulae, but only of those that are required to justify the presence of others. This is in line with how labels of the tableau are constructed, as noted above. Note that the empty set is a valid atom in our definition. Single atoms are not useful by themselves, but have to be arranged in sequences, called *pre-models*, which are an abstract representation of the models of a formula.

**Definition 6.5** — Pre-model.
*Let* $\phi$ *be a* LTL *formula. A* pre-model *of* $\phi$ *is an infinite sequence* $\overline{\Delta} = \langle \Delta_0, \Delta_1, \ldots \rangle$ *of* minimal *atoms for* $\phi$ *such that, for all* $i \geq 0$*:*

1. $\phi \in \Delta_0$;
2. *if* $\mathsf{X}\psi \in \Delta_i$, *then* $\psi \in \Delta_{i+1}$;
3. *if* $\psi_1 \mathcal{U} \psi_2 \in \Delta_i$, *then there is a* $j \geq i$ *with* $\psi_2 \in \Delta_j$ *and* $\psi_1 \in \Delta_k$ *for all* $i \leq k < j$;

Pre-models take their name from the fact that they abstractly represent a model for their formula, and thus the existence of a pre-model witnesses the satisfiability of the formula.

**Lemma 6.6** — Extraction of a model from a pre-model.
*Let $\phi$ be a* LTL *formula. If $\phi$ has a pre-model, then $\phi$ is satisfiable.*

*Proof.* Let $\overline{\Delta}$ be a pre-model of $\phi$ and let $\overline{\sigma}$ be a state sequence such that $p \in \Delta_i$ if and only if $\overline{\sigma}, i \models p$. Then, we show that $\overline{\sigma} \models \phi$ and thus the formula is satisfiable. Note that this definition makes the precise choice of setting as *false* all the literals that are missing from the atoms of the pre-model, but any arbitrary choice would work for such literals. For any $\psi \in \mathcal{C}(\phi)$, let the *nesting degree* $\mathsf{deg}(\psi)$ of $\psi$ be defined inductively as $\mathsf{deg}(p) = \mathsf{deg}(\neg p) = 0$ for $p \in \Sigma$, $\mathsf{deg}(\mathsf{X}\psi) = \mathsf{deg}(\mathsf{Y}\psi) = \mathsf{deg}(\psi)+1$, and $\mathsf{deg}(\phi_1 \circ \phi_2) = \mathsf{max}(\mathsf{deg}(\psi_1), \mathsf{deg}(\psi_2))+1$, with $\circ \in \{\wedge, \vee, \mathcal{U}, \mathcal{S}, \mathcal{R}, \mathcal{T}\}$. We prove by induction on $\mathsf{deg}(\psi)$ that if $\psi \in \Delta_i$, then $\overline{\sigma}, i \models \psi$ for any $\psi \in \mathcal{C}(\phi)$ and any $i \geq 0$. The thesis then follows immediately, since $\phi \in \Delta_0$ by definition.

As for the base case, if $p \in \Delta_i$ or $\neg p \in \Delta_i$, then the thesis follows by the definition of $\overline{\sigma}$. As for the inductive step, we go by cases:

1. if $\psi_1 \vee \psi_2 \in \Delta_i$ (resp., $\psi_1 \wedge \psi_2 \in \Delta_i$), then by definition of atom and by the inductive hypothesis, either $\overline{\sigma}, i \models \psi_1$ or $\overline{\sigma}, i \models \psi_2$ (resp., both), and thus $\overline{\sigma}, i \models \psi_1 \vee \psi_2$ (resp., $\overline{\sigma}, i \models \psi_1 \wedge \psi_2$);

2. if $\mathsf{X}\psi \in \Delta_i$, then, by Item 2 of Definition 6.5, it holds that $\psi \in \Delta_{i+1}$. Since $\mathsf{deg}(\psi) < \mathsf{deg}(\mathsf{X}\psi)$, by the inductive hypothesis it follows that $\overline{\sigma}, i+1 \models \psi$. Then, by the semantics of the *tomorrow* operator, we have $\overline{\sigma}, i \models \mathsf{X}\psi$;

3. if $\psi_1 \,\mathcal{U}\, \psi_2 \in \Delta_i$, then, by definition of atom, there exists $j \geq i$ such that $\psi_2 \in \Delta_j$ and $\psi_1 \in \Delta_k$, for all $i \leq k < j$. Then, by the inductive hypothesis, $\overline{\sigma}, j \models \psi_2$ and $\overline{\sigma}, k \models \psi_1$ for all $i \leq k < j$, hence by the semantics of the *until* operator, we have $\overline{\sigma}, i \models \psi_1 \,\mathcal{U}\, \psi_2$;

4. the argument is similar to Item 1 when $\psi_1 \,\mathcal{R}\, \psi_2 \in \Delta_i$, $\psi_1 \,\mathcal{S}\, \psi_2 \in \Delta_i$, or $\psi_1 \,\mathcal{T}\, \psi_2 \in \Delta_i$, following the corresponding expansion rules. ∎

Note that, on the contrary, a pre-model can be obtained straightforwardly from any model $\overline{\sigma} \models \phi$ of a formula by simply setting $\phi \in \Delta_0$, and then following the definition obtaining a sequence $\overline{\Delta} = \langle \Delta_0, \Delta_1, \ldots \rangle$ of atoms where each $\Delta_i$ is the *minimal* atom whose every formula holds at $\overline{\sigma}_i$.

## 3.2 SOUNDNESS

Here we prove that the tableau system is *sound*, that is, if a complete tableau for a formula has a successful branch, then the formula is satisfiable. Moreover, a model for the formula can be effectively obtained from any successful branch. The proof shows how a pre-model for $\phi$ can be extracted from a complete tableau, which then lets us obtain one or more models of the formula. Let us first define how an atom can be constructed on top of a tableau poised node.

**Definition 6.7** — Atom of a tableau node.
*Let $T$ be a tableau for an* LTL *formula $\phi$ and let $u_i$ be a* non-crossed *poised node of $T$. The* atom *of $u_i$, written $\Delta(u_i)$, is the closure by logical entailment of $\Gamma(u_i)$.*

The exclusion of crossed poised nodes is essential to guarantee that the resulting set does not contain contradictions and hence the definition is well-defined. Now we can state how exactly to extract a model for a formula from a successful branch of its tableau. Let $\overline{u} = \langle u_0, \ldots, u_n \rangle$ be a branch of the tableau, and let $\overline{\pi} = \langle \pi_0, \ldots, \pi_m \rangle$ be the sequence of its poised nodes. Then, given a poised node $u_i$, let $\Gamma^*(u_i) = \bigcup_{j<k\leq i} \Gamma(u_k)$, where $u_j$ is the child of the closest node preceding $u_i$ where the STEP rule was applied, or the root if the STEP rule was never applied before $u_i$. In other terms, $\Gamma^*(u_i)$ contains all the formulae that were expanded to build the poised label of $u_i$.

**Lemma 6.8** — Extraction of a pre-model from a successful tableau.
*Let $\phi$ be a* LTL *formula and $T$ a complete tableau for $\phi$. If $T$ has a successful branch, then there exists a pre-model for $\phi$.*

*Proof.* Let $\overline{u} = \langle u_0, \ldots, u_n \rangle$ be a successful branch of $T$ and let $\overline{\pi} = \langle \pi_0, \ldots, \pi_m \rangle$ be the subsequence of poised nodes of $\overline{u}$. Intuitively, a pre-model for $\phi$ can be obtained from $\overline{u}$ by building the atoms from the labels of the poised nodes, and extending them to an infinite sequence. If the branch has been accepted by the LOOP rule, we can identify a position $0 \leq k \leq m$ in $\overline{\pi}$ such that $\Delta(\pi_k) = \Delta(\pi_m)$ and all the X-eventualities requested in $\pi_k$ are fulfilled in $\pi_{[k+1\ldots m]}$. If instead $\overline{u}$ has been accepted by the EMPTY rule, then $\Gamma(\pi_m) = \varnothing$, and in particular there are no X-eventualities requested, hence setting $k = m$ we obtain the same effect. Therefore, we can extract from $\overline{\pi}$ the *periodic* sequence of atoms $\overline{\Delta} = \overline{\Delta}_0 \overline{\Delta}_T^{\omega}$, where $\overline{\Delta}_0 = \langle \Delta(\pi_0), \ldots, \Delta(\pi_k) \rangle$, and either $\overline{\Delta}_T = \langle \Delta(\pi_{k+1}), \ldots, \Delta(\pi_m) \rangle$ or $\overline{\Delta}_T = \langle \pi_m \rangle$ depending on which rule accepted the branch, respectively the LOOP or the EMPTY rule. In other words, we build a periodic pre-model that infinitely repeats the fulfilling loop identified by the LOOP rule, or the last empty node otherwise. Then let $K : \mathbb{N} \to \mathbb{N}$ be the map from positions in the pre-model to their original positions in the branch, which is defined as $K(i) = i$ for $0 \leq i < k$, and for $i \geq k$ is defined either as $K(i) = k + ((i-k) \mod T)$, with $T = m - k$ (LOOP rule), or as $K(i) = k$ (EMPTY rule).

We can now show that $\overline{\Delta}$ is indeed a pre-model for $\phi$. First, observe that $\phi \in \Delta_0$ because $\phi \in \Gamma(\pi_0)$ by construction, thus Item 1 of Definition 6.5 is satisfied. Then, we check Items 2 and 3 of Definition 6.5.

2. consider any formula $\mathsf{X}\psi \in \Delta_i$. Being an elementary formula, we can observe that that $\mathsf{X}\psi \in \Gamma(\pi_{K(i)})$. Two cases have to be considered. If $\pi_{K(i+1)} = \pi_{K(i)+1}$, *i.e.*, the next atom comes from the actual successor of the current one in the tableau branch, then, by the STEP rule, we know $\psi \in \Gamma(\pi_{K(i+1)}) \subseteq \Delta_{i+1}$. Otherwise, $\Delta_i = \Delta_m = \Delta(\pi_m)$ and $\pi_m$ was ticked by the LOOP rule (because $\Delta_i$ is not empty), and thus $\Delta_{i+1} = \Delta(\pi_{k+1})$ for some $k < m$ such that $\Gamma(\pi_k) = \Gamma(\pi_m)$. Hence, $\mathsf{X}\psi \in \Gamma(\pi_k)$ as well, and, by the STEP rule applied to $\pi_k$, we know that $\psi \in \Gamma^*(\pi_{k+1}) \subseteq \Delta(\pi_{k+1}) = \Delta_{i+1}$.

3. The other cases, such as if $\psi_1 \mathcal{U} \psi_2 \in \Delta_i$, are straightforward in view of how expansion rules are defined. ■

The above results let us conclude the soundness of the tableau system.

**Theorem 17** — Soundness.
Let $\phi$ be a LTL formula, and let $T$ be a complete tableau for $\phi$. If $T$ has a successful branch, then $\phi$ is satisfiable.

*Proof.* Extract a pre-model for $\phi$ from the successful branch of $T$ as shown in Lemma 6.8, and then obtain from it an actual model for the formula as shown by Lemma 6.6. ■

## 3.3 COMPLETENESS

We now prove the completeness of the tableau system, *i.e.*, if a formula $\phi$ is satisfiable, then any complete tableau $T$ for it has an accepting branch.

The proof uses a pre-model for $\phi$, which we know to exist if the formula is satisfiable, as a guide to suitably descend through the tableau to look for an accepted branch. Then, we will show how to make sure that this descent must obtain an accepted branch. The descent is performed as follows.

**Lemma 6.9** — Extraction of the branch.
*Let $\overline{\Delta} = \langle \Delta_0, \Delta_1, \ldots \rangle$ be a pre-model for a formula $\phi$. Then, any complete tableau for $\phi$ has a branch $\overline{u}$, with sequence of poised nodes $\overline{\pi} = \langle \pi_0, \ldots, \pi_m \rangle$, such that $\Delta(\pi_i) = \Delta_i$ for all $0 \le i \le m$.*

*Proof.* To find $\overline{u}$, we traverse the tree using $\overline{\Delta}$ as a guide, starting from the root $u_0$, building a sequence of branch prefixes $\overline{u}_i = \langle u_0, \ldots, u_i \rangle$, suitably choosing $u_{i+1}$ at each step among the children of $u_i$. Meanwhile, we maintain a non-decreasing function $J : \mathbb{N} \to \mathbb{N}$ that maps positions in $\overline{u}_i$ to positions in $\overline{\Delta}$ such that $\Gamma(u_k) \subseteq \Delta_{J(k)}$ for each $0 \le k \le i$, starting from $\overline{u}_0 = \langle u_0 \rangle$ and $J(0) = 0$. With this base case the invariant holds since $\Gamma(u_0) = \{\phi\}$ and $\phi \in \Delta_0$ by definition. Then, at each step $i \ge 0$, we choose $u_{i+1}$ among the children of $u_i$ as follows:

1. if $u_i$ is a poised node but not a leaf, then it has a single child which is chosen as $u_{i+1}$, defining $J(i+1) = J(i) + 1$, since we need to advance to the next position in the pre-model as well;

2. if $u_i$ is not a poised node, then it has two children $u_i'$ and $u_i''$ and a formula $\phi$ such that $\Gamma(u_i') = \Gamma(u_i)\setminus\{\phi\}\cup\Gamma_1(\phi)$ and $\Gamma(u_i'') = \Gamma(u_i)\setminus\{\phi\}\cup\Gamma_2(\phi)$, where $\Gamma_1(\phi)$ and $\Gamma_2(\phi)$ are defined in Table 6.1. Since we maintained that $\Gamma(u_i) \subseteq \Delta_{J(i)}$, and thus $\phi \in \Delta_{J(i)}$, by Definition 6.4 we know that either $\Gamma_1(\phi) \sqsubseteq \Delta_{J(i)}$ or $\Gamma_2(\phi) \sqsubseteq \Delta_{J(i)}$, hence either $\Gamma_1(u_i') \subseteq \Delta_{J(i)}$ or $\Gamma_2(u_i'') \subseteq \Delta_{J(i)}$, so we can choose $u_{i+1}$ accordingly. Note that both might be suitable choices, in which case, which one is chosen is not important. In any case, we set $J(i+1) = J(i)$, since we do not need to advance in the pre-model.

Now, let $\overline{u} = \langle u_0, \ldots, u_n \rangle$ be the branch found as described above, and let $\overline{\pi} = \langle \pi_0, \ldots, \pi_m \rangle$ the sequence of its poised nodes. Since in the traversal the value of $J(i)$ is incremented only when an application of the STEP rule is traversed, it holds that $\Gamma(\pi_i) \subseteq \Delta_i$. Since $\Delta(\pi_i)$ is by definition the minimal atom including $\Gamma(\pi_i)$, it follows that $\Delta(\pi_i) \subseteq \Delta_i$. Now, consider the set of formulae $X_i$ such that $X_0 = \{\phi\}$, and $X_{i+1} = \{\psi \mid \mathsf{X}\psi \in \Delta_i\}$, and note that, by construction it holds that $X_i \subseteq \Delta(\pi_i)$ for each $0 \le i \le m$, and that all the formulae $\psi \in \Delta_i$ that are the result of expansion of $X_i$ are in $\Delta(\pi_i)$ as well, because of how we followed $\overline{\Delta}$ during the descent. Moreover, by Definitions 6.4 and 6.5, any formula in $\Delta_i$ must be either the result of the expansion of $X_i$ or the logical deduction of some other formulae of the set hence we can conclude that $\Delta(\pi_i) = \Delta_i$. ■

The particular branch found as described above might, in general, be crossed. However, it is immediate to note that it cannot possibly have been crossed by an application of the CONTRADICTION rule, since this would imply the existence of some $\{p, \neg p\} \subseteq \Delta_i$ for some $i$, which cannot be the case. Hence, if a crossed leaf is found, it has been crossed by the PRUNE rule. The novelty of the proof presented here is how we can select a proper class of models (and their pre-models) such that the descent described by Lemma 6.9, applied on one of these particular models, cannot possibly find a node crossed by the PRUNE rule, neither.

The key concept behind our proof is that of *greedy pre-model*. Given a pre-model $\overline{\Delta} = \langle \Delta_0, \Delta_1, \ldots \rangle$, an $\mathsf{X}$-eventuality $\psi \equiv \mathsf{X}(\psi_1 \mathcal{U} \psi_2)$ is *requested* at position $i \ge 0$ if $\psi \in \Delta_i$, and *fulfilled* at $j > i$ if $j$ is the *first* position where $\psi_2 \in \Delta_j$ and $\psi_1 \in \Delta_k$, for all $i < k < j$. Let $\mathcal{U}(\phi) = \{\psi \in \mathcal{C}(\phi) \mid \psi \equiv \mathsf{X}(\psi_1 \mathcal{U} \psi_2)\}$ be the set of $\mathsf{X}$-eventualities in the closure of $\phi$. For each position $i \ge 0$, we define the *delay* at position $i$ as a function $\mathsf{d}_i : \mathcal{U}(\phi) \to \mathbb{N}$ providing a natural number for each eventuality in $\mathcal{U}(\phi)$, as follows:

$$\mathsf{d}_i(\psi) = \begin{cases} 0 & \text{if } \psi \text{ is not requested at position } i \\ n & \text{if } \psi \text{ is requested at } i \text{ and fulfilled at } j \text{ such that } n = j - i \end{cases}$$

Intuitively, $\mathsf{d}_i(\psi)$ is the number of states elapsed between the request and the fulfilment of $\psi$. Let $\mathbb{D}$ be the set of all possible delays. Then, we can define a partial order $(\mathbb{D}, \preceq)$ between delays by comparing them component-wise, *i.e.*, for any $\mathsf{d}, \mathsf{d}' \in \mathbb{D}$, $\mathsf{d}(\psi) \leq \mathsf{d}'(\psi)$ for each $\psi \in \mathcal{U}(\phi)$. Note that $\mathbb{D}$ is just another way to denote $\mathbb{N}^{|\mathcal{U}(\phi)|}$, and $(\mathbb{D}, \preceq)$ is just $(\mathbb{N}^{|\mathcal{U}(\phi)|}, \leq)$, *i.e.*, tuples of $|\mathcal{U}(\phi)|$ natural numbers ordered component-wise. In particular, this means that $(\mathbb{D}, \preceq)$ is *well-founded*, *i.e.*, there are no infinite descending chains of elements. Given two infinite sequences of delays $\overline{\mathsf{d}} = \langle \mathsf{d}_0, \mathsf{d}_1, \ldots \rangle$ and $\overline{\mathsf{d}}' = \langle \mathsf{d}'_0, \mathsf{d}'_1, \ldots \rangle$, we can compare them lexicographically, hence defining a partial order $(\mathbb{D}^\omega, \preceq_{lex})$ such that $\overline{\mathsf{d}} \preceq_{lex} \overline{\mathsf{d}}'$ iff either $\mathsf{d}_0 \preceq \mathsf{d}'_0$ or $\mathsf{d}_0 = \mathsf{d}_0$ and $\overline{\mathsf{d}}_{\geq 1} \preceq_{lex} \overline{\mathsf{d}}'_{\geq 1}$.

A pre-order can instead be defined over pre-models, by defining $\overline{\Delta} \preceq \overline{\Delta}'$ iff $\overline{\mathsf{d}} \preceq_{lex} \overline{\mathsf{d}}'$, where $\overline{\mathsf{d}}$ and $\overline{\mathsf{d}}'$ are the sequences of delays of $\overline{\Delta}$ and $\overline{\Delta}'$, respectively. This is only a pre-order instead of a partial order because $\overline{\Delta} \preceq \overline{\Delta}'$ and $\overline{\Delta}' \preceq \overline{\Delta}$ do not imply that $\overline{\Delta} = \overline{\Delta}'$, as two different pre-models may have the same delays. Minimal elements in this pre-order are called *greedy pre-models*.

**Definition 6.10** — Greedy pre-models.
*A pre-model $\overline{\Delta}$ for a formula $\phi$ is* greedy *if there is no pre-model $\overline{\Delta}'$ such that $\overline{\Delta}' \prec \overline{\Delta}$.*

Intuitively, in greedy pre-models all the requested X-eventualities are always fulfilled as soon as possible. We will show that starting from one such pre-model ensures we avoid crossed nodes when extracting a branch from the tableau as described in Lemma 6.9. Therefore we need first to ensure that a greedy pre-model always exists, which follows from the lexicographic ordering used to compare delays.

**Lemma 6.11** — Existence of greedy pre-models.
*Let $\overline{\Delta}$ be a pre-model for a formula $\phi$. Then, there is a* greedy *pre-model $\overline{\Delta}' \preceq \overline{\Delta}$.*

*Proof.* We distinguish two cases. If there is a finite sequence $\overline{\Delta}_1 \succ \overline{\Delta}_2 \succ \ldots \succ \overline{\Delta}_n$, with $\overline{\Delta} = \overline{\Delta}_1$ and $n \geq 1$, which is maximal with respect to $\succ$, *i.e.*, it cannot be further extended, then $\overline{\Delta}' = \overline{\Delta}_n$ is a greedy model with $\overline{\Delta}' \preceq \overline{\Delta}$. Otherwise, let $\overline{\Delta}_1$ $(= \overline{\Delta}) \succ \overline{\Delta}_2 \succ \ldots$ be an infinite sequence of pre-models. We prove that its limit is a greedy pre-model $\overline{\Delta}'$. To this end, it suffices to show that for every $n \in \mathbb{N}$ (prefix length), there is $m \in \mathbb{N}$ (pre-model index) such that the prefix up to position $n$ of pre-models $\overline{\Delta}_m, \overline{\Delta}_{m+1}, \ldots$ is the same.

For $i \geq 1$, let $\mathsf{d}^i = \langle \mathsf{d}^i_0, \mathsf{d}^i_1, \ldots \rangle$ be the sequence of delays of $\overline{\Delta}_i$. Let us consider the $j$th pre-model $\overline{\Delta}_j$, for some $j \geq 1$. By definition of $\succ$, there is a position $n_j \geq 0$ such that $\mathsf{d}^{j+1}_{n_j} < \mathsf{d}^j_{n_j}$, and $\mathsf{d}^{j+1}_m = \mathsf{d}^j_m$, for all $0 \leq m < n_j$. We show that there are finitely many indices $l > j$ for which there exists a position $n_k$, with $n_k \leq n_j$, such that $\mathsf{d}^{l+1}_{n_k} < \mathsf{d}^l_{n_k}$, and $\mathsf{d}^{l+1}_m = \mathsf{d}^j_m$, for all $0 \leq m < n_k$. Let $\overline{l}$ be the largest of such indices $l$. We prove it by contradiction. Assume that there are infinitely many. Let $n_h$ be the leftmost position that comes into play infinitely many times. If $n_h = 0$, then there is an infinite strictly decreasing sequence of delays $\mathsf{d}^{h_1}_0 > \mathsf{d}^{h_2}_0 > \mathsf{d}^{h_3}_0 > \ldots$, with $j < h_1 < h_2 < h_3 < \ldots$, which cannot be the case since the ordered set $(\mathbb{N}^{|\mathcal{U}(\phi)|}, \leq)$ is well-founded (the definition of temporal shift operators ensures that the closure set of $\phi$ is finite, and thus $\mathcal{U}(\phi)$ is finite as well). Let $0 < n_h \leq n_j$. Since the positions to the left of $n_h$ are chosen only finitely many times, there exists a tuple $(\mathsf{d}_0,\ldots,\mathsf{d}_{n_h-1})$ which is paired with an infinite strictly decreasing sequence of delays $\mathsf{d}_{n_h}^{h_1} > \mathsf{d}_{n_h}^{h_2} > \mathsf{d}_{n_h}^{h_3} > \ldots$, with $j < h_1 < h_2 < h_3 < \ldots$, which again cannot be the case since the ordered set $(\mathbb{N}^{|\mathcal{U}(\phi)|},\leq)$ is well-founded. This allows us to conclude that the prefix up to position $n_j$ of all pre-models of index greater than or equal to $\bar{l}$ is the same. ■

Now, we can introduce the connection between the PRUNE rule and greedy pre-models. To this aim, we define a similar contraction rule on pre-models.

**Definition 6.12** — Redundant segments.
*Let $\overline{\Delta} = \langle\Delta_0,\Delta_1,\ldots\rangle$ be a pre-model for $\phi$, and let $i < j < k$ be three positions such that $\Delta_i = \Delta_j = \Delta_k$. Then, the subsegment $\overline{\Delta}_{[j+1\ldots k]}$ of $\overline{\Delta}$ is* redundant *if not all the* X*-eventualities requested in the atoms are fulfilled between in $\overline{\Delta}_{[j+1\ldots k]}$, and all those fulfilled in $\overline{\Delta}_{[j+1\ldots k]}$ are fulfilled in $\overline{\Delta}_{[i+1\ldots j]}$ as well.*

Intuitively, a redundant segment is one where the pre-model is doing some useless work, because there are no new X-eventualities fulfilled in the segment that were not already fulfilled before, among those recurrently requested. It can be recognised that the condition of Definition 6.12 is similar to the one checked by the PRUNE rule on a branch of the tableau, but transferred to pre-models. The important feature of redundant segments is that they can be safely removed obtaining another pre-model which, most importantly, has lower delays than the original.

**Lemma 6.13** — Removal of redundant segments.
*Let $\overline{\Delta}$ be a pre-model for a formula $\phi$ with a redundant segment $\overline{\Delta}_{[j+1\ldots k]}$. Then, the sequence of atoms $\overline{\Delta}' = \overline{\Delta}_{\leq j}\overline{\Delta}_{>k}$ is a pre-model for $\phi$, and $\overline{\Delta}' \prec \overline{\Delta}$.*

*Proof.* First of all, we check that $\overline{\Delta}' = \overline{\Delta}_{\leq j}\overline{\Delta}_{>k}$ is still a pre-model for $\phi$. This can be seen by observing that, since $\Delta_j$ and $\Delta_k$ are equal, $\psi \in \Delta_{k+1}$ for any $\mathsf{X}\psi \in \Delta_j$.

Now, let $\overline{\mathsf{d}}$ and $\overline{\mathsf{d}}'$ be the sequences of delays of $\overline{\Delta}$ and $\overline{\Delta}'$, respectively. Since $\overline{\Delta}_{[j+1\ldots k]}$ is a redundant segment, there is an atom $\Delta_i$ with $i < j$ such that $\Delta_i = \Delta_j (= \Delta_k)$. We proceed by showing that $\mathsf{d}_i' \prec \mathsf{d}_i$, while $\mathsf{d}_n' \preceq \mathsf{d}_n$ for all $n < i$, thus implying that $\overline{\Delta}' \prec \overline{\Delta}$. To this aim we show that there is at least one X-eventuality $\psi \equiv \mathsf{X}(\psi_1 \,\mathcal{U}\, \psi_2)$ for which $\mathsf{d}_i'(\psi) < \mathsf{d}_i'(\psi)$ while the other values of the delay vector for the other eventualities at worst do not increase. Now, let $\psi \equiv \mathsf{X}(\psi_1 \,\mathcal{U}\, \psi_2)$ be any X-eventuality requested in $\Delta_i$ (and $\Delta_j$ and $\Delta_k$), and consider the position $h > i$ where $\psi$ is first fulfilled (such that $\mathsf{d}_i(\psi) = h - i$). Note that it cannot be the case that $h$ appears inside the redundant segment, *i.e.*, that $j < h \leq k$, since $\psi$, by Definition 6.12, would have to be fulfilled between $\Delta_i$ and $\Delta_j$ as well, hence $h$ would not be the point of its first fulfilment. Hence, there are two cases. If $h < j$, then the point of first fulfilment of $\psi$ is not affected by the cut and the delay cannot decrease because of it. Otherwise, if $h > k$, the cut decreases the delay of $\psi$, hence $\mathsf{d}_i'(\psi) = \mathsf{d}_i(\psi) - (k - j)$. Note that at least one

X-eventuality fulfilled after $k$ exists because they cannot be neither all fulfilled in $\overline{\Delta}_{[i+1\ldots j]}$, nor inside $\overline{\Delta}_{[j+1\ldots k]}$ by Definition 6.12, hence $\mathsf{d}_i \prec \mathsf{d}'_i$.

Now, consider any position $n < i$. In any of those positions, $\mathsf{d}_n(\psi)$, for any X-eventuality $\psi_1 \,\mathcal{U}\, \psi_2$, cannot increase because of the cut, otherwise the first fulfilment of $\psi$ would have been in $\overline{\Delta}_{[j+1\ldots k]}$ (thus postponing its first fulfilment to a later point), which cannot be the case because all the eventualities fulfilled there are fulfilled also in $\overline{\Delta}_{[i+1\ldots j]}$. Hence, $\mathsf{d}_n \preceq \mathsf{d}'_n$ for all $n < i$, thus $\overline{\mathsf{d}} \prec_{lex} \overline{\mathsf{d}}'$, which implies $\overline{\Delta}' \prec \overline{\Delta}$. ■

Now we can exploit the results obtained so far to prove the completeness of the tableau system, proving that a complete tableau for a satisfiable formula contains at least an accepted branch.

■ **Theorem 18** — Completeness.
Let $\phi$ be a closed LTL formula and let $T$ be a complete tableau for $\phi$. If $\phi$ is satisfiable, then $T$ contains a successful branch.

*Proof.* Let $\overline{\sigma}$ be a model for $\phi$. As already noted, it is straightforward to build a pre-model for $\phi$ from $\overline{\sigma}$. Then, given a pre-model for $\phi$, Lemma 6.11 ensures that a *greedy* pre-model for $\phi$ exists. We can thus consider $\overline{\Delta} = \langle \Delta_0, \Delta_1, \ldots \rangle$ to be a greedy pre-model for $\phi$. Now, given a complete tableau $T$ for $\phi$, thanks to Lemma 6.9 we can obtain a branch from $T$, with sequence of poised nodes $\overline{\pi} = \langle \pi_0, \ldots, \pi_m \rangle$ such that $\Delta(\pi_k) = \Delta_k$ for all $0 \leq k \leq m$. As already noted, we know that if $\pi_m$ is crossed, then it has to have been crossed by the PRUNE rule. If this was the case, however, it would mean there are other two poised nodes $\pi_i$ and $\pi_j$ with $i < j < m$ and $\Gamma(\pi_i) = \Gamma(\pi_j) = \Gamma(\pi_m)$, and such that all the X-eventualities requested in the three nodes and fulfilled between $\pi_{j+1}$ and $\pi_m$ are fulfilled between $\pi_{i+1}$ and $\pi_j$ as well. Since $\Delta(\pi_k) = \Delta_k$ for all $0 \leq k \leq m$, this fact reflects onto the pre-model, hence $\Delta_i = \Delta_j = \Delta_m$, and all the X-eventualities requested in these atoms and fulfilled in $\overline{\Delta}_{[j+1\ldots m]}$ are fulfilled in $\overline{\Delta}_{[i+1\ldots j]}$ as well. In other words, $\overline{\Delta}_{[j+1\ldots m]}$ is a redundant segment, but this, by Lemma 6.13 contradicts the assumption that $\overline{\Delta}$ is greedy. ■

## 4 EXPERIMENTAL EVALUATION

This section surveys the main results of the experimental evaluation of a tool for LTL satisfiability, called Leviathan, which implements the above system [17]. As it turns out, the simplicity of its rule-based tree search structure pays off, allowing for a very efficient and low-overhead implementation. The tool was also parallelised with great results [97]. In our evaluation, we compared the tool on a standard set of benchmarks, not only against other tableau-based tools but also against tools employing different techniques. The performance of the tool turned out to be competitive with other tools on many classes of formulae, while better than other tableau-based tools most of the cases. Notably, the tool obtains this result by implementing the method *as-is*, without any sort of search heuristics. Here, we illustrate the main techniques used in the tool to obtain such efficiently, and then describe the benchmark results more in details.

## 4.1 IMPLEMENTATION

Despite the simplicity of the system's rules, finding the most efficient way to implement them is not trivial, requiring a subtle balance between execution speed and memory consumption. The most important ingredient is the data memory layout of the structure used to represent the current explored branch and its nodes.

Each input formula, before being given as input to the main algorithm, passes through several preprocessing steps which syntactically simplify the formula while maintaining logical equivalence. The preprocessing phase is used to desugar derived logical syntax and to turn the formula into Negation Normal Form, as assumed in Section 2, but also to remove or transform a few kinds of trivial subformulae with common propositional and temporal equivalences, most of which can be found in [69].

To achieve as much space efficiency as possible, formulae are represented in a compact way during the search. In the preprocessing phase, all the subformulae that will be needed for the application of expansion rules are extracted. The resulting set of formulae is then ordered in such a way that, for each formula $\phi$, if $\phi$ is at position $i$, then formulae $\neg\phi$ and $\mathsf{X}\phi$, if present, are at positions $i+1$ and $i+2$, respectively. The ordered set is not represented explicitly. Instead, a few *bitset* data structures are created, one for each syntactic type of formula, such that the $i$th bit in the bitset $T$ is set to 1 if and only if the $i$th formula in our ordering is of type $T$. To complete the picture, two vectors, respectively called `lhs` and `rhs`, are used to get the index of the left and right subformulae of each formula.

This compact representation also provides an efficient way to test the conditions of the tableau rules. As an example, consider the CONTRADICTION rule, which crosses a branch if occurrences of both $p$ and $\neg p$, for some $p$, are detected in a node's label. Such an operation can be efficiently implemented as follows. Let `formulae` be the bitset corresponding to the current label and `neg_lits` be the bitset that specifies which subformulae are *negative literals*. Then, consider the following expression of bitwise operations:

```
((formulae & neg_lits) << 1) & formulae
```

The first *bitwise and* operation intersects the current label with the set of all the negative literals. The *shift* moves of one position all those bits, and in the result of the second *bitwise and* there will be a bit set to 1 only if both were set, *i.e.*, only if both positive and negative literals of the same atom were present. This exploits the fact that $\phi$ and $\neg\phi$ are consecutive in the ordering. Another example is an expansion rule such as CONJUNCTION, whose condition can be tested by a *bitwise and* operation between the bitsets representing the current node's label, the set of formulae in the label that still have to be processed, and the set of all the subformulae of *conjunction* type. Finally, during the preprocessing step all the subformulae corresponding to X-eventualities are discovered and saved in a vector for later uses, together with a lookup table that links each eventuality to the corresponding index in the bitset representation.

The total independence of each branch of the tree from each other allows the tool to keep in memory only a single branch at any given time. Since a run of the procedure resembles a pre-order depth-first visit of the tree, a stack data structure is sufficient to maintain the state of the search. Similarly to the implementation of logic programming languages such as *prolog*, two different types of frame are interleaved in the stack: *choice* and *step* frames. The former are pushed when an expansion rule has been applied and a new branching point has been created. Additional information is held by the frame to make it possible to *rollback* the choice and to descend through the other branch. The latter are pushed when a STEP rule has been applied and thus a temporal step has been made. These are the frames corresponding to the nodes which have a *poised label* in the tableau and they bring with them information about the X-eventuality satisfied at the current step. Note that only the expansion rules that create a two children for the current node have a corresponding choice frame in the stack. The others are expanded in-place in the current frame.

Each frame of the stack records the set of formulae belonging to the corresponding tableau node, and it keeps track of those formulae that have been already expanded by expansion rules. Both these pieces of information are stored in two bitsets similar to those described previously. Moreover, each frame keeps track of which eventualities have been fulfilled. Finally, each frame stores three pointers to previous frames in the stack, used to check the PRUNE rule: to the last step frame, to the last occurrence of its label, and to its first, earliest occurrence.

## 4.2 EXPERIMENTAL RESULTS

We can now outline the experimental evaluation of the tool against a number of other LTL satisfiability checkers. In order to obtain significant data and to reduce chances of misinterpretation, we relied on *StarExec* [131], an online testing and benchmarking infrastructure specifically designed to measure performance of tools for logic-related problems like SAT, SMT, CLP, *etc*. The use of a common infrastructure increases the reproducibility of the experimental results, and minimises the risk of configuration errors of the tools that could lead to misleading results.

Complete and detailed surveys of the performance of available LTL satisfiability checkers appeared in the last few years [71, 87, 121, 124]. The following analysis used the reference set of testing formulae available on StarExec, which came from the survey by Schuppan and Darmawan, including a total of 3723 formulae collected from a number of different sources. The following tools were available in StarExec and were included in the comparison: Aalta [86], based on Büchi automata, TRP++ [76] and LS4 [132], which are based on temporal resolution, the NuSMV symbolic model checker [45], and PLTL, another tableau-based tool. NuSMV, configured in BDD-based symbolic model checking mode, has been chosen among other model checkers, as a single proponent of this class of tools, mainly because it was the one used in the aforementioned survey by Schuppan and Darmawan. The PLTL tool implements two different kinds of tableau-based algorithms for LTL satisfiability, an *on-the-fly* graph-shaped tableau system [1], and the already mentioned tree-like tableau system by Schwendimann [126].

A plot of the comparison results can be seen in Figure 6.3, where test formulae are grouped by type and displayed horizontally, and vertical bars of different shades represent the relative performance of different tools. The top and bottom plots show time and memory usage, respectively, obtained in two different runs with 500MB maximum memory limit in the second case, and a 30 minutes timeout in both (which is the maximum timeout allowed by StarExec). Experiments are basically limited to solvers already available in the StarExec infrastructure, to ensure repeatability and availability of results, but other tools may be considered in the future [23, 72].

In summary, the results show that Leviathan's performance is comparable in most cases with other tools, both regarding time and memory usage, with both dark and bright corners. Diving deeper, a case-by-case analysis has to be done looking at different kinds of test formulae. While data confirm that LS4, Aalta and NuSMV are likely the best tools available, our tool is competitive in a number of cases. Notable examples are the `anzu`, `forobots`, and `rozier` datasets (excluding counters), where Leviathan's running time is one of the lowest, and the `schuppan` and `rozier-counter` datasets, where the memory usage was very low. The `rozier` dataset is also notable for being much easier for all the tested tableau-based approaches, including Leviathan, than for other tools. The `alaska` dataset is very difficult for most of the tested tools, and Leviathan times out on all of it. However, it is a curious fact that, for the time it has been running before the timeout, its memory usage in this dataset was very low compared to other tools.

Another interesting point of view is the comparison of Leviathan with PLTL, as both are tableau-based tools. Leviathan performs better on the `anzu` dataset, uses less memory in the `trp` and `shuppan` datasets, and performance is comparable in other datasets.

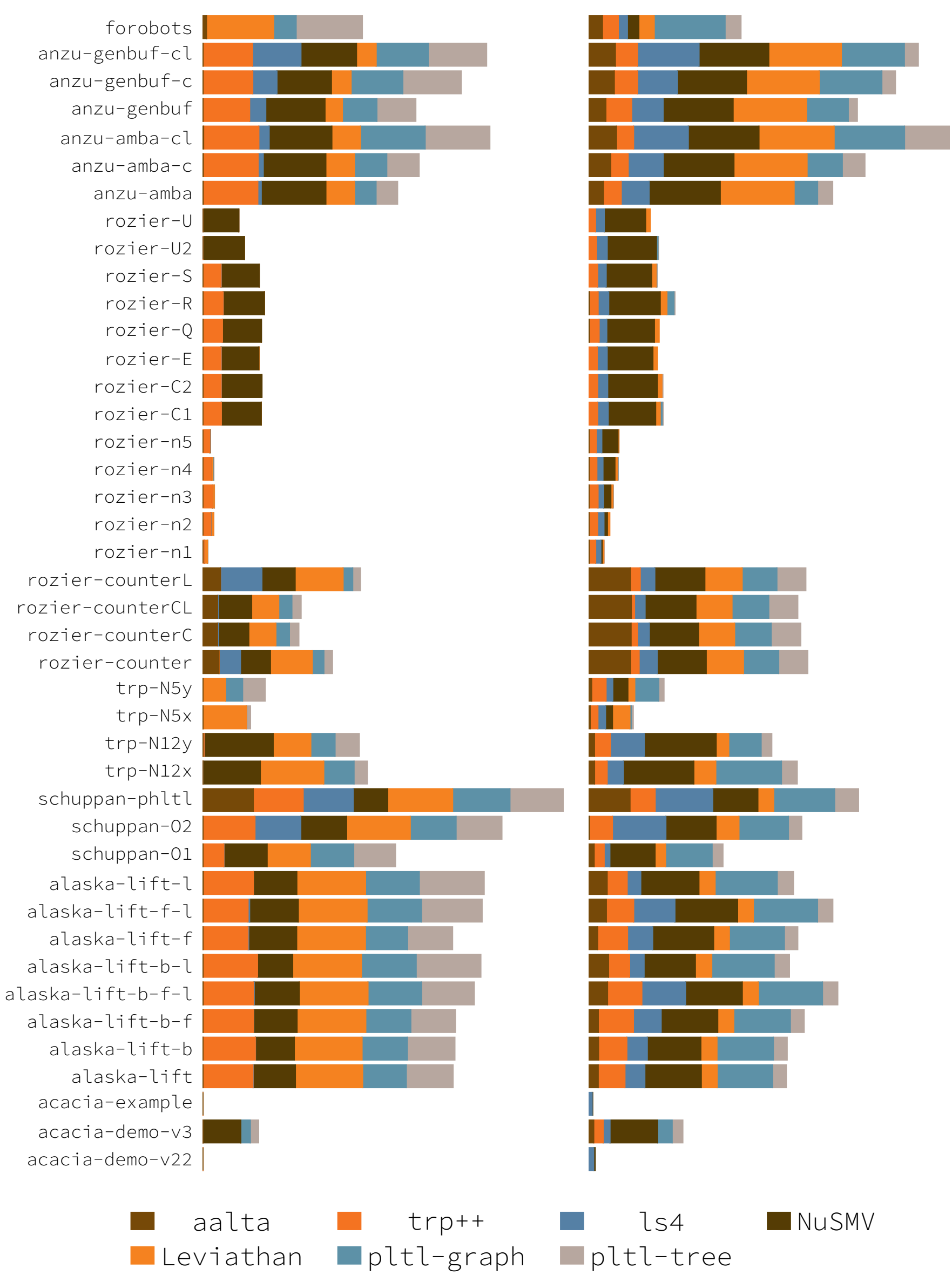

**Figure 6.3:** Benchmark results about time (left) and memory consumption (right)

# 5 THE ONE-PASS TREE-SHAPED TABLEAU FOR LINEAR TEMPORAL LOGIC WITH PAST

This section shows how the tableau system described in Section 2 can be extended to support *past operators*, hence providing a one-pass tree-shaped tableau system for *Linear Temporal Logic with Past* (LTL+P).

In contrast to classic graph-shaped tableaux like the one discussed in Section 1, where one only has to ensure that edges between nodes are consistent with the past requests of each state, in the one-pass tree-shaped system each branch of the tree is committed to the choices done in its own history, hence it is not clear *a priori* how such system could support past operators without making substantial changes to its structure. Nevertheless, our system extends the original future-only one in a perfectly modular way, by only adding a few rules which handle past operators orthogonally. The resulting system can be cut back to the original one by simply ignoring these rules, and, furthermore, running our system on a future-only formula leads to exactly the same computation as with the future-only system.

The soundness and completeness of the system are proved. Notably, the model-theoretic argument based on greedy pre-models, employed in Section 3, also applies in this case, hence the proofs are a simple extension of those for the future-only case.

## 5.1 HANDLING THE PAST

The one-pass tree-shaped tableau system for LTL+P works in the same way as the one for LTL, with the addition of a few rules specific to past operators.

Past operators supported by LTL+P are specular to future ones supported by LTL. In particular, as a formula employing the *until* operator $\psi_1 \mathcal{U} \psi_2$ holds if either $\psi_2$ or $\psi_1 \wedge \mathsf{X}(\psi_1 \mathcal{U} \psi_2)$ hold at the current state, so the *since* operator can be recursively expressed in terms of the *yesterday* operator, as $\psi_1 \mathcal{S} \psi_2$ holds if either $\psi_2$ or $\psi_1 \wedge \mathsf{Y}(\psi_1 \mathcal{S} \psi_2)$ hold. Hence, as a first step in extending the tableau system to LTL+P, we can extend the set of *expansion rules* of Table 6.1 to a set of specular rules involving past operators, as shown in Table 6.2.

Formulae of the form $\mathsf{Y}\psi$ are considered *elementary*, in the same way as *tomorrow* operators and literals, since they cannot be further expanded. Hence, the *yesterday* operator needs to be specifically handled, and the YESTERDAY rule is added for this purpose. Let $\overline{u} = \langle u_0, \ldots, u_n \rangle$ be a poised branch.

| Rule | $\phi \in \Gamma$ | $\Gamma_1(\phi)$ | $\Gamma_2(\phi)$ |
|---|---|---|---|
| SINCE | $\alpha \mathcal{S} \beta$ | $\{\beta\}$ | $\{\alpha, \mathsf{Y}(\alpha \mathcal{S} \beta)\}$ |
| TRIGGERED | $\alpha \mathcal{T} \beta$ | $\{\alpha, \beta\}$ | $\{\beta, \mathsf{Y}(\alpha \mathcal{S} \beta)\}$ |
| PAST | $\mathsf{P}\alpha$ | $\{\alpha\}$ | $\{\mathsf{YP}\alpha\}$ |
| HISTORICALLY | $\mathsf{H}\alpha$ | $\{\alpha, \mathsf{YH}\alpha\}$ | |

**Table 6.2:** Additional expansion rules supporting LTL+P past operators.

YESTERDAY If $\mathsf{Y}\alpha \in \Gamma(u_n)$, then let $Y_n = \{\psi \mid Y\psi \in \Gamma(u_n)\}$, and let $u_k$ be the closest ancestor of $u_n$ where the STEP rule was applied.

Then, the node $u_n$ is *crossed* either if $u_k$ does not exist because there is no application of the STEP rule preceding $u_n$, or if $Y_n \not\subseteq \Gamma^*(u_k)$.

Intuitively, the YESTERDAY rule fulfils the purpose of Item 2 of Definition 6.3, that is, to check whether the transitions from a state to the next are consistent with the coming past requests. It is also in this sense specular to the STEP rule, but differs significantly from the latter because instead of actively building the model such that the requests are satisfied, it is forced to retroactively check this consistency.

Whenever a formula $\mathsf{Y}\alpha$ is found in a poised label, the previous state is checked for the presence of $\alpha$. Note that, of course, the rule could not conceivably test the exact satisfaction of the formula, because that would require testing whether $\alpha$ is a logical consequence of the formulae found in the labels, which would be as hard as the satisfiability problem itself. Instead, the presence of the formula in the labels is taken as a good approximation of whether the formula holds in the state or not.

However, the YESTERDAY rule alone would result into an incomplete system. Given a formula $\mathsf{Y}\alpha$ in a given node, most of the times there are no other reasons for $\alpha$ to be forced to hold at the previous state. Hence, if not explicitly accounted for, the formula may be never expanded and any instance of the YESTERDAY rule would fail. Hence, we need a further rule, running on *poised nodes* before the application of the STEP rule, which *guesses* which formulae must be true at the current state because of yet-unseen past requests coming from future nodes. This guess is made explicit in the tableau by adding a number of children whose label contain the additional formulae. Hence, let $\overline{u} = \langle u_0, \ldots, u_n \rangle$ be a poised branch.

FORECAST Let $G_n = \{\alpha \in \mathcal{C}(\phi) \mid \mathsf{Y}\alpha \text{ is a subformula of any } \psi \in \Gamma(u_n)\}$ be the set of formulae involved in any *yesterday* operator appearing inside the formulae of $\Gamma(u_n)$.

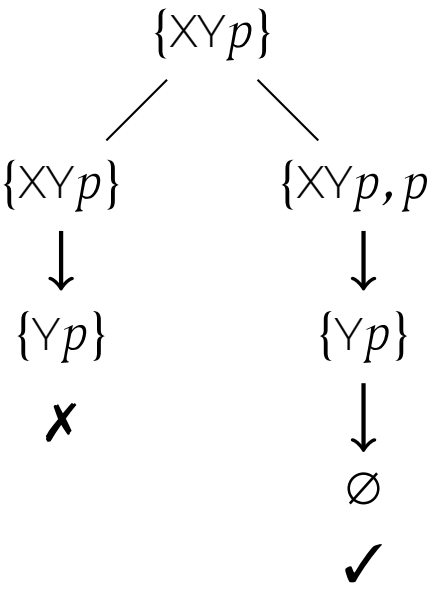

**Figure 6.4:** Example tableau for the formula XY$p$, applying the additional rules.

Then, for each subset $G'_n \subseteq G_n$ (including the empty set), a child $u'_n$ is added to $u_n$ such that $\Gamma(u'_n) = \Gamma(u_n) \cup G'_n$. This is done at most once before each application of the STEP rule.

The added children will not be, in general, poised nodes, hence the expansion continues as usual and their whole subtrees are explored. The expansion eventually leads to further poised nodes, which then can be subject to the STEP rule in order to advance in time. In this way, we ensure that for any potentially failing instance of the YESTERDAY rule, another branch exists where the rule does not fail (if this is possible at all).

Figure 6.4 shows an example of the use of the new rules on the very simple formula XY$p$. Before the first application of the STEP rule, the root node $u_0$, whose label is $\Gamma(u_0) = \{\mathsf{XY}p\}$, triggers the FORECAST rule, which adds to it two children, one with the same label, and one with an additional $p$. Then, proceeding with the expansion of such children as usual, a STEP rule is performed in each branch and the following instance of the YESTERDAY rule fails in one case but succeeds in the other. In the succeeding case, a second STEP rule then produces an empty label, as no occurrence of *tomorrow* is present, accepting the branch. Note that, in this example, the branch where $p$ is not added might seem useless. Indeed in more complex formulae, some of the added children might be redundant, but in general there is no way to know which formulae will be actually needed without expanding until the failure of the YESTERDAY rule, at which time it would be too late. For this reason, any possible combination is tried beforehand.

Note that the number of children added by the FORECAST rule is finite. Then, with an argument totally similar to Theorem 16, the finite branching factor guarantees the termination of the construction.

**Theorem 19** — Termination of the tableau for LTL+P.
Given an LTL+P formula $\phi$, the construction of a (complete) tableau $T$ for $\phi$

will always terminate in a finite number of steps. ■

## 5.2 SOUNDNESS AND COMPLETENESS

Soundness and completeness of the system will be now proved, extending the proofs shown in Section 3. Notably, the approach based on *greedy models* can be easily extended to cope with the YESTERDAY and FORECAST rules.

Let us start by adapting the concept of pre-model. The *atoms* for an LTL+P formula are defined exactly in the same way as in Definition 6.4, but considering the extended set of expansion rules of Tables 6.1 and 6.2.

**Definition 6.14** — Pre-models for LTL+P.
*Let $\phi$ be an LTL+P formula. A* pre-model *for $\phi$ is an infinite sequence $\overline{\Delta}$ of* minimal *atoms for $\phi$ such that, for all $i \geq 0$:*

1. *Items 1, 2 and 3 of Definition 6.5 hold;*
2. *if $\mathsf{Y}\psi \in \Delta_i$, then $i > 0$ and $\psi \in \Delta_{i-1}$.*

Hence, what holds for pre-models defined in Section 3 holds here as well, and the soundness of the system follows easily by simply extending the arguments shown in Lemmata 6.6 and 6.8. The presence of the FORECAST rule implies a little difference, though. Since the rule might add some children to any poised node, with the expansion going on from there after the addition, not all poised nodes have to be considered as representing distinct states of the found model. Hence, when considering a branch, we have to distinguish between those poised nodes where the STEP rule was applied, creating a single child node, and those where the FORECAST rule were applied, creating a number of children which however do not represent an advancement in time.

**Definition 6.15** — Step nodes.
*Let $u$ be a poised node of a complete tableau $T$. Then, $u$ is said to be a* step node *if either $u$ is a leaf, or it has a child added by an application of the STEP rule.*

Then, the soundness proof can proceed by extracting a pre-model for $\phi$ from the *step nodes* of any successful branch.

**Theorem 20** — Soundness.
Let $\phi$ be a LTL+P formula, and let $T$ be a complete tableau for $\phi$. If $T$ has a successful branch, then $\phi$ is satisfiable.

*Proof.* Let $\overline{u} = \langle u_0, \ldots, u_n \rangle$ be an accepted branch, and let $\overline{\pi} = \langle \pi_0, \ldots, \pi_m \rangle$ be the sequence of its step nodes. As in the proof of Theorem 17, we can build a periodic pre-model $\overline{\Delta} = \langle \Delta_0, \Delta_1, \ldots \rangle$ from the sequence of atoms $\langle \Delta(\pi_0), \ldots, \Delta(\pi_m) \rangle$, infinitely repeating the fulfilling segment identified by the LOOP rule. Only *step nodes* are considered in the construction, instead of any poised node. Recall that the branch and the constructed pre-model are bound by a function $K : \mathbb{N} \to \mathbb{N}$ that maps the atom $\Delta_i$ to its original step node $\Delta_{K(i)}$. Then, we can check that the process actually yields a pre-model for $\phi$. We know that $\phi \in \Delta_0$ by construction, and if $\mathsf{X}\phi \in \Delta_i$ or $\psi_1 \mathcal{U} \psi_2 \in \Delta_i$, then Items 2 and 3 of Definition 6.5, recalled by Definition 6.14, are satisfied as shown in the proof of Lemma 6.8. Then, consider any formula $\mathsf{Y}\psi \in \Delta_i$ and thus $\mathsf{Y}\psi \in \Gamma(\pi_{K(i)})$. By the YESTERDAY rule, $i > 0$. Then, either $\Delta_{i-1}$ is the atom coming from the previous poised node, and thus $\psi \in \Delta_{i-1}$ by the YESTERDAY rule, or $\Delta_i = \Delta(\pi_{k+1})$ for some $k$ that triggered the LOOP rule because $\Gamma(\pi_k) = \Gamma(\pi_m)$, which implies that $\Delta_k = \Delta_m$. Hence, by the YESTERDAY rule, we have $\psi \in \Delta_k = \Delta_m = \Delta_{i-1}$.

Once a pre-model has been obtained from the successful branch, an actual model for $\phi$ can be extracted as shown in Lemma 6.6, where a state sequence $\sigma = \langle \sigma_0, \sigma_1, \ldots \rangle$ is extracted from $\overline{\Delta}$ by stating that $\overline{\sigma}, i \models p$ if and only if $p \in \Delta_i$ for any $p \in \Sigma$. It can be checked, by induction over the nesting degree of the formulae, that $\sigma, i \models \psi$ for each $\psi \in \Delta_i$ and all $i \geq 0$. In particular, if $\mathsf{Y}\psi \in \Delta_i$, then $i > 0$ and $\psi \in \Delta_{i-1}$ by Definition 6.14, and by the induction hypothesis we know $\overline{\sigma}, i-1 \models \psi$, hence $\overline{\sigma}, i \models \mathsf{Y}\psi$, and $\overline{\sigma}$ is a sound model for $\phi$. ■

Hence, the soundness proof has been adapted to LTL+P quite straightforwardly. Similarly, the completeness proof from Theorem 18 can be extended to handle the new rules.

■ **Theorem 21** — Completeness.
Let $\phi$ be a closed LTL+P formula and let $T$ be a complete tableau for $\phi$. If $\phi$ is satisfiable, then $T$ contains a successful branch.

*Proof.* Similarly to Theorem 18, the proof proceeds in the following way: starting from a model $\overline{\sigma}$ for $\phi$, which is supposed to exist by hypothesis, we consider a corresponding greedy pre-model $\overline{\Delta}$, which we use as a guide to traverse the tableau to find an accepted branch. The tree is traversed to extract such branch in the same way as described in the proof of Lemma 6.9, excepting for when a poised node is found. Recall that the descent through the tree finds a branch $\overline{u} = \langle u_0, \ldots, u_n \rangle$ while maintaining a mapping $J : \mathbb{N} \to \mathbb{N}$ such that $\Gamma(u_k) \in \Delta_{J(k)}$ for all $0 \leq k \leq n$. We maintain this property also in the case where the FORECAST rule is applied to a poised node $u_i$. In this case, $u_i$ has a number of children $\{u_i^0, \ldots, u_i^k\}$, such that $\Gamma(u_i) \subseteq \Gamma(u_i^j)$ for all $0 \leq j \leq k$. Then, we set $J(i+1) = J(i)$, and we choose as $u_{i+1}$ the child $u_i^j$ with the *maximal* label such that $\Gamma(u_i^j) \subseteq \Delta_{J(i)}$. Note that at least one such child exists since one among them has the same label as $u_i$. The rest of the descent proceeds as shown in the proof of Lemma 6.9. Then, consider the sequence $\pi = \langle \pi_0, \ldots, \pi_m \rangle$ of step nodes of the branch $\overline{u}$. Then, similarly to Lemma 6.9, we can show that $\Delta(\pi_i) = \Delta_i$ for all $0 \leq i \leq m$. To do this, consider again the set $X_i$ such that $X_0 = \{\phi\}$ and $X_{i+1} = \{\phi \mid \mathsf{X}\phi \in \Delta_i\}$. In contrast to the future-only case, we also need to define a set $Y_i = \{\phi \mid \mathsf{Y}\phi \in \Delta_{i+1}\}$. Note that $Y_i \subseteq \Delta_i$ by Definition 6.14 and that, crucially, it holds that $Y_i \subseteq \Delta(\pi_i)$ as well, because of the FORECAST rule and the way we choose the children of a non-step poised node in the descent described above. Now, $\Delta_i$ is by definition the minimal closure by expansion and logical deduction of the formulae in $X_i$ and $Y_i$ (Definition 6.4), and so is $\Delta(\pi_i)$ by the way we descend through the tree, hence $\Delta_i = \Delta(\pi_i)$. Now, observe that, by construction, the branch found in this way cannot be crossed by an application of the YESTERDAY rule (nor by the CONTRADICTION rule, as in the future-only case), hence it might only have been crossed by the PRUNE rule. However this would reach a contradiction, exactly as in Theorem 18, because of the assumption that $\overline{\Delta}$ is a greedy pre-model. $\blacksquare$

# 6 CONCLUSIONS AND FURTHER WORK

This chapter introduces Reynolds' one-pass tree-shaped tableau method for satisfiability checking of LTL formulae. The system is presented with a novel simpler proof of completeness, and is extended to support *past operators*, obtaining a one-pass tree-shaped tableau system for LTL+P. The approach is of interest for us for a number of reasons. The reported experimental results are promising, and encouraged us into looking at it more deeply, but the simplicity of the rules is the real feature of this system, which allowed us to extend it to other logics while maintaining the same basic structure.

In particular, note that the extension to LTL+P presented here consists only of a few additional rules, and the resulting system runs exactly as the future-only counterpart when applied to a future-only formula. Proof techniques employed for the LTL tableau also directly apply to the past extension. The modularity of the design of the system, and the resulting extendability, will be further demonstrated in the next chapter where the system will be extended to a real-time logic used to encode timeline-based planning problems.

When comparing this system with the one-pass tableau by Schwendimann [126], where the acceptance or rejection of a branch required to wait for the exploration of a whole subtree, we can realise that the PRUNE rule is, in the end, what allows for the pure tree-shaped structure and the total independence of the exploration of each branch. Indeed, the PRUNE rule were the main novelty of the system when originally described [116]. In our novel proof of completeness, we shed some light on how the rule works under the hood, providing a model-theoretic interpretation of its role, *i.e.*, trying to avoid non-greedy models. A more detailed and complete study of the notion of *greedy-model* is due, to better understand the rule and possibly find improvements to the system or useful heuristics.

# 7 A LOGICAL CHARACTERISATION OF TIMELINE-BASED PLANNING

This chapter addresses the issue of the *expressiveness* of timeline-based planning problems in *logical* terms, as opposed to the comparative take of Chapter 3. As we know that classical action-based planning can be capture by LTL formulae, we pursue a similar result for timeline-based planning as well. To this end, we introduce the *Bounded* TPTL *with Past* logic ($\mathsf{TPTL_b{+}P}$), a fragment of TPTL+P, the classic *Timed Propositional Temporal Logic* (TPTL) augmented with past operators. We motivate its introduction, we show how (most) timeline-based planning problems can be expressed by $\mathsf{TPTL_b{+}P}$, and prove that its satisfiability problem is EXPSPACE-complete. Then, we adapt to $\mathsf{TPTL_b{+}P}$ the one-pass tree-shaped tableau method described in the previous chapter.

# 1 INTRODUCTION

In our study of the expressiveness of timeline-based planning languages described in Chapter 3, we mainly approached the problem from the point of view of the comparison with action-based languages. Starting to fill the theoretical understanding of the timeline-based planning paradigm, a comparison with more mainstream approaches was certainly due. In this chapter, we again look at expressiveness issues from a different point of view, looking for a *logical characterisation* of what timeline-based planning languages are able to model. In other words, we will answer the question of which *logic* is expressive enough to describe solution plans of a given timeline-based planning problem with an equivalent formula.

In the world of action-based planning, some results exist that relate PDDL and similar languages to common temporal logics. In particular, Cialdea Mayer et al. [42] proved that any classical STRIPS-like planning problem can be encoded as a *Linear Temporal Logic* (LTL) formula $\phi$, such that $\phi$ is satisfiable if and only if the original planning problem admits a solution plan. Their work shows how, once the problem has been encoded, the formula can be then augmented with a variety of *control knowledge*, otherwise not easily expressible as PDDL, that can help guide the search for a solution.

Having a similar result for timeline-based planning, identifying a temporal logic able to capture timeline-based planning problems, is useful for a number of reasons. First of all, it eases any further expressiveness comparison with other formalisms, using the firm ground of logic as a kind of *Rosetta* stone. Furthermore, given the definition of *timeline-based games* provided in Chapter 5, a logical encoding of timeline-based planning problems can be taken a starting point to approach the problem of the *synthesis* of a controller for timeline-based games, exploiting the great amount of literature available in the field of reactive synthesis of logical specifications [56]. In sight of a practical use of such a logical encoding, in this chapter we are interested, similarly to Chapter 3, in translations that *preserve the solutions*, meaning that we do not accept simple reductions between the respective decision problems, but we want to be able to extract solution plans for the original timeline-based planning problem from the models of the corresponding logical formula. Moreover, when possible, we want *polynomial-size* translations, even though, from the point of view of pure expressiveness, any translation would be perfectly fine, notwithstanding the size of the resulting formulae.

However, in contrast to classical and temporal planning, timeline-based planning languages immediately appear to be syntactically richer and, because of that, much more difficult to encode. In particular, simple LTL formulae fall short in capturing the semantics of synchronisation rules if the corresponding *rule graph* (see Chapter 4) is not a tree. Abstractly, the inability to use LTL for a polynomial-size translation of timeline-based planning problems comes directly from the mismatch in computational complexity of the two formalism, since LTL satisfiability is PSPACE-complete [127], while plan existence for timeline-based planning problems is EXPSPACE-complete (Theorem 8 in Chapter 4).

By looking in more detail, however, we can identify two main concrete obstacles. The first obvious one is the absence of *metric* capabilities, in contrast to the synchronisation rules which allow bounded atoms to impose real-time constraints. This problem, however, would be easily circumventable by adopting *Metric Temporal Logic* (MTL) [82], which allows to succinctly specify a bound on the number of the considered states, such as in formulae like $\mathsf{F}_n\psi$, that holds if $\phi$ holds before $n$ steps from now. MTL, interpreted over *discrete-time* structures, is EXPSPACE-complete, which closes the gap in computational complexity. However, a second obstacle affects MTL as well, and still prevents us from obtaining a polynomial-size translation of timeline-based planning problems. To see it, consider the following synchronisation rule:

$$a[x = v] \to \exists b[y = u]c[z = w]\,.$$
$$\mathsf{start}(a) \le_{[0,10]} \mathsf{start}(b) \wedge \mathsf{start}(b) \le_{[0,10]} \mathsf{start}(c) \wedge \mathsf{start}(a) \le_{[0,15]} \mathsf{start}(c)$$

The rule graph corresponding to this rule is a triangle between the three nodes $\mathsf{start}(a)$, $\mathsf{start}(b)$ and $\mathsf{start}(c)$. Note that, because of the finite bounds on atoms, the $\mathsf{start}(a) \le_{[0,10]} \mathsf{start}(c)$ atom is *not* made redundant by the other two. Trying to encode a rule of this kind with MTL, one would probably start with a formula such as $\mathsf{G}(a \to \mathsf{F}_{10}(b \wedge \mathsf{F}_{10}(c \wedge \cdots)))$. That is, supposing to have somehow encoded as the propositions $a$, $b$ and $c$ the starting point of the considered tokens, we would say that every occurrence of $a$ is followed after at most ten steps by an occurrence of $b$, which in turn is followed after at most ten steps by an occurrence of $c$. However, when trying to complete the formula to encode the last requirement that $c$ must be at most ten steps away from $a$, one cannot talk again about to the point where $a$ was originally found. An encoding of this scenario in MTL is indeed possible, but requires an expensive enumeration of all the different possible ordering of the events. Intuitively, what lacks is the ability to talk multiple times about previously visited points in the model.

This observation led us to focus on *Timed Propositional Temporal Logic* (TPTL) [6, 5], a *real-time* extension of LTL originally introduced for the formal verification of real-time systems. TPTL augments LTL with the addition of a *freeze quantifier* $x.\phi$, which binds the variable $x$ with the *timestamp* of the current state, and with *timing constraints*, such as $x \le y + 5$, which allow one to compare the timestamp of different states where the corresponding freeze quantifiers where previously interpreted. TPTL is known to be strictly more expressive of MTL [75], and in light of the observations made above, it is potentially a great fit to encode timeline-based planning problems. Under the same assumptions of the previous example, the above synchronisation rule could be encoded as the following TPTL formula:

$$\mathsf{G}x.(a \to \mathsf{F}y.(b \wedge \mathsf{F}z.(c \wedge y \leq x + 10 \wedge z \leq y + 10 \wedge z \leq x + 10)))$$

However, even TPTL is not enough to encode any possible synchronisation rule, because of the lack of *past operators*. Indeed, synchronisation rules can arbitrarily look back in the past and forward in the future from the point where they get triggered, hence, a slightly modified version of the above rule where $b$ instead of $a$ is the trigger token would not be expressible without looking at the past. Adding past operators to TPTL, however, is far from trivial. As noted in Chapter 6, adding past operators to LTL makes no difference regarding the computational complexity of the satisfiability problem, which remains PSPACE-complete for LTL+P [89]. However, this is not the case for TPTL, as the satisfiability problem of TPTL+P is known to be *non-elementary* [6].

In this chapter, we circumvent this problem by isolating a fragment of TPTL+P, called *Bounded Propositional Temporal Logic with Past* ($\mathsf{TPTL_b}$+P), which allows for the use of past operators, but put some restrictions that allow the computational complexity of its satisfiability problem to be brought back to be EXPSPACE-complete again. Given this match in computational complexity, it makes sense to use $\mathsf{TPTL_b}$+P to encode timeline-based planning problems, and we show that, although $\mathsf{TPTL_b}$+P appears to be still not sufficient to provide a polynomial-size encoding of the general formalism, it can compactly capture a broad fragment of the syntax of synchronisation rules.

The complexity of the satisfiability problem of $\mathsf{TPTL_b}$+P is proved by adapting the tableau method originally provided by Alur and Henzinger [6] for TPTL. Then, this graph-shaped tableau is adapted to a one-pass tree-shaped tableau system in the style of Chapter 6, showing another example of the extensibility of the framework.

The chapter is structured as follows. Section 2 introduces syntax and semantics of TPTL, TPTL+P and $\mathsf{TPTL_b}$+P, and Section 3 shows how $\mathsf{TPTL_b}$+P can be used to capture timeline-based planning problems. Then, Section 4 proves that the satisfiability problem for $\mathsf{TPTL_b}$+P is EXPSPACE-complete by showing a graph-shaped tableau system adapted from the one originally introduced for TPTL. Finally, Section 5 adapts it to obtain a one-pass tree-shaped tableau system, and Section 6 wraps up with some conclusive remarks and a discussion of open questions and future lines of work.

# 2 BOUNDED TPTL WITH PAST

This section recalls syntax and semantics of TPTL and TPTL+P, and introduces their restriction $\mathsf{TPTL_b{+}P}$. TPTL is an extension of LTL originally introduced in the area of formal verification to model properties of real-time systems [6]. In its original definition, the logic only supports *future* temporal operators, because the addition of *past* modalities makes the complexity of the problem for the resulting logic go from EXPSPACE-complete to *non-elementary* [5]. Nevertheless, we introduce TPTL+P first, to be able later to restrict it to obtain $\mathsf{TPTL_b{+}P}$, which will then used and studied in the next sections.

## 2.1 TPTL AND TPTL WITH PAST

Let $\Sigma$ be a set of *proposition letters* and $\mathcal{V}$ be a set of *variables*. A TPTL+P formula $\phi$ over $\Sigma$ and $\mathcal{V}$ is recursively defined as follows:

$$
\begin{aligned}
\phi := p \mid \neg\phi_1 \mid \phi_1 \vee \phi_2 \mid \phi_1 \wedge \phi_2 & \qquad \text{boolean connectives}\\
\mid \mathsf{X}\phi_1 \mid \phi_1 \,\mathcal{U}\, \phi_2 \mid \phi_1 \,\mathcal{R}\, \phi_2 & \qquad \text{future temporal operators}\\
\mid \mathsf{Y}\phi_1 \mid \phi_1 \,\mathcal{S}\, \phi_2 \mid \phi_1 \,\mathcal{T}\, \phi_2 & \qquad \text{past temporal operators}\\
\mid x.\phi_1 \mid x \le y + c \mid x \le c & \qquad \text{freeze quantifier and timing constraints}
\end{aligned}
$$

where $p \in \Sigma$, $\phi_1, \phi_2$ are TPTL+P formulae, $x, y \in \mathcal{V}$, $c \in \mathbb{Z}$. Formulae of the form $x.\phi$ are called *freeze quantifications*, while those of the forms $x \le y + c$ and $x \le c$ are called *timing constraints*. A formula $\phi$ is *closed* if any variable $x$ occurring inside $\phi$ occurs inside a freeze quantification $x.\psi$. TPTL is the fragment of TPTL+P obtained by removing the past operators.

One can note how the syntax directly extends that of LTL+P (see Section 1 of Chapter 6), with the addition of the freeze quantifier and the timing constraints. Similarly to Chapter 6, we prefer to consider the *release* and *triggered* operators as primitive, so that each formula can have a corresponding *negated normal form*, even though they can be defined in terms of the respective dual operators as $\psi_1 \,\mathcal{R}\, \psi_2 \equiv \neg(\neg\psi_1 \,\mathcal{U}\, \neg\psi_2)$ and $\psi_1 \,\mathcal{T}\, \psi_2 \equiv \neg(\neg\psi_1 \,\mathcal{S}\, \neg\psi_2)$. Moreover, similarly to LTL+P, standard logical and temporal shortcuts are used, such as $\top$ for $p \vee \neg p$, for some $p \in \Sigma$, $\bot$ for $\neg\top$, $\phi_1 \wedge \phi_2$ for $\neg(\neg\phi_1 \vee \neg\phi_2)$, $\mathsf{F}\phi$ for $\top \,\mathcal{U}\, \phi$, $\mathsf{G}\phi$ for $\neg\mathsf{F}\neg\phi$, and $\mathsf{P}\phi$ for $\top \,\mathcal{S}\, \phi$, as well as constraint shortcuts, such as, *e.g.*, $x \le y$ for $x \le y + 0$, $x > y$ for $\neg(x \le y)$, $x = y$ for $x \le y \wedge y \le x$, and others.

TPTL+P formulae are interpreted over *timed state sequences*, *i.e.*, structures $\rho = (\overline{\sigma}, \overline{\tau})$, where $\overline{\sigma} = \langle \sigma_0, \sigma_1, \ldots \rangle$ is an infinite sequence of states $\sigma_i \in 2^{\Sigma}$, with $i \ge 0$, and $\overline{\tau} = \langle \tau_0, \tau_1, \ldots \rangle$ is an infinite sequence of *timestamps* $\tau_i \in \mathbb{N}$, with $i \ge 0$, such that: (1) $\tau_{i+1} \ge \tau_i$ (*monotonicity*), and (2) for all $t \in \mathbb{N}$, there is some $i \ge 0$ such that $\tau_i \ge t$ (*progress*).

Formally, the semantics of TPTL+P is defined as follows. An *environment* is a function $\xi : \mathcal{V} \to \mathbb{N}$ that interprets a given variable as a timestamp. A timed state sequence $\rho = (\overline{\sigma}, \overline{\tau})$ satisfies a TPTL+P formula $\phi$ at position $i \geq 0$, under the environment $\xi$, written $\rho, i \models_\xi \phi$, if the following conditions are met:

1. $\rho, i \models_\xi p$ iff $p \in \sigma_i$;
2. $\rho, i \models_\xi \phi_1 \vee \phi_2$ iff either $\rho, i \models_\xi \phi_1$ or $\rho, i \models_\xi \phi_2$;
3. $\rho, i \models_\xi \neg\phi_1$ iff $\rho, i \not\models_\xi \phi_1$;
4. $\rho, i \models_\xi x \leq y + c$ iff $\xi(x) \leq \xi(y) + c$;
5. $\rho, i \models_\xi x \leq c$ iff $\xi(x) \leq c$;
6. $\rho, i \models_\xi x.\phi_1$ iff $\rho, i \models_{\xi'} \phi_1$ where $\xi' = \xi[x \leftarrow \tau_i]$;
7. $\rho, i \models_\xi \mathsf{X}\phi_1$ iff $\rho, i+1 \models_\xi \phi_1$;
8. $\rho, i \models_\xi \phi_1 \,\mathcal{U}\, \phi_2$ iff there exists $j \geq i$ such that (a) $\rho, j \models_\xi \phi_2$, and (b) $\rho, k \models_\xi \phi_1$ for all $k$ such that $i \leq k < j$;
9. $\rho, i \models_\xi \phi_1 \,\mathcal{R}\, \phi_2$ iff either $\rho, j \models_\xi \phi_2$ for all $j \geq i$, or there is a $k \geq i$ with $\rho, k \models_\xi \phi_1$ and $\rho, j \models_\xi \phi_2$ for all $i \leq j \leq k$;
10. $\rho, i \models_\xi \mathsf{Y}\phi_1$ iff $i > 0$ and $\rho, i-1 \models_\xi \phi_1$;
11. $\rho, i \models_\xi \phi_1 \,\mathcal{S}\, \phi_2$ iff there exists $j \leq i$ such that (a) $\rho, j \models_\xi \phi_2$, and (b) $\rho, k \models_\xi \phi_1$ for all $k$ such that $j < k \leq i$;
12. $\rho, i \models_\xi \phi_1 \,\mathcal{T}\, \phi_2$ iff either $\rho, j \models_\xi \phi_2$ for all $0 \leq j \leq i$, or there is a $k \leq i$ with $\rho, k \models_\xi \phi_1$ and $\rho, j \models_\xi \phi_2$ for all $i \geq j \geq k$.

As already noted, the satisfiability problem for TPTL is EXPSPACE-complete, while the same problem for TPTL+P is *non-elementary* [6]. Finding models for TPTL+P can be so hard because, by a combination of future and past operators, we can turn the freeze quantifier into a full-fledged first-order existential quantifier, capturing the first-order logic of timed state sequences as follows:

$$\exists x.\phi(x) \equiv y.\mathsf{FP}x.(\mathsf{FP}z.(z = y \wedge \phi(x)))$$

Intuitively, this encoding works because the $\mathsf{FP}\psi$ combination of future and past operators allows us to find a point anywhere in the model ($y.\mathsf{FP}x.(\cdots)$), and then, with the freeze quantifier, go back exactly at the point where the whole formula was originally being interpreted ($\mathsf{FP}z.(z = y \wedge \cdots)$), to interpret $\phi$ in the correct context. This mechanism thus requires two capabilities to work properly: to move forth and back *arbitrarily far*, and to use the freeze quantifier to refer to the timestamps of states found in that way.

## 2.2 BOUNDED TPTL WITH PAST

The $\mathsf{TPTL_b{+}P}$ logic, that we are now going to introduce, restricts TPTL+P to forbid the kind of constructs exploited above, by limiting the use of freeze quantifiers only to formulae that can guarantee a *bound* on the distance between different quantified variables. Intuitively, each temporal operator such as $X\phi$ is replaced by a *bounded* version $\mathsf{X}_w\phi$, with $w \in \mathbb{N} \cup \{+\infty\}$, such that $\mathsf{X}_w\phi$ holds if the *timestamp* of the next state differs by at most $w$ from the current one, and $\phi$ there holds. Hence, $\mathsf{X}_{+\infty}\phi$ is equivalent to $\mathsf{X}\phi$ from TPTL+P, but we impose an additional restriction that $w$ can be infinite only if $\phi$ is a *closed* formula, *i.e.*, we can look *arbitrarily far* in the model only if the formula will not refer back to the starting point.

More formally, an $\mathsf{TPTL_b{+}P}$ formula $\phi$ is recursively defined as follows:

$$
\begin{aligned}
\phi := {} & p \mid \neg\phi_1 \mid \phi_1 \vee \phi_2 \mid \phi_1 \wedge \phi_2 && \text{boolean connectives} \\
& \mid \mathsf{X}_w\phi_1 \mid \widetilde{\mathsf{X}}_w\phi_1 \mid \phi_1\, \mathcal{U}_w \phi_2 \mid \phi_1\, \mathcal{R}_w \phi_2 && \text{future temporal operators} \\
& \mid \mathsf{Y}_w\phi_1 \mid \widetilde{\mathsf{Y}}_w\phi_1 \mid \phi_1\, \mathcal{S}_w \phi_2 \mid \phi_1\, \mathcal{T}_w \phi_2 && \text{past temporal operators} \\
& \mid x.\phi_1 \mid x \leq y + c \mid x \leq c && \text{freeze quantifier and constraints}
\end{aligned}
$$

where $w \in \mathbb{N} \cup \{+\infty\}$, and in any temporal operator, if the enclosed formulae are not closed then $w \neq +\infty$. A noteworthy addition with respect to TPTL+P is the *weak* version of the *tomorrow* and *yesterday* operators, $\widetilde{\mathsf{X}}_w\psi$ and $\widetilde{\mathsf{Y}}_w\psi$. While $\mathsf{X}_w\psi$ mandates the next state to satisfy $\psi$ and have a timestamp less than $w$ greater than the current one, the weak version $\widetilde{\mathsf{X}}_w\psi$ states that *if* the timestamp bound is satisfied, then $\psi$ has to hold. The *tomorrow* and *weak tomorrow* operators are negated duals, since $\mathsf{X}_w\psi \equiv \neg\widetilde{\mathsf{X}}_w\neg\psi$, and the same applies to the corresponding past operators, so this allows us to have a *negated normal form* for $\mathsf{TPTL_b{+}P}$ formulae as well.

The semantics of $\mathsf{TPTL_b{+}P}$ is defined as follows. A $\mathsf{TPTL_b{+}P}$ formula $\phi$ is satisfied by a timed state sequence $\rho$ at position $i$ with the environment $\xi$ if the following conditions are met:

1. same as TPTL+P for all kind of formulae not explicitly considered;
2. $\rho, i \models_\xi \mathsf{X}_w\phi_1$ iff $\tau_{i+1} - \tau_i \leq w$ and $\rho, i+1 \models_\xi \phi_1$;
3. $\rho, i \models_\xi \widetilde{\mathsf{X}}_w\phi_1$ iff $\tau_{i+1} - \tau_i \leq w$ implies $\rho, i+1 \models_\xi \phi_1$;
4. $\rho, i \models_\xi \phi_1\, \mathcal{U}_w \phi_2$ iff there exists $j \geq i$ such that:
   (a) $\tau_j - \tau_i \leq w$;
   (b) $\rho, j \models_\xi \phi_2$;
   (c) $\rho, k \models_\xi \phi_1$ for all $k$ such that $i \leq k < j$;
5. $\rho, i \models_\xi \phi_1\, \mathcal{R}_w \phi_2$ iff either (a) $\tau_j - \tau_i \leq w$ implies $\rho, j \models_\xi \phi_2$ for all $j \geq i$, or (b) there is a $k \geq i$ such that $\tau_k - \tau_i \leq w$ and $\rho, k \models_\xi \phi_1$, and $\rho, j \models_\xi \phi_2$ for all $i \leq j \leq k$;

6. $\rho, i \models_\xi \mathsf{Y}_w \phi_1$ iff $i > 0$, $\tau_i - \tau_{i-1} \le w$, and $\rho, i-1 \models_\xi \phi_1$;
7. $\rho, i \models_\xi \widetilde{\mathsf{Y}}_w \phi_1$ iff $i > 0$ and $\tau_i - \tau_{i-1} \le w$ imply $\rho, i-1 \models_\xi \phi_1$;
8. $\rho, i \models_\xi \phi_1 \mathcal{S}_w \phi_2$ iff there exists $j \le i$ such that:
   (a) $\tau_i - \tau_j \le w$;
   (b) $\rho, j \models_\xi \phi_2$;
   (c) $\rho, k \models_\xi \phi_1$ for all $k$ such that $j < k \le i$;
9. $\rho, i \models_\xi \phi_1 \mathcal{T}_w \phi_2$ iff either (a) $\tau_i - \tau_j \le w$ implies $\rho, j \models_\xi \phi_2$ for all $j \le i$, or (b) there is a $k \le i$ such that $\tau_i - \tau_k \le w$ and $\rho, k \models_\xi \phi_1$, and $\rho, j \models_\xi \phi_2$ for all $i \ge j \ge k$.

## 3 CAPTURING TIMELINES WITH TPTL$_b$+P

This section shows how timeline-based planning problems can be captured by suitable TPTL$_b$+P formulae. Unfortunately, TPTL$_b$+P is still not sufficiently expressive to *compactly* capture the whole syntax of synchronisation rules used in timeline-based planning problems, but the encoding shown here is able to obtain formulae of polynomial size for a broad fragment of the whole formalism. Nevertheless, we also show how the entire language can be captured by TPTL$_b$+P if we allow a single exponential increase of the size of the formulae.

To identify such syntactic fragment, and in the description of the encoding itself, we will make use once again of the concept of *rule graph*, introduced in Chapter 4 to reason about the computational complexity of the plan existence problem. Recall that any existential statement $\mathcal{E}$ of any rule $\mathcal{R}$ can be associated with a *rule graph* $G_\mathcal{E}$ (Definition 4.5), that represents the semantics of the statement in graph form. As in Chapter 4, we can make a few assumptions that simplify the exposition without loss of generality. In particular, we can assume that any considered timeline-based planning problem does *not* make use of *pointwise atoms* nor of *triggerless rules*, and uses only *trivial duration functions* (Theorems 1, 2 and 3 of Chapter 3). Furthermore, we can assume *w.l.o.g.* that the rule graphs of any considered synchronisation rules satisfy the conditions of Lemma 4.11, and, in particular, that they are *acyclic* (Lemma 4.12).

In Chapter 4 we defined the notion of *bounded component* (Definition 4.13), as subgraphs $B \subseteq G_\mathcal{E}$ formed only by *bounded* edges. If $\overline{B} = \{B_0, \ldots, B_n\}$ are the bounded components of a rule graph $G_\mathcal{E}$, we can define an *undirected* graph $\mathcal{B}_\mathcal{E} = (B, E)$, where there is an edge between $B_i$ and $B_j$ if there is any *unbounded* edge between a node of $B_i$ and a node of $B_j$, or *vice versa*. Then, the timeline-based planning problems that TPTL$_b$+P can capture with polynomial-size formulae can be identified as those where the rule graphs of any synchronisation rule satisfy the following property.

**Definition 7.1** — Forest of bounded components.
*Let $G_{\mathcal{E}}$ be the rule graph of an existential statement $\mathcal{E}$. Then, we say that $G_{\mathcal{E}}$ is a* forest of bounded components *if any bounded component of $G_{\mathcal{E}}$ is connected to any other by* at most *one unbounded edge, and $\mathcal{B}_{\mathcal{E}}$ is a forest.*

We can now finally show how TPTL$_b$+P can capture timeline-based planning problems, where all the rule graphs are forests of bounded components, with polynomial-size formulae. Let $P = (\mathsf{SV}, S)$ be a timeline-based planning problem. We will build a TPTL$_b$+P formula $\phi_P$ such that $\phi_P$ is satisfiable if and only if $P$ admits a solution plan.

Recall that plans over SV, and in particular solution plans for $P$, can be represented as *event sequences* (Definition 4.2), which are sequences $\overline{\mu} = \langle \mu_1, \ldots, \mu_n \rangle$ of events $\mu_i = (A_i, \delta_i)$, where $A_i \subseteq \mathcal{A}_{\mathsf{SV}}$ is a set of *actions* and $\delta_i \in \mathbb{N}_+$ is the time distance between the event and the previous one. Looking at the definition of *timed state sequences* given in the previous section, one can easily interpret any such event sequence as a timed state sequence $\rho$ over the alphabet made of actions from $\mathcal{A}_{\mathsf{SV}}$, where the timestamps can be defined on top of the $\delta_i$. However, TPTL$_b$+P models are infinite timed state sequences, while event sequences are finite. The formula $\phi_P$ will be the conjunction of a formula $\phi_0$, enforcing each considered timed state sequence to represent a valid event sequence, with a formula $\phi_{\mathcal{R}}$ for any $\mathcal{R} \in S$, which encode the semantics of each rule.

When representing an event sequence as a timed state sequence we need an additional symbol end representing the *end* of the interesting prefix of the model, so we define the alphabet as $\Sigma = \mathcal{A}_{\mathsf{SV}} \cup \{\mathsf{end}\}$. Symmetrically, we also use the shortcut $\mathsf{start} \equiv \neg \mathsf{Y}\top$, to identify the first state of the state sequence. Recall that actions in $\mathcal{A}_{\mathsf{SV}}$, which are now our proposition letters, are either $\mathsf{start}(x, v)$ or $\mathsf{end}(x, v)$ for some $x \in \mathsf{SV}$ and some $v \in V_x$. We will use the shortcuts $\mathsf{start}(x) \equiv \bigvee_{v \in V_x} \mathsf{start}(x, v)$ and $\mathsf{end}(x) \equiv \bigvee_{v \in V_x} \mathsf{end}(x, v)$ to identify the start or the end of a token for the variable $x$ independently from the value.

The structure of timed state sequences as proper event sequences, following Definition 4.2, and the meaning of the end symbol, can be encoded as the conjunction of the following clauses. First of all, we need to state that time strictly increases at each state/event of the sequence:

$$\mathsf{G}x.\mathsf{X}y.(y > x)$$

Then, Items 1 and 2 of Definition 4.2 are stated by conjuncting the following clauses for all $x \in \mathsf{SV}$ and all $v \in V_x$:

$$\mathsf{G}\Big(\mathsf{start}(x, v) \rightarrow \mathsf{X}\Big(\neg \mathsf{start}(x)\, \mathcal{U}\, \mathsf{end}(x, v)\Big)\Big)$$

$$\mathsf{G}\Big(\mathsf{end}(x, v) \rightarrow \mathsf{Y}\Big(\neg \mathsf{end}(x)\, \mathcal{S}\, \mathsf{start}(x, v)\Big)\Big)$$

Items 3 and 4 of Definition 4.2 are stated as follows, for all $x \in \mathsf{SV}$ and $v \in V_x$:

$$\mathsf{G}\Big(\mathsf{end}(x,v) \rightarrow \Big(\mathsf{end} \vee \bigvee_{v' \in T_x(v)} \mathsf{start}(x,v')\Big)\Big)$$

$$\mathsf{G}\Big(\mathsf{start}(x,v) \rightarrow \Big(\mathsf{start} \vee \mathsf{end}(x)\Big)\Big)$$

Then, the semantics of the end symbol is expressed as follows:

$$\mathsf{G}\Big(\mathsf{end} \rightarrow \mathsf{G} \bigwedge_{\substack{x \in \mathsf{SV} \\ v \in V_x}} \Big(\neg \mathsf{start}(x,v) \wedge \neg \mathsf{end}(x,v)\Big)\Big) \wedge \mathsf{F}\,\mathsf{end}$$

Let $\phi_0$ be the conjunction of all the clauses above, expressing the basic structure of event sequences. Note that most of the clauses in $\phi_0$ use unbounded temporal operators, but only on closed formulae. It can be easily checked that for any event sequence $\overline{\mu}$ there is a timed state sequence $\rho_{\overline{\mu}}$ such that $\rho_{\overline{\mu}} \models \phi_0$.

We can now encode the synchronisation rules of the problem, by means of their rule graphs. Thus, let $\mathcal{R} \equiv a_0[x_0 = v_0] \rightarrow \mathcal{E}_1 \vee \cdots \vee \mathcal{E}_m \in S$ be a synchronisation rule. We can write a formula $\phi_{\mathcal{R}}$ such that $\mathcal{R}$ is satisfied by an event sequence $\overline{\mu}$ if and only if $\phi_{\mathcal{R}}$ is satisfied by $\rho_{\overline{\mu}}$, as follows:

$$\phi_{\mathcal{R}} \equiv \mathsf{G} t_0. \Big(\mathsf{start}(x_0, v_0) \rightarrow \phi_{\mathcal{E}_1}(t_0) \vee \cdots \vee \phi_{\mathcal{E}_m}(t_0)\Big)$$

where $\phi_{\mathcal{E}_i}(t_0)$ are formulae, with a free variable $t_0$, encoding the semantics of each existential statement $\mathcal{E}_i$. To write $\phi_{\mathcal{E}}(t_0)$ for some existential statement $\mathcal{E}$ of $\mathcal{R}$, consider its rule graph $G_{\mathcal{E}}$ and its bounded components $\overline{B} = \{B_0, \ldots, B_n\}$. Since the bounded components form a forest as described in Definition 7.1, we can separate them into subsets $\overline{B}_0, \ldots, \overline{B}_k$, each forming a tree of bounded components. Let us suppose *w.l.o.g.* that the forest is a single tree, *i.e.*, that it is connected. If not, we can encode each tree separately and conjunct the resulting formulae. Each bounded component $B_i$ will be encoded by a formula $\phi_{B_i}$. All such formulae will be *closed*, with the exception of the formula that encodes the trigger component of $G_{\mathcal{E}}$, say $B_0$, which will contain a free appearance of the variable $t_0$. We choose $B_0$ as the root of the tree, and proceed encoding the tree from there, hence $\phi_{\mathcal{E}}(t_0) \equiv \phi_{B_0}(t_0)$.

Each $B_i$ is encoded by looking for all the events matching with the nodes of the components, and then imposing the timing constraints between them. Some of the nodes of the component might be the endpoints of unbounded edges connecting the component to another one. When one such node is found, the existence of the corresponding other component is requested, by looking at the other endpoint node and matching the component from there. Hence, each component, excepting the trigger component, will be found starting from a specific node that we call the *anchor* of the component. As a last remark, note that in any formula we need to apply special care to ensure that nodes corresponding to the endpoints of the same token are matched accordingly.

We can now show the formulae backing this intuition. Let $B = (V, E, \beta)$ be one of the bounded components. Each node $T \in V$ is identified by a formula $\phi_T$. Suppose $T = \mathsf{start}(a)$, with $a$ quantified as $a[x = v]$. The case where $T = \mathsf{end}(a)$ will be symmetrical. Then, $\phi_T$ is defined as just $\phi_T \equiv \mathsf{start}(x, v)$, or, if $T$ is the endpoint of an *unbounded* edge connecting $T$ with a node $T'$ in another component $B'$ (hence $T'$ will be the anchor of $B'$), then $\phi_T$ is defined as follows:

1. if the edge goes from a node $T'$ to $T$, then:

$$\phi_T \equiv \mathsf{start}(x, v) \wedge \mathsf{P}\phi_{B'}$$

2. if the edge goes from $T$ to $T'$, and $T' \neq \mathsf{end}(a)$, then:

$$\phi_T \equiv \mathsf{start}(x, v) \wedge \mathsf{F}\phi_{B'}$$

3. otherwise, if the edge goes from $T$ to $T'$, and $T' = \mathsf{end}(a)$, then:

$$\phi_T \equiv \mathsf{start}(x, v) \wedge \neg\mathsf{start}(x)\,\mathcal{U}\,\phi_{B'}$$

Note that, since we will build all the $\phi_B$ as closed formulae (excepting $\phi_{B_0}$, which however is taken as the root of the tree thus it does not come up in this context), the application of the unbounded temporal operators in the formulae above is well-formed. Now, to encode the whole component, we will need to apply temporal operators to formulae containing free variables, hence requiring a bound. From Chapter 4 we know that the maximum distance between any node of a component is bounded above by $\mathsf{window}(P)$, hence, in the following formulae, we will consider the bound $w = \mathsf{window}(P)$ for any bounded temporal operator.

Then, if $T_0$ is the anchor of $B$, fix an arbitrary order between the nodes $T_i \in V$, say $\overline{T} = \langle T_0, T_1, \ldots, T_n \rangle$, such that $T_0$ is the first, if present, and if $T_i = \mathsf{start}(a)$ and $T_j = \mathsf{end}(a)$, then $j = i + 1$, *i.e.*, the end and the start of the same token, if they are both part of the component, are placed one after the other. Then, we define $\phi_B$ as $\phi_{\overline{T}}$, where $\phi_{\overline{T}}$ is defined recursively as follows. Let $\overline{T} = \langle T_i, T_{i+1}, \ldots \rangle$. If $T_i = \mathsf{start}(x, v)$ and $T_{i+1} = \mathsf{end}(x, v)$, then:

$$\phi_{\langle T_i, T_{i+1}, \ldots \rangle} \equiv t_i.\Big(\phi_{T_i} \wedge \neg\mathsf{start}(x)\,\mathcal{U}_w \phi_{\langle T_{i+1}, \ldots \rangle}\Big)$$

or, otherwise:

$$\phi_{\langle T_i, T_{i+1}, \ldots \rangle} \equiv t_i.\Big(\phi_{T_i} \wedge \mathsf{F}_w \mathsf{P}_w \phi_{\langle T_{i+1}, \ldots \rangle}\Big)$$

For the base case when $\overline{T} = \langle T \rangle$ is made of a single node, we start checking that all the temporal constraints between the found time points are satisfied, hence $\phi_{\langle T \rangle} \equiv \phi_T \wedge \phi_C$, where $\phi_C = \bigwedge_{e \in E} \phi_e$ encodes each edge $e \in E$ of the current component, that is, if $e = (T_i, T_j) \in E$ with $\beta(e) = (l, u)$, then the constraint is encoded as $\phi_e \equiv t_j \geq t_i + l \wedge t_j \leq t_i + u$.

Since all the variables $t_i$ used in these timing constraints were quantified in previous steps of the recursive definition of $\phi_{\overline{T}}$, the resulting formula has no free variables, eventually excepting for $t_0$ in the case of the trigger component, thus the whole construction is well-formed from the point of view of the $\mathsf{TPTL_b{+}P}$ syntax. It can be checked that the semantics of $\phi_P$ precisely matches the semantics of $P$, hence we can state the following result.

**Theorem 22** — $\mathsf{TPTL_b{+}P}$ captures timeline-based planning with forest rules.
Let $P$ be a timeline-based planning problem whose rule graphs consist of forests of bounded components. Then, a $\mathsf{TPTL_b{+}P}$ formula $\phi_P$ can be built, of size polynomial in the size of $P$, such that $\phi_P$ is satisfiable if and only if $P$ admits a solution plan.

Besides the technicalities involved in its definition, the encoding shown above is quite natural: each time a rule is triggered by the start of a token, the endpoints of all the tokens involved in the rule are looked for with temporal operators, their timestamps are recorded with freeze quantifiers, and the temporal constraints are then expressed using timing constraints. The subdivision of the rule graphs into bounded components allows us to isolate pieces of the rules that can be bounded in size and thus encoded with an unrestricted use of nested freeze quantifications, while unbounded edges between the components are expressed by unbounded temporal operators.

Now, the meaning of Definition 7.1 becomes clear. If multiple unbounded edges connect two components, or if the undirected graph of the bounded components were not a tree (or forest), then the schema used above would not work: the comparison of the timestamps of a component to another would violate the syntax of $\mathsf{TPTL_b{+}P}$. Unfortunately, whether $\mathsf{TPTL_b{+}P}$ admits a way to compactly encode timeline-based planning problems without restrictions, or whether it can be extended to do so without increasing its computational complexity, are still open questions.

It is worth to note that the fragment captured here is broad enough to comprehend the formulae used to encode action-based temporal problems in Chapter 3, whose rule graphs actually consist of the trigger component only. Furthermore, note that the formulae described above, which are of polynomial size, can also in particular be built in polynomial time. Hence, we can exploit the hardness result of Corollary 3.10, and state the following.

**Theorem 23** — $\mathsf{TPTL_b{+}P}$ satisfiability is EXPSPACE-hard.
Finding whether a $\mathsf{TPTL_b{+}P}$ formula is satisfiable is EXPSPACE-hard.

# 4 COMPLEXITY OF TPTL$_b$+P SATISFIABILITY

This section proves that the problem of finding whether a TPTL$_b$+P formula is satisfiable is EXPSPACE-complete. Since we got the hardness result from the encoding of timeline-based planning problems shown in the previous section, here we will provide an exponential-space decision procedure for the problem.

In particular, we present a *tableau system* for TPTL$_b$+P, which adapts and extends the tableau system originally provided for TPTL by Alur and Henzinger [6]. It is a *graph-shaped* tableau, very similar in basic principles to the one for LTL described in Section 1.2. The next section will show how it can be adapted to a one-pass tree-shaped system *à la* Reynolds, as those described in Chapter 6.

## 4.1 THE TABLEAU SYSTEM

For ease of exposition, we can assume *w.l.o.g.* any formula to have a top-level freeze quantifier, that we always mention explicitly, so any formula will be referred to as $x.\phi$. Given such a formula, we can define two quantities useful in the definition of the tableau. Given $x.\phi$, let $m$ be the number of temporal modalities with *finite* bounds used in $x.\phi$, and let $\{w_1,\ldots,w_m\}$ be the set of all such finite bounds. Among those, let $w_{\mathsf{max}}$ be the maximum one, and let $W = w_{\mathsf{max}} \cdot (m+1)$. Intuitively, $W$ is a broad upper bound on how far any bounded temporal operator in the formula can look. Then, let $\delta_{\mathsf{max}} = \prod_i |c_i|$ for all the *non-zero* coefficients $c_i$ appearing in the timing constraints of the formula, *e.g.*, in $x \leq y + c_i$. Note that, since coefficients are succinctly encoded, the size of both $W$ and $\delta_{\mathsf{max}}$ is exponential in the size of $x.\phi$. Adapting the argument employed for TPTL by [6], and for event sequences in Chapter 4, we can suppose *w.l.o.g.* that any satisfiable formula $x.\phi$ has a model $\rho = (\overline{\sigma}, \overline{\tau})$ where $\tau_{i+1} - \tau_i \leq \delta_{\mathsf{max}}$ for all $i \geq 0$. Furthermore, we will suppose all the formulae to be in *negated normal form*.

The basic mechanics of the system is similar to the graph-shaped tableau for LTL from Section 1.2: a graph is built, where each node represents a possible state of a model, and then a model is searched among the infinite paths of this graph. In contrast to the usual LTL tableaux, however, a tableau for TPTL and TPTL+P has also to keep track, in addition to the truth assignment of any given node, of how much time has to pass between two different states, and handle the freeze quantifications accordingly. The key ingredient of the original tableau for TPTL, adapted here to the TPTL$_b$+P case, is the *temporal shift*, a transformation of formulae that allows the tableau handle the binding of variables to timestamps, *i.e.*, the freeze quantifiers, in an implicit way, without keeping track of the environment explicitly.

**Definition 7.2** — Temporal shift.
*Let $z.\phi$ be a* closed TPTL *formula, $\delta \in \mathbb{N}$, and $x.\psi \in \mathcal{C}(z.\phi)$. Then, $x.\psi^{\delta}$ is the formula obtained by applying the following steps:*

1. *replace any timing constraint of the forms $x \le y + c$, and $y \le x + c$, for any other variable $y \in V$, by, respectively, $x \le y + c'$, and $y \le x + c''$, where $c' = c + \delta$ and $c'' = c - \delta$; and then*

2. *replace any timing constraint of the forms $x \le y + c$ and $y \le x + c$ either by $\top$, if $c \ge W$, or by $\bot$, if $c < -W$.*

Intuitively, the temporal shift $x.\psi^{\delta}$ of a formula $x.\psi$ allows one to interpret the formula as if the timestamp of the current state were shifted by $\delta$ time steps. However, the transformation also recognises when a temporal shift makes a timed constraint trivially valid or unsatisfiable, because the involved constants have grown too much. The formal counterpart of this intuition will be found in Section 4.2, along with the proofs of soundness and completeness.

We can now define the *closure* of a formula.

**Definition 7.3** — Closure.
*The* closure *of an* $\mathsf{TPTL_b{+}P}$ *formula $x.\phi$ is the set $\mathcal{C}(x.\phi)$ of formulae defined as:*

1. *$x.\phi \in \mathcal{C}(x.\phi)$;*

2. *if $z.\psi \in \mathcal{C}(x.\phi)$, then $\mathsf{nnf}(\neg z.\psi) \in \mathcal{C}(x.\phi)$;*

3. *if $z.y.\psi \in \mathcal{C}(x.\phi)$, then $z.\psi[y/z] \in \mathcal{C}(x.\phi)$*

4. *if $\mathsf{X}_w \psi \in \mathcal{C}(x.\phi)$, then $\{x.\psi^{\delta} \mid \delta \in \mathbb{N}\} \subseteq \mathcal{C}(x.\phi)$;*

5. *if $\mathsf{Y}_w \psi \in \mathcal{C}(x.\phi)$, then $\{x.\psi^{-\delta} \mid \delta \in \mathbb{N}\} \subseteq \mathcal{C}(x.\phi)$;*

6. *if $\phi_1 \circ \phi_2 \in \mathcal{C}(x.\phi)$, with $\circ \in \{\wedge, \vee\}$, then $\{\phi_1, \phi_2\} \subseteq \mathcal{C}(x.\phi)$;*

7. *if $\phi_1 \circ_w \phi_2 \in \mathcal{C}(x.\phi)$, with $\circ \in \{\mathcal{U}, \mathcal{R}\}$, then $\{\phi_1, \phi_2\} \subseteq \mathcal{C}(x.\phi)$ and $\mathsf{X}_w(\phi_1 \circ_{w-\delta} \phi_2) \in \mathcal{C}(x.\phi)$ for all $\delta \le w$;*

8. *if $\phi_1 \circ_w \phi_2 \in \mathcal{C}(x.\phi)$, with $\circ \in \{\mathcal{S}, \mathcal{T}\}$, then $\{\phi_1, \phi_2\} \subseteq \mathcal{C}(x.\phi)$ and $\mathsf{Y}_w(\phi_1 \circ_{w-\delta} \phi_2) \in \mathcal{C}(x.\phi)$ for all $\delta \le w$;*

Note that in the closure of a formula of the form $x.\phi$, all the formulae have a top-level freeze quantification. Moreover, if $x.\phi$ is *closed*, then its closure contains only closed formulae. In particular, any timing constraint in $\mathcal{C}(x.\phi)$ is of the form $x.(x \le x + c)$ with $|c| \le \delta_{\mathsf{max}}$.

It is worth to discuss Items 4 and 5 of Definition 7.3. As all the possible temporal shifts of a formula will be needed to construct the tableau, these are included in the closure. However, since timing constraints with too big or too little coefficients are flattened to $\top$ or $\bot$, in Definition 7.2, the set $\{x.\psi^{\delta} \mid \delta \in \mathbb{N}\}$ actually turns out to be finite.

Note that the TPTL version of the system [6] adopted a definition of temporal shift, with the same property, which however was simpler as it does not need to deal with any bound. The TPTL definition of temporal shift was based on the observation that in a formula such as $\mathsf{F}x.(x \leq y - c)$, with a free variable $y$, because the logic only supports future operators we can assume that $x$ is going to be bound to a timestamp surely greater than $y$, hence the timing constraint can be considered unsatisfiable. With past operators, this assumption is invalid, and without any other truncation condition for the temporal shift operation the closure set of the formula becomes infinite. Here, thanks to the bounds adopted by $\mathsf{TPTL_b{+}P}$ temporal operators, we can be sure that the distance between any bound variable appearing in the same formula cannot be greater than $W$, hence obtaining a different truncating condition that recovers a finite (and exponentially sized) closure set.

We now show how the tableau for $\phi$ is built. To this end, let us define $\mathcal{C}^*(x.\phi) = \mathcal{C}(x.\phi) \cup \{\mathsf{Prev}_{\delta} \mid 0 \leq \delta \leq \delta_{\mathsf{max}}\} \cup \{\mathsf{Succ}_{\gamma} \mid 0 \leq \gamma \leq \delta_{\mathsf{max}}\}$ to be an extension of the closure set of $x.\phi$ with fresh proposition letters $\mathsf{Prev}_{\delta}$ and $\mathsf{Succ}_{\gamma}$, which keep track of the time between the current state and, respectively, the previous and the next one.

**Definition 7.4** — Atom for the $\mathsf{TPTL_b{+}P}$ tableau.
*An* atom *for $x.\phi$ is a maximal subset $\Delta$ of $\mathcal{C}^*(\phi)$ such that:*

- $\mathsf{Prev}_{\delta} \in \Delta$ *and* $\mathsf{Succ}_{\gamma} \in \Delta$ *for exactly one $\delta$ and exactly one $\gamma$ between 0 and $\delta_{\mathsf{max}}$, denoted respectively as $\delta_{\Delta}$ and $\gamma_{\Delta}$, with $\delta_{\Delta_i,\Delta_j} = \sum_{i<k\leq j} \delta_{\Delta_k}$ for all $i \leq j$;*

- *$z.(z \leq z + c) \in \Delta$ iff $c \geq 0$;*

- *if $z.y.\psi \in \Delta$, then $z.\psi[y/z] \in \Delta$;*

- *$z.\psi \in \Delta$ iff* $\mathsf{nnf}(\neg z.\phi) \notin \Delta$*;*

- *if $z.(\psi_1 \vee \psi_2) \in \Delta$, then $z.\psi_1 \in \Delta$ or $z.\psi_2 \in \Delta$;*

- *if $z.(\psi_1 \,\mathcal{U}_w \psi_2) \in \Delta$, then $z.\psi_2 \in \Delta$ or $\{z.\psi_1, z.\mathsf{X}_w(\psi_1 \,\mathcal{U}_{w-\gamma_{\Delta}} \psi_2)\} \subseteq \Delta$;*

- *if $z.(\psi_1 \,\mathcal{R}_w \psi_2) \in \Delta$, then $\{z.\psi_1, z.\psi_2\} \subseteq \Delta$ or $\{z.\psi_1, z.\mathsf{X}_w(\psi_1 \,\mathcal{R}_{w-\gamma_{\Delta}} \psi_2)\} \subseteq \Delta$;*

- *if $z.(\phi_1 S_w \phi_2) \in \Delta$, then $z.\psi_2 \in \Delta$ or $\{z.\phi_1, z.\mathsf{Y}_w(\phi_1 \,\mathcal{S}_{w-\delta_{\Delta}} \psi_2)\} \subseteq \Delta$.*

- *if $z.(\phi_1 T_w \phi_2) \in \Delta$, then $\{z.\psi_1, z.\psi_2\} \subseteq \Delta$ or $\{z.\phi_1, z.\mathsf{Y}_w(\phi_1 \,\mathcal{T}_{w-\delta_{\Delta}} \psi_2)\} \subseteq \Delta$.*

**Definition 7.5** — Tableau for $\mathsf{TPTL_b{+}P}$.
*The* tableau *for $\phi$ is a graph where the nodes are all the possible atoms for $\phi$ and there is an edge between $\Delta$ and $\Delta'$ iff the following conditions hold:*

1. $\gamma_\Delta = \delta'_\Delta$;

2. $z.\mathsf{X}_w\psi \in \Delta$ *iff* $\gamma_\Delta \leq w$ *and* $z.\psi^{\gamma_\Delta} \in \Psi$;

3. $z.\mathsf{Y}_w\psi \in \Psi$ *iff* $\delta_{\Delta'} \leq w$ *and* $z.\psi^{-\delta_\Psi} \in \Delta$.

Similarly to the graph-shaped tableau for LTL+P, Items 2 and 3 of Definition 7.5 handle the temporal operators *tomorrow* and *yesterday*, by ensuring that whenever there is an edge between two atoms, the formulae requested by temporal operators in the two atoms are present. However, this is also the point where the tableau handles the binding of variables without explicitly keeping track of any environment. The freeze quantifications are pushed to the next state by shifting the formula of the right amount, preserving the semantics. Then, as in the tableaux for LTL and TPTL, the search for a model for $\phi$ is reduced to the search for a particular infinite path.

**Definition 7.6** — Fulfilling paths.
*Given the tableau for* $x.\phi$*, a fulfilling path is an* infinite *path* $\overline{\Delta} = \langle \Delta_0, \Delta_1, \ldots \rangle$ *of atoms from the tableau such that:*

1. $x.\phi \in \Delta_0$;

2. *there are no* $z.\mathsf{Y}_w\psi \in \Delta_0$;

3. $\delta_{\Delta_i} > 0$ *for infinitely many* $i \geq 0$;

4. *for all* $i \geq 0$ *and all* $z.(\phi_1\, \mathcal{U}_w \phi_2) \in \Delta_i$*, there is* $k \geq i$ *such that* $\delta_{\Delta_i,\Delta_k} \leq w$ *and* $z.\psi_2^{\delta_{\Delta_i,\Delta_k}} \in \Delta_k$;

As proved in the next section, a formula $x.\phi$ is satisfiable if and only if its tableau contains a fulfilling path. As the number of possible atoms is exponential in the size of $x.\phi$, one can prove a contraction argument similar to that employed by Sistla and Clarke [127] for LTL and by Alur and Henzinger [6] for TPTL, proving the complexity of the problem.

**Theorem 24** — Complexity of $\mathsf{TPTL_b}$+P satisfiability.
Finding whether a $\mathsf{TPTL_b}$+P formula is satisfiable is EXPSPACE-complete.

*Proof.* Let $\overline{\Delta} = \langle \Delta_0, \Delta_1, \ldots \rangle$ be any fulfilling path in the tableau for a $\mathsf{TPTL_b}$+P formula $x.\phi$. It can be checked that the total number of possible different atoms in the tableau is exponential in the size of $x.\phi$. Moreover, the number of *eventualities* of the form $z.(\psi_1 \mathcal{U}_{w'} \psi_2)$ is finite and exponentially bounded. Hence, there must be two position $i < j$ such that $\Delta_i = \Delta_j$, and for all $z.(\psi_1\, U_w \psi_2) \in \Delta_i$, there is a $i \leq k \leq j$ such that $\delta_{\Delta_i,\Delta_k} \leq w$ and $z.\psi_2 \in \Delta_k$. Then, we can obtain a *periodic* fulfilling path of the form $\overline{\Delta} = \overline{\Delta}_{\leq i}(\overline{\Delta}_{[i+1\ldots j]})^\omega$, of doubly-exponential length, which can be found in (singly) exponential space by a nondeterministic procedure similar to those employed for LTL [127] and TPTL [6]. ■

## 4.2 SOUNDNESS AND COMPLETENESS

This section proves that the tableau system for TPTL$_b$+P described above is sound and complete. The structure of the proofs is similar to that of other graph-shaped tableau systems, but the additional ingredient of the handling of time makes it a bit more involved.

The key step is to formalise and prove the effects of the *temporal shifting* operation that is applied to the formulae during the construction of the tableau.

**Lemma 7.7** — Semantics of the temporal shifts.
*Let $\rho = (\overline{\sigma}, \overline{\tau})$ be a timed state sequence and $\xi$ be an environment. Consider a position $i \geq 0$ and $\delta \in \mathbb{Z}$ such that $\delta \leq \tau_i$.*

*Then, for any formula $z.\psi \in \mathcal{C}(x.\phi)$, it holds that:*

$$\rho, i \models_{\xi} z.\psi^{\delta} \textit{ iff } \rho, i \models_{\xi'} \psi,$$

*where $\xi' = \xi[x \leftarrow \tau_i - \delta]$.*

*Proof.* Let us first define some notation. For each TPTL$_b$+P formula $\psi$, let $\mathsf{deg}(\psi)$ be the *temporal nesting* of $\psi$, *i.e.*, the nesting degree defined counting only *bounded* temporal operators. If $\psi$ is a subformula of $x.\phi$, let $\mathsf{d}(\psi) = \mathsf{deg}(x.\phi) - \mathsf{deg}(\psi) + 1$. Observe that $\mathsf{d}(x.\phi) = 1$ and that the value of $\mathsf{d}(\psi)$ is maximal when $\psi$ is atomic (*i.e.*, a literal or a timing constraint). Note, moreover, that since the closure of an *until* or *since* operator can increase the temporal nesting by one, it may be that $\mathsf{d}(z.\psi) = 0$ (*e.g.*, when $z.\psi$ is $x.\mathsf{X}_w\phi$). Recall that $W = w_{\mathsf{max}} \cdot (m+1)$ where $m$ is the number of temporal operators with finite bound that appear in $x.\phi$, and $w_{\mathsf{max}}$ is the maximum one. Thus, since $\mathsf{deg}(z.\psi) \leq m+1$, observe that $W \geq w_{\mathsf{max}} \cdot \mathsf{deg}(z.\psi)$.

We first prove a more general claim, namely that if $\psi$ is a *subformula, not necessarily closed*, of some formula $x.\psi' \in \mathcal{C}(x.\phi)$ (including itself), and $\xi$ is an environment such that $|\xi(y) - \tau_i| \leq w_{\mathsf{max}} \cdot \mathsf{d}(\psi)$ for any variable $y$ that is *free* in $\psi$, then $\rho, i \models_{\xi} x.\psi^{\delta}$ iff $\rho, i \models_{\xi[x \leftarrow \tau_i - \delta]} \psi$. The thesis then follows as a special case, since all the $x.\psi \in \mathcal{C}(\phi)$ are closed and the above condition on $\xi$ is trivially satisfied if there are no free variables.

The claim is proved by structural induction on $x.\psi$. The first base case $x.p$ for $p \in \Sigma$ is trivial since $x.p^{\delta} \equiv p$. The interesting base case is thus when $x.\psi$ is a timing constraint involving $x$. If $x.\psi \equiv x.(x \leq z + c)$, there are a few cases:

1. if $|c + \delta| \leq W$, then $x.\psi^{\delta} \equiv x.(x \leq z + (c + \delta))$. In this case, $\rho, i \models_{\xi} x.\psi^{\delta}$ iff $\rho, i \models_{\xi[x \leftarrow \tau_i]} x \leq z + (c + \delta)$. But the constraint $x \leq z + (c + \delta)$ is equivalent to $x - \delta \leq z + c$, thus $\rho, i \models_{\xi[x \leftarrow \tau_i - \delta]} x \leq z + c \equiv \psi$.

2. if $c > 0$ and $c + \delta > W$, then $x.\psi^{\delta} \equiv \top$, thus we have to show that it cannot be the case that $\rho, i \not\models_{\xi[x \leftarrow \tau_i - \delta]} \psi$. By contradiction, that would mean that $\tau_i - \delta > \xi(z) + c$, which means $\tau_i > \xi(z) + W$ is impossible since we assumed $|\xi(z) - \tau_i| \leq w_{\mathsf{max}} \cdot \mathsf{d}(x.\psi) \leq W$.

If $x.\psi \equiv x.(z \leq x + c)$ the argument is symmetrical. Hence the base case holds, and we can consider the inductive step. The cases of boolean connectives come directly from the inductive hypothesis, hence we focus on temporal operators.

If $x.\psi \equiv x.\mathsf{X}_w \psi'$, then $x.\psi^{\delta} \equiv x.\mathsf{X}_w \psi'^{\delta}$. Thus, by considering the semantics of the *tomorrow* operator, and the inductive hypothesis, we obtain:

$$
\begin{aligned}
\rho, i &\models_{\xi} x.\mathsf{X}_w \psi'^{\delta} && \\
\rho, i+1 &\models_{\xi[x \leftarrow \tau_i]} \psi'^{\delta} && \text{at the next state} \\
\rho, i+1 &\models_{\xi[x \leftarrow \tau_{i+1} - \delta']} \psi'^{\delta} && \text{where } \delta' = \tau_{i+1} - \tau_i \\
\rho, i+1 &\models_{\xi} x.\psi'^{\delta + \delta'} && \text{by the ind. hyp.} \\
\rho, i+1 &\models_{\xi[x \leftarrow \tau_{i+1} - \delta - \delta']} \psi' && \text{by the ind. hyp.} \\
\rho, i+1 &\models_{\xi[x \leftarrow \tau_i - \delta]} \psi' && \text{since } \tau_{i+1} - \delta' = \tau_i \\
\rho, i &\models_{\xi[x \leftarrow \tau_i - \delta]} \mathsf{X}_w \psi' && \text{back one state}
\end{aligned}
$$

We still have to check that the inductive hypothesis was applicable, by showing that $|\xi(y) - \tau_i| \leq w_{\mathsf{max}} \cdot \mathsf{d}(x.\psi)$ implies that $|\xi(y) - \tau_{i+1}| \leq w_{\mathsf{max}} \cdot \mathsf{d}(x.\psi')$ for any variable $y$ that is free in $x.\psi$, and thus free in $x.\psi'$. Observe that, if $w = \infty$, $\psi'$ has to be a closed formula, so there are no free variables $y$ whatsoever, and this condition on $\xi$ is trivially satisfied.

Otherwise, we know that $\delta' \leq w \leq w_{\mathsf{max}}$, thus we obtain:

$$
\begin{aligned}
|\xi(y) - \tau_i| &\leq w_{\mathsf{max}} \cdot \mathsf{d}(\psi) && \\
|\xi(y) - \tau_{i+1} - \delta'| &\leq w_{\mathsf{max}} \cdot \mathsf{d}(\psi) && \text{because } \tau_i = \tau_{i+1} - \delta' \\
|\xi(y) - \tau_{i+1}| &\leq w_{\mathsf{max}} \cdot \mathsf{d}(\psi) + \delta' && \text{because } \delta' \geq 0 \\
&\leq w_{\mathsf{max}} \cdot \mathsf{d}(\psi) + w_{\mathsf{max}} && \text{because } \delta' \leq w_{\mathsf{max}} \\
&\leq w_{\mathsf{max}} \cdot (\mathsf{d}(\psi) + 1) && \\
&\leq w_{\mathsf{max}} \cdot \mathsf{d}(\psi') &&
\end{aligned}
$$

The converse is symmetric, and the argument is similar for other operators. ■

With the shifting operator in place, we can now prove that the system is sound and complete.

**Theorem 25** — The $\mathsf{TPTL_b{+}P}$ tableau is sound and complete.
A $\mathsf{TPTL_b{+}P}$ formula is satisfiable if and only if its tableau has a fulfilling path.

*Proof (soundness).* We show that, given a fulfilling path $\Delta = \langle \Delta_0, \Delta_1, \ldots \rangle$ in the tableau for $x.\phi$, we can build a timed state sequence $\rho = (\overline{\sigma}, \overline{\tau})$ such that $\rho \models x.\phi$. The model $\rho$ can be extracted in an easy way: for each $p \in \Sigma$, $p \in \sigma_i$ iff $x.p \in \Delta_i$, and $\tau_i = \sum_{0 \leq k \leq i} \delta_{\Delta_i}$. Note that the *progress* and *monotonicity* conditions on the timed state sequence so obtained are satisfied by construction. We will now show that, for each formula $x.\psi \in \mathcal{C}(x.\phi)$ and all $i \geq 0$, if $x.\psi \in \Delta_i$ then $\rho, i \models x.\psi$, from which the thesis follows since $x.\phi \in \Delta_0$. This is done by structural induction on $x.\psi$.

- if $x.p \in \Delta_i$, the thesis holds by construction;
- if a timing constraint $x.(x \leq x + c) \in \Delta_i$, by Definition 7.4 we know that $c \geq 0$, thus $\rho, i \models x.(x \leq x + c)$.
- If $x.\neg\psi \in \Delta_i$, Definition 7.4 implies that $x.\psi \notin \Delta_i$, which by the inductive hypothesis implies that $\rho, i \models x.\neg\psi$.
- If $x.(\psi_1 \vee \psi_2) \in \Delta_i$, then either $x.\psi_1 \in \Delta_i$ or $x.\psi_2 \in \Delta_i$, thus either $\rho \models x.\psi_1$ or $\rho \models x.\phi_2$, which implies $\rho \models x.(\psi_1 \vee \psi_2)$.
- If $x.\mathsf{X}_w\psi \in \Delta_i$, then $\gamma_{\Delta_i} \leq w$ and $x.\psi^{\gamma_{\Delta_i}} \in \Delta_{i+1}$ by Definition 7.5, which implies $\rho, i+1 \models_\xi x.\psi^{\gamma_{\Delta_i}}$ for any $\xi$. Then, by Lemma 7.7, we know that $\rho, i+1 \models_{\xi[x \leftarrow \tau_{i+1} - \gamma_{\Delta_i}]} \psi$. But $\tau_i = \tau_{i+1} - \gamma_{\Delta_i}$, thus $\tau_{i+1} \leq \tau_i + w$, and we have $\rho, i \models_\xi x.\mathsf{X}_w\psi$ for any $\xi$.
- If $x.\mathsf{Y}_w\psi \in \Delta_i$, then by Definitions 7.5 and 7.6 we know $i > 0$, $\delta_{\Delta_i} \leq w$, and $x.\psi^{-\delta_{\Delta_i}} \in \Delta_{i-1}$, which implies $\rho, i-1 \models_\xi x.\psi^{-\delta_{\Delta_i}}$ for any $\xi$. Then, by Lemma 7.7, we have $\rho, i-1 \models_{\xi[x \leftarrow \tau_{i-1} + \delta_{\Delta_i}]} \psi$. But $\tau_i = \tau_{i-1} + \delta_{\Delta_i}$, thus $\tau_i \leq \tau_{i-1} + w$, and we have $\rho, i \models_\xi x.\mathsf{Y}\psi$ for any $\xi$.
- If $x.(\psi_1 \,\mathcal{U}_w \psi_2) \in \Delta_i$, then by Definition 7.6 there is a $k \geq i$ such that $x.\psi_2^{\delta_{\Delta_i, \Delta_k}} \in \Delta_k$ and $x.\psi_1^{\delta_{\Delta_i, \Delta_j}} \in \Delta_j$ for all $i \leq j < k$. Thus, by the induction hypothesis, $\rho, k \models_\xi x.\psi_2^{\delta_{\Delta_i, \Delta_k}}$ and $\rho, j \models_\xi x.\psi_2^{\delta_{\Delta_i, \Delta_j}}$ for all $i \leq j < k$, for any $\xi$. By Lemma 7.7, it implies that $\rho, k \models_{\xi[x \leftarrow \tau_k - \delta_{\Delta_i, \Delta_k}]} \psi_2$ and $\rho, j \models_{\xi[x \leftarrow \tau_j - \delta_{\Delta_i, \Delta_j}]} \psi_1$ for all $i \leq j < k$. But, $\tau_k = \tau_i + \delta_{\Delta_i, \Delta_k}$ and $\tau_j = \tau_i + \delta_{\Delta_i, \Delta_j}$ for all $i \leq j < k$, thus $\tau_k \leq \tau_i + w$, and we have $\rho, k \models_\xi x.\psi_1 \,\mathcal{U}_w \psi_2$.
- The case of $x.(\psi_1 \,\mathcal{S}_w \psi_2) \in \Delta_i$ mirrors the previous one.

(*completeness*) We now show that, given a timed state sequence $\rho = (\sigma, \tau)$ such that $\rho \models \phi$, then there exists a fulfilling path $\Delta = \langle \Delta_0, \Delta_1, \ldots \rangle$ in the tableau for $\phi$. The construction, again, is simple: the atom $\Delta_i$, for each $i$, is built so that $x.\psi \in \Delta_i$ if and only if $\rho, i \models x.\psi$ for each $x.\psi \in \mathcal{C}(x.\phi)$. Then, $\mathsf{Prev}_0 \in \Delta_0$ and $\mathsf{Prev}_\delta \in \Delta_i$ with $\delta = \tau_i - \tau_{i-1}$ for each $i > 0$, and $\mathsf{Succ}_\gamma \in \Delta_i$ with $\gamma = \tau_{i+1} - \tau_i$ for each $i \geq 0$. It is easy to verify that the consistency conditions of Definition 7.4 are satisfied, so that $\Delta_0, \Delta_1, \ldots$ are indeed atoms.

To see that $\Delta$ is indeed a path in the tableau, we need to show that there is an edge between $\Delta_i$ and $\Delta_{i+1}$ for each $i$. It is easy to check that $\gamma_{\Delta_i} = \delta_{\Delta_{i+1}}$ by construction. Then, consider a formula $x.\mathsf{X}_w\psi \in \Delta_i$. By construction we know that $\rho, i \models_\xi x.\mathsf{X}_w\psi$, for any $\xi$, thus $\rho, i+1 \models_{\xi[x\leftarrow\tau_i]} \psi$ and $\tau_{i+1} \le \tau_i + w$. Thus, since $\tau_i = \tau_{i+1} - \gamma_{\Delta_{i+1}}$ we have $\gamma_{\Delta_{i+1}} \le w$, and by Lemma 7.7 we obtain that $\rho, i+1 \models_\xi x.\psi^{\gamma_{\Delta_i}}$, which implies that $x.\psi^{\gamma_{\Delta_i}} \in \Delta_{i+1}$. With a similar argument we know that $x.\mathsf{Y}_w\psi \in \Delta_{i+1}$ implies $x.\psi^{-\delta_{\Delta_{i+1}}} \in \Delta_i$.

To see that $\Delta$ is, in particular, a fulfilling path, first observe that $\phi \in \Delta_0$ because $\rho, 0 \models \phi$, that no $x.\mathsf{Y}_w\psi$ can be in $\Delta_0$ since it cannot be the case that $\rho, 0 \models x.\mathsf{Y}_w\psi$, and that $\Delta$ satisfies the progress condition because the timed state sequence does, thus Items 1, 2 and 3 of Definition 7.6 are satisfied. For Definition 7.6, consider a formula $x.(\psi_1 \mathcal{U}_w \psi_2) \in \Delta_i$ for some $i \ge 0$. By construction, $\rho, i \models_\xi x.(\psi_1 \mathcal{U}_w \psi_2)$, so there is a $k$ such that $\tau_k \le \tau_i + w$ and $\rho, k \models_{\xi[x\leftarrow\tau_i]} \psi_2$. But $\tau_i = \tau_k - \delta_{\Delta_i,\Delta_k}$, thus by Lemma 7.7 we have that $\rho, k \models_\xi x.\psi_2^{\delta_{\Delta_i,\Delta_k}}$, thus by construction $x.\psi_2^{\delta_{\Delta_i,\Delta_k}} \in \Delta_k$. Similarly, $x.\psi_1^{\delta_{\Delta_i,\Delta_j}} \in \Delta_j$ for all $i \le j < k$. ■

# 5 A ONE-PASS TREE-SHAPED TABLEAU FOR TPTL$_b$+P

This section shows how the *one-pass tree-shaped* tableau system for LTL+P described in Section 5 of Chapter 6 can be adapted to TPTL$_b$+P. The result is a one-pass tree-shaped tableau system *à la* Reynolds for this logic.

The tableau system shown here does not handle directly TPTL$_b$+P formulae. Rather, the formulae are translated into a proper fragment of TPTL+P, and the tableau rules are defined to handle such fragment. The fragment of TPTL+P identified here corresponds exactly to TPTL$_b$+P, thus providing also a characterisation of the expressiveness of TPTL$_b$+P in comparison with the whole TPTL+P. The system will be proved to be sound and complete by sketching how to adapt to it the same argument based on greedy models employed in Chapter 6 for the LTL and LTL+P case.

## 5.1 TPTL$_b$+P AS A GUARDED FRAGMENT OF TPTL+P

We now show how TPTL$_b$+P can be identified as a *guarded* fragment of TPTL+P, that is, a syntactic fragment of the logic, that we call G(TPTL+P), where each occurrence of any temporal operator is guarded by an additional formula which implements the bounded semantics of TPTL$_b$+P operators.

$\mathsf{G(TPTL{+}P)}$ is the fragment of $\mathsf{TPTL{+}P}$ defined as follows:

$$\begin{aligned}\phi := p \mid \neg p \mid \phi_1 \vee \phi_2 \mid x \leq y + c \mid x \leq c \\ \mid x.\mathsf{X}y.(\gamma_w^{x,y} \wedge \phi_1) \mid x.\mathsf{X}y.(\gamma_w^{x,y} \rightarrow \phi_1) \\ \mid x.\mathsf{Y}y.(\gamma_w^{x,y} \wedge \phi_1) \mid x.\mathsf{Y}y.(\gamma_w^{x,y} \rightarrow \phi_1) \\ \mid x.\big(z.(\gamma_w^{x,z} \rightarrow \phi_1)\,\mathcal{U}\,y.(\gamma_w^{x,y} \wedge \phi_2)\big) \mid x.\big(z.(\gamma_w^{x,z} \wedge \phi_1)\,\mathcal{R}\,y.(\gamma_w^{x,y} \rightarrow \phi_2)\big) \\ \mid x.\big(z.(\gamma_w^{x,z} \rightarrow \phi_1)\,\mathcal{S}\,y.(\gamma_w^{x,y} \wedge \phi_2)\big) \mid x.\big(z.(\gamma_w^{x,z} \wedge \phi_1)\,\mathcal{T}\,y.(\gamma_w^{x,y} \rightarrow \phi_2)\big)\end{aligned}$$

where $\gamma_w^{x,y} = y \leq x + w$, if $w \neq +\infty$, and $\gamma_w = \top$ otherwise, with $w \in \mathbb{N} \cup \{+\infty\}$ and $x$ and $y$ fresh in $\phi_1$ and $\phi_2$. Moreover, as in $\mathsf{TPTL_b{+}P}$, each temporal operator can appear with $w = +\infty$ only if the corresponding formula is closed. All the temporal operators where $w \neq +\infty$ are called *guarded*.

One can check that the *negated normal form* of a $\mathsf{G(TPTL{+}P)}$ formula is still a $\mathsf{G(TPTL{+}P)}$ formula, and, as shown below, that any $\mathsf{TPTL_b{+}P}$ formula has an equivalent $\mathsf{G(TPTL{+}P)}$ one.

**Lemma 7.8** — Translation of $\mathsf{TPTL_b{+}P}$ into $\mathsf{G(TPTL{+}P)}$.
*Let $\phi$ be a $\mathsf{TPTL_b{+}P}$ formula. Then, there exists a $\mathsf{G(TPTL{+}P)}$ formula $\phi'$ such that for any timed state sequence $\rho$, any environment $\xi$, and any $i \geq 0$, it holds that $\rho, i \models_\xi \phi$ if and only if $\rho, i \models_\xi \phi'$.*

*Proof.* The semantics of most temporal operators of $\mathsf{TPTL_b{+}P}$ exactly matches the corresponding guarded form in the $\mathsf{G(TPTL{+}P)}$ syntax, hence the translation is straightforward. The only exception is the *tomorrow* and *yesterday* operators, whose strong and weak versions are both mapped to the $\mathsf{G(TPTL{+}P)}$ *tomorrow* and *yesterday* operators. However, $\mathsf{G(TPTL{+}P)}$ supports two possible ways to guard these operators, one where the guard is conjuncted to, and one where the guard implies, the target formula. Hence, the *tomorrow* operator is translated as $\mathsf{X}_w\psi \equiv x.\mathsf{X}y.(y \leq x{+}w \wedge \psi)$ for the strong version, and $\widetilde{\mathsf{X}}_w\psi \equiv x.\mathsf{X}y.(y \leq x{+}w \rightarrow \psi)$, for the weak version, if $w \neq +\infty$, and simply $\mathsf{X}_{+\infty}\psi \equiv \widetilde{\mathsf{X}}_{+\infty}\psi \equiv \mathsf{X}\psi$ otherwise. ∎

## 5.2 THE TABLEAU SYSTEM FOR G(TPTL+P)

Here we can finally show how to adapt to $\mathsf{TPTL_b{+}P}$ the one-pass tree-shaped tableau for $\mathsf{LTL{+}P}$ shown in Section 5 of Chapter 6. As anticipated, the tableau does not handle directly $\mathsf{TPTL_b{+}P}$ formulae. Rather, it handles generic $\mathsf{TPTL{+}P}$ formulae, but embedding the guarded semantics of $\mathsf{TPTL_b{+}P}$. Hence, when applying the system to a $\mathsf{G(TPTL{+}P)}$ formula, the result is sound and complete.

The first step is thus the definition of the *closure* of a $\mathsf{TPTL{+}P}$ formula. The definition will again use the *temporal shift* operation defined for the graph-shaped tableau for $\mathsf{TPTL_b{+}P}$. Although applied on $\mathsf{TPTL{+}P}$ formulae, the operation is defined exactly as in Definition 7.2.

**Definition 7.9** — Closure of a TPTL+P formula.
*The* closure *of a* TPTL+P *formula* $x.\phi$ *is the set* $\mathcal{C}(z.\phi)$ *of formulae defined as:*

1. $z.\phi \in \mathcal{C}(z.\phi)$;
2. *if* $x.(\psi_1 \wedge \psi_2) \in \mathcal{C}(z.\phi)$, *then* $\{x.\psi_1, x.\psi_2\} \subseteq \mathcal{C}(z.\phi)$;
3. *if* $x.(\psi_1 \vee \psi_2) \in \mathcal{C}(z.\phi)$, *then* $\{x.\psi_1, x.\psi_2\} \subseteq \mathcal{C}(z.\phi)$;
4. *if* $x.\mathsf{X}\psi \in \mathcal{C}(z.\phi)$, *then* $x.\psi^{\delta} \in \mathcal{C}(z.\phi)$, *for all* $\delta \geq 0$;
5. *if* $x.\mathsf{Y}\psi \in \mathcal{C}(z.\phi)$, *then* $x.\psi^{-\delta} \in \mathcal{C}(z.\phi)$, *for all* $\delta \geq 0$;
6. *if* $x.(\psi_1 \circ \psi_2) \in \mathcal{C}(z.\phi)$, *where* $\circ \in \{\mathcal{U}, \mathcal{R}, \mathcal{S}, \mathcal{T}\}$, *then* $\{x.\psi_1, x.\psi_2, x.\mathsf{X}(\psi_1 \circ \psi_2)\} \subseteq \mathcal{C}(z.\phi)$;
7. *if* $x.y.\psi \in \mathcal{C}(z.\phi)$, *then* $x.\psi[y/x] \in \mathcal{C}(z.\phi)$.

The basic structure of the system is very similar to the LTL+P tableau. A tableau for a G(TPTL+P) formula $x.\phi$ is a tree where any node $u$ is labelled by a subset $\Gamma(u) \subseteq \mathcal{C}(x.\phi)$ of the closure of $x.\phi$. The tree is built, starting from the root $u_0$ with $\Gamma(u_0) = \{x.\phi\}$, by applying a set of rules to each leaf. Each rule can potentially create some children, thus creating new leaves from which the process can continue, or close the branch by accepting or rejecting it.

The set of rules is similar to those used in the LTL+P tableau. The *expansion rules* for TPTL+P are shown in Table 7.1, and are very similar to those for LTL+P. They look for a formula $x.\psi$ in $\Gamma(u)$, and create two children $u'$ and $u''$ such that $\Gamma(u') = \Gamma(u) \setminus \{x.\psi\} \cup \Gamma_1(u)$ and $\Gamma(u'') = \Gamma(u) \setminus \{x.\psi\} \cup \Gamma_2(u)$, or a single child $u'$ if $\Gamma_2(u)$ is empty. Apart from the different syntax of the formulae themselves, it can be seen that the expansions of each formula are mostly unchanged with regards to LTL+P.

Atomic formulae such as propositions and timing constraints, and *tomorrow* and *yesterday* operators, are considered *elementary* formulae. Nodes whose labels contain only elementary formulae are called *poised* nodes. The major difference of this system in contrast to the LTL+P one is the STEP rule, which here does not only have to propagate the *tomorrow* requests from a state to the next, but also has to choose how much time has to pass between the two states. This is done by simply creating as many children as are the possible time advancements. Recall that we can assume that the maximum time gap between two states is bounded by $\delta_{\mathsf{max}}$, hence the number of such choices is finite.

STEP Let $u$ be a poised node. Then, $\delta_{\mathsf{max}} + 1$ children nodes $u_0, \ldots, u_{\delta_{\mathsf{max}}}$ are added to $u$, such that $\Gamma(u_\delta) = \{x.\psi^{\delta} \mid x.\mathsf{X}\psi \in \Gamma(u)\}$ for all $0 \leq \delta \leq \delta_{\mathsf{max}}$.

| Rule | $\phi \in \Gamma$ | $\Gamma_1(\phi)$ | $\Gamma_2(\phi)$ |
|---|---|---|---|
| DISJUNCTION | $x.\psi_1 \vee x.\psi_2$ | $\{x.\psi_1\}$ | $\{x.\psi_2\}$ |
| CONJUNCTION | $x.\psi_1 \wedge x.\psi_2$ | $\{x.\psi_1, x.\psi_2\}$ | |
| UNTIL | $x.(\psi_1 \mathcal{U} \psi_2)$ | $\{x.\psi_2\}$ | $\{x.\psi_1, x.\mathsf{X}(\psi_1 \mathcal{U} \psi_2)\}$ |
| RELEASE | $x.(\psi_1 \mathcal{R} \psi_2)$ | $\{x.\psi_1, x.\psi_2\}$ | $\{x.\psi_2, x.\mathsf{X}(\psi_1 \mathcal{R} \psi_2)\}$ |
| EVENTUALLY | $x.\mathsf{F}\psi_1$ | $\{x.\psi_1\}$ | $\{x.\mathsf{XF}\psi_1\}$ |
| ALWAYS | $x.\mathsf{G}\psi_1$ | $\{x.\psi_1, x.\mathsf{XG}\psi_1\}$ | |
| SINCE | $x.(\psi_1 \mathcal{S} \psi_2)$ | $\{x.\psi_2\}$ | $\{x.\psi_1, x.\mathsf{Y}(\psi_1 \mathcal{S} \psi_2)\}$ |
| TRIGGERED | $x.(\psi_1 \mathcal{T} \psi_2)$ | $\{x.\psi_1, x.\psi_2\}$ | $\{x.\psi_2, x.\mathsf{Y}(\psi_1 \mathcal{S} \psi_2)\}$ |
| PAST | $x.\mathsf{P}\psi_1$ | $\{x.\psi_1\}$ | $\{x.\mathsf{YP}\psi_1\}$ |
| HISTORICALLY | $x.\mathsf{H}\psi_1$ | $\{x.\psi_1, x.\mathsf{YH}\psi_1\}$ | |

**Table 7.1:** Expansion rules for the G(TPTL+P) tableau.

Similarly to the LTL+P tableau, the STEP rule is not always applied to all poised nodes, but rather the FORECAST rule is applied first to guess which formulae will be needed for future instances of the YESTERDAY rule. Only when the expansion of such nodes leads again to poised nodes, the STEP rule is applied to them. As a consequence, any poised node has either a number of children added by the FORECAST rule or $\delta_{\mathsf{max}}$ children added by the STEP rule. As in the LTL+P case, given a branch $\overline{u} = \{u_0, \ldots, u_n\}$ of the tableau, we define the *step nodes* as those poised nodes $u_i$ where $u_{i+1}$ is one of the children $u_\delta$, for some $0 \leq \delta \leq \delta_{\mathsf{max}}$, added by the STEP rule. We denote such value of $\delta$ as $\delta(u_i)$. For each node $u_i$ in the branch we can thus define a quantity $\mathsf{time}(u_i) = \sum_{0<j\leq i} \delta(u_i)$, which will correspond to the timestamp of the state corresponding to $u_i$ in the model extracted from the branch. With this notation in place, we can show the remaining rules of the system. The rules are shown in the order they have to be checked on any given poised node. Hence, let $\overline{u} = \langle u_0, \ldots, u_n \rangle$ be a poised branch of the tableau:

EMPTY If $\Gamma(u) = \emptyset$, then, $u_n$ is *ticked*.

CONTRADICTION If $\{x.p, x.\neg p\} \subseteq \Gamma(u_n)$, for some $p \in \Sigma$, then $u$ is *crossed*.

SYNC If either $x.(x \leq x + c) \in \Gamma(u_n)$ or $x.\neg(x \leq x + c) \in \Gamma(u_n)$, but, respectively, $c < 0$ or $c \geq 0$, then $u_n$ is *crossed*.

FORECAST Let $G_n = \{\alpha \in \mathcal{C}(\phi) \mid x.\mathsf{Y}\psi \text{ is a subformula of any } \psi \in \Gamma(u_n)\}$ be the set of formulae involved in any *yesterday* operator appearing inside the formulae of $\Gamma(u_n)$.

Then, at most once before any application of the STEP rule, for each subset $G'_n \subseteq G_n$ (including the empty set), a child $u'_n$ is added to $u_n$ such that $\Gamma(u'_n) = \Gamma(u_n) \cup G'_n$.

YESTERDAY If $x.\mathsf{Y}\psi \in \Gamma(u_n)$, let $u_k$ be the closest ancestor of $u_n$ where the STEP rule was applied, and let $Y_n = \{x.\psi \mid x.Y\psi \in \Gamma(u_n)\}$.

Then, the node $u_n$ is *crossed* if $u_k$ does not exists because there is no application of the STEP rule preceding $u_n$, or if $Y_n \not\subseteq \Gamma^*(u_k)$.

LOOP If there is a step node $u_i < u_n$ such that $\Gamma(u_i) = \Gamma(u_n)$, and all the X-eventualities requested in $u_i$ are fulfilled in $\overline{u}_{[i+1\dots n]}$, then:

1. if $\mathsf{time}(u_i) = \mathsf{time}(u_n)$, then $u_n$ is *crossed*;
2. if $\mathsf{time}(u_i) < \mathsf{time}(u_n)$, then $u_n$ is *ticked*.

PRUNE If there are two positions $i < j \leq n$, such that $\Gamma(u_i) = \Gamma(u_j) = \Gamma(u_n)$, and among the X-eventualities requested in these nodes, all those fulfilled in $\overline{u}_{[j+1\dots n]}$ are fulfilled in $\overline{u}_{[i+1\dots j]}$ as well, then $u_n$ is *crossed*.

Let us discuss this set of rules. In comparison with the rules for the LTL+P tableau, an additional SYNC rule is present, which similarly to the CONTRADICTION rule checks for contradictions, but regarding the timing constraints. The EMPTY, FORECAST, YESTERDAY and PRUNE rules are unchanged. The LOOP rule works similarly, but now has to distinguish two cases, depending on whether time has passed between the two repeating nodes. If the difference in time between the nodes is zero, then the loop cannot be accepted because it would result into a timed state sequence that violates the *progress* condition. It can be seen that the tree has a finite branching factor, hence adapting the arguments used in Theorem 19 of Chapter 6, we can see that the construction always terminates. Moreover, adapting the arguments used in Theorems 20 and 21 of Chapter 6, we can prove the soundness and completeness of the system.

We will again base our arguments on the notion of *pre-model*, adapted to G(TPTL+P) from Section 5.2. Here, an atom is thus a set $\Delta \subseteq \mathcal{C}(x.\phi)$ of formulae from the closure of $x.\phi$ that are closed by expansion (by Table 7.1) and by logical deduction, similarly to Definition 6.4. Then, pre-models are infinite sequences of such atoms, defined similarly to Definition 6.14, but suitably adapted to the structure of G(TPTL+P) and of our tableau.

**Definition 7.10** — Pre-models for G(TPTL+P).
*Let $x.\phi$ be a* G(TPTL+P) *formula. A* pre-model *for $x.\phi$ is a pair $\Pi = (\overline{\Delta}, \delta)$, where $\overline{\delta} = \langle \delta_0, \delta_1, \dots \rangle$ is an infinite sequence of non-negative integers $\delta_i \in \mathbb{N}$, and $\overline{\Delta}$ is an infinite sequence $\overline{\Delta} = \langle \Delta_0, \Delta_1, \dots \rangle$ of* minimal *atoms for $x.\phi$ such that, for all $i \geq 0$:*

1. $x.\phi \in \Delta_0$;

2. *if* $x.\mathsf{X}\psi \in \Delta_i$, *then* $x.\psi^{\delta_{i+1}} \in \Delta_{i+1}$*;*

3. *if* $x.\mathsf{Y}\psi \in \Delta_i$, *then* $i > 0$ *and* $x.\psi^{-\delta_i} \in \Delta_{i-1}$*;*

4. *if* $x.(\psi_1 \mathcal{U} \psi_2) \in \Delta_i$, *there is a* $j \geq i$ *with* $x.\psi^{\delta_{ij}} \in \Delta_j$ *and* $x.\psi_1^{\delta_i k} \in \Delta_k$ *for all* $i \leq k < j$, *where* $\delta_{ij} = \sum_{i<k\leq j} \delta_k$*;*

5. *there are infinite positions* $i$ *such that* $\delta_i > 0$.

With this definition we can adapt the arguments of Theorems 17 and 18 to prove the following result.

**Theorem 26** — The tree-shaped tableau for $\mathsf{TPTL_b{+}P}$ is sound and complete. A $\mathsf{TPTL_b{+}P}$ formula is *satisfiable* if and only if the tableau built on its $\mathsf{G(TPTL{+}P)}$ translation has an accepted branch.

*Proof* (soundness). Let $x.\phi$ be the $\mathsf{G(TPTL{+}P)}$ translation of a $\mathsf{TPTL_b{+}P}$ formula. For Lemma 7.8 we know $x.\phi$ is satisfiable if and only if the original formula is, hence let us focus on the tableau built on $x.\phi$.

To show the soundness of the system, *i.e.*, that if the tableau for $\phi$ has an accepted branch then the formula is satisfiable, we look at one such accepted branch $\overline{u} = \langle u_0, \ldots, u_n \rangle$ and extract a model for $x.\phi$. Let $\overline{\pi} = \langle \pi_0, \ldots, \pi_m \rangle$ be the sequence of step nodes of $\overline{u}$. A pre-model $\Pi = (\overline{\Delta}, \delta)$ can be extracted from an accepted branch. A suitable periodic sequence of atoms $\overline{\Delta}$ can be extracted in the same way as in Lemma 6.8, and $\delta$ can be defined such that $\delta_i = \mathsf{time}(\pi_i)$, where $\pi_i$ is the tableau node corresponding to $\Delta_i$, for the prefix, and consequently in the period. We can check that thanks to the definition of $\mathsf{LOOP}$ rule, an accepted branch can only lead to a pre-model satisfying Item 5 of Definition 7.10. An actual model for $x.\psi$ can then be extracted from $\Pi$ as in Lemma 6.6, with arguments totally similar to those used in Theorem 25, and suitably computing from $\overline{\delta}$ the absolute timestamps of the timed state sequence, proving the soundness of the system.

(completeness). To show the completeness, *i.e.*, that the tableau for a satisfiable formula has at least an accepted branch, the argument based on *greedy pre-models* used for $\mathsf{LTL}$ and $\mathsf{LTL{+}P}$ in Chapter 6 can again be adapted to the $\mathsf{G(TPTL{+}P)}$ case. In particular, the definition of *delays* of the requests of $\mathsf{X}$-eventualities is totally similar to that employed in Section 3 of Chapter 6. Note that these delays still only count the number of atoms from the request of an $\mathsf{X}$-eventuality to its satisfaction, disregarding the timestamps of such atoms. Following the argument used in the proof of Theorem 21, knowing that $x.\phi$ is satisfiable we can suppose to have a *greedy* pre-model $\Pi = (\overline{\Delta}, \delta)$ for $x.\phi$, and traverse the tree to obtain a branch $\overline{u} = \langle u_0, \ldots, u_n \rangle$, as in Lemma 6.9. In this traversal, when descending from a step node $u_i$ through the application of the $\mathsf{STEP}$ rule, we have to choose the child $u_i^{\delta_{i+1}}$, matching the time advancement made by the pre-model and in the branch. If $\overline{\pi} = \langle \pi_0, \ldots, \pi_m \rangle$ is the sequence of step nodes of $\overline{u}$, then, as in the LTL+P case, by construction we have that $\Delta_i = \Delta(\pi_i)$. In addition to the argument employed in the LTL+P case, we only have to observe that if $\overline{u}$ is not accepting, then $u_n$ cannot have been crossed by the SYNCH rule, nor by the LOOP rule, since this would contradict the fact that $\Pi$ is a pre-model for $\Pi$. Then, as in Theorem 21, the node cannot have been crossed by the PRUNE rule either, because it would contradict the assumption that $\Pi$ is a greedy pre-model. Hence $\overline{u}$ must be an accepted branch, completing the proof. ■

# 6 CONCLUSIONS AND OPEN QUESTIONS

This chapter approached the expressiveness of timeline-based planning problems from a *logical* perspective, as opposed to the comparative point of view adopted in Chapter 3, by exhibiting a temporal logic capable of expressing a broad fragment of the whole formalism. The $\mathsf{TPTL_b}$+P logic, a guarded fragment of TPTL+P, has been defined and proved to have an EXPSPACE-complete satisfiability problem. This is in contrast with full TPTL+P which is known to be non-elementary. Furthermore, a one-pass tree-shaped tableau for $\mathsf{TPTL_b}$+P has been shown, extending the one for LTL+P described in Chapter 6.

In contrast to action-based planning languages, which can be easily captured by LTL formulae [42], capturing timeline-based planning with temporal logic proved to be a challenging task. As has been shown, the use of the *freeze quantifier* of $\mathsf{TPTL_b}$+P is essential in expressing the structure of arbitrary rules, but the restrictions needed to keep the complexity of the logic under control make it still not expressive enough to capture the whole formalism, forcing us to define the concept of *forest of bounded components*, restricting the encoding to timeline-based planning problems whose rule graphs satisfy such property. The question of whether a more complex encoding could reach our goal, or whether $\mathsf{TPTL_b}$+P could be extended to capture the whole formalism while keeping the same computational complexity, are still open questions. Past operators are essential in the encoding of synchronisation rules, which can arbitrarily look forward or backward from their trigger. However there might be other ways to add past operators to TPTL while restricting their usage so to recover a good complexity. A possible way might be to retain the full TPTL+P syntax but restricting the maximum *quantifier alternation depth* of the formulae. Given the strict $\forall\exists^*$ structure of synchronisation rules, an encoding similar to that shown here can be defined to capture the whole formalism with formulae of TPTL+P with the form G(FP), *i.e.*, a single universal modality encompassing an arbitrary combination of existential past or future modalities. The conjecture is that this fixed-alternation fragment of TPTL+P may still be EXPSPACE-complete.

The primary motivation behind the pursue of a logical characterisation of timelines is the possibility to leverage the substantial corpus of research devoted to *synthesis* of controllers from temporal logic specifications [56]. Synthesising a controller from a $\mathsf{TPTL}_b$+P specification may represent an alternative and promising way to implement synthesis for controllers of the timeline-based games defined in Chapter 5.

# LIST OF PUBLICATIONS

*Time is an illusion.*
*Lunchtime doubly so.*

---

Douglas Adams
*The Hitchhiker's Guide to the Galaxy*